\documentclass[a4paper]{article}
\usepackage[dvipsnames]{xcolor}
\usepackage[margin=1in]{geometry}
\usepackage{multirow}
\usepackage{stfloats}
\usepackage{silence}
\usepackage{afterpage}
\usepackage[inline]{enumitem}
\usepackage{aux/custom_macros}
\usepackage{algorithm}
\usepackage{algpseudocode}

\usepackage[dvipsnames]{xcolor}

\usepackage[capitalise]{cleveref}

\crefformat{section}{\S#2\color{blue}#1#3}
\crefformat{subsection}{\S#2\color{blue}#1#3}
\crefformat{subsubsection}{\S#2\color{blue}#1#3}

\def\citep{\cite}
\title{Adversarial Robustness of Kernel Machines\\
{\large A Comprehensive Analysis}}

\author{Rahul K. Yadav$^{\dagger}$$\qquad$ Tunir Ghosh$^{\dagger}$$\qquad$ Sudhendu Ahir$^{\dagger}$\\[2ex]
$\qquad$ Parthe Pandit$^{\dagger}$\thanks{Correspondence to \texttt{pandit@iitb.ac.in} and \texttt{arjunp@iitb.ac.in}}$\quad$ Arjun Bhagoji$^{\dagger}$\footnotemark[1]\\[2ex]
$^{\dagger}$Center for Machine Intelligence and Data Science, IIT Bombay}
\date{}

\begin{document}

\maketitle

\begin{abstract}
Kernel machines are an important class of models in machine learning. However, the question of how \emph{robust} modern kernel machines are to adversarial examples has not been studied comprehensively. 
In this paper, 
we provide a thorough empirical investigation of the adversarial robustness of kernel-based models for both binary and multiclass classification tasks, via white-box and black-box evasion attacks. 
We evaluate the native robustness of kernel machines, random feature classifiers, and the relatively newer Recursive Feature Machines. In addition, we provide an algorithm for approximate adversarial training of kernel machines.
Our experiments on classification tasks on image and tabular datasets strongly indicate that kernel machines are natively more robust than neural networks, \textit{i.e.}, their performance degrades more gracefully against adversarial perturbations.
\end{abstract}

\tableofcontents
\section{Introduction} \label{sec: intro}

In domains such as vision and language, machine learning (ML) models based on neural networks (NNs) are often purported to have so-called \textit{superhuman} capabilities. However, this claim is brought into question by the prevalence of test-time attacks known as adversarial examples~\citep{goodfellow2014explaining,carlini2017towards}, even in limited-information settings~\citep{liu2016delving,bhagoji2018practical}, and the lack of effective defenses against them~\citep{athalye2018obfuscated,croce2020robustbench}. Adversarial examples~(\cref{subsec: adv_perturbations}) are minimally-perturbed inputs that induce misclassification in otherwise well-performing classification models. A vast array of attacks and defenses~\citep{croce2020robustbench} have been proposed over the years, with a bulk of them focusing on NNs. Apart from robust training, in both its empirical~\citep{madry2017towards} and certified variants~\citep{li2023sok}, few defenses have been shown to be effective. Robust training which solves a min-max problem, however, is significantly more computationally expensive than standard training, and can lead to unfavourable trade-offs~\citep{zhang2019theoretically}.

In that light, we revisit the question of whether there exists a class of models that is inherently more robust than neural networks. Kernel Machines (KMs) (\cref{subsec: kernel_classifiers}) are a popular class of ML models which are seeing a renewed interest, following results showing that in certain regimes, NNs are approximated by KMs~\citep{jacot2018neural}. In this paper, we empirically compare neural networks and kernel machines, seeking an answer to: 
\begin{center}
\textit{Is there a hierarchy of adversarial robustness (possibly dataset dependent) between function classes?}
\end{center}
In other words, are some classes of models fundamentally harder to attack for a test-time adversary? 

\afterpage{%
\begin{table*}[t]
\centering
\caption{Overview of datasets, models, and attacks compared in this paper.}
\label{tab:overview}
\begin{tabular*}{\textwidth}{@{\extracolsep{\fill}}lll}
\toprule
Choices & Image data & Tabular data\\
\midrule
Datasets & FashionMNIST, KMNIST, QMNIST, SVHN, CIFAR-10 & {\small Adult,Phishing,BankMarketing}\\
Models & {\footnotesize CNN, FCNN, FCNN-Adv, KRR, KLR, KSVM, RFC, RFM, Adv-KLR} & FCNN, KRR, RFM \\
Attack &  $\ell_\infty,\ell_2$-balls with white-box (\texttt{autoattack}), black-box (\texttt{spsa}) & $\ell_\infty$-ball with \texttt{cafa} \\
\bottomrule
\end{tabular*}
\end{table*}
}

Three popular versions of kernel machines exist with different qualities: (i) the standard KMs trained using kernel ridge regression (KRR), Kernel Support Vector Machines (KSVM) or Kernel Logistic Regression (KLR), (ii) Random feature classifiers (RFCs) which approximate KMs in a scalable manner~\citep{rahimi2007random}, and (iii) Recursive Feature Machines (RFMs) which enhance performance via feature learning~\citep{radhakrishnan2024mechanism}. Apart from performance in terms of accuracy, other important properties of KMs such as adversarial robustness have received little attention~\citep{wang2021adversarial}. 

\begin{figure*}[!htbp]
    \centering
    \includegraphics[width=\textwidth]{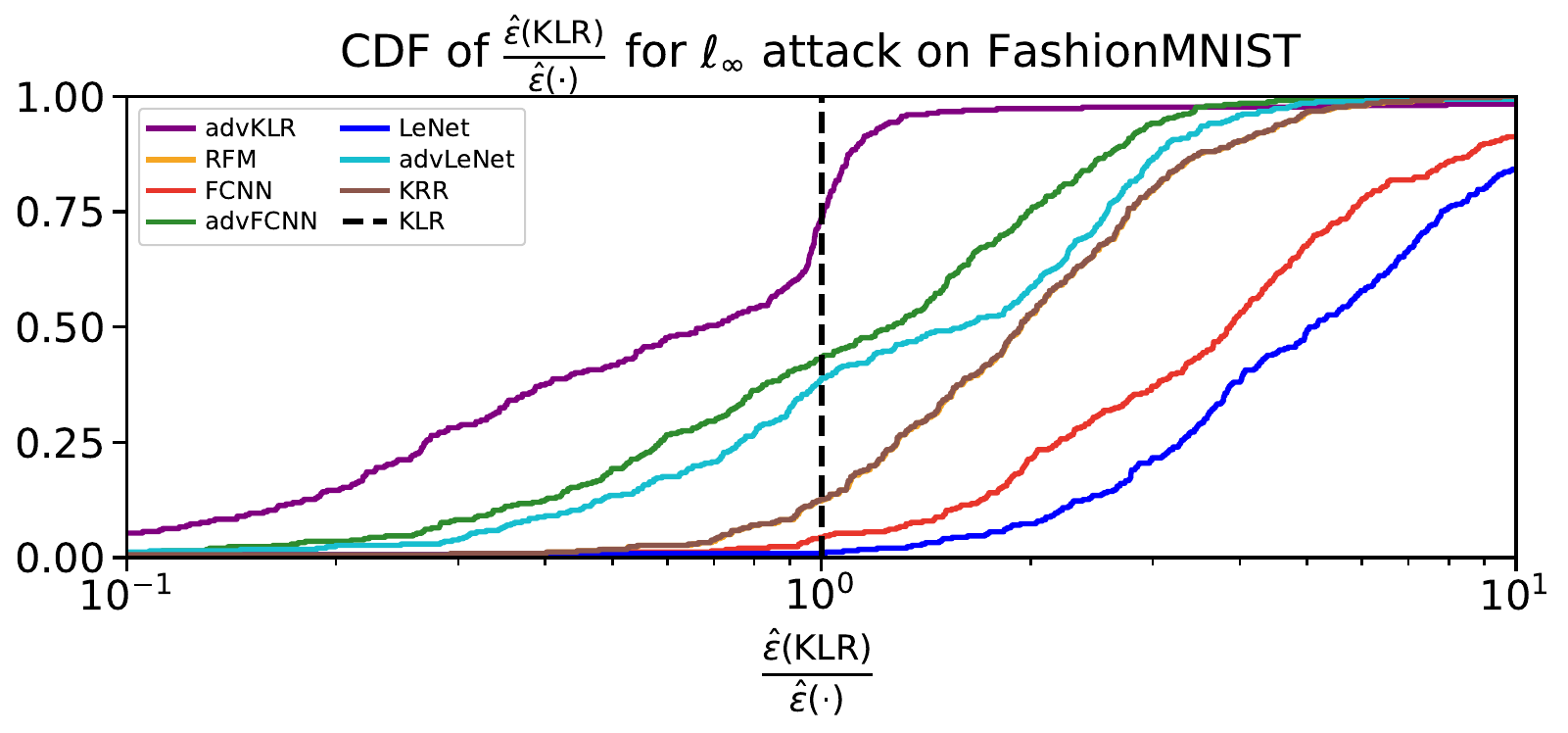}
    \caption{
Test for robustness as a pointwise property. We plot the empirical CDF of the ratio (KLR/other models) of the smallest $\ell_\infty$ perturbations $\wh\eps$ needed to successfully attack 388 random samples which were classified correctly by all the models from the test set of FashionMNIST. Note the x-axis is in log-scale. See \cref{subsec:pointwise_robustness} for more details.\\
For most samples, KRR is harder to attack than the FCNN. For the median datum, KLR is more robust than Adv-FCNN and Adv-LeNet. The adversarially trained FCNN, LeNet and KLR were trained with an $\ell_\infty$ (matched) adversary with a budget of $\eps=\frac{8}{255}$. Adversarially trained KM remains most robust. In the case of FMNIST, KRR and RFM overlap. We have used Gaussian KLR, Laplacian KRR, and Laplacian RFM.}
\label{fig:fmnist_pointwise}
\end{figure*}

We present a thorough empirical comparison of the robustness of different kernel-based classifiers (See \Cref{tab:abbr} in the Appendix for abbreviations.)\footnote{Code at ~\href{https://github.com/irohs-lab/robust-kms/tree/main}{https://github.com/irohs-lab/robust-kms/tree/main}.}: KMs, RFCs, and RFMs against fully connected neural networks (FCNNs) and convolutional neural networks (CNNs), namely LeNet-5~\citep{lecun1998gradient} and WideResNet-28-10 (WRN-28-10)~\citep{zagoruyko2016wide}, as well as their robustly trained variants across 8 datasets and several different attacks~(\cref{tab:overview}). We provide a fine-grained analysis at 3 levels: 
(i) robust accuracy, which is the aggregate behaviour across the test dataset across all classes, (ii) \textit{one-v-one} robust accuracy, which breaks down the robust accuracy across $\binom{C}2$ binary classification tasks, and (iii) pointwise robustness, which estimates the distance of an individual datum from the decision boundary. 

While the adversarial training of KMs is a challenging non-convex problem, we derive an approximate formulation for the case of Lipschitz loss functions (\cref{sec:mark}). For binary classification this is a convex formulation, however for multiclass problems this yields a minimization problem of a convex function over rank-1 matrices, which is a non-convex domain. We provide an iterative algorithm for training these models along with an empirical analysis.

\subsection{Main findings}
\begin{figure}[t]

\includegraphics[width=\columnwidth,keepaspectratio]
{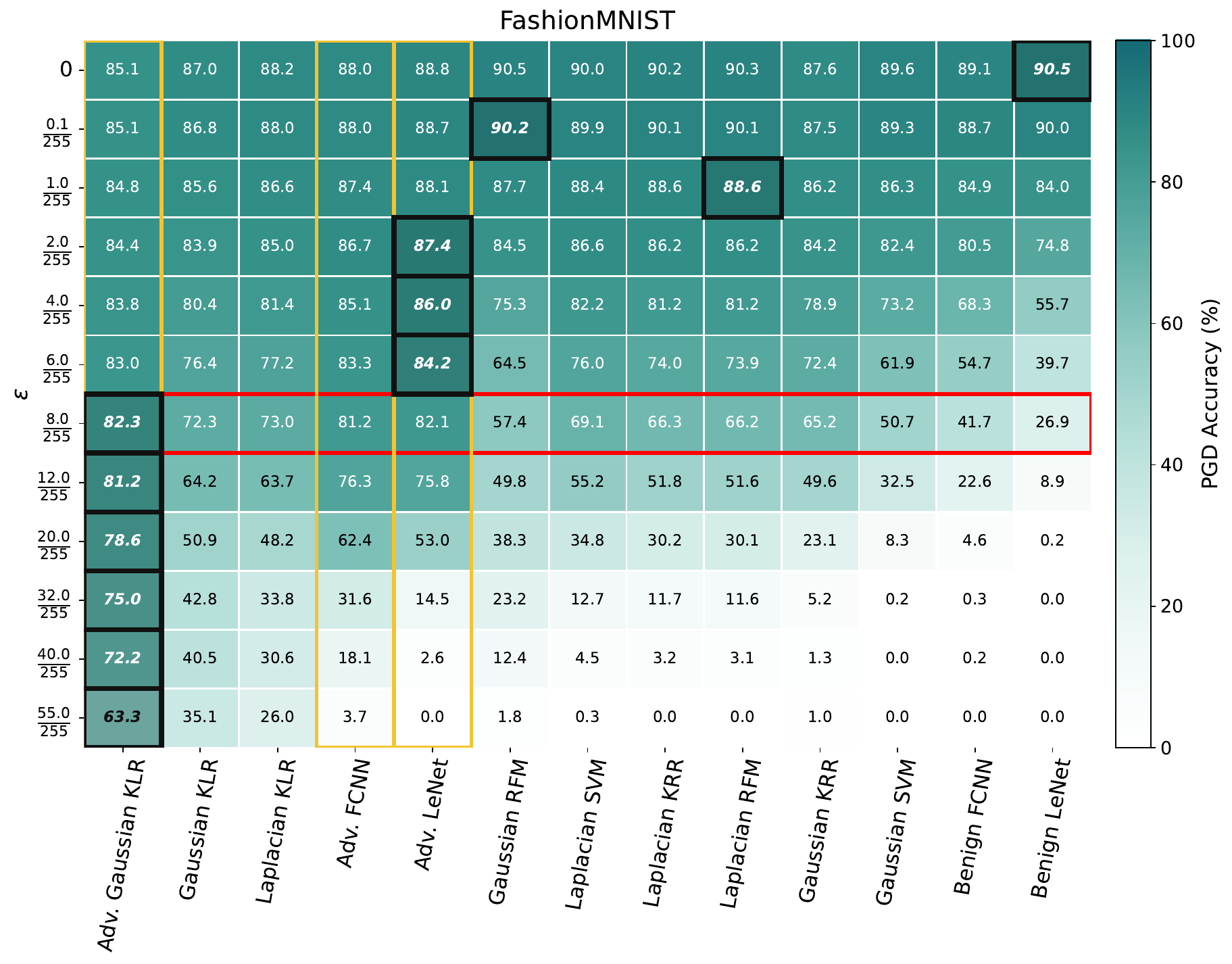}
\caption{\label{fig:fmnist_mc} Summary of results for the multiclass FashionMNIST dataset. We compare robust accuracy of 13 models (3 adversarially trained, 10 benign) on FashionMNIST (10 classes). Columns with yellow borders correspond to adversarially trained models with training adversary's attack level highlighted by the red row. Black boxes mark the best robust accuracy across each row. Note the best models for high $\eps$ are kernel machines with logistic loss.}
\end{figure}

\paragraph{Kernel machines are more robust than neural networks} (\cref{subsec: kernels_vs_neural_nets}):
    The robustness of KMs degrades more gracefully than NNs as the attack budget $\varepsilon$ increases. KMs are consistently more robust than NNs trained in the standard fashion, \emph{i.e.} without adversarial training. 
    In some cases, the difference is up to $40$ percentage points in multiclass classification tasks and $70$ percentage points in binary classification tasks, with KMs providing near equivalent performance on clean data (\cref{fig:fmnist_mc}). 
\paragraph{Recursive Feature Machines retain the robustness of the initial KMs} (\cref{subsec: rfm_robustness}): RFMs are a variant of KMs equipped with feature learning. Our experiments show that RFMs retain nearly all of the robustness of their native KMs, but their robustness can degrade with additional iterations of feature learning.
    \paragraph{Random Feature classifiers can be robust at a sufficiently high number of features} (\cref{subsec: rff_robustness}): RFCs are a scalable approximation to KMs which can match their performance in terms of zeroth order metrics such as generalisation error and accuracy. We observe that increasing the number of random features consistently makes these models more robust, except in the case of Laplacian RFCs trained using square loss. This behaviour is stable for Gaussian RFCs whereas it exhibits a large variability for Laplacian RFCs, in part due to the heavy tailed nature of the weight matrices appearing in the Random Fourier map for the Laplacian kernel.
    \paragraph{Kernel Machines can be nearly as robust as adversarially trained NNs} (\cref{subsec: kms_vs_adv_nns}): On certain datasets, KMs, even without adversarial training, have a robustness profile similar to adversarially trained NNs. While this does not hold across all datasets and training budgets, it is a surprising finding even with these caveats. Their robustness can be further improved using our approximate adversarial training method, providing high robust accuracy even under powerful attacks.
    \paragraph{Higher robustness of KMs is likely a point-wise property} (\cref{subsec:pointwise_robustness}): It is instructive to understand, on a per-sample basis, how far a particular data point is from the classification boundary. For a median datum, we find that KLR is more robust than adversarially trained FCNN on FashionMNIST. We use bisection coupled with AutoAttack~\citep{croce2020reliable} to find the boundary distance empirically.

 In Appendix \ref{appsec: further_ablations}, we carry out extensive ablations. We first \begin{enumerate*}[label=(\roman*)]
  \item check for the presence of gradient masking, and
  \item  evaluate our models on tabular datasets. 
\end{enumerate*}
We also ablate over kernel classifier parameters such as the \begin{enumerate*}[label=(\alph*)]
  \item smoothness parameter $\nu$ in the Mat\'ern kernel,
  \item RFM iterations,
  \item length scale ($\sigma$),
  \item ridge penalty ($\lambda$),
  and \item number of features used in RFC.
\end{enumerate*}
Overall, this paper motivates further rigorous analyses of the robustness of KMs. It is interesting that interpolation or near-interpolation on training data, which is the behaviour of KRR, may not be at odds with adversarial robustness with respect to test data. In future work, a satisfactory theoretical result would provide a pointwise gap between the margins of KMs and NNs, which could perhaps be due to a \textit{flatter} loss landscape for the adversary to climb out of.

\subsection{Prior work on robustness of KMs}

The problem of the robustness of kernel-based  classifiers has received some limited attention. However, a comprehensive comparative analysis of the adversarial robustness of this class of models across the 3 variants- KMs, RFC, and RFM, listed above has not been carried out before.

\paragraph{Robustness to contamination:}  KMs with outlier-resilient loss functions have been studied. These include the Cauchy loss \citep{wen2025robustness}, Huber loss \citep{wang2022huber}, $\ell_1$ loss \citep{wang2014least}, among others \citep{feng2015learning}. For example, Li et al. \citep{li2020gradient} show robustness to label noise for KMs with the Neural Tangent Kernel~\citep{jacot2018neural} trained using early stopping.

\paragraph{Test-time perturbations against kernel machines:} Xu et al.~\citep{xu2009robustness} showed that KSVMs are robust to point-wise uncertainty at test time, by using an upper bound to the hinge loss with a regularized variant. However, they do not provide any empirical evidence of robustness, and only operate with perturbations in feature space. Hassani and Javanmard~\citep{hassani2024curse} carried out a theoretical analysis of the robustness of random feature maps in the overparametrized setting, and show that overparametrization hurts robust generalization in a proportional asymptotic regime where $d,p,n\to\infty$. However, this analysis is more attuned to understanding the RF approximation of KSVMs. Wu \emph{et al.}~\citep{wu2021fast} provide a method to adversarially train KMs in feature space via doubly stochastic gradients over a convex upper bound of the adversarial loss on the RFC approximation. In the work closest to ours, Wang~\citep{wang2021adversarial} analyses the empirical robustness of kernel SVMs trained with the linear and Gaussian kernels, and finds they are not robust. Han and Rubinstein~\citep{han2017adequacy} study gradient-descent evasion attacks on Gaussian SVMs and find that carefully reduced kernel smoothness substantially increases robustness. We find that KLR, KRR and KSVM with the Laplacian kernel, and the Gaussian kernel, with proper tuning, are robust. 

\section{Preliminaries}\label{sec: preliminaries}
We focus on the setting of multiclass classification with $C$ classes. For $\Omega\subset\Real^d$, let $f:\Omega\to\Real^C$ be a model associated with a classifier  $x\mapsto \displaystyle
\arg\max_{i\in[C]} f_i(x)$ for $x\in\Omega.$ Note $f(x)$ are the logits for a test sample $x$, and $f_i(x)=e_i\tran f(x)$, the $i^{\rm th}$ logit, for $i\in[C]$. The loss function is represented as
$\mathsf{L}:\Real^C \times [C] \mapsto \Real$.

\subsection{Adversarial examples}
\label{subsec: adv_perturbations}
    At a point $x,$ we say $\delta$ is an \textit{adversarial perturbation}  within a $\ell_p$-\text{budget} of $\eps$ if, $\norm{\delta}_p \leq \eps$ and 

$\arg\max_{i\in[C]}f_i(x)\neq\arg\max_{i\in[C]}f_i(x+\delta)$.

We also call $x+\delta$ an \textit{adversarial example}. A popular class of methods for finding adversarial perturbations~\citep{carlini2017towards,goodfellow2014explaining,madry2017towards} is to solve the following non-convex optimisation
\begin{subequations}\label{eq:adv_opt}

\begin{align}
&\max_{\delta\in\Real^d}~ \ell(\delta):=\mathsf{L}(f(x+\delta),\arg\max_{i\in C}f_i(x))
\\
&\text{subject to } \norm{\delta}_p\leq \eps,\quad x+\delta\in\Omega
\end{align}
\end{subequations}
for some loss function $\mathsf{L}$, using a local search method. 

The typical choice for $\mathsf{L}$ is the cross-entropy loss.
Note that even though $u\mapsto\mathsf{L}(u,v)$ may be convex for all $v,$ the $\delta\mapsto \mathsf{L}(f(x+\delta),v)$ is not convex in general, unless of course $f$ is linear in its input $x$. In this paper we focus on  models that are nonlinear in the input such as NNs and KMs. Note KMs, \cref{eq:kernel_machine}, are linear in the parameters $\alpha$, not in the input $x$.

Tools for finding adversarial perturbations (also called \textit{attacks}) are classified based on whether they provide the adversary input-output access $f$, called \textit{black-box}~\citep{bhagoji2018practical,liu2016delving} attacks, or access to the implementation of $f$, called \textit{whitebox} attacks. Most white-box attacks, typically solve \cref{eq:adv_opt} approximately via an iterative first order method starting from $\delta=0$. In this paper we use a popular tool for white-box attacks: AutoPGD~\cite{croce2020reliable} which adaptively selects the step size for the non-convex optimisation in~\cref{eq:adv_opt}. Athalye et al.\citep{athalye2018obfuscated} recommend also evaluating with black-box attacks to ensure there is no inadvertent gradient masking. We use SPSA~\citep{spall1992spsa,uesato2018adversarial} (details in~\cref{apd: spsa_attack}). On tabular data, we use the Cost-aware Feasible Attack (CaFA) ~\citep{bentov2024cafa} which ensures input validity.

\subsection{Kernel-based models}\label{subsec: kernel_classifiers}

For a compact domain $\Omega\subset\Real^d$, let $K:\Omega\times\Omega\to\Real$ be a kernel, \emph{i.e.}, a bivariate positive definite map. For a collection of vectors $X=\set{x_i}_{i=1}^n$, we denote the kernel matrix with $K(X,X)=(K(x_i,x_j))_{1\leq i,j\leq n}.$

\paragraph{Kernel machines (KM):} A kernel machine is a model with parameter $\alpha\in\Real^{n\times C}$, and takes the form
\begin{align}\label{eq:kernel_machine}
    f(x) = \sum_{i=1}^n \alpha_i K(x,x_i).
\end{align}
where $\alpha_i\in\Real^C$. This class of models are of interest in machine learning because the representer theorem \citep{scholkopf2001generalized} tells us that any solution to a non-parametric statistical estimation problem, 
\begin{align}\label{eq:representer_theorem}
    \min_{f\in\Hilbert_K} J\round{\set{x_i,f(x_i),y_i}_{i=1}^n} + \lambda\cdot R(\norm{f}^2_{\Hilbert_K}),
\end{align}
with $\lambda>0$, must have the form in \cref{eq:kernel_machine}. Here the search space $\Hilbert_K$ is the reproducing kernel hilbert space (RKHS) corresponding to $K.$ Note here that $J$ is an arbitrary loss function, and $R$ is any increasing regularization function. We refer the reader to~\citep{hofmann2008kernel,scholkopf2002learning,steinwart2008support} for a review of kernel machines. We note that the case of $C>1,$ calls for the use of so-called vector valued RKHSs corresponding to vector valued kernels $K_{vv}$, such that $K_{vv}(x_i,x_j)$ is a $C\times C$ matrix and $K_{vv}(X,X)\in\Real^{nC\times nC}$ is a symmetric positive definite matrix with $C\times C$ blocks. However, in what follows, we have considered $K_{vv}(x_i,x_j)=K(x_i,x_j) I_C$. We train KMs using the square loss (Kernel Ridge Regression), hinge loss (Kernel Support Vector Machine) and logistic loss (Kernel Logistic Regression) (\cref{apd: matern_kernels_background}).

\paragraph{Derivatives:} We define $G_1(x,z):=\frac{\partial}{\partial x}K(x,z)\in\Real^d$. Similarly $G_2(x,z):=\frac{\partial}{\partial z}K(x,z),$ and $G_{12}(x,z):=\frac{\partial^2}{\partial x\partial z}K(x,z)\in\Real^{d\times d}$. Observe that $G_1(x,z)=G_2(z,x)$, and $G_{12}(x,z)=G_{12}(z,x)\tran$. We use these for adversarial training of kernel machines later in \Cref{sec:mark}.

\paragraph{Recursive feature machines (RFM):} A new family of kernel machines was proposed by~\citep{radhakrishnan2024mechanism} where the kernel function was parameterized by a positive semidefinite matrix $M\in\Real^{d\times d}$ as
    $K_M(x,z) = K (\sqrt{M}x,\sqrt{M}z),$
where $\sqrt{M}$ is the unique symmetric square root matrix of a positive semidefinite matrix $M$, and $K$ is some base kernel. The RFM is the model \cref{eq:kernel_machine} with $K$ replaced by $K_M$, and the trainable parameter $M$ in RFMs is updated recursively using the average gradient outer product (AGOP) of the model, using the training data $\set{x_i}_{i=1}^n$
\begin{align}\label{eq:km_RFM}
    M_{t+1} \gets \tfrac1n\sum_{i=1}^n\grad_x f_{M_t}(x_i) \round{\grad_x f_{M_t}(x_i)}\tran
\end{align}
where $f_{M}$ denotes the model in \cref{eq:kernel_machine} with kernel $K=K_M$, and $\alpha\in\Real^{n\times C}$ obtained by solving \cref{eq:krr} with $K=K_M$, and $\grad_{x}f_M\in\Real^{d\times C}$ is the Jacobian. RFMs provide the model a property formally studied as \textit{feature learning}, which is natively absent in KMs. RFMs are state of the art for certain tasks, such as prediction on tabular datasets~\citep{beaglehole2025xrfm}.

\paragraph{Random feature classifiers (RFCs):} Kernel machines for a large class of kernels can be learned via the so-called random features approximation~\citep{rahimi2007random,kar2012random,yu2016orthogonal}. This approximation makes inference independent of the size of the dataset $n$, unlike \cref{eq:kernel_machine}, which may become prohibitive for tasks with large datasets. If $K$ has a special form 
\begin{align*}
    K(x,z)=\Exp_{w\sim \Prob_w} \varsigma\round{\inner{w, x}_{\Real^d}}\cdot\varsigma\round{\inner{w, z}_{\Real^d}},
\end{align*}
then \cref{eq:kernel_machine} can be approximated as 
$f_\theta (x) = \inner{\theta,\phi(x)}_{\Real^m},$
where $\phi(x)=\frac{1}{\sqrt{m}}\varsigma(Wx)$, the vector $\theta=\sum_{i=1}^n\alpha_i \phi(x_i)$, and the rows of $W\in\Real^{m\times d}$ are $m$ i.i.d. samples from $\Prob_w$, where $(\varsigma,\Prob_w)$ depend on $K$. Several classes of kernel functions such as the Gaussian and Laplacian kernel, can be approximated in this manner.

\section{Main Experimental Results}\label{sec: main_results}

Our goal is to test whether kernel-based classifiers, in any of the 3 variants described above, are more robust than neural networks, whether fully connected or convolutional.
A summary of the settings considered is in~\cref{tab:overview}. Our key findings are i) KMs are more robust than both FCNNs and CNNs (LeNet) on average~(\cref{subsec: kernels_vs_neural_nets}) and point-wise~(\cref{subsec:pointwise_robustness}); ii) RFMs are nearly as robust as KMs~(\cref{subsec: rfm_robustness}); and iii) RFCs become robust with an increase in the number of features~(\cref{subsec: rff_robustness}). These results are captured succinctly in~\cref{fig:fmnist_mc,fig:kmnist_mc,fig:qmnist_mc,fig:fmnist_kmnist_qmnist_mc_l2} for the case of $\ell_\infty$ and $\ell_2$ attacks on the FashionMNIST, KMNIST and QMNIST dataset.

\subsection{Setup Summary}\label{subsec: setup}
We briefly summarise our experimental setup here. All experiments were run on a cluster with 8 NVIDIA RTX 6000 GPUs. All further details on hyperparameter searches and attack budgets used are in~\cref{appsec:background_experiments}.

\noindent \textbf{Datasets:} Our experiments for image data are with {\small FashionMNIST}~\citep{xiao2017fashion}, {\small KMNIST}~\citep{clanuwat2018deep}, {\small QMNIST}~\citep{qmnist-2019}, {\small SVHN}~\citep{netzer2011reading} and {\small CIFAR-10}~\citep{krizhevsky2009learning}. For tabular data, we use the Adult~\citep{adult_2}, Bank Marketing~\citep{bank_marketing_222} and Phishing~\citep{phishing_websites_327} binary classification datasets.

\noindent \textbf{Choice of \textit{best} model to attack:} We define the best model to be the one with the highest benign accuracy on the validation set, across all tunable hyperparameters.

\noindent \textbf{Measurements and comparisons:} We measure the robust accuracy of a model given $f$ via 
\begin{align*}
    \text{ra}_\eps(f)\!:=\!1 - \Prob\set{\exists\delta:\norm{\delta}_p\leq\eps,\arg\max_{i\in[C]}f_i(x+\delta)\neq y}.
\end{align*}
using a plug-in estimator for $\Prob$ and commonly used search criteria for finding $\delta$. 
We report robust accuracies for both settings - multiclass classification and binary classification. For a $C$ class classification problem we have $\binom{C}{2}$ binary classification problems, which allow us to provide a fine-grained analysis.

\noindent \textbf{Kernel machines:} We consider kernel machines trained using kernels from the Mat\'ern family~(\cref{apd: matern_kernels_background}), of which both the Laplacian and Gaussian kernels are a part. KMs considered in this section were trained using kernel logistic regression (KLR), kernel ridge regression (KRR) and kernel SVM (KSVM), with $M = I_d$. KLR uses the cross-entropy loss, KRR uses the square loss and KSVM uses the hinge loss. For KRR and KSVM, we tuned the length scale $\sigma$ and the regularisation strength $\lambda$ by grid search. For KLR, we searched over length scale $\sigma$ and learning rate. Detailed results for the Laplacian ($\gamma=1$) and Gaussian ($\gamma=2$) kernels are in Appendix~\ref{appsec: full_results}. 

\noindent \textbf{RFCs and RFMs:} The RFCs were trained using the square loss as well as the cross-entropy loss, each with ridge regularisation. For RFMs, the matrix $M$ is set adaptively rather than fixed to $I_d$ (\cref{subsec: kernel_classifiers}), and the models were trained with the square loss. The RFMs were tuned for up to $5$ iterations, with search over length scale $\sigma$ and regularisation strength $\lambda$ being the same as in KRR. 

\noindent \textbf{Neural networks.} We train both fully connected neural networks (FCNNs) and Convolutional neural networks (CNNs). For the latter, the modified LeNet5 (using ReLU activation and max-pooling layers)~\citep{lecun1998gradient}, and WideResNet-28-10~\citep{zagoruyko2016wide} architectures were used. Runtime comparisons are in~\cref{tab:computational_complexity}. 

\noindent \textbf{Attacks:} For a rigorous evaluation of robustness, and so as not to fall into common pitfalls~\citep{carlini2019evaluating}, we use two different white-box attacks, as well as a black box attack. In all cases, we use the cross-entropy loss. In the white-box case, we consider both i) a fixed iteration projected gradient descent (PGD) attack~\citep{madry2017towards} and ii) \texttt{AutoAttack}, using apgd-ce~\citep{croce2020reliable,croce2021mind} (the standard in benchmarks such as RobustBench~\citep{croce2020robustbench}), and report the \emph{lower} of the two robust accuracy numbers.

\subsection{KMs are more robust than neural nets}\label{subsec: kernels_vs_neural_nets}
For multi-class classification problems, we see that the robustness of both Laplacian and Gaussian KRR degrades more gracefully than that of both FCNNs and LeNets~(\cref{fig:fmnist_mc,fig:kmnist_mc,fig:qmnist_mc,fig:fmnist_kmnist_qmnist_mc_l2}), particularly in the case of FMNIST. A comprehensive evaluation for all the $1\text{v}1$ sub-problems for FashionMNIST, KMNIST and QMNIST is presented in \cref{fig:gap_heatmaps_lap_vs_fcnn,fig:gap_heatmaps_gauss_vs_fcnn}, which clearly establishing that, as the attack budget increases for both $\ell_\infty$ and $\ell_2$ attacks, kernel machines are more robust in a majority of cases. These demonstrate that while the Laplace kernel seems to be better choice for robustness, the Gaussian kernel also offers robustness benefits. Perhaps surprisingly, fine-grained class-wise analysis reveals that the \emph{robustness of Laplacian KRR is even higher than of the LeNet} as the perturbation budget increases~(\cref{fig:gap_heatmaps_lap_vs_lenet,fig:gap_heatmaps_gauss_vs_lenet}). In terms of the benign accuracy ($\eps=0$), we find that the LeNet is marginally better on these datasets, but KRR can outperform the FCNN. For example, in the case of $5\text{v}9$ binary classification problem on FMNIST, KRR has $0.8$ percentage points better benign accuracy. We further compare LeNet with the substantially larger WRN-28-10~(\cref{subsec:wrn_vs_lenet}). Despite its higher clean accuracy, WRN-28-10 loses robust accuracy faster than LeNet as the perturbation budget increases~(\cref{tab:pgd_combined}). Taken together, these results suggest that the robustness advantage of KRR over LeNet is not due to the model size used for comparison. 

\noindent \textbf{KSVMs:} Kernel SVMs (both Laplacian and Gaussian) also outperform FCNNs and LeNets in terms of robust accuracies for both multiclass (\cref{fig:fmnist_mc,fig:kmnist_mc,fig:qmnist_mc,fig:fmnist_kmnist_qmnist_mc_l2}) and binary classification problems (\cref{fig:gap_heatmaps_lapsvmvsfcnn,fig:gap_heatmaps_lapsvmvslenet}). At the same time they can be better than their corresponding KRR. For example, in~\cref{fig:fmnist_mc}, Laplacian SVM performs better than the Laplacian KRR for FMNIST, while for QMNIST~(\cref{fig:qmnist_mc}), Gaussian SVM is better than the Gaussian KRR. KSVMs also remain competitive in terms of benign accuracy. For instance, in FMNIST, the Laplacian SVM is $0.6$ percentage points better than FCNN on the same task.

\noindent \textbf{KLR:} Like KRR and KSVM, the robust accuracy of KLR degrades quite gracefully as compared to FCNNs and LeNets. They are also quite robust in comparison to KRRs and KSVMs as seen in~\cref{fig:fmnist_mc,fig:kmnist_mc,fig:qmnist_mc}. For example, on KMNIST at $\eps=40/255$, Gaussian KLR retains $73.4\%$ accuracy, whereas adversarially trained LeNet, adversarially trained FCNNs, Gaussian KRR, and Gaussian SVM retain $28.0\%$, $20.1\%$, $12.5\%$, and $16.2\%$ respectively. Laplacian KLRs also show the similar behaviour, but are less robust in comparison to Gaussian KLR. In terms of benign accuracy, however, KLR lags behind almost all the other models. For example, in the case of FashionMNIST, adversarially trained Gaussian KLR has the lowest benign accuracy followed by Gaussian KLR, Gaussian KRR, adversarially trained FCNN and Laplacian KLR. Similar observations can be seen in the case of KMNIST and QMNIST. The 1v1 results show both Laplacian and Gaussian KLR being more robust than all the other benign trained models at larger perturbation radii. In addition, for the 0v8 problem on FashionMNIST, Gaussian KLR is having $70.3$ percentage points more accuracy than FCNN at the $\varepsilon=\frac{20}{255}$. The 1v1 benign accuracies are competitive with FCNN and LeNet in most cases as seen in~\cref{fig:gap_heatmaps_lapklr_vs_fcnn,fig:gap_heatmaps_lapklr_vs_lenet,fig:gap_heatmaps_gaussklr_vs_fcnn,fig:gap_heatmaps_gaussklr_vs_lenet}. Overall, we find that the Gaussian KLR is more robust than the Laplacian KLR.  Similar observations hold across the other datasets.

\subsection{RFMs are nearly as robust as base KMs} \label{subsec: rfm_robustness}
RFMs allow for feature learning in kernel machines. As observed in previous work~\citep{radhakrishnan2024mechanism}, we find this can lead to improved accuracy on the base benign task.  However, the improved feature learning does not always translate to improved robustness~(\cref{fig:gap_heatmaps_lapkervsrfm} for Laplacian and \cref{fig:gap_heatmaps_gauss_rfmvsker} for Gaussian). In particular, we find that KRR (with $M=I$ in~\cref{eq:km_RFM}) is often more robust when the RFM has improved benign accuracy, \emph{i.e.} some feature learning has taken place. In the case of multi-class classification~(\cref{fig:fmnist_mc,fig:kmnist_mc,fig:qmnist_mc,fig:fmnist_kmnist_qmnist_mc_l2}), there is a clear separation, with KRR being more robust for the KMNIST and QMNIST datasets across budgets and attack types. For FMNIST, the RFM has competitive robust accuracy.

\noindent \textbf{RFMs vs NNs:} Fine-grained class-wise analysis~(\cref{fig:gap_heatmaps_laprfmvsnn,fig:gap_heatmaps_gauss_rfmvsnn}) shows that RFMs are almost always more robust compared to the FCNNs. Additionally, unlike in the case of KRR, RFMs can have higher benign accuracy than the FCNNs, and are competitive with the LeNet. Thus, RFMs are a good choice of model, at least for simpler datasets, offering attractive best-of-both worlds performance. 

\subsection{Robustness of RFCs increases with number of random features} \label{subsec: rff_robustness}

We observe that random feature classifiers (RFCs) suffer in terms of benign and robust accuracies compared to the base KMs when the number of features is inadequate. To compare RFCs with other models, we use $5000$ and $20000$ features for binary classification problems using square loss~(\cref{fig:gap_rfc_summary_linf}), while in the multiclass case we carry out an extensive ablation over the number of features~(\cref{apd: num_of_ftr_vs_robust_results}) to study its effect on robust accuracy. In multiclass case, RFCs are trained using both cross-entropy and square loss. 

For Gaussian RFCs, trained with either the cross-entropy or the square loss, robustness is low when the number of features is small, but increases steadily as the number of features grows, in both the 1v1 and multiclass settings. The same trend holds under both SPSA and PGD, and the robust accuracy under SPSA is consistently higher hence, we can say for Gaussian RFCs, no gradient masking is happening. With enough features, Gaussian RFCs become more robust than FCNNs and LeNet in many cases.

However, we find that Laplacian RFCs trained with either loss exhibit significant gradient masking. SPSA attacks reduce their robust accuracy substantially compared to white-box attacks~(\cref{apd: bb_results}). Robust accuracy under first-order gradient-based attacks therefore overestimates their true robustness and should not be taken at face value. This includes the high robust accuracy of the cross-entropy-trained Laplacian RFC in~\cref{fig:num_of_robust_lap_ce_fmnist}, which holds up under white-box attacks even at very large budgets. For the cross-entropy loss, the robustness still increases with the number of features (when attacked against SPSA and PGD). For the square loss the robust accuracy of Laplacian RFCs varies erratically with the number of features, and more features do not yield more robustness, unlike for Gaussian RFCs. These observations warrant further theoretical investigation.

Overall, these results call into question the practical validity of the results from Hassani and Javanmard~\cite{hassani2024curse}, which indicates that overparametrisation hurts robustness for RFCs. A plausible explanation is that their feature isotropy assumption yields pessimistic results. Thus, the implications of overparametrisation for robustness must be revisited. 

\subsection{Robustness of KMs is comparable with adversarially trained neural networks} \label{subsec: kms_vs_adv_nns}
The dominant approach to training robust NNs has been adversarial training~\citep{madry2017towards} and its variants~\citep{zhang2019theoretically,wang2019improving}. While adversarial training is significantly more computationally expensive than benign training, it is instructive to compare whether NNs, particularly FCNNs, match the intrinsic robustness of KMs post adversarial training. Experimental details for adversarial training are in~\cref{apd: adv_training}. In the multiclass setting for $\ell_\infty$ in~\cref{fig:fmnist_mc,fig:kmnist_mc,fig:qmnist_mc}, we see that for the FMNIST, KMNIST and QMNIST datasets that Gaussian KLR outperforms adversarially trained neural networks. In the case of $\ell_2$ attacks, as in~\cref{fig:fmnist_kmnist_qmnist_mc_l2}, we can see that Laplacian KRR performs at par with adversarially trained FCNN and LeNet (both of which are trained at $\ell_2$ perturbation radius of $1$). Fine-grained class-wise analysis for the Laplacian kernel compared to $\ell_2$ training at $\eps=1$~ and $\ell_\infty$ training at $\eps=\frac{6}{255}$~(\cref{fig:gap_lapvsadvnn_l2_1}) reveal that depending on the threat model, and the specific class pair, KRR can be more robust than adversarially trained neural networks.

\section{Adversarial training of kernel classifiers}
\label{sec:mark}
A particularly important problem which remains open is that of adversarially training kernel machines. Prior approaches~\cite{wang2021adversarial,wu2021fast} considered the problem in the feature space, and hence do not perform adversarial training in the input space. Both these methods use the Xu et al.~\cite{xu2009robustness} upper bound to create an Euclidean boundary in feature space which is often over-conservative.
In what follows, we provide a method for adversarially training KMs by approximation, via an upper bound on the robust loss.

Consider adversarial training over the RKHS, for $\eps>0$,
\begin{subequations}\label{eq:adv_train_rkhs}
\begin{align}
&\min_{f\in\Hilbert_K}\sum_{i=1}^n \max_{\delta_i\in\Real^d} \ell_i(f(x_i+\delta_i)) +\tfrac\lambda2\norm{f}_{\Hilbert_K}^2\\
&\text{subject to }\norm{\delta_i}_p\leq \eps, ~\forall\,i\in[n]
\end{align}
\end{subequations}
$p\geq 1$. We look at binary classification and multiclass classification separately below.

\subsection{Convex approximation for binary classification}
The objective function in \cref{eq:adv_train_rkhs} no longer has the form of \cref{eq:representer_theorem} whereby there is no guarantee of a finitely represented solution like \cref{eq:kernel_machine}. However one could consider an approximation by minimizing a linearization for small $\eps$. By H\"olders inequality, for a twice differentiable loss function and predictor function, this yields
\begin{align}
\label{eq:apx_adv_train_rkhs} 
\min_{f\in\Hilbert}\!\sum_{i=1}^n  \ell_i(\!f(x_i))\! +\! \eps\norm{\ell_i'(f(x_i))^{\!\top}\!\grad\! f(x_i)}_{q}\!+\!\tfrac\lambda2\norm{f}_{\Hilbert}^2
\end{align}
where $\tfrac1q+\tfrac1p=1$, and $\ell_i'(f(x_i))\in\Real^{C}$ and $\grad f(x_i)\in\Real^{C\times d}$. The following proposition provides a way to represent the solution to problem \eqref{eq:apx_adv_train_rkhs}. The proof is a direct consequence of \cite[Thm. 2]{zhou2008derivative} with $s=1$. 

\begin{proposition}\label{prop:adv_kms}
Consider \cref{eq:adv_train_rkhs}, where $K\in C^2$ and the loss function $\ell\in C^2$. Then, any solution $f$ to \cref{eq:apx_adv_train_rkhs}, satisfies
\begin{align}\label{eq:adv_km}
   f(x) = \sum_{i=1}^n \alpha_i K(x,x_i) + \beta_i\tran G_2(x,x_i)
\end{align}
for some coefficients $\alpha_i\in\Real$ and $\beta_i\in\Real^{d}$. Furthermore, for such an $f$, we have that $\norm{f}_{\Hilbert_K}^2$ equals
$$
\sum_{i,j}\alpha_i\alpha_j K(x_i,x_j) + 
2\alpha_i \beta_j\tran G_2(x_i,x_j)\\ +
\beta_i\tran G_{12}(x_i,x_j)\beta_j.$$
\end{proposition}
The assumption $K\in C^2$ implies that $f\in\mc H$ is twice differentiable. Along with $\ell'$ bounded, this allows the 2nd-order remainder term in the Taylor series expansion to be bounded and can vanish for small $\eps$.

A key point to note however, is that solving \eqref{eq:apx_adv_train_rkhs} with parameterization in \cref{eq:adv_km} may lead to a non-convex optimization problem, since the second term in the objective function is non-convex in $f$. With an additional assumption: $\norm{ \ell_i'(f(x_i))}\leq L$, which holds for Lipschitz losses, one may obtain the convex approximation below which can be solved efficiently. 
\begin{align}
\label{eq:cvx_adv_train_rkhs}
\min_{f\in\Hilbert_K}\; &
\sum_{i=1}^n
    \ell_i(f(x_i))
    + L\eps\norm{\grad f(x_i)}_{q}
+\frac{\lambda}{2}\norm{f}_{\Hilbert_K}^2
\end{align}
We take a closer look at the case of $p=\infty$ below.

\paragraph{$\ell_\infty$ perturbations:} Consider a bounded $\ell_\infty$ adversary, and the logistic loss $\ell(\hat y,y)=\log(1+e^{-y \hat y})$ for $y\in\set{-1,1}$, for which $|\ell_i'(f(x_i))|\leq 1$,
\begin{align}
  \min_{f\in\Hilbert}
  \sum_{i=1}^n
    \ell_i(f(x_i))
  + \varepsilon \|\nabla f(x_i)\|_1
  + \frac{\lambda}{2}\|f\|_{\Hilbert}^2.
\end{align}
The dual formulation of this is derived as follows.
Define $Ef=(f(x_i))_{i=1}^n$ and
$Df=(\nabla f(x_i))_{i=1}^n$.
Using $\mathcal C=\{\delta\in\mathbb R^{n\times d}:|\delta_{ij}|\leq\eps\},$ and
\(
  \eps\|Df\|_1
  = \max_{\delta\in\mathcal C}\langle Df,\delta\rangle,\)
we obtain the saddle point problem
\[
  \min_{f\in\Hilbert}\max_{\delta\in\mathcal C}
  \mathcal L(f,\delta)
  =
  \sum_i\ell_i(f(x_i))
  +\tfrac{\lambda}{2}\|f\|_{\Hilbert}^2
  +\langle Df,\delta\rangle.
\]
Starting from zero, all primal iterates have the form
\(
  f_t(x)
  =\sum_i\alpha_{t,i}K(x,x_i)
   +\beta_{t,i}\,
    G_2(x,x_i).
\)
Consequently, we update coefficients as follows. Sample a uniform subset $B_t\subseteq\{1,\ldots,n\}$ of size $b$.
Let $P_{B_t}$ retain the sampled rows and set the others to zero.
\begin{subequations}
\label{eq:adv_klr}
\begin{align}
  \alpha_{t+1}
    &=\alpha_t
      -\eta_t\tfrac nb P_{B_t}\bigl(s(Ef_t)-y\bigr),\\
  \beta_{t+1}
    &=\beta_t
      -\eta_t\tfrac nb P_{B_t}\delta_t,\\
\delta_{t+1}[B_t]
  &=
    \underset{[-\eps,\eps]}{\text{clip}}
      \left(\delta_{t}[B_t]+\eta_t\tfrac nb\nabla f_t(X[B_t])\right).
\end{align}
\end{subequations}
$s(t)=(1+e^{-t})^{-1}$. \Cref{apd: adv_train_kms} provides details on how this in fact tracks a SGD in the RKHS. We have a matrix-free implementation of the above updates which evaluates $f_t(x_i)$ and
$\nabla f_t(x_i)$ only on the minibatch $x_i\in B_t$ which can be implemented in $O(n(d+1))$ memory.

We use a diminishing step size
$\eta_t=\eta_0(t+1)^{-p}$ and optionally return the weighted average
\(
  \bar f_T
  =\frac{\sum_{t=0}^{T-1}\eta_t f_t}
         {\sum_{t=0}^{T-1}\eta_t}.
\)
Convergence is only guaranteed for $\lambda>0$. For $\lambda=0$ along with early-stopping may be implemented, but needs a separate analysis. 
We also note that $\eps=0$ recovers the kernel logistic regression which has the form of a usual kernel machine since $\beta=0$.

\subsection{Nonconvex approximation for multiclass}
The same procedure along with \cite[Thm. 2]{zhou2008derivative} applied to \eqref{eq:apx_adv_train_rkhs} for vector-valued RKHSs, where $K_C(x,z)=K_C(z,x)\tran\in\Real^{C\times C}$, which consists of multioutput functions $f:\Real^d\to\Real^C$ from an RKHS $\Hilbert^C_{K_C}$ would have representation of the form,
   $\wh f(x) = \sum_{i=1}^n K(x,x_i) \alpha_i + G_2(x,x_i)\tran \beta_i\in\Real^C$
where $\alpha_i\in\Real^C$, $G_2(x,z)\in\Real^{d\times C\times C}$ is the Jacobian, and $\beta_i\in\Real^{d\times C}$. Note the resulting function has $nc(d+1)$ unknown coefficients that need to be trained.

However for $K_C(x,z)=K(x,z)I_C$, i.e., isotropic and separable kernels, \cref{eq:apx_adv_train_rkhs} yields simpler solutions with $2nc+nd$ coefficients.
Suppose $\ell$ is a twice differentiable loss function, and $K$ is twice differentiable,
consider the problem of learning $f:\Real^d\to\Real^C$ via \cref{eq:apx_adv_train_rkhs}
where $\grad f:\Real^d\to\Real^{C\times d}$ is the Jacobian of $f$, and $\ell':\Real^C\times[C]\to\Real^C$ is the gradient of the loss function.
Finally, define,
    \(\bm{M} = \begin{bmatrix}
        \bm{K} & \bm{G}\\
        \bm{G}\tran & \bm{H}
    \end{bmatrix}\in\Real^{n\times n\times (d+1)\times (d+1)}\)
where $\bm{K}\in\Real^{n\times n}$ is the kernel matrix, $\bm{G}\in\Real^{n\times n\times d}$ is the gradient tensor $\frac{\partial}{\partial z}K(x,z)\big|_{z=x_i}$, and $\bm{H}\in\Real^{n\times n\times d\times d}$ is the second-derivative tensor $\frac{\partial^2}{\partial x\partial z}K(x,z)\big|_{\substack{x=x_i\\z=x_j}}$.
We also use \texttt{einsum} -- Einstein notation for expressing summation over tensors. We now state our result.

\begin{proposition}\label{prop:adv_multiclass_clf} 
There exists an optimal solution $f$ to problem \eqref{eq:apx_adv_train_rkhs} of the form
\begin{align}\label{eq:adv_km_rank1}
f(x) = \sum_{i=1}^n \mu_i K(x,x_i) + \gamma_i\theta_i\tran G_2(x,x_i) \in\Real^C,
\end{align}
where $\mu_i,\gamma_i\in\Real^C$ and $\theta_i\in\Real^d$.
Furthermore, if $\bm{M}$ is positive definite, then every optimal solution must have the above form. For such a solution, $\norm{f}_\Hilbert^2$ equals
\(
\sum_{i,j} 
\gamma_i\tran \gamma_j\cdot \theta_i\tran G_{12}(x_i,x_j)\theta_j+\mu_i\tran\mu_jK(x_i,x_j)+\mu_i\tran\gamma_j 
\theta_j\tran (G_2(x_i,x_j)+G_1(x_i,x_j)).\)
\end{proposition}
Observe that the solution above only has $2nc+nd$ coefficients, far fewer than the general case $nc(d+1)$.
Since the objective function in problem \eqref{eq:apx_adv_train_rkhs} is not convex in $f$, we consider a convex upper bound under the assumption of Lipschitz losses whereby $\norm{\ell'}\leq L$.

\paragraph{$\ell_\infty$ adversary:} For the crossentropy loss we have $\ell'(\hat y,y)=\texttt{softmax}(\hat y) - \texttt{onehot}(y)$, which satisfies $\norm{\ell'(\hat y,y)}_1\leq 2$, whereby we can approximate
\(
    \norm{\ell_i'(f(x_i))\tran \grad f(x_i)}_1 \leq 2\norm{\grad f(x_i)\tran}_{1}
\)
where $\norm{A}_{1}$ is the induced $1\mapsto1$-norm. Thus $\norm{\grad f(x_i)\tran}_1$ is the maximum absolute row sum of $\grad f(x_i)$. We propose solving
\begin{align}\label{eq:cvx_adv_train_rkhs_multiclass}
&\min_{f\in\Hilbert}
  \sum_{i=1}^n
    \ell_i(f(x_i))
  + 2\varepsilon \|\nabla f(x_i)\tran\|_1
  + \tfrac{\lambda}{2}\|f\|_{\Hilbert}^2,\\
  &\text{subject to constraint \cref{eq:adv_km_rank1}}.\nonumber
\end{align}
Observe that although the above problem has a convex objective function, the constraint set is non-convex. Our iterative algorithm, derived below, to solve the above problem is based on delayed projections similar to \cite{abedsoltan2024fast}. We propose \Cref{algo:mark} to solve this problem. It involves a projection step every few epochs, which requires solving a non-convex problem solved approximately using \Cref{algo:rank1_projection}. More details on this methodology are provided in \Cref{apd: adv_train_kms}

\subsection{Evaluation of adversarially trained KMs}

We use the stochastic primal--dual procedure described above to train the adversarially regularized kernel classifier, which we denote by \texttt{MARK}. Starting from zero primal and dual variables, at each iteration we sample a minibatch of training examples and update the predictor coefficients according to Eqs.~\eqref{eq:adv_klr}. The dual variables are updated using the input gradients of the current classifier and are subsequently projected element-wise onto the admissible $\ell_\infty$ perturbation set $[-\varepsilon,\varepsilon]$. We use the matrix-free implementation described above, which requires evaluating the predictor and its input gradients only on the current minibatch. The results in this section apply only to twice differentiable losses and kernels like Gaussian KLR. \cref{fig:fmnist_pointwise,fig:kmnist_pointwise,fig:qmnist_pointwise,fig:fmnist_mc,fig:kmnist_mc,fig:qmnist_mc} show that adversarial training of KM boosts their robustness as expected. We also have a direct comparison between KLR and the adversarially trained KLR in \cref{fig:adv_klr_minus_klr,fig:adv_klr_fcnn,fig:adv_klr_adv_fcnn}.

\section{Discussion}\label{sec: discussion}

Through our experiments we   established that KMs are more robust than both FCNN and CNNs over most vision datasets we consider.

\subsection{Summary of Ablations}
We also carried out a limited number of experiments on the SVHN and CIFAR-10 datasets (\cref{apd:svhn_linf_l2}). We find that KMs are consistently more robust than FCNN and CNN over CIFAR-10. We show that in 4 out of 5 randomly chosen $1\text{v}1$ classifications from CIFAR-10, KMs out-perform FCNNs by upto $4\%$ (\cref{fig:cifar_lap_linf,fig:cifar_lap_l2,fig:cifar_gaus_linf,fig:cifar_gaus_l2}). LeNets however, show comparable performance on CIFAR-10 for $\ell_2$ and improved robustness on $\ell_\infty$ attacks. We also find that the Gaussian RFM outperforms all the models on SVHN (\cref{fig:svhn_linf_l2}), as it improves accuracy of the already robust base kernel by $20$ percentage points (\cref{tab:svhn_gaussian_pgd}).

To measure robustness accurately, we check for gradient masking, as sometimes robustness measured under white-box attacks can be overestimated when a model produces uninformative or misleading gradients, causing gradient-based attacks to fail. To validate our PGD results, we perform black-box SPSA attacks (\cref{apd: bb_results}), finding no evidence for gradient masking overall, except in the case of Laplacian RFCs, where \textit{SPSA yields lower robust accuracy than white-box attacks}. We attribute this to the heavy-tailed spectral distribution of the Laplace kernel. Laplace RFCs sample frequencies from a heavy-tailed (Cauchy) distribution. Gaussian RFCs also sample some large frequencies, but PGD can recover from uninformative gradients over multiple steps. For Laplace RFCs this recovery is much weaker, and we expect the same for RFCs drawn from other heavy-tailed distributions. The gradient estimation done by SPSA avoids this issue. We also investigate the effect of variation in kernel parameters like the $\nu$ in Mat\'ern Kernels~(\cref{apd: mat_results}), length scale ($\sigma$)~(\cref{apd:  sigma_robust_results}), ridge regularisation ($\lambda$)~(\cref{apd: ridge_robust_results}). For RFMs we additionally study the effect of feature learning by varying number of iterations~(\cref{apd: rfm_iter_results}). We also show variation in robustness with number of features in RFC models~(\cref{apd: num_of_ftr_vs_robust_results}).

For changes in kernel parameters we see that increase in smoothness ($\nu$) reduces robustness~(\cref{fig:matern}) while a lower length scale increases robustness~(\cref{fig:length_scale_vs_robustness,fig:length_scale_vs_robustness_gaus}). The ridge parameter ($\lambda$) is not particularly consequential for robustness except at much larger values, where it aids robustness at the cost of benign accuracy~(\cref{fig:ridge_vs_robust}), though the exact values are dataset dependent. In the case of RFMs more feature learning at a higher number of iterations adversely affects robustness as demonstrated for the Laplacian RFM~(\cref{fig:rfm_v_iter}).

\subsection{Are KMs more robust point-wise?} \label{subsec:pointwise_robustness}

Ideally, we would like a certificate that some types of KMs are provably more robust than NNs. However, this certification is known to be hard for NNs~\citep{katz2017reluplex,li2023sok}, and made challenging for KMs by the non-convex nature of the boundary they induce. We take a direct, but approximate, approach. Let us define the minimal perturbation needed for a successful attack,
\begin{align*}
    \wh\eps(f,(x,y)) = \min_{\delta\in\Real^d} \set{\norm{\delta}_p~:~ \arg\max_{i\in C}f_i(x+\delta)\neq y}
\end{align*}
The quantity above is calculated approximately by using a bisection method on top of \cref{eq:adv_opt}.
For a fixed $f$ and $g$, and $(X,Y)\sim\Prob$, the ratio $\frac{\wh\eps(f,(X,Y))}{\wh\eps(g,(X,Y))}$ is a random variable. We plot the CDF of this random variable in~\cref{fig:fmnist_pointwise,fig:kmnist_pointwise,fig:qmnist_pointwise} for different pairs $(f,g)$, where $f$ is the KLR for 3 datasets. We see that adversarially trained KMs are most pointwise robust of all the models we consider.

 \subsection{Limitations}
 We identify 5 key limitations of our exploration in this paper. First, since the main attacks considered have been tuned to effectively attack NNs, they may over-estimate the robustness of KMs. However we believe this is unlikely, as KMs should have smooth loss surfaces for the adversary, and first-order attacks would work well. Second, we only consider $\ell_p$ adversaries for vision data, and it would be instructive to consider others like StAdv~\citep{xiao2018spatially} and ReColor~\citep{laidlaw2019functional}. Third, we only consider less complex datasets where KMs are competitive with NNs. This is, however, a limitation that is out of the scope of this paper, and we believe there the tasks we consider are those where lightweight KMs are a good alternative. Additionally, we do not provide any theoretical guarantees that KMs are more robust than NNs. Finally, we do not fully characterise the phenomena of gradient masking in Laplacian RFCs, and the impact of training loss on robustness with respect to the number of features in RFCs.

\subsection{Future Work}
In addition to directly addressing the limitations discussed above, we believe there are several fruitful directions for future work. From a theoretical perspective, obtaining a version of the Representer Theorem for exact robust training and finding KM-specific bounds on robustness are particularly important. 

On tabular data, it would be interesting to compare KMs against tree-based methods such as gradient-boosted trees, which outperform neural networks~\citep{grinsztajn2022tree}. Finally, analysing the adversary's loss landscape, and the confidence of KMs on adversarial examples, and out-of-distribution examples in general, can provide further evidence of their enhanced robustness.

\subsection*{Acknowledgements} All authors acknowledge funding from the SBI Foundation Hub at IIT Bombay. PP acknowledges funding for compute from DST INSPIRE faculty fellowship, Schmidt Sciences, Amazon, and the Nilekani Research Fund. AB acknowledges funding from the ANRF ARG MATRICS program.

\subsection*{AI Use Statement}
In this work, we used generative AI tools for coding an implementation of \Cref{algo:mark} and \Cref{alg:rkhs-projection}. A very preliminary draft of \cref{sec:mark} was generated using a generative AI tool, which was then completely rewritten. Apart from this,
we have not used generative AI tools for writing, and the rest of the required disclosure tasks are not applicable to this work. We did not use generative AI tools for any tasks with recommended disclosure.
We have reviewed all AI-assisted work.
LLM-generated code was verified and tested for correctness by two authors. We take responsibility for the final content of this work, including text, claims, or artifacts produced with the aid of generative AI.

\bibliographystyle{abbrv}
\bibliography{aux/ref}

\begin{thebibliography}{10}

\bibitem{abedsoltan2024fast}
A.~Abedsoltan, S.~Ma, P.~Pandit, and M.~Belkin.
\newblock Fast training of large kernel models with delayed projections.
\newblock {\em arXiv preprint arXiv:2411.16658}, 2024.

\bibitem{athalye2018obfuscated}
A.~Athalye, N.~Carlini, and D.~Wagner.
\newblock Obfuscated gradients give a false sense of security: Circumventing
  defenses to adversarial examples.
\newblock In {\em International conference on machine learning}, pages
  274--283. PMLR, 2018.

\bibitem{beaglehole2025xrfm}
D.~Beaglehole, D.~Holzm{\"u}ller, A.~Radhakrishnan, and M.~Belkin.
\newblock xrfm: Accurate, scalable, and interpretable feature learning models
  for tabular data.
\newblock {\em arXiv preprint arXiv:2508.10053}, 2025.

\bibitem{adult_2}
B.~Becker and R.~Kohavi.
\newblock {Adult}.
\newblock UCI Machine Learning Repository, 1996.
\newblock {DOI}: https://doi.org/10.24432/C5XW20.

\bibitem{bentov2024cafa}
M.~Ben-Tov, D.~Deutch, N.~Frost, and M.~Sharif.
\newblock {CaFA}: Cost-aware, feasible attacks with database constraints
  against neural tabular classifiers.
\newblock In {\em 2024 IEEE Symposium on Security and Privacy (SP)}, pages
  1345--1364. IEEE, 2024.

\bibitem{bhagoji2018practical}
A.~N. Bhagoji, W.~He, B.~Li, and D.~Song.
\newblock Practical black-box attacks on deep neural networks using efficient
  query mechanisms.
\newblock In {\em Proceedings of the European conference on computer vision
  (ECCV)}, pages 154--169, 2018.

\bibitem{carlini2019evaluating}
N.~Carlini, A.~Athalye, N.~Papernot, W.~Brendel, J.~Rauber, D.~Tsipras,
  I.~Goodfellow, A.~Madry, and A.~Kurakin.
\newblock On evaluating adversarial robustness.
\newblock {\em arXiv preprint arXiv:1902.06705}, 2019.

\bibitem{carlini2017towards}
N.~Carlini and D.~Wagner.
\newblock Towards evaluating the robustness of neural networks.
\newblock In {\em IEEE Symposium on Security and Privacy}, pages 39--57, 2017.

\bibitem{clanuwat2018deep}
T.~Clanuwat, M.~Bober-Irizar, A.~Kitamoto, A.~Lamb, K.~Yamamoto, and D.~Ha.
\newblock Deep learning for classical japanese literature, 2018.

\bibitem{croce2020robustbench}
F.~Croce, M.~Andriushchenko, V.~Sehwag, E.~Debenedetti, N.~Flammarion,
  M.~Chiang, P.~Mittal, and M.~Hein.
\newblock Robustbench: a standardized adversarial robustness benchmark.
\newblock {\em Advances in Neural Information Processing Systems}, 2020.

\bibitem{croce2020reliable}
F.~Croce and M.~Hein.
\newblock Reliable evaluation of adversarial robustness with an ensemble of
  diverse parameter-free attacks.
\newblock In {\em ICML}, 2020.

\bibitem{croce2021mind}
F.~Croce and M.~Hein.
\newblock Mind the box: $l_1$-apgd for sparse adversarial attacks on image
  classifiers.
\newblock In {\em ICML}, 2021.

\bibitem{feng2015learning}
Y.~Feng, X.~Huang, L.~Shi, Y.~Yang, and J.~A. Suykens.
\newblock Learning with the maximum correntropy criterion induced losses for
  regression.
\newblock {\em Journal of Machine Learning Research}, 16(30):993--1034, 2015.

\bibitem{goodfellow2014explaining}
I.~J. Goodfellow, J.~Shlens, and C.~Szegedy.
\newblock Explaining and harnessing adversarial examples.
\newblock {\em International Conference on Learning Representations}, 2015.

\bibitem{grinsztajn2022tree}
L.~Grinsztajn, E.~Oyallon, and G.~Varoquaux.
\newblock Why do tree-based models still outperform deep learning on typical
  tabular data?
\newblock {\em Advances in neural information processing systems}, 35:507--520,
  2022.

\bibitem{han2017adequacy}
Y.~Han and B.~I.~P. Rubinstein.
\newblock Adequacy of the gradient-descent method for classifier evasion
  attacks.
\newblock {\em arXiv preprint arXiv:1704.01704}, 2017.

\bibitem{hassani2024curse}
H.~Hassani and A.~Javanmard.
\newblock The curse of overparametrization in adversarial training: Precise
  analysis of robust generalization for random features regression.
\newblock {\em The Annals of Statistics}, 52(2):441--465, 2024.

\bibitem{hofmann2008kernel}
T.~Hofmann, B.~Sch{\"o}lkopf, and A.~J. Smola.
\newblock Kernel methods in machine learning.
\newblock {\em The Annals of Statistics}, 36(3):1171--1220, 2008.

\bibitem{jacot2018neural}
A.~Jacot, F.~Gabriel, and C.~Hongler.
\newblock Neural tangent kernel: Convergence and generalization in neural
  networks.
\newblock {\em Advances in neural information processing systems}, 31, 2018.

\bibitem{kar2012random}
P.~Kar and H.~Karnick.
\newblock Random feature maps for dot product kernels.
\newblock In {\em Artificial intelligence and statistics}, pages 583--591.
  PMLR, 2012.

\bibitem{katz2017reluplex}
G.~Katz, C.~Barrett, D.~L. Dill, K.~Julian, and M.~J. Kochenderfer.
\newblock Reluplex: An efficient smt solver for verifying deep neural networks.
\newblock In {\em International conference on computer aided verification},
  pages 97--117. Springer, 2017.

\bibitem{krizhevsky2009learning}
A.~Krizhevsky and G.~Hinton.
\newblock Learning multiple layers of features from tiny images.
\newblock Technical report, University of Toronto, 2009.

\bibitem{laidlaw2019functional}
C.~Laidlaw and S.~Feizi.
\newblock Functional adversarial attacks, 2019.

\bibitem{lecun1998gradient}
Y.~LeCun, L.~Bottou, Y.~Bengio, and P.~Haffner.
\newblock Gradient-based learning applied to document recognition.
\newblock {\em Proceedings of the IEEE}, 86(11):2278--2324, 1998.

\bibitem{li2023sok}
L.~Li, T.~Xie, and B.~Li.
\newblock Sok: Certified robustness for deep neural networks.
\newblock In {\em IEEE Symposium on Security and Privacy}, pages 1289--1310,
  2023.

\bibitem{li2020gradient}
M.~Li, M.~Soltanolkotabi, and S.~Oymak.
\newblock Gradient descent with early stopping is provably robust to label
  noise for overparameterized neural networks.
\newblock In {\em International conference on artificial intelligence and
  statistics}, pages 4313--4324. PMLR, 2020.

\bibitem{liu2016delving}
Y.~Liu, X.~Chen, C.~Liu, and D.~Song.
\newblock Delving into transferable adversarial examples and black-box attacks.
\newblock {\em International Conference on Learning Representations}, 2017.

\bibitem{madry2017towards}
A.~Madry, A.~Makelov, L.~Schmidt, D.~Tsipras, and A.~Vladu.
\newblock Towards deep learning models resistant to adversarial attacks.
\newblock {\em International Conference on Learning Representations}, 2018.

\bibitem{phishing_websites_327}
R.~Mohammad and L.~McCluskey.
\newblock {Phishing Websites}.
\newblock UCI Machine Learning Repository, 2012.
\newblock {DOI}: https://doi.org/10.24432/C51W2X.

\bibitem{bank_marketing_222}
S.~Moro, P.~Rita, and P.~Cortez.
\newblock {Bank Marketing}.
\newblock UCI Machine Learning Repository, 2014.
\newblock DOI: \url{https://doi.org/10.24432/C5K306}.

\bibitem{netzer2011reading}
Y.~Netzer, T.~Wang, A.~Coates, A.~Bissacco, B.~Wu, and A.~Y. Ng.
\newblock Reading digits in natural images with unsupervised feature learning.
\newblock In {\em NIPS Workshop on Deep Learning and Unsupervised Feature
  Learning 2011}, 2011.

\bibitem{radhakrishnan2024mechanism}
A.~Radhakrishnan, D.~Beaglehole, P.~Pandit, and M.~Belkin.
\newblock Mechanism for feature learning in neural networks and
  backpropagation-free machine learning models.
\newblock {\em Science}, 383(6690):1461--1467, 2024.

\bibitem{rahimi2007random}
A.~Rahimi and B.~Recht.
\newblock Random features for large-scale kernel machines.
\newblock {\em Advances in neural information processing systems}, 20, 2007.

\bibitem{scholkopf2001generalized}
B.~Sch{\"o}lkopf, R.~Herbrich, and A.~J. Smola.
\newblock A generalized representer theorem.
\newblock In {\em International conference on computational learning theory},
  pages 416--426. Springer, 2001.

\bibitem{scholkopf2002learning}
B.~Sch{\"o}lkopf and A.~J. Smola.
\newblock {\em Learning with kernels: support vector machines, regularization,
  optimization, and beyond}.
\newblock MIT press, 2002.

\bibitem{spall1992spsa}
J.~C. Spall.
\newblock Multivariate stochastic approximation using a simultaneous
  perturbation gradient approximation.
\newblock {\em IEEE Transactions on Automatic Control}, 37(3):332--341, 1992.

\bibitem{steinwart2008support}
I.~Steinwart and A.~Christmann.
\newblock {\em Support Vector Machines}.
\newblock Springer Publishing Company, Incorporated, 1st edition, 2008.

\bibitem{uesato2018adversarial}
J.~Uesato, B.~O'Donoghue, P.~Kohli, and A.~van~den Oord.
\newblock Adversarial risk and the dangers of evaluating against weak attacks.
\newblock In {\em Proceedings of the 35th International Conference on Machine
  Learning}, volume~80 of {\em Proceedings of Machine Learning Research}, pages
  5025--5034. PMLR, 2018.

\bibitem{wang2014least}
K.~Wang, J.~Zhang, Y.~Chen, and P.~Zhong.
\newblock Least absolute deviation support vector regression.
\newblock {\em Mathematical Problems in Engineering}, 2014(1):169575, 2014.

\bibitem{wang2021adversarial}
T.~T. Wang.
\newblock Adversarial examples in simpler settings.
\newblock M.eng. thesis, Massachusetts Institute of Technology, 2021.

\bibitem{wang2022huber}
Y.~Wang, B.~Wang, C.~Peng, X.~Li, and H.~Yin.
\newblock Huber regression analysis with a semi-supervised method.
\newblock {\em Mathematics}, 10(20):3734, 2022.

\bibitem{wang2019improving}
Y.~Wang, D.~Zou, J.~Yi, J.~Bailey, X.~Ma, and Q.~Gu.
\newblock Improving adversarial robustness requires revisiting misclassified
  examples.
\newblock In {\em International conference on learning representations}, 2019.

\bibitem{wen2025robustness}
H.~Wen, A.~Betken, and W.~Koolen.
\newblock On the robustness of kernel ridge regression using the cauchy loss
  function.
\newblock {\em arXiv preprint arXiv:2503.20120}, 2025.

\bibitem{wu2021fast}
H.~Wu, Z.~Hu, and B.~Gu.
\newblock Fast and scalable adversarial training of kernel svm via doubly
  stochastic gradients.
\newblock In {\em Proceedings of the AAAI conference on artificial
  intelligence}, volume~35, pages 10329--10337, 2021.

\bibitem{xiao2018spatially}
C.~Xiao, J.-Y. Zhu, B.~Li, W.~He, M.~Liu, and D.~Song.
\newblock Spatially transformed adversarial examples.
\newblock {\em arXiv preprint arXiv:1801.02612}, 2018.

\bibitem{xiao2017fashion}
H.~Xiao, K.~Rasul, and R.~Vollgraf.
\newblock Fashion-mnist: a novel image dataset for benchmarking machine
  learning algorithms.
\newblock {\em arXiv preprint arXiv:1708.07747}, 2017.

\bibitem{xu2009robustness}
H.~Xu, C.~Caramanis, and S.~Mannor.
\newblock Robustness and regularization of support vector machines.
\newblock {\em Journal of machine learning research}, 10(7), 2009.

\bibitem{qmnist-2019}
C.~Yadav and L.~Bottou.
\newblock Cold case: The lost mnist digits.
\newblock In {\em Advances in Neural Information Processing Systems 32}. Curran
  Associates, Inc., 2019.

\bibitem{yu2016orthogonal}
F.~X.~X. Yu, A.~T. Suresh, K.~M. Choromanski, D.~N. Holtmann-Rice, and
  S.~Kumar.
\newblock Orthogonal random features.
\newblock {\em Advances in neural information processing systems}, 29, 2016.

\bibitem{zagoruyko2016wide}
S.~Zagoruyko and N.~Komodakis.
\newblock Wide residual networks.
\newblock In {\em Proceedings of the British Machine Vision Conference (BMVC)},
  pages 87.1--87.12. BMVA Press, 2016.

\bibitem{zhang2019theoretically}
H.~Zhang, Y.~Yu, J.~Jiao, E.~Xing, L.~El~Ghaoui, and M.~Jordan.
\newblock Theoretically principled trade-off between robustness and accuracy.
\newblock In {\em International conference on machine learning}, pages
  7472--7482. PMLR, 2019.

\bibitem{zhou2008derivative}
D.-X. Zhou.
\newblock Derivative reproducing properties for kernel methods in learning
  theory.
\newblock {\em Journal of computational and Applied Mathematics},
  220(1-2):456--463, 2008.

\end{thebibliography}

\clearpage
\appendix 
{\Huge \textbf{Appendix}}

\setcounter{figure}{0}
\renewcommand{\thefigure}{\arabic{figure}A}
\renewcommand{\theHfigure}{A\arabic{figure}} 

\setcounter{table}{0}
\renewcommand{\thetable}{\arabic{table}A}
\renewcommand{\theHtable}{A\arabic{table}}

\begin{table}[h]
\centering
\caption{Abbreviations (sorted in alphabetical order)}
\label{tab:abbr}
\begin{tabular}{lll}
\toprule
advKLR & adversarially trained KLR \\
CAFA & Cost-aware Feasible Attack & \cite{bentov2024cafa}\\
CNN & convolutional neural network\\
FCN & fully connected neural network (a.k.a. MLP)\\
KM & kernel machine\\
KRR & kernel ridge regression\\
KSVM & kernel support vector machines \\
KLR & kernel logistic regression\\
MLP & multi-layer perceptron (a.k.a. FCN)\\
NN & neural network\\
PGD & projected gradient descent\\
RFM & recursive feature machine\\
RFC & random feature classifier\\
SPSA & Simultaneous Perturbation
Stochastic Approximation & \cite{spall1992spsa}\\
WRN & Wide Residual Network \\
\bottomrule
\end{tabular}
\end{table}

\begin{table*}[b]
\centering
\caption{Overview of figures showing results regarding
$\ell_2$ and $\ell_\infty$ attacks on Laplacian and Gaussian
kernel machines compared with neural networks and different variants of kernel machines for FashionMNIST,
KMNIST, and QMNIST. Each link is clickable and takes you
to the relevant figure.}
\label{tab:kernel_appendix_overview}

\setlength{\tabcolsep}{3.2pt}
\footnotesize

\resizebox{\textwidth}{!}{%
\begin{tabular}{lllllllllllllll}
\toprule

& \multicolumn{7}{c}{Laplacian Kernel}
& \multicolumn{7}{c}{Gaussian Kernel}\\

\cmidrule(lr){2-8}
\cmidrule(lr){9-15}

& KLR & KRR & RFM & KSVM & RFC & FCNN & LeNet
& KLR & KRR & RFM & KSVM & RFC & FCNN & LeNet\\

\midrule

KLR
& $\ddots$
& \ref{fig:gap_heatmaps_lapklrvskrr}
& \ref{fig:gap_heatmaps_lapklrvsrfm}
& \ref{fig:gap_heatmaps_lapklrvssvm}
& \ref{fig:gap_lapheatmaps_klrvsrff5k}, \ref{fig:gap_lapheatmaps_klrvsrff20k}
& \ref{fig:gap_heatmaps_lapklr_vs_fcnn}
& \ref{fig:gap_heatmaps_lapklr_vs_lenet}
& $\ddots$
& \ref{fig:gap_heatmaps_gauss_klrvskrr}
& \ref{fig:gap_heatmaps_gauss_klrvsrfm}
& \ref{fig:gap_heatmaps_gauss_klrvssvm}
& \ref{fig:gap_heatmaps_gauss_rff5kvsklr},\ref{fig:gap_heatmaps_gauss_rff20kvsklr}
& \ref{fig:gap_heatmaps_gaussklr_vs_fcnn}
& \ref{fig:gap_heatmaps_gaussklr_vs_lenet}
\\

KRR
&
& $\ddots$
& \ref{fig:gap_heatmaps_lapkervsrfm}
& \ref{fig:gap_heatmaps_lapkrr_vs_svm}
& \ref{fig:gap_heatmaps_rff5kvsker},
  \ref{fig:gap_heatmaps_rff20kvsker}
& \ref{fig:gap_heatmaps_lap_vs_fcnn}
& \ref{fig:gap_heatmaps_lap_vs_lenet}
&
& $\ddots$
& \ref{fig:gap_heatmaps_gauss_rfmvsker}
& \ref{fig:gap_heatmaps_gauss_kervssvm}
& \ref{fig:gap_heatmaps_gauss_rff5kvsker},
  \ref{fig:gap_heatmaps_gauss_rff20kvsker}
& \ref{fig:gap_heatmaps_gauss_vs_fcnn}
& \ref{fig:gap_heatmaps_gauss_vs_lenet}
\\

RFM
&
&
& $\ddots$
& \ref{fig:gap_heatmaps_laprfm_vs_svm}
& \ref{fig:gap_heatmaps_rff_5k_vsrfm},
  \ref{fig:gap_heatmaps_rff_20k_vsrfm}
& \ref{fig:gap_heatmaps_laprfmvsnn}
& \ref{fig:gap_heatmaps_laprfmvslenet}
&
&
& $\ddots$
& \ref{fig:gap_heatmaps_gauss_rfmvssvm}
& \ref{fig:gap_heatmaps_gauss_rff5kvsrfm},
  \ref{fig:gap_heatmaps_gauss_rff20kvsrfm}
& \ref{fig:gap_heatmaps_gauss_rfmvsnn}
& \ref{fig:gap_heatmaps_leNetvsgaussrfm_ker}
\\

KSVM
&
&
&
& $\ddots$
& \ref{fig:gap_heatmaps_rff5k_vssvm},
  \ref{fig:gap_heatmaps_rff20k_vssvm}
& \ref{fig:gap_heatmaps_lapsvmvsfcnn}
& \ref{fig:gap_heatmaps_lapsvmvslenet}
&
&
&
& $\ddots$
& \ref{fig:gap_heatmaps_gauss_rff5kvssvm},
  \ref{fig:gap_heatmaps_gauss_rff20kvssvm}
& \ref{fig:gap_heatmaps_gauss_svmvsfcnn}
& \ref{fig:gap_heatmaps_leNetvsgausssvm}
\\

RFC
&
&
&
&
& $\ddots$
& \ref{fig:gap_heatmaps_rff5kvsfcnn},
  \ref{fig:gap_heatmaps_rff20kvsfcnn}
& \ref{fig:gap_heatmaps_laprff_5k_vslenet},
  \ref{fig:gap_heatmaps_laprff_20k_vslenet}
&
&
&
&
& $\ddots$
& \ref{fig:gap_heatmaps_gauss_rff5kvsnn},
  \ref{fig:gap_heatmaps_gauss_rff20kvsnn}
& \ref{fig:gap_heatmaps_leNetvsgaussrfc5k},
  \ref{fig:gap_heatmaps_leNetvsgaussrfc20k}
\\

\bottomrule
\end{tabular}%
}
\end{table*}

\section{Further Technical Details and Setup}\label{appsec: background}

\subsection{Gradient-free (black-box) attacks using SPSA}\label{apd: spsa_attack}
Let $\mathsf{L}(x,y)$ denote the loss at input \(x \in \mathbb{R}^d\) with label \(y\). At a given iterate, SPSA samples a random perturbation vector $\Delta \in \{-1,+1\}^d$. Using a two-sided finite-difference approximation, the SPSA gradient estimator for the \(i\)-th coordinate is
\[
\hat{g}_i(x)
=
\frac{\mathsf{L}(x+\alpha \Delta, y)-\mathsf{L}(x-\alpha \Delta, y)}{2\alpha \,\Delta_i},
\qquad i=1,\dots,d,
\]
where \(\alpha>0\) is a small perturbation magnitude. Since \(\Delta_i \in \{-1,+1\}\), this can equivalently be written as
\[
\hat{g}_i(x)
=
\frac{\mathsf{L}(x+\alpha \Delta, y)-\mathsf{L}(x-\alpha \Delta, y)}{2\alpha}\,\Delta_i.
\]
In vector form, the estimator is
\[
\hat{g}(x)
=
\frac{\mathsf{L}(x+\alpha \Delta, y)-\mathsf{L}(x-\alpha \Delta, y)}{2\alpha}\,\Delta.
\]
In practice, we use a vectorised, variance-reduced estimate of the gradient over \(m\) independent perturbation vectors \(\Delta^{(1)},\dots,\Delta^{(m)}\):
\[
\hat{g}(x)
=
\frac{1}{m}\sum_{j=1}^m
\frac{\mathsf{L}(x+\alpha \Delta^{(j)}, y)-\mathsf{L}(x-\alpha \Delta^{(j)}, y)}{2\alpha}\,\Delta^{(j)}.
\]

\subsection{Attacks on Tabular datasets using CaFA}\label{apd: cafa_attack}
The objective in the CaFA attacks is as follows:
\begin{align*}
    \text{Find} \, \, \delta \in \mathbb{R}^d \\
    \text{s.t.} \, \, f(x+\delta) \neq y \\
    x+\delta \vDash T \\
    \text{Cost}(\delta) \leq \eps
\end{align*}
Note here the additional constraint denoted by $x+\delta \vDash T$, which indicates that the generated adversarial example must satisfy feature space constraints. Unlike in the case of images, where simple boundary clipping is sufficient, tabular data is more complex as it has a combination of continuous, integer, and categorical features. In addition, the range of each feature can be different. CaFA proceeds in the following 3 steps:
\begin{enumerate}
    \item \textbf{Mining constraints}: There are two main types of constraints to be considered. The first are structural constraints, which are available via feature engineering. The second type are relation constraints, which in CaFA are set to be denial constraints. These prevent pairs of samples from having a set of predicates that hold simultaneously. These are difficult to specify and are mined from the data.
    \item \textbf{Perturbing samples}: The samples are first perturbed using a method known as TabPGD, which works by first creating temporary adversarial examples. These are then modified feature by feature depending on the type of the feature.
    \item  \textbf{Constraint projection}: The generated adversarial examples are checked for compliance with the set of mined constraints $T$ using a SAT solver. If the constraints are not satisfied, the samples are then projected by relaxing constraints on certain features. The final obtained adversarial example should now satisfy both structural and relational constraints.
\end{enumerate}

In our experiments we sweep over $\eps$ in $\{0.01,0.03,0.075\}$.

\subsection{Kernels preliminaries}\label{apd: matern_kernels_background}
Matérn Kernels are a class of kernels parameterized by a smoothness parameter $\nu > 0$ and
length scale $\sigma > 0$. 
The general Matérn kernel is defined as
\begin{align}
K_{\nu}(x,z)
=
\frac{2^{1-\nu}}{\Gamma(\nu)}
\left(
\tfrac{\sqrt{2\nu}\,\norm{x-z}}{\sigma}
\right)^{\nu}
J_{\nu}
\left(
\tfrac{\sqrt{2\nu}\,\norm{x-z}}{\sigma}
\right),
\end{align}
where $\Gamma(\cdot)$ is the Gamma function and 
$J_{\nu}$ denotes the modified Bessel function 
of the second kind. Here $\sigma$ is the
length scale, while $\nu$ defines the smoothness of functions associated 
with the kernel, with larger $\nu$ corresponding to smoother functions. The 
Laplacian kernel ($\nu=1/2$), denoted $K^{\sf lap}$, and the Gaussian kernel ($\nu \to \infty$), denoted $K^{\sf gauss}$, are also members of the Mat\'ern family. In addition, Mat\'ern Kernels with $\nu=(2k+1)/2$ for $k\in\Natural$ have known closed form expressions. For example, $\nu=3/2$ and $\nu=5/2$ are given below:
\begin{align*}
K^{\sf lap}_M(x,z) &=\exp\round{-\tfrac{\norm{x-z}_M^\gamma}{\sigma}},
K^{\sf g}_M(x,z) = \exp\round{-\tfrac{\norm{x-z}_M^2}{2\sigma^2}}\\
K_{3/2}(x,z) &= \left(1 + \tfrac{\sqrt{3}\norm{x-z}}{\sigma}\right) \exp\left(-\tfrac{\sqrt{3}\norm{x-z}}{\sigma}\right), \\
K_{5/2}(x,z) &=\left(1 + \tfrac{\sqrt{5}\norm{x-z}}{\sigma} + \tfrac{5\norm{x-z}^2}{3\sigma^2}\right) \exp\left(-\tfrac{\sqrt{5}\norm{x-z}}{\sigma}\right).
\end{align*}

\paragraph{Kernel Ridge Regression (KRR), Kernel SVM (KSVM) and Kernel Logistic Regression (KLR):} While the Representer Theorem allows for arbitrary loss functions, in practice most KMs are trained for the square loss $J\round{\set{x_i,f(x_i),y_i}_{i=1}^n}=\sum_{i=1}^n\norm{y_i-f(x_i)}^2$, due to the simplicity in implementation. KRR is the solution to the linear equation
\begin{align}\label{eq:krr}
    Y = (K(X,X)+\lambda I_n)\alpha
\end{align}
where $Y\in\Real^{n\times C}$ is the matrix of targets in the training data. We also trained KSVM to minimize  hinge loss, or soft-margin SVM, $J\round{\set{x_i,f(x_i),y_i}_{i=1}^n}=\sum_{i=1}^n\max{(0,1-y_if(x_i))}$, using one-vs-one for multiclass classification problems.

\subsection{Adversarial training of NNs}\label{apd: adv_training}
To train neural networks adversarially, we use the standard min-max optimisation approach from~\citep{madry2017towards}:
\begin{align*}
f_{\text{robust}}=\mathop{\arg\min}\limits_{f \in \mathcal{F}} \mathbb{E}_{(x,y) \sim \mathbb{P}_{\text{train}}} \left[ \sup_{\delta: \|\delta\|_q \leq \eps }\mathsf{L} \left(x+\delta,y \right) \right].
\end{align*}
In practice, an alternating minimisation approach is used to obtain $f_{\text{robust}}$. First, $\delta$ is set to $0$, and the standard empirical risk is minimised. The function $f$ thus obtained is fixed, and the inner supremum computed with respect to it, typically using Projected Gradient Descent. The empirical risk with respect to the resulting $\delta$ per sample is then computed, and a new $f$ is obtained. This procedure continues until the training loss plateaus or a sufficiently high robust accuracy is obtained.

\subsection{Adversarial training of Kernel Machines}\label{apd: adv_train_kms}

\noindent \textbf{Minibatch stochastic updates for binary classification:}
The derivatives of the loss function are
\[
  \nabla_f\mathcal L
  = E^*\bigl(s(Ef)-y\bigr)+\lambda f+D^*\delta,
  \qquad
  \nabla_\delta\mathcal L=Df.
\]

We use the following procedure for training. At iteration $t$, sample a uniform subset
$B_t\subseteq\{1,\ldots,n\}$ of size $b$.
Let $P_{B_t}$ retain the sampled rows and set the others to zero.
Since $\mathbb E[(n/b)P_{B_t}]=I_n$, unbiased stochastic
primal and dual directions give
\begin{align*}
  f_{t+1}
  &= f_t
     -\eta_t\frac nb
       \left[
         E^*P_{B_t}\bigl(s(Ef_t)-y\bigr)
         +D^*P_{B_t}\delta_t
       \right],\\
  \delta_{t+1}
  &=\Pi_{\mathcal C}
       \left(\delta_t+\eta_t\frac nb P_{B_t}Df_t\right).
\end{align*}
Thus only the sampled rows of $\delta$ change. Clipping acts component-wise. Both updates use the old iterates.

\paragraph{Approximate RKHS projection:}
Let $\mc F$ be the subspace of functions $f$ of the form $\sum_{i=1}^n\alpha_i K(x,x_i) + \beta_i\tran G_2(x,x_i)$ where $\alpha_i\in\Real^C,\beta_i\in\Real^{d\times C}$, and let $\mc F_1$ be the cone of functions $f$ of the form \eqref{eq:adv_km_rank1}. 
\begin{proposition}
Suppose $g\in\mc F$ with coefficients $(\alpha,\beta)\in\Real^{n\times c}\times\Real^{n\times cd}$, then solving
\begin{align*}
    \min_{f\in\mc F_1}\norm{f-g}_\Hilbert
\end{align*} is equivalent to solving
\begin{align}\label{eq:rkhs_projection_rank1}
\min_{\mu,\theta,\gamma} \sum_{i,j=1}^n\tfrac{1}{2}\inner{\begin{pmatrix}
    \mu_i-\alpha_i\\
    \theta_i\gamma_i\tran-\beta_i
\end{pmatrix},\bm{M}_{ij}
\begin{pmatrix}
    \mu_j-\alpha_j\\
    \theta_j\gamma_j\tran-\beta_j
\end{pmatrix}},
\end{align}
where
$(\mu,\theta,\gamma)\in\Real^{n\times c}\times\Real^{n\times d}\times\Real^{n\times c}$.
\end{proposition}
We provide \Cref{algo:rank1_projection} to solve problem \eqref{eq:rkhs_projection_rank1} approximately.

\begin{algorithm}[t]
\caption{\label{algo:mark}\texttt{MARK}: Multiclass Adversarially Robust Kernel machines}
\label{alg:rkhs-training}
\begin{algorithmic}[1]
\Require Data $\{(x_i,y_i)\}_{i=1}^n$, kernel functions $(K,G_1,G_2,G_{12})$, budget $\eps$, Epochs $E$, learning rate $\eta$,
projection interval $T$
\State Initialize $\mu,\gamma=0_{n\times c}$, $\theta=0_{n\times d}$ (if not provided) 
\State Initialize $\alpha_j=\mu_j,\beta_j=\theta_j\gamma_j\tran$
\State $r_{0}\!:=\!{\tt einsum(``bn, nc\!\to\! bc")}$
\State $r_{2}\!:=\!{\tt einsum(``bnd, ndc\!\to\! bc")}$
\State $r_{1}\!:=\!{\tt einsum(``bnd, nc\!\to\! bdc")}$
\State $r_{12}\!:=\!{\tt einsum(``bndD, nDc\!\to\! bdc")}$
\For{$e=1,\ldots,E$}
    \For{minibatch $b \subset [n]$ w/o replacement}
    \State Fetch minibatch inputs $B \gets X[b]$
        \State $\bm{F}\gets r_0(K(B,X),\alpha)+r_2(G_2(B,X),\beta)$
        \State $\bm{R} \gets \texttt{softmax}(\bm{F})-\texttt{one\_hot}(Y[B])$
        \State
        $\bm{J}\gets r_1(G_1(B,X),\alpha)+r_{12}(H(B,X),\beta)$
        \State $k_i \gets \arg\max_{1\le k\le C} \sum_{a=1}^d |\bm{J}_{ika}|$\quad $\forall\,i\in B$
        \State $\bm{Q} \gets 2\eps\cdot\round{\operatorname{sign}(J_{ik_i,:}^{\top}) e_{k_i}^{\top}}_i$
        \Comment{rank-1 dual maximizer}
        \State $\alpha[b] \gets
        \alpha[b]
        -\eta\frac{n}{\abs{b}}
         \bm{R}$
        \State $ \beta[b] \gets
        \beta[b]-\eta\frac{n}{\abs{b}}
         \bm{Q}$
    \EndFor
    \If{$e = 0\bmod T$}
        \State $(\mu,\theta,\gamma)\gets\Call{Project}{\alpha, \beta}$ \Comment{solve \eqref{eq:rkhs_projection_rank1}}
        \State reset $\alpha_j=\mu_j,$ and $\beta_j=\theta_j\gamma_j\tran$ for $j\in[n]$
    \EndIf
\EndFor
\State \Return
$\displaystyle
 f=\sum_{j=1}^n
K(\cdot,x_j)\mu_j+
\theta_j\tran G_2(\cdot,x_j)\gamma_j$
\end{algorithmic}
\end{algorithm}

\begin{algorithm}[t]
\caption{\label{algo:rank1_projection}Approximate rank-1 projection in RKHS}
\label{alg:rkhs-projection}
\begin{algorithmic}[1]
\Require Coefficients $(\alpha,\beta)$, $\bm K,\bm G,\bm H$,
refinement steps $L$
\State $(\theta_i,\gamma_i)\gets\texttt{rank1-SVD}(\beta_i)$
       \quad $\forall i\in[n]$
\For{$l=1,\ldots,L$}
    \State $E_i\gets\theta_i\gamma_i\tran-\beta_i$
           \quad $\forall i\in[n]$
    \State $\bm Z\gets\texttt{solve}(\bm K,r_{2}(G_2(X,X),\bm E))$ \Comment{EigenPro}
    \State $\bm R\gets r_{12}(\bm H,\bm E)-r_{1}(G_1(X,X),\bm Z)$
    \State $P_i\gets R_i\gamma_i,\quad
           Q_i\gets R_i\tran\theta_i$
           \quad $\forall i\in[n]$
    \State $\eta\gets\texttt{backtrack}(\theta,\gamma;\bm P,\bm Q)$
           \Comment{Armijo rule}
    \State $\theta_i\gets\theta_i-\eta P_i,\quad
           \gamma_i\gets\gamma_i-\eta Q_i$
           \quad $\forall i\in[n]$
\EndFor
\State $E_i\gets\theta_i\gamma_i\tran-\beta_i$
       \quad $\forall i\in[n]$
\State $\mu\gets\alpha-\texttt{solve}(\bm K,r_{2}(G_2(X,X),\bm E))$\Comment{EigenPro}
\State \Return $(\mu,\theta,\gamma)$
\end{algorithmic}
\end{algorithm}

\section{Detailed Experimental Setup}\label{appsec:background_experiments}

\subsection{Kernel Machines training details}

\noindent \textbf{Base KM}: For KRR and RFM, we searched for $\lambda \in \{$1, 0.1, 0.01, 0.001, 0.0001$\}$ and $\sigma \in \{$0.05, 0.125, 0.25, 0.5, 1, 2, 4, 6$\}$. For KSVM, the search over $\sigma$ is the same as for KRR, while $\lambda$ was searched over the set $\{$10000, 1000, 100, 10, 1, 0.1, 0.01$\}$. For KLR, we searched for length scale over the set $\{0.5, 1.0, 2.0, 4.0\}$, while the learning rate is searched over $\{0.1, 0.5, 1.0, 2.0, 5.0\}$.

\noindent \textbf{RFCs:} For binary classification problems, we use $5000$ and $20000$ random features. The 1v1 results for RFCs are accuracies averaged over the seeds in $\{$100, 150, 200, 250, 300, 350, 400, 450, 500, 550$\}$. For multiclass classification, we searched over the number of features in $\{$10{\sf k}, 20{\sf k}, 30{\sf k}, 40{\sf k}, 50{\sf k}, 60{\sf k}$\}$, using the same seeds as in the 1v1 problem, to study the dependence between the number of features and robust accuracy in RFCs. We searched the length scale over $\sigma \in \{0.1, 0.3, 0.5, 0.75, 1, 2, 3, 5, 7, 10\}$ and the ridge parameter $\lambda$ such that $-\log_{10}\lambda \in \{0, 1, 2, 3\}$.

\subsection{Neural Networks training details}
 The FCNNs were trained using stochastic gradient descent (SGD) with momentum to minimise the cross-entropy loss. We perform an exhaustive search over the network's widths in $\set{64,128,256}$, depths in $\set{2,3,4,5}$, learning rates in $\set{0.01,0.03, 0.1}$, and weight decay rates in $\set{0.01,0.001,0.0001}$, with momentum fixed at $0.9$. We did the above search for both robust training (using Projected Gradient Descent) and benign training. Convolutional neural networks (CNNs) with the modified LeNet5 architecture (using ReLU activation and max-pooling layers)~\citep{lecun1998gradient}, and WideResNet-28-10~\citep{zagoruyko2016wide} were trained with standard training hyperparameters. We additionally evaluated the sensitivity of the FCNN baselines to random initialization across multiple training seeds. We observed only minor variations in clean and adversarial accuracy, with the overall robustness trends remaining consistent across runs. Therefore, we present the results for FCNNs trained using seed 42.

\subsection{Attacks}
For the $\ell_\infty$ constrained attacks, we use perturbation budgets $\eps = \frac{\eps_0}{255}$ with $\eps_0 \in \{0,\allowbreak .1,\allowbreak 1,\allowbreak 2,\allowbreak 4,\allowbreak 6,\allowbreak 8,\allowbreak 12,\allowbreak 20\}$ for 1v1 comparison, while for multiclass we used these $\ell_\infty$ perturbation budgets, $\eps = \frac{\eps_0}{255}$ with $\eps_0 \in \{0,\allowbreak .1,\allowbreak 1,\allowbreak 2,\allowbreak 4,\allowbreak 6,\allowbreak 8,\allowbreak 12,\allowbreak 20,\allowbreak 32, \allowbreak 40, \allowbreak 55\}$ for multiclass classification. For $\ell_2$ constrained attacks, we use $\eps \in \set{0, .5, 1, 2, 4, 6, 8}$ for 1v1, and $\eps \in \set{0, .5, 1, 2, 4, 6}$. The fixed iteration attack uses $100$ iterations, with step size $\eps/10$. For details on other attacks used, see black-box attacks~(\cref{apd: bb_results}) and attacks on tabular data~(\cref{apd: tabular_data}).

\subsection{Adversarial training of neural networks}

We have trained our models for multiclass with $\ell_\infty$ perturbation for FashionMNIST with $\frac{8}{255}$, for KMNIST it is $\frac{16}{255}$, for QMNIST it is $\frac{24}{255}$. For $\ell_2$ training, we have used perturbation budget $\{\varepsilon=1\}$. For binary classification we have used $\ell_\infty$ perturbation budget $\{\varepsilon=\frac{12}{255}\}$ for FashionMNIST, $\{\varepsilon=\frac{16}{255}\}$ for KMNIST, $\{\varepsilon=\frac{24}{255}\}$ budget for QMNIST for 200 epochs while doing robust training. Depth is searched over $\{2,4\}$ for pairwise classification, width is searched over $\{128, 256\}$, learning rate is in $\{0.03,0.1\}$, and weight decay in $\{0.0001, 0.001\}$, momentum was fixed as $0.9$, dropout was fixed at $0$.
While for multiclass adversarial training, we searched over depth $\in \{3,5\}$, width over $\{128, 256\}$, and searched for learning rate in $\{0.03,0.1\}$, and weight decay in $\{0.0001, 0.001\}$, with momentum kept at 0.9, and with $0$ dropout. For adversarial LeNets we searched over the learning rate in $0.03,0.1$, and weight decay in $\{0.0001, 0.001\}$, with momentum kept at 0.9, and dropout being $0$. LeNets were trained for 50 epochs, while WRNs were trained for 20 epochs.

\begin{table*}[b]
\vspace{-10pt}
    \centering
    \caption{Computational complexity of training on a single NVIDIA RTX PRO 6000 Blackwell Max-Q Workstation GPU.
    ($^\dagger$time in seconds(s) to train a single model on Fashion MNIST. Note that KRR and RFM have fewer hyperparameters to tune, hence the total time is smaller.)}
    \label{tab:computational_complexity}
    \begin{tabular}{llll}
    \toprule 
    model & hyperparameters & time ($1\text{v}1$)$^\dagger$ & time (multi-class)$^\dagger$ \\
    \midrule
    FCNN     &  width, depth, lr, weight decay & $\approx40$s &  $\approx300$s  \\
    FCNN-adv &  $\cdots$ + $\eps$ & $\approx300$s &  $\approx900$s  \\
    LeNet   &  N.A. & $\approx40$s & $\approx240$s\\
    adv-LeNet   &  N.A. & $\approx120$s & $\approx720$s\\
    \midrule
    KRR & $\lambda$, $\sigma$ & $\approx5$s &  $\approx30$s \\
    RFM & $\cdots$ + \#iterations & $\approx25$s & $\approx100$s \\
    RFC & $\cdots$ + \#features & $\approx0.5$s & $\approx1$s \\
   Kernel SVM & $\cdots$  & $\approx2$s & $\approx45$s \\
   KLR & learning rate, $\sigma$  & $\approx5$s &  $\approx30$s \\
       adv-KLR & learning rate, $\sigma$, $\lambda$ & $\approx900$s & $\approx86{,}000$s$$ \\
    \bottomrule
    \end{tabular}
\end{table*}

\subsection{Adversarial training of KLR}
\paragraph{Adversarial training of kernel machines.}
For binary adversarial KLR, we used Gaussian kernels with length scales explored over $\{1,2,4\}$ across class pairs and swept the RKHS regularization coefficient over $\lambda\in\{10^{-4},10^{-3},10^{-2}\}$, selecting $\lambda$ by clean validation accuracy. Training used 500 stochastic primal--dual updates, a minibatch size of 128, an initial learning rate of $0.1$, a learning-rate decay exponent of $0.75$, weighted iterate averaging, and random seed 0. The gradient-penalty coefficient $\rho$ was set to $12/255$, $20/255$, and $24/255$ for Fashion-MNIST, KMNIST, and QMNIST, respectively.

For multiclass adversarial KLR, we compared Gaussian length scales $\{1,2\}$ and selected 2 based on improved clean accuracy and adversarial robustness. Training used 10 epochs, a minibatch size of 128, an initial learning rate of $0.01$, a learning-rate decay exponent of $0.75$, RKHS regularization $\lambda=10^{-4}$, and random seed 42. The radius parameter $\epsilon$ was set to $8/255$, $16/255$, and $24/255$ for Fashion-MNIST, KMNIST, and QMNIST, respectively, with a penalty multiplier of 2. Projection intervals of one and two epochs yielded comparable performance; the reported checkpoints used one epoch for Fashion-MNIST and two epochs for KMNIST and QMNIST. Projection used at most 100 refinement steps and a tolerance of $10^{-4}$. The EigenPro solver used relative and absolute tolerances of $10^{-3}$ and $10^{-8}$, respectively, at most 100 solver epochs, a subsample size of 1024, rank 100, a minibatch size of 128, and matrix-free storage. Query and center tile sizes were 64 and 1024, respectively.

\subsection{\texorpdfstring{$\nu$}{nu} vs. robustness of Mat\'ern Kernels}\label{apd:mat_kernel}
In our experiments, we fine-tuned over ridge $\lambda \in \{10^{-6}, 10^{-5}, 10^{-4}, 10^{-3}, 10^{-2}, 10^{-1}, 10^{0}\}$ and length scale $\sigma \in \{0.05, 0.125, 0.25, 0.5, 1, 2, 4, 6, 8, 10\}$, for each value of $\nu \in \{0.5, 1.5, 2.5, \infty\}$. We chose models corresponding to each $\nu$ such that their $\lambda$ and $\sigma$ are the same with benign accuracy being close to the best benign accuracy found in the grid search, so as to isolate the effect of $\nu$ on robustness. Square loss is used for the training.

\subsection{Computational efficiency details}
\noindent \textbf{Model training runtimes:} Kernel Machines take around $5$ seconds for each dataset to train for one 1v1 problem, while these take around $30$ seconds for multiclass classification problem. RFMs take around $25$ seconds to train for one 1v1 classification problem, while for multiclass classification they take around $2$ minutes of time. Training for RFC is instant no matter the type of problem. Kernel SVM takes around $45$ seconds to train using scikit-learn. Benign training of FCNNs for one 1v1 classification problems without hyperparameter training took approximately 40 seconds for FashionMNIST, 25 seconds for KMNIST, and 50 seconds for QMNIST. While for the multiclass classification over FashionMNIST, KMNIST and QMNIST, FCNN took around 5 minutes to train for 200 epochs. The robust training took around 15 to 30 minutes for all the datasets, no matter the training budget. LeNets benign training for one 1v1 classification problem takes around 40 seconds, while the robust training for 1v1 takes around 2 minutes for 50 epochs. Training of adv-LeNet for multiclass takes around 12 minutes, and for 1v1 it is around 2 minutes. WRN benign training took around 40 minutes to get trained for 20 epochs. Training for robust adversarial KLR for multiclass classification took around 24 hours for 10 epochs, for projection at every step. Robust binary KLR took around 15 minutes for training at 500 iterations.

\noindent \textbf{Attack runtimes:}
For multiclass FCNNs and CNNs the attack takes around $15-30$ minutes, for RFC the attack takes around $5-10$ minutes, for Kernel machines and RFMs the attack takes around $15-20$ minutes. For one 1v1 attack for any model the time is less than $2$ minutes. These timings are irrespective of the norm being used, and these are the time to attack a particular model on all the perturbation budgets.

\clearpage
\section{Detailed Experimental Results}\label{appsec: full_results}

\Cref{tab:kernel_appendix_overview} provides references to all comparisons for the Laplacian and Gaussian KMs.

\vspace{10mm}
\subsection{Adversarial Robustness under \texorpdfstring{$\ell_2$}{ell2} and  \texorpdfstring{$\ell_\infty$}{ell2} Perturbations}
Here we provide result for the multiclass problem on FMNIST,  KMNIST, QMNIST, and 1v1 for the FMNIST dataset. In the $\ell_2$ attack, the error bars for different seeded attacks have been provided, albeit they are too small to be relevant. Similar small error bars for attacks were observed for $\ell_\infty$ attacks, but due to them being quite small only values corresponding to seed 42 have been plotted. 

\vspace{10mm}
\begin{figure*}[!htbp]
\centering
\caption{Comparison of the robust accuracy against $\ell_2$ perturbations ($\uparrow$ is better) for 10 models across 3 datasets. Results were obtained using \texttt{autoattack}, using apgd-ce.}
\includegraphics[
    width=\textwidth,
    height=0.45\textheight,
    keepaspectratio
]{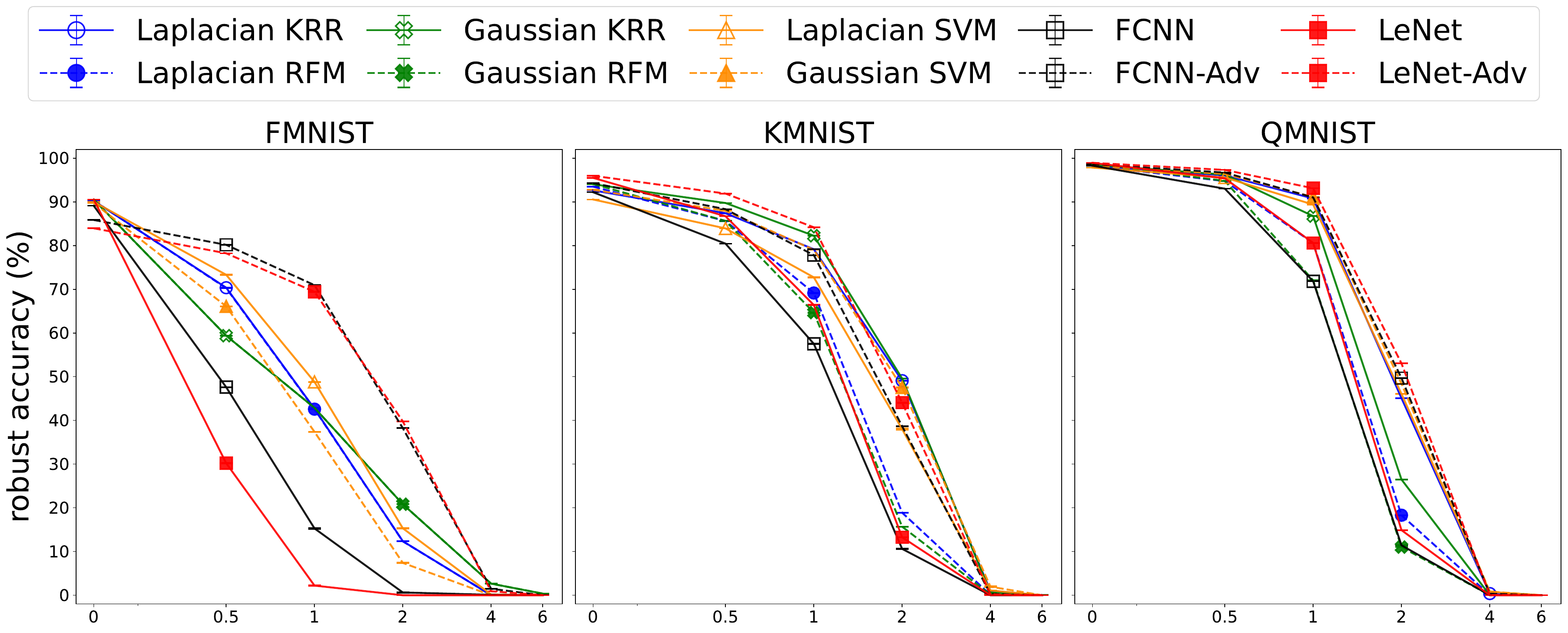}
\label{fig:fmnist_kmnist_qmnist_mc_l2}
\end{figure*}

\begin{figure*}[!htbp]
    \centering
\caption{\label{fig:gap_heatmaps_summary_l2} \textbf{Gap in robust accuracy (ra) for $\ell_2$ perturbation on 45 $1\text{v}1$ binary classification tasks from FashionMNIST} comparing 2 neural networks (columns) and 3 kernel machines (rows) with the Laplacian kernel. 
    Perturbation size increases as we go down each heatmap, and values correspond to \textrm{ra}(row model) - \textrm{ra}(col model).
    The \textcolor{teal}{teal} colour indicates that the kernel machine is more robust, whereas the \textcolor{Bittersweet}{orange} color indicates that the neural network is more robust. Each of the 6 heatmaps compares 45 pairs of models, each finetuned for best accuracy on clean data.
    For each heatmap, class pairs for binary classification are sorted according to values in the bottom row. PGD attack is used to attack the models.
    }
\includegraphics[width=\textwidth,keepaspectratio]{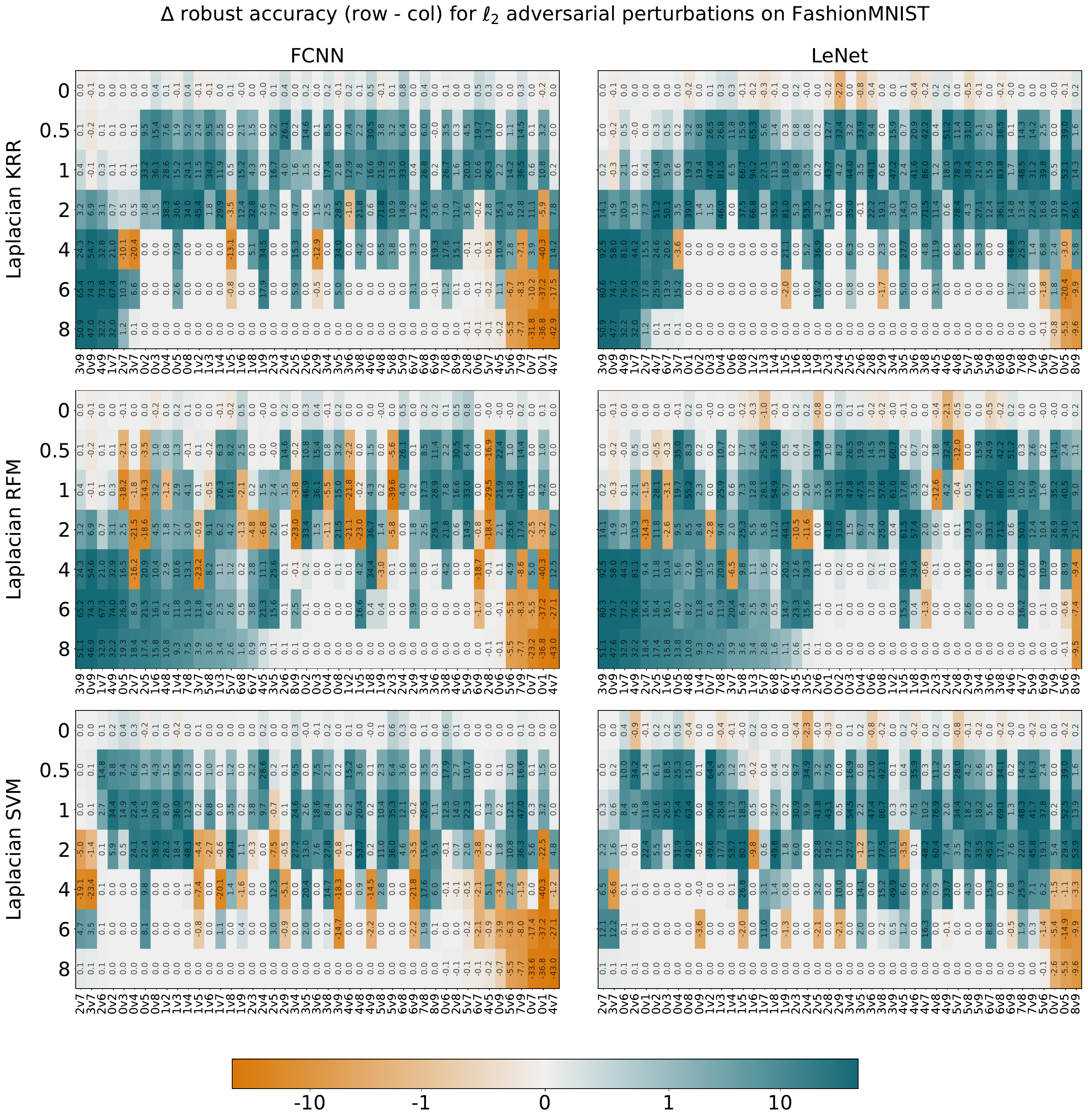}
\end{figure*}

\begin{figure*}[!htbp]
\caption{\label{fig:gap_rfc_summary_fmnist_l2} 
\textbf{Gap in robust accuracy (ra) for $\ell_2$ perturbation} on 45 $1\text{v}1$ binary classification tasks from FashionMNIST comparing 2 neural networks (columns) and 2 RFCs with different number of features (rows) with the Laplacian kernel. 
Perturbation size increases as we go down each heatmap, and values correspond to \textrm{ra}(row model) - \textrm{ra}(col model).
The \textcolor{teal}{teal} colour indicates that the RFC is more robust, whereas the \textcolor{Bittersweet}{orange} color indicates that the neural network is more robust. Each of the 4 heatmaps compares 45 pairs of models.
For each heatmap, class pairs for binary classification are sorted according to values in the bottom row.     
(Top) RFC with 5000 random features 
(Bottom) RFC with 20000 random features. PGD attack is used to attack the models.
}
\includegraphics[width=\textwidth,keepaspectratio]
{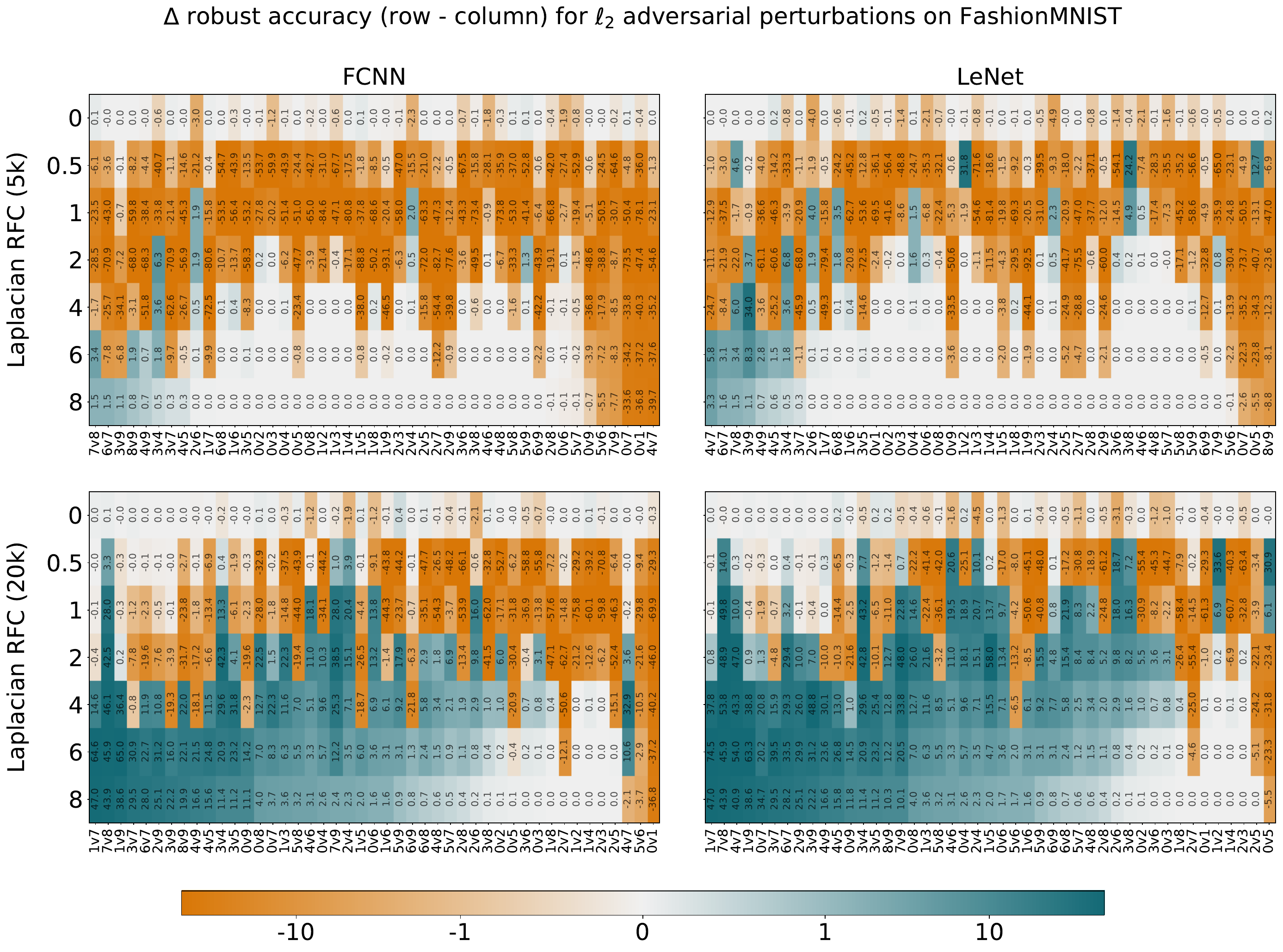}
\vspace{-6ex}
\end{figure*}

\subsection{Linf summary for KMNIST/QMNIST vs FCNN and LeNet}
\begin{figure*}[!htbp]
    \centering
    \includegraphics[width=0.75\textwidth]{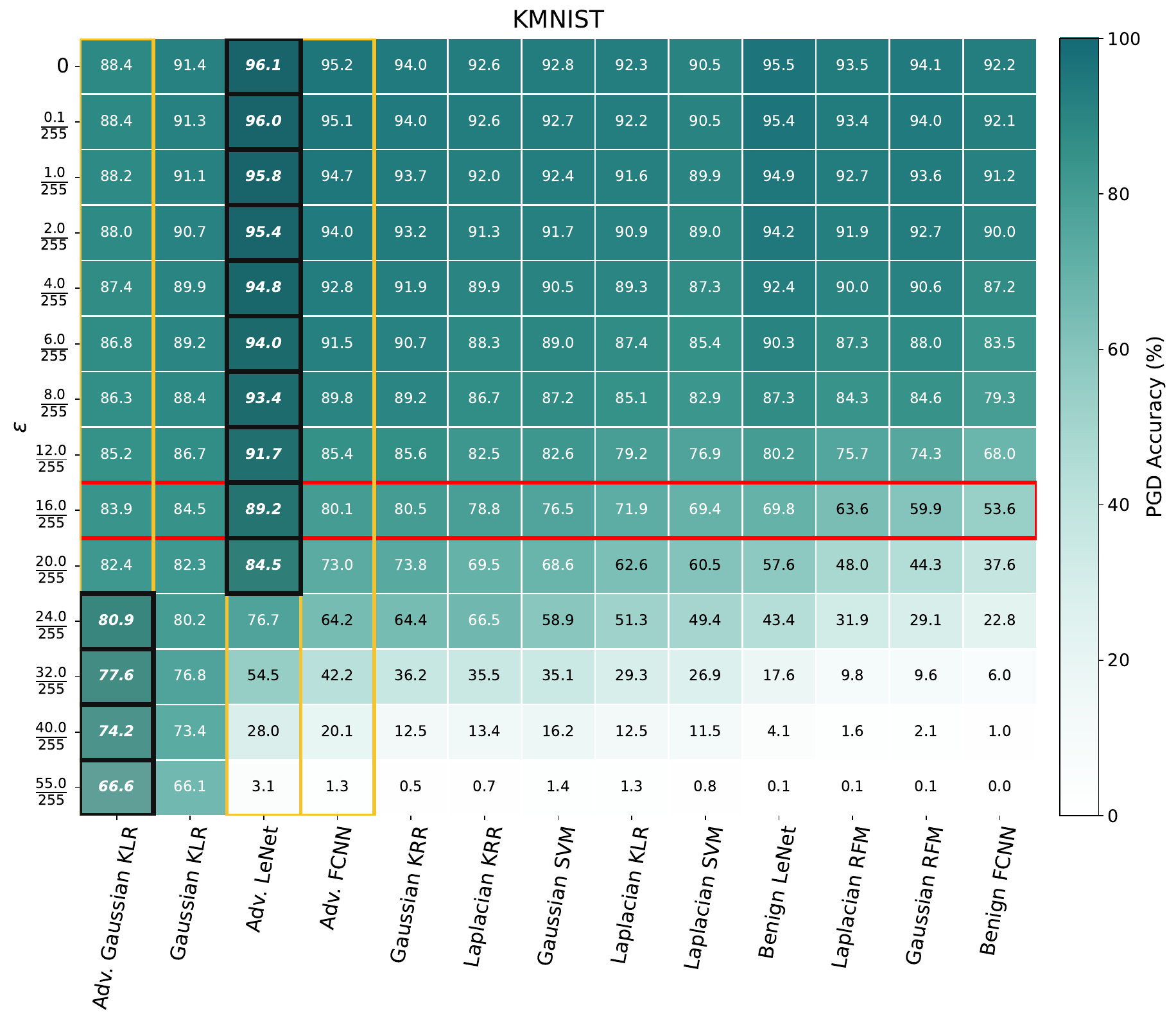}
    \caption{
Summary of results for the multiclass KMNIST dataset. We compare robust accuracy of 13 models (3 adversarially trained, 10 benign) on KMNIST (10 classes) attacked via PGD using $\ell_\infty$ perturbations. Columns with yellow borders correspond to adversarially trained models with training adversary's attack level highlighted by the red row. Black boxes mark the best robust accuracy across each row. Note the best models for high $\eps$ are kernel machines with logistic loss.
    }
    \label{fig:kmnist_mc}
\end{figure*}

\begin{figure*}[!htbp]
    \centering
    \includegraphics[width=0.75\textwidth]{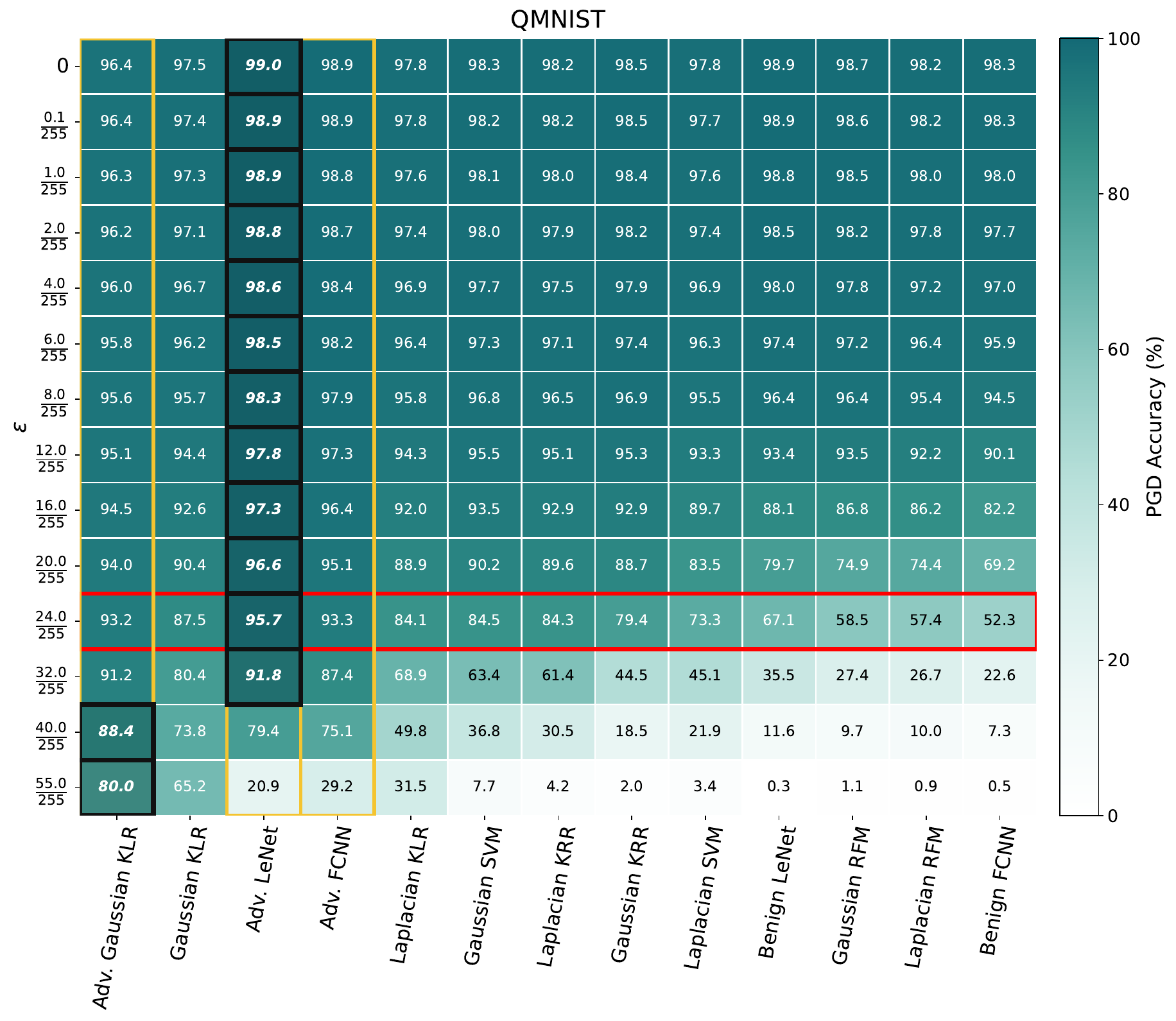}
    \caption{
Summary of results for the multiclass QMNIST dataset. We compare robust accuracy of 13 models (3 adversarially trained, 10 benign) on QMNIST (10 classes), attacked via PGD using $\ell_\infty$ perturbations.. Columns with yellow borders correspond to adversarially trained models with training adversary's attack level highlighted by the red row. Black boxes mark the best robust accuracy across each row. Note the best models for high $\eps$ are kernel machines with logistic loss.
    }
    \label{fig:qmnist_mc}
\end{figure*}

$\ell_\infty$ attack multiclass results are in~\cref{fig:kmnist_mc,fig:qmnist_mc}.

\begin{figure*}[!htbp]
    \centering
    \includegraphics[width=0.75\textwidth]{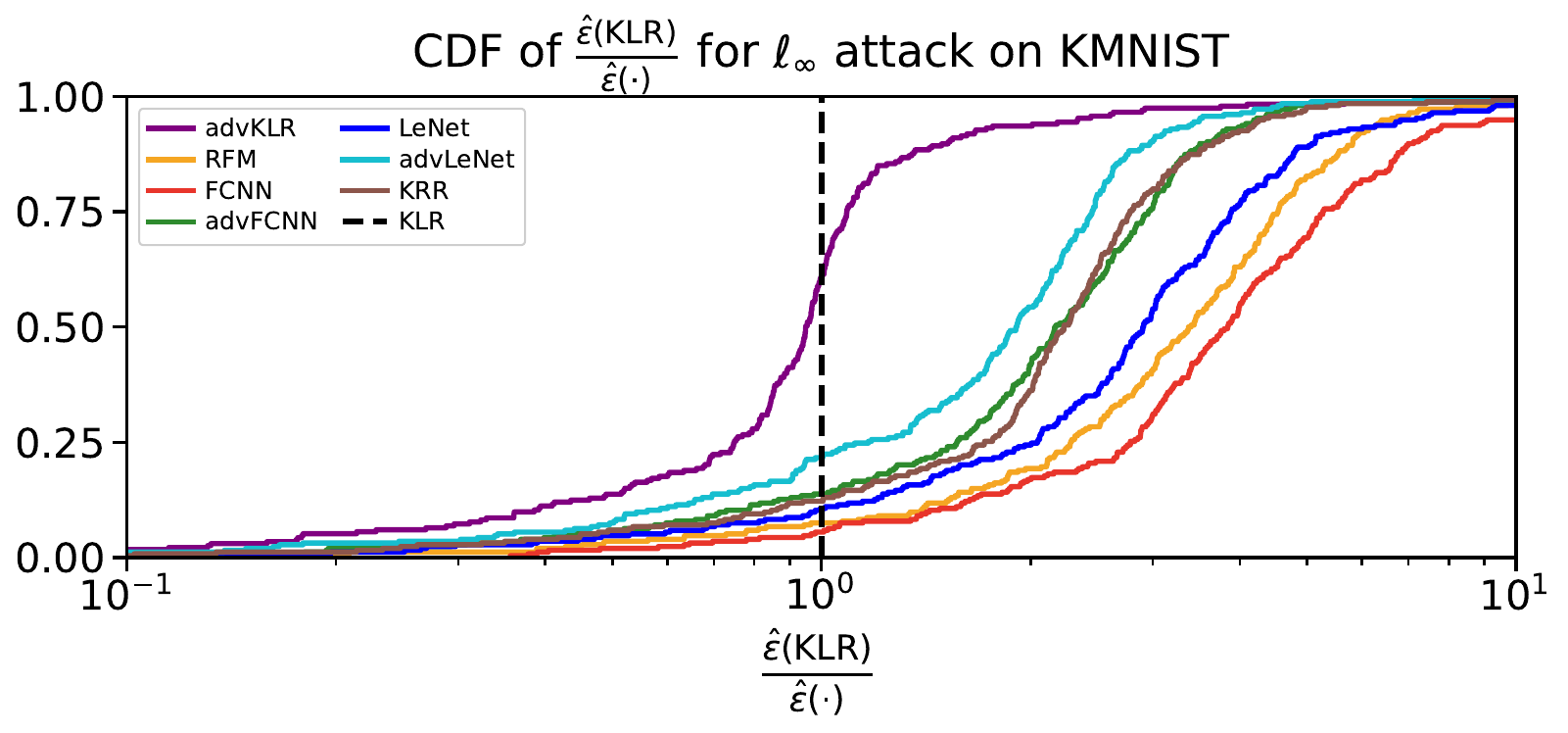}
    \caption{
Test for robustness as a pointwise property. We plot the empirical CDF of the ratio (KLR/other models) of the smallest $\ell_\infty$ perturbations $\wh\eps$ needed to successfully attack 423 random samples from the test set of KMNIST. Note the x-axis is in log-scale. See \cref{subsec:pointwise_robustness} for more details.\\
KRR is more robust than both FCNN and LeNet. For the median datum, KLR is nearly 2x more robust than Adv-FCNN and Adv-LeNet. The adversarially trained FCNN, LeNet and KLR were trained with an $\ell_\infty$ (matched) adversary with a budget of $\eps=\frac{16}{255}$. Adversarially trained KM remains most robust.}
    \label{fig:kmnist_pointwise}
\end{figure*}

\begin{figure*}[!htbp]
    \centering
    \includegraphics[width=0.75\textwidth]{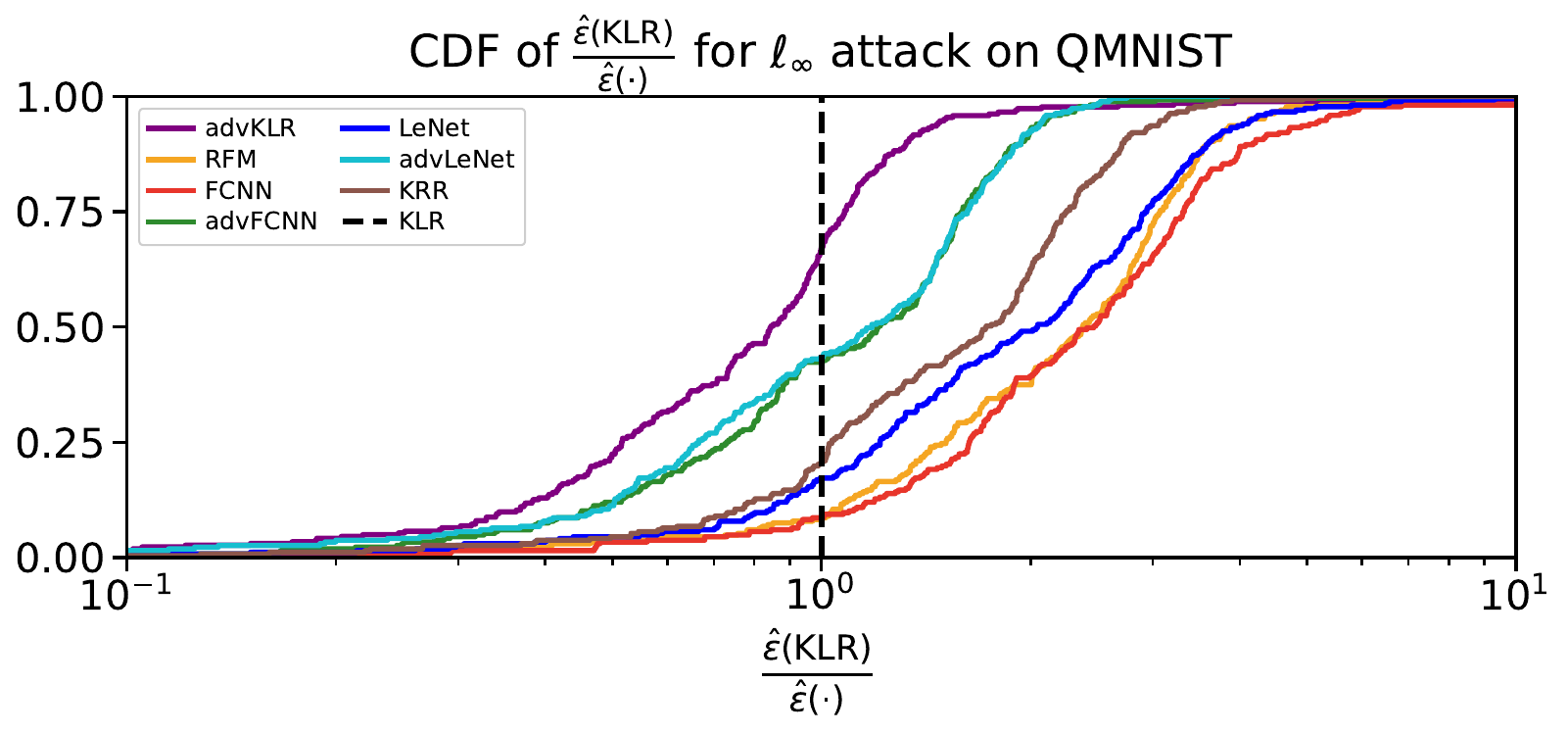}
    \caption{
Test for robustness as a pointwise property. We plot the empirical CDF of the ratio (KLR/other models) of the smallest $\ell_\infty$ perturbations $\wh\eps$ needed to successfully attack 479 random samples from the test set of QMNIST. Note the x-axis is in log-scale. See \cref{subsec:pointwise_robustness} for more details.\\
For most samples, KRR is harder to attack than the FCNN. For the median datum, KLR is more robust than Adv-FCNN and Adv-LeNet. The adversarially trained FCNN, LeNet and KLR were trained with an $\ell_\infty$ (matched) adversary with a budget of $\eps=\frac{24}{255}$. Adversarially trained KM remains most robust.}
    \label{fig:qmnist_pointwise}
\end{figure*}

\clearpage
\subsection{Comparison of KMs with FCNN and LeNet}
\begin{figure*}[!htbp]
    \caption{ \label{fig:gap_heatmaps_summary_linf} 
    \textbf{Gap in robust accuracy (ra) for $\ell_\infty$ perturbation on 45 $1\text{v}1$ binary classification tasks from FashionMNIST} comparing 2 neural networks (columns) and 3 kernel machines (rows) with the Laplacian kernel. 
    Perturbation size increases as we go down each heatmap, and values correspond to \textrm{ra}(row model) - \textrm{ra}(col model).
    The \textcolor{teal}{teal} colour indicates that the kernel machine is more robust, whereas the \textcolor{Bittersweet}{orange} color indicates that the neural network is more robust. Each of the 6 heatmaps compares 45 pairs of models.
    For each heatmap, class pairs for binary classification are sorted according to values in the bottom row. 
    (Top) Kernel machines are more robust than fully connected neural networks and convolutional networks. (Middle) Recursive Feature Machines retain much of the robustness of the base kernel machine ($M=I_d$). 
    (Bottom) Laplacian SVM are more robust to both FCNN and LeNet. PGD attack is used to attack the models.
    }

\includegraphics[width=\textwidth,keepaspectratio]
{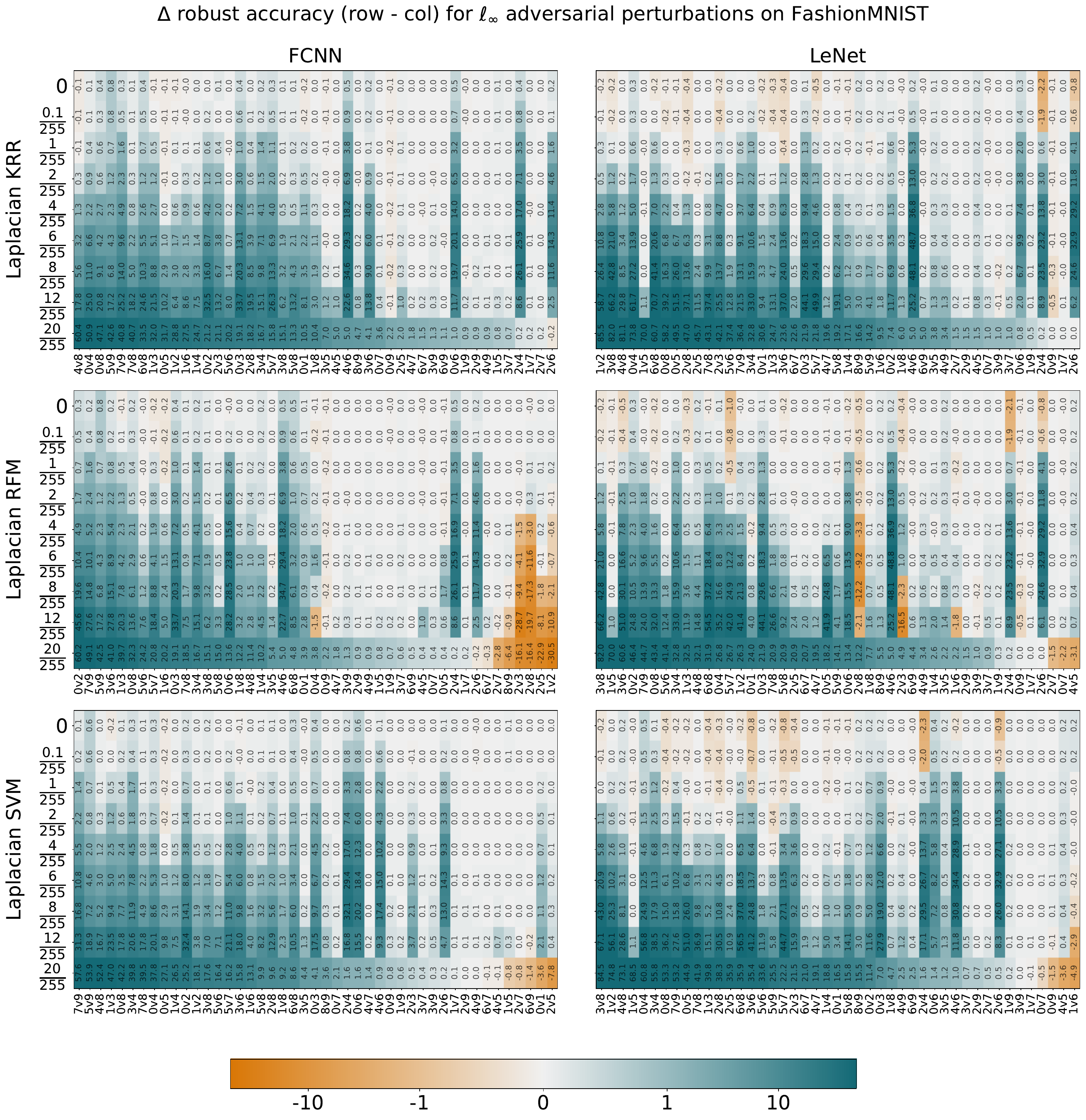}
\vspace{-3ex}
\end{figure*}

\clearpage
\subsection{Laplacian Results for FMNIST, KMNIST, QMNIST}
Here we provide results comparing laplacian kernels in any type of model over the datasets FMNIST, KMNIST, QMNIST.

\vspace{2.5mm}

\begin{figure*}[!htbp]
    \centering
\caption{\label{fig:gap_heatmaps_lapklr_vs_fcnn} \textbf{Laplacian KLR versus FCNN:} $\ell_2$ and $\ell_\infty$ perturbation for FashionMNIST, KMNIST and QMNIST. (Top) row corresponds to the results for FashionMNIST dataset, (Middle) row corresponds to the KMNIST dataset, and (Bottom) row corresponds to the QMNIST dataset. Of these, (First) column corresponds to the robust accuracies for $\ell_\infty$ adversaries while (Second) column portrays the robust accuracies for $\ell_2$ adversaries. PGD attack is used to attack the models.}
\includegraphics[width=1\textwidth]{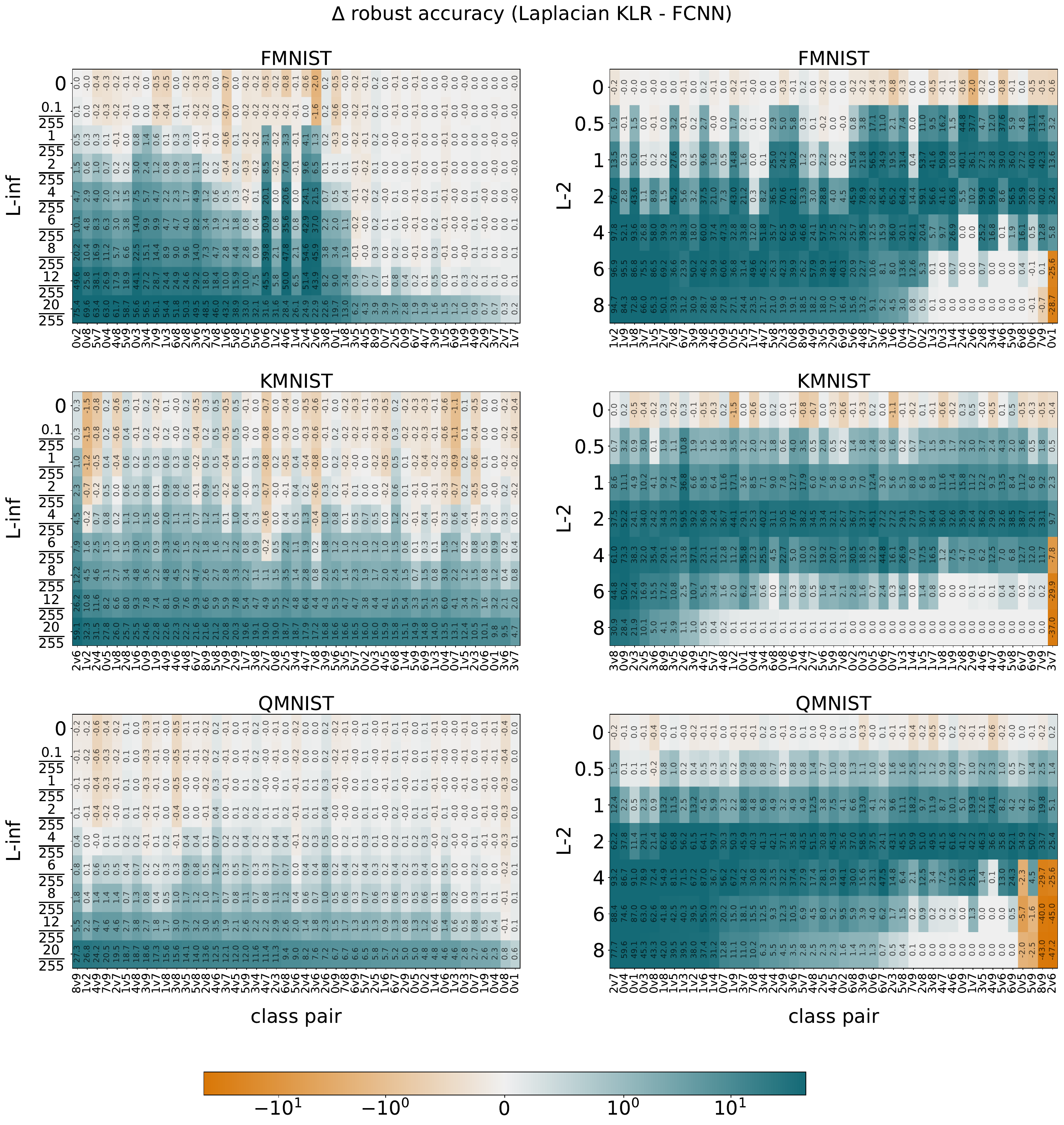}
\end{figure*}

\begin{figure*}
    \centering
    \caption{\label{fig:gap_heatmaps_lapklr_vs_lenet} \textbf{Laplacian KLR versus LeNet:} $\ell_2$ and $\ell_\infty$ perturbation for FashionMNIST, KMNIST and QMNIST. (Top) row corresponds to the results for FashionMNIST dataset, (Middle) row corresponds to the KMNIST dataset, and (Bottom) row corresponds to the QMNIST dataset. Of these, (First) column corresponds to the robust accuracies for $\ell_\infty$ adversaries while (Second) column portrays the robust accuracies for $\ell_2$ adversaries. PGD attack is used to attack the models.}
\includegraphics[width=\textwidth]{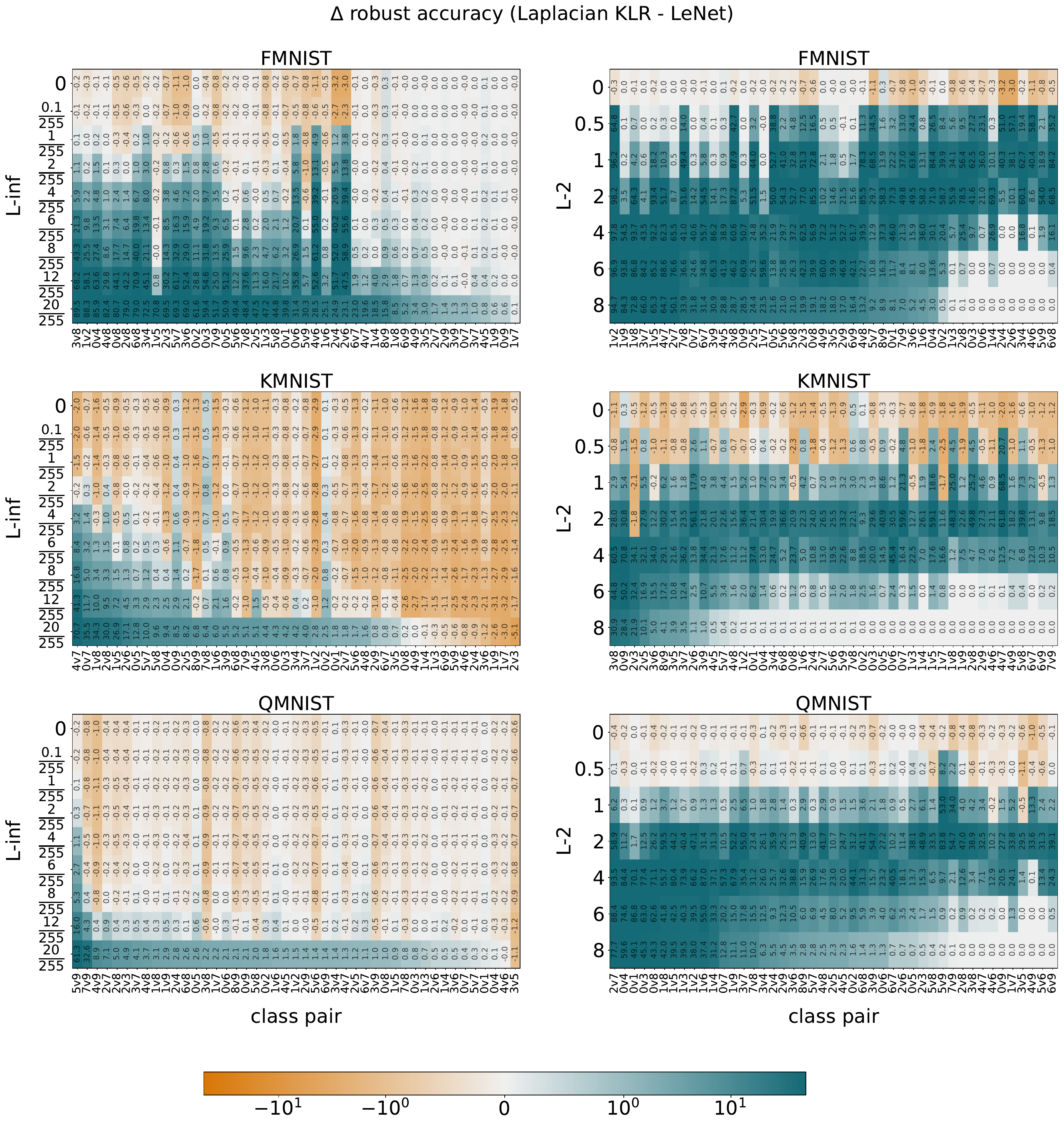}
\end{figure*}

\begin{figure*}[!htbp]
    \centering
\caption{\label{fig:gap_heatmaps_lap_vs_fcnn} \textbf{Laplacian KRR versus FCNN:} $\ell_2$ and $\ell_\infty$ perturbation for FashionMNIST, KMNIST and QMNIST. (Top) row corresponds to the results for FashionMNIST dataset, (Middle) row corresponds to the KMNIST dataset, and (Bottom) row corresponds to the QMNIST dataset. Of these, (First) column corresponds to the robust accuracies for $\ell_\infty$ adversaries while (Second) column portrays the robust accuracies for $\ell_2$ adversaries. PGD attack is used to attack the models.}
\includegraphics[width=1\textwidth]{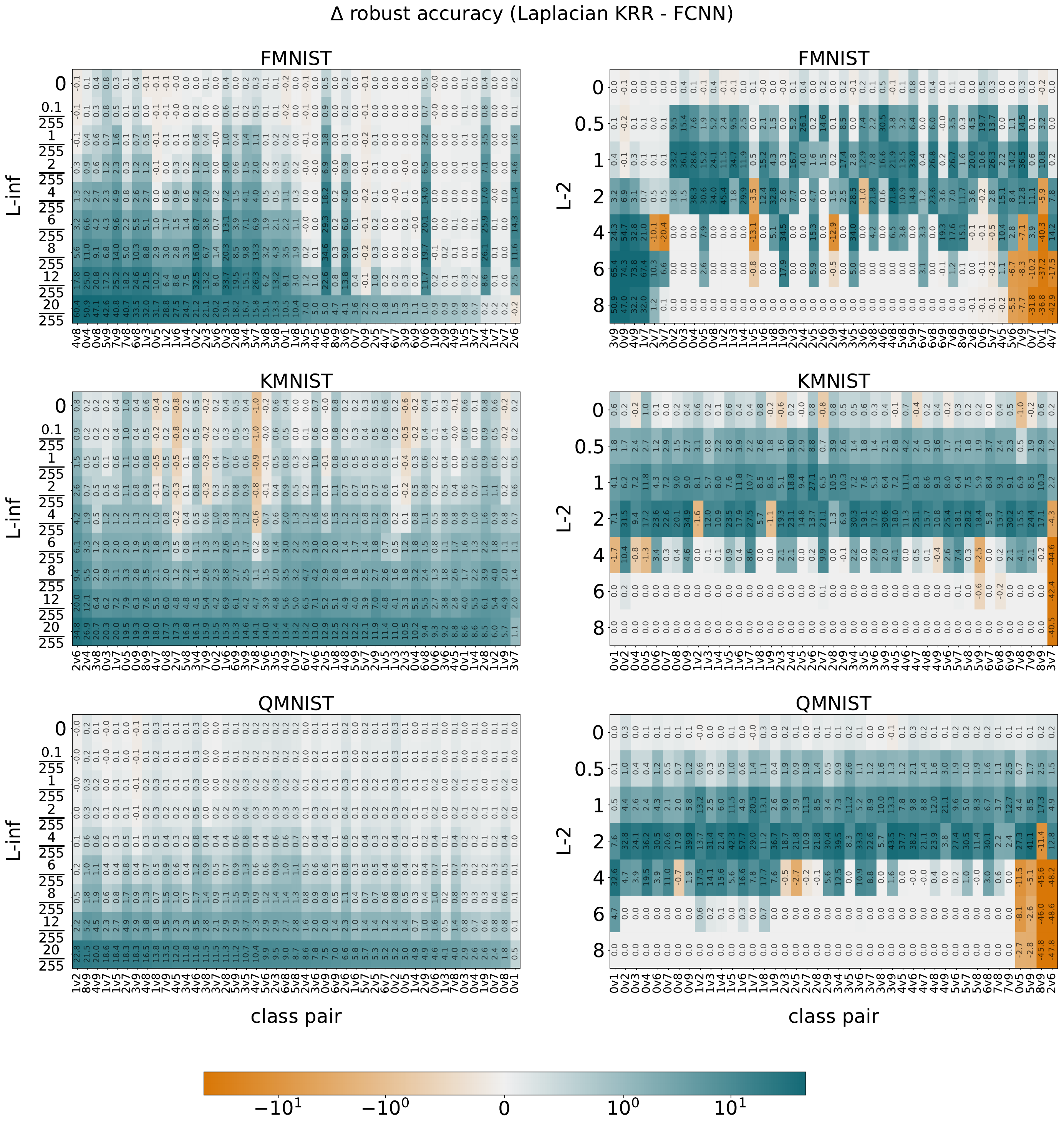}
\end{figure*}

\begin{figure*}
    \centering
    \caption{\label{fig:gap_heatmaps_lap_vs_lenet} \textbf{Laplacian KRR versus LeNet:} $\ell_2$ and $\ell_\infty$ perturbation for FashionMNIST, KMNIST and QMNIST. (Top) row corresponds to the results for FashionMNIST dataset, (Middle) row corresponds to the KMNIST dataset, and (Bottom) row corresponds to the QMNIST dataset. Of these, (First) column corresponds to the robust accuracies for $\ell_\infty$ adversaries while (Second) column portrays the robust accuracies for $\ell_2$ adversaries. PGD attack is used to attack the models.}
\includegraphics[width=\textwidth]{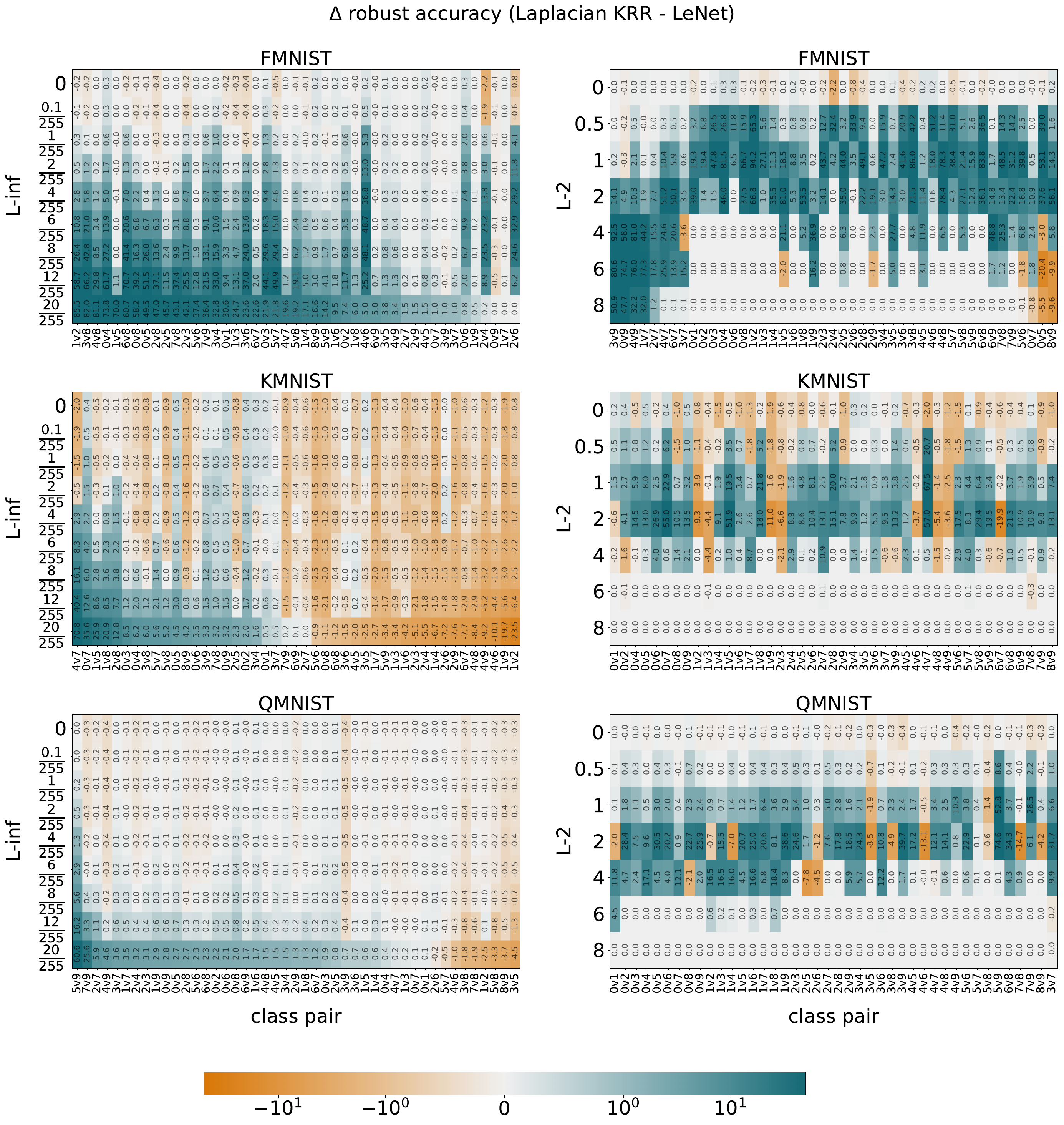}

\end{figure*}

\begin{figure*}
    \centering
    \caption{\label{fig:gap_heatmaps_laprfmvsnn} \textbf{Laplacian RFM versus FCNN:} $\ell_2$ and $\ell_\infty$ perturbation for FashionMNIST, KMNIST and QMNIST. (Top) row corresponds to the results for FashionMNIST dataset, (Middle) row corresponds to the KMNIST dataset, and (Bottom) row corresponds to the QMNIST dataset. Of these, (First) column corresponds to the robust accuracies for $\ell_\infty$ adversaries while (Second) column portrays the robust accuracies for $\ell_2$ adversaries. PGD attack is used to attack the models.}
\includegraphics[width=1\textwidth]{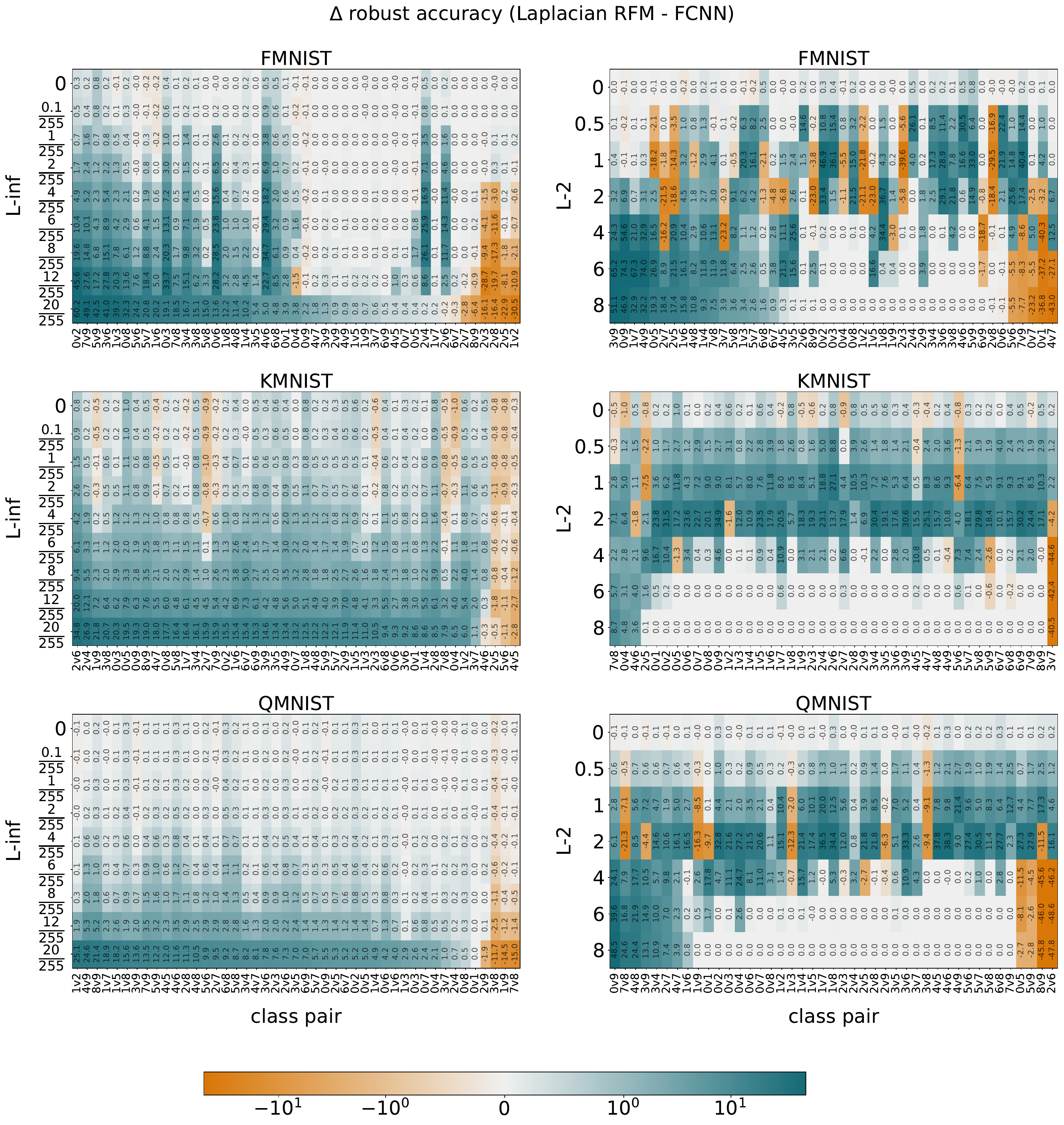}

\end{figure*}

\begin{figure*}
    \centering
    \caption{\label{fig:gap_heatmaps_laprfmvslenet} \textbf{Laplacian RFM versus LeNet:} $\ell_2$ and $\ell_\infty$ perturbation for FashionMNIST, KMNIST and QMNIST. (Top) row corresponds to the results for FashionMNIST dataset, (Middle) row corresponds to the KMNIST dataset, and (Bottom) row corresponds to the QMNIST dataset. Of these, (First) column corresponds to the robust accuracies for $\ell_\infty$ adversaries while (Second) column portrays the robust accuracies for $\ell_2$ adversaries. PGD attack is used to attack the models.}
\includegraphics[width=1\textwidth]{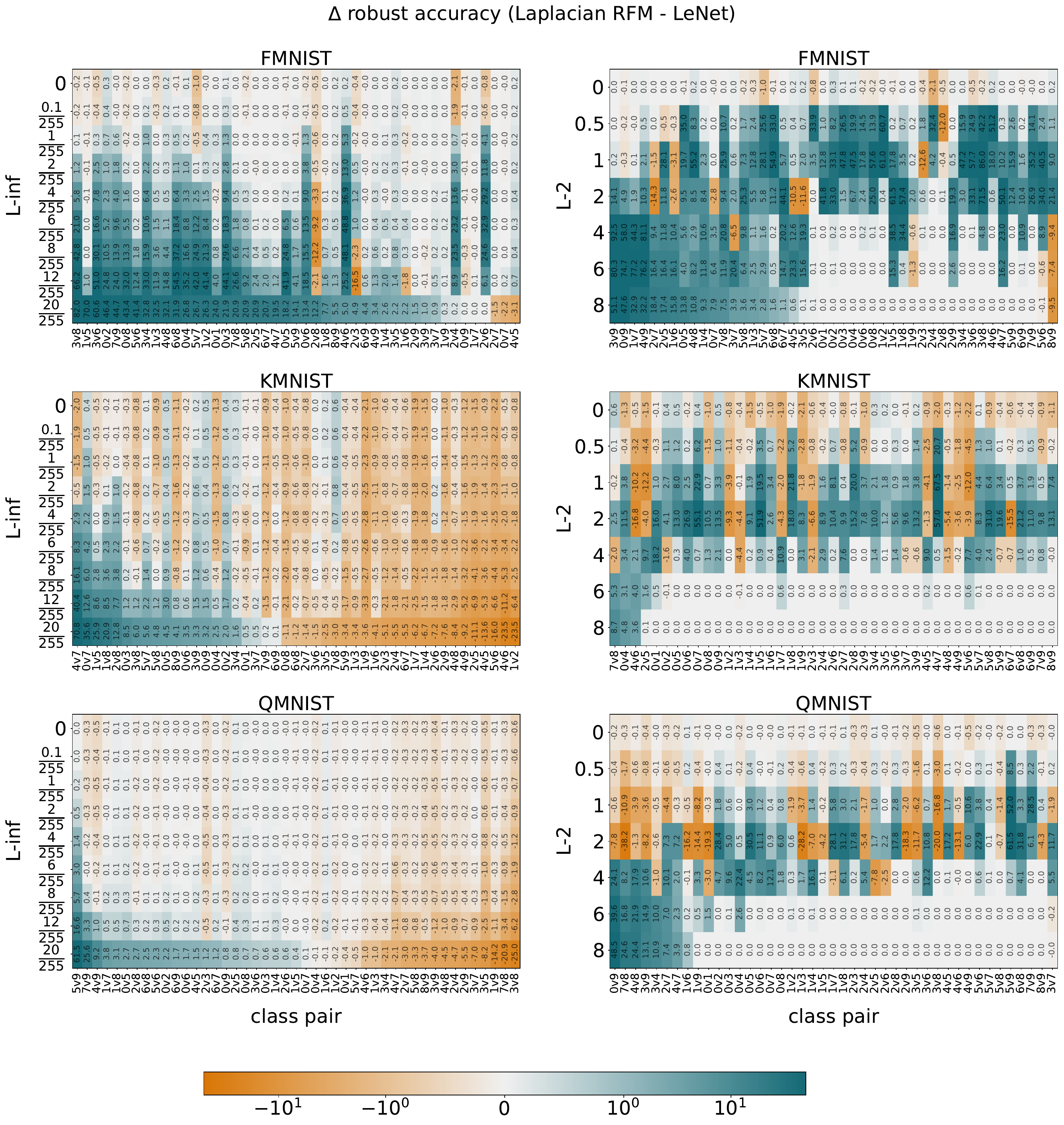}

\end{figure*}

\begin{figure*}
    \centering
    \caption{\label{fig:gap_heatmaps_lapsvmvsfcnn} \textbf{Laplacian SVM versus FCNN:} $\ell_2$ and $\ell_\infty$ perturbation for FashionMNIST, KMNIST and QMNIST. (Top) row corresponds to the results for FashionMNIST dataset, (Middle) row corresponds to the KMNIST dataset, and (Bottom) row corresponds to the QMNIST dataset. Of these, (First) column corresponds to the robust accuracies for $\ell_\infty$ adversaries while (Second) column portrays the robust accuracies for $\ell_2$ adversaries. PGD attack is used to attack the models.}
\includegraphics[width=1\textwidth]{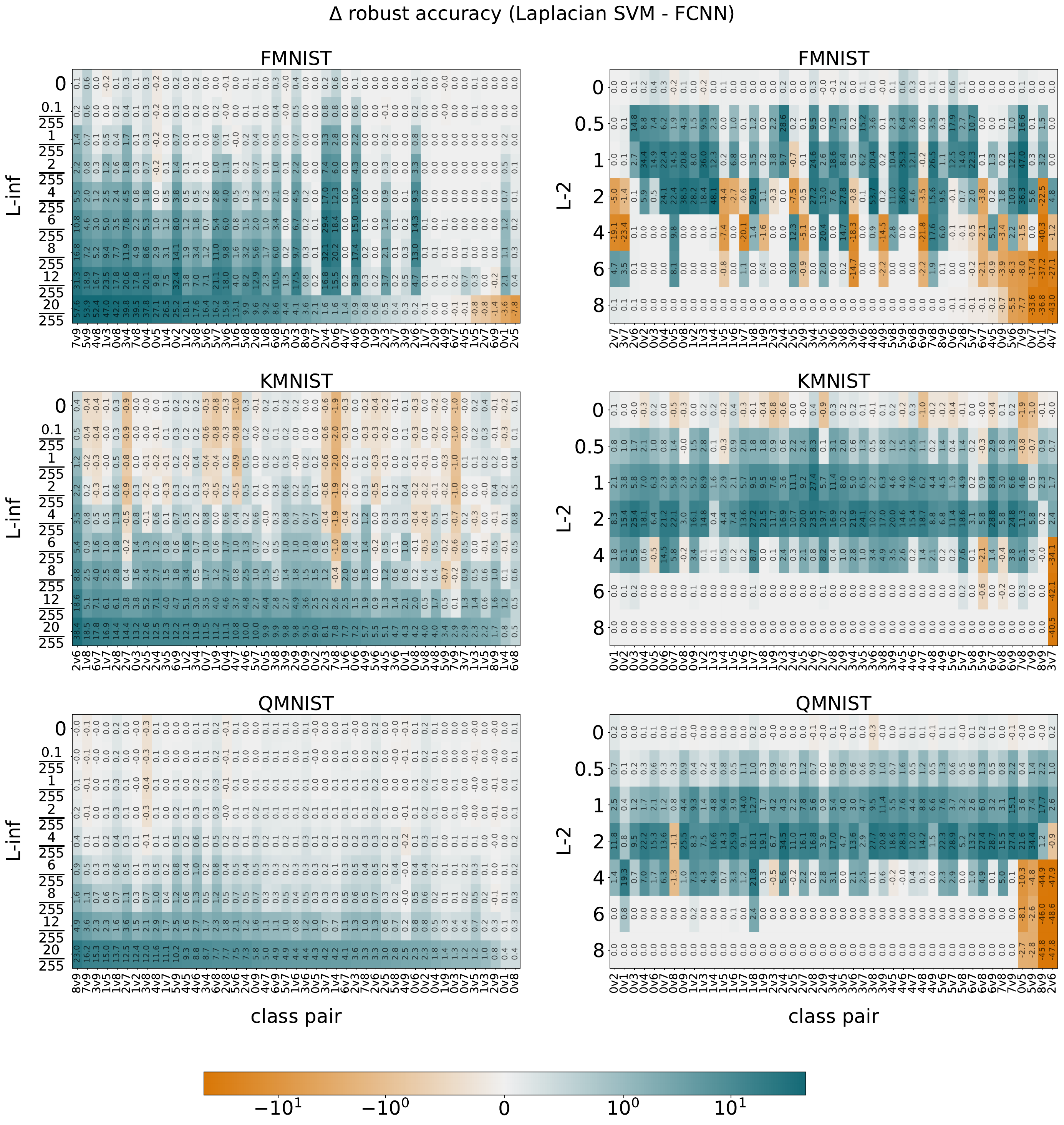}

\end{figure*}

\begin{figure*}
    \centering
    \caption{\label{fig:gap_heatmaps_lapsvmvslenet} \textbf{Laplacian SVM versus LeNet:} $\ell_2$ and $\ell_\infty$ perturbation for FashionMNIST, KMNIST and QMNIST. (Top) row corresponds to the results for FashionMNIST dataset, (Middle) row corresponds to the KMNIST dataset, and (Bottom) row corresponds to the QMNIST dataset. Of these, (First) column corresponds to the robust accuracies for $\ell_\infty$ adversaries while (Second) column portrays the robust accuracies for $\ell_2$ adversaries. PGD attack is used to attack the models.}
\includegraphics[width=1\textwidth]{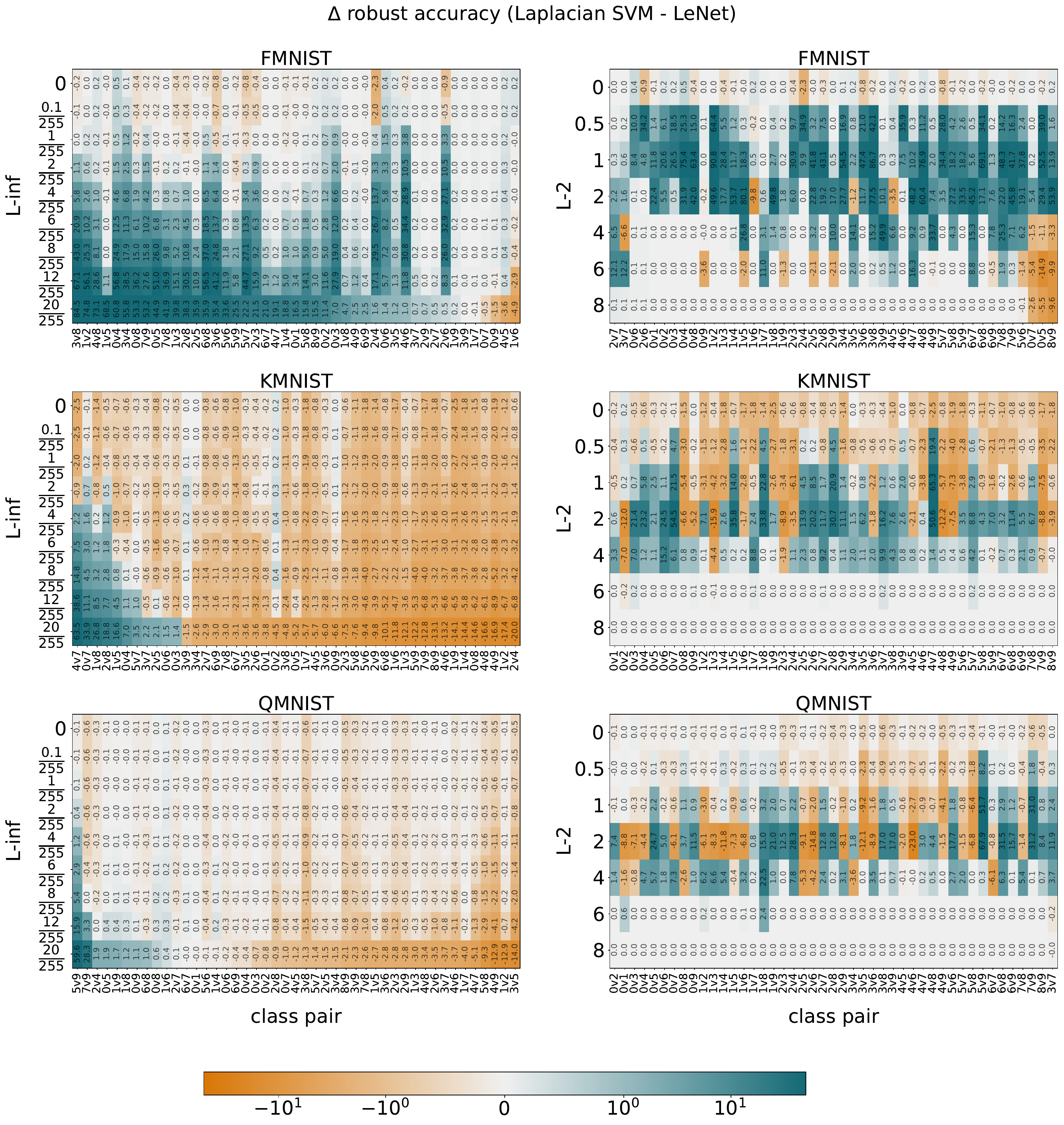}

\end{figure*}

\begin{figure*}[!htbp]
    \centering
    \caption{\label{fig:gap_heatmaps_lapklrvskrr} \textbf{Laplacian KLR vs Laplacian KRR:} $\ell_2$ and $\ell_\infty$ perturbation for FashionMNIST, KMNIST and QMNIST. (Top) row corresponds to the results for FashionMNIST dataset, (Middle) row corresponds to the KMNIST dataset, and (Bottom) row corresponds to the QMNIST dataset. Of these, (First) column corresponds to the robust accuracies for $\ell_\infty$ adversaries while (Second) column portrays the robust accuracies for $\ell_2$ adversaries. PGD attack is used to attack the models.}
\includegraphics[width=1\textwidth]{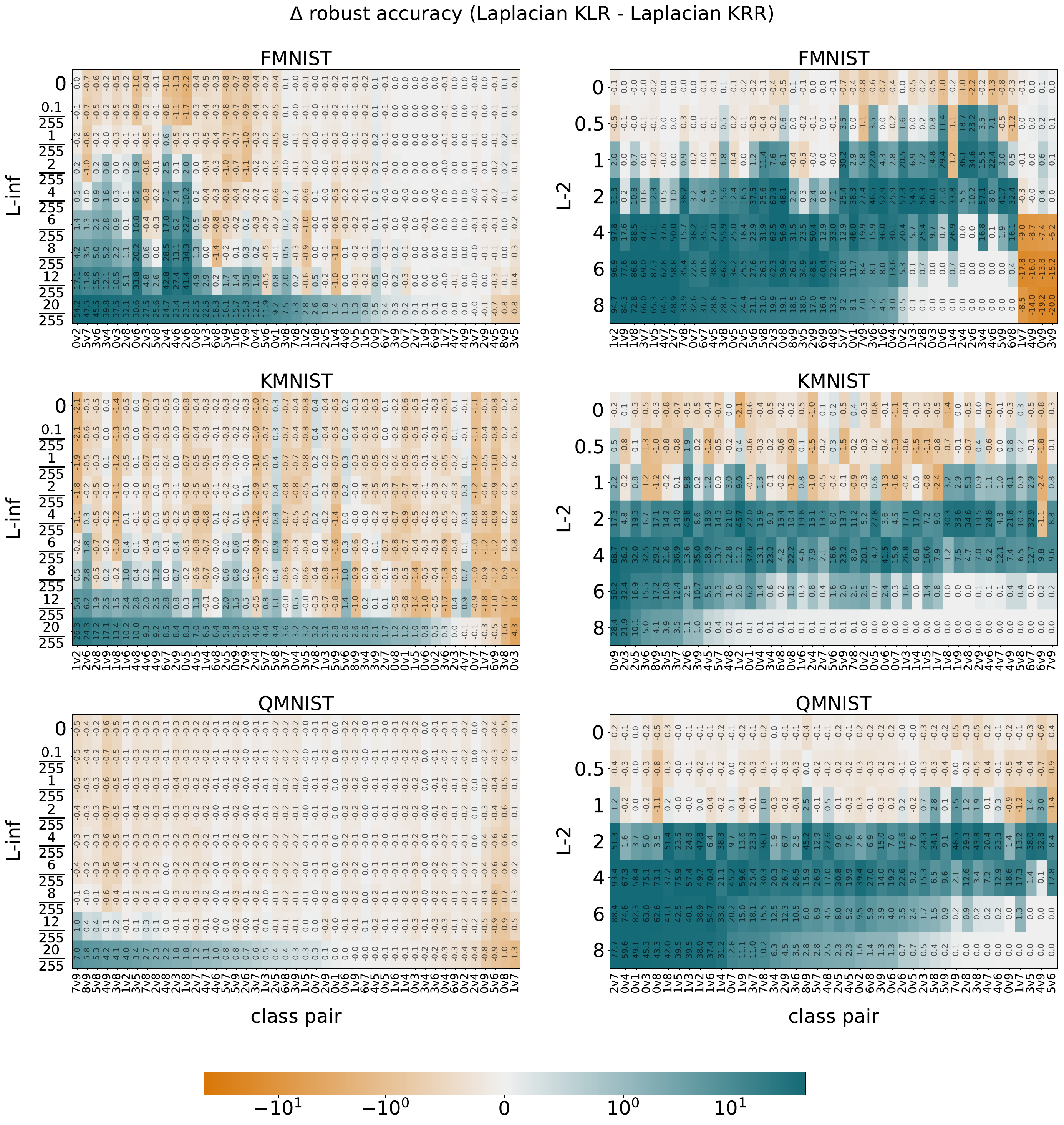}
\end{figure*}

\begin{figure*}[!htbp]
    \centering
    \caption{\label{fig:gap_heatmaps_lapklrvsrfm} \textbf{Laplacian KLR vs Laplacian RFM:} $\ell_2$ and $\ell_\infty$ perturbation for FashionMNIST, KMNIST and QMNIST. (Top) row corresponds to the results for FashionMNIST dataset, (Middle) row corresponds to the KMNIST dataset, and (Bottom) row corresponds to the QMNIST dataset. Of these, (First) column corresponds to the robust accuracies for $\ell_\infty$ adversaries while (Second) column portrays the robust accuracies for $\ell_2$ adversaries. PGD attack is used to attack the models.}
\includegraphics[width=1\textwidth]{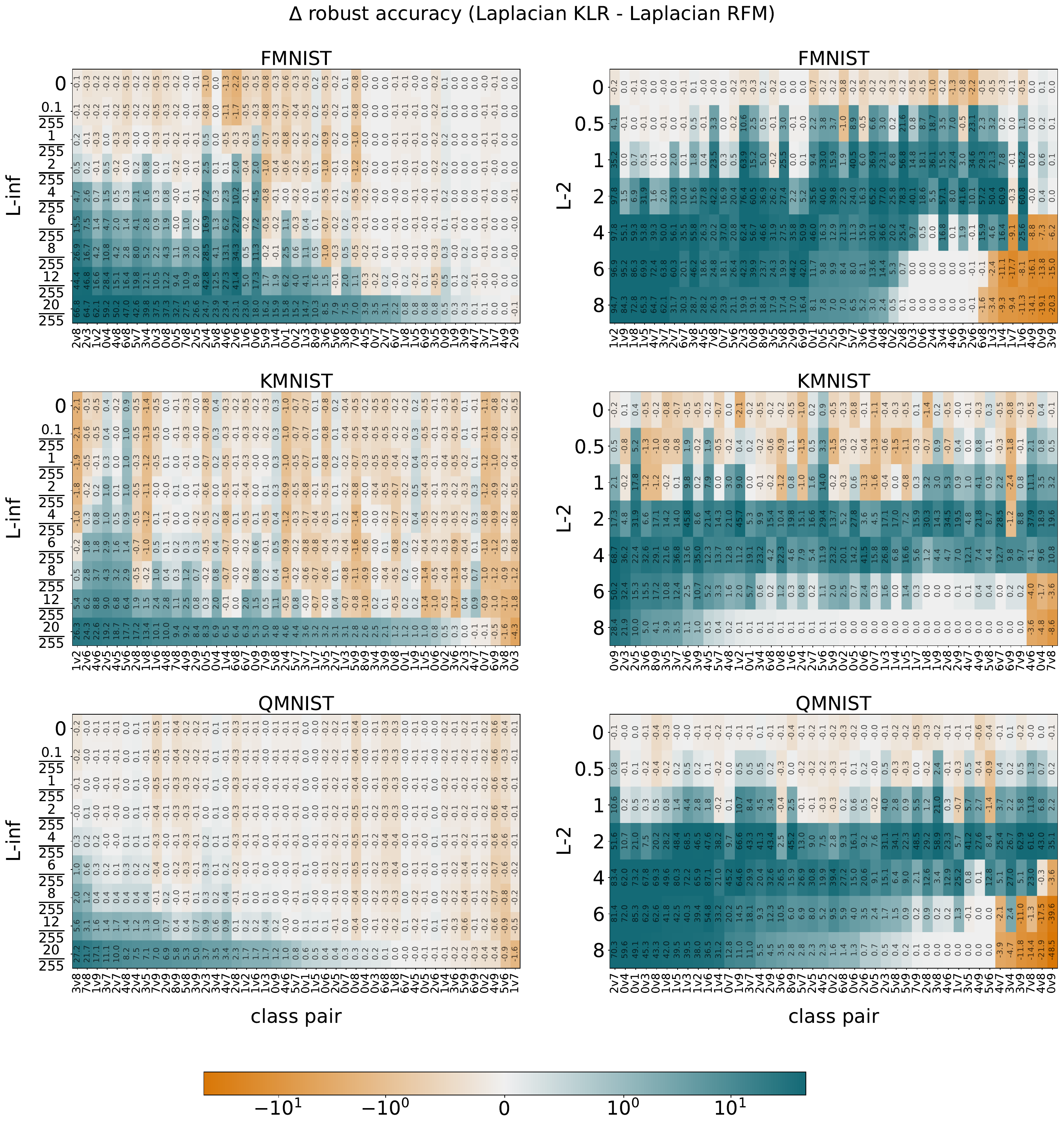}
\end{figure*}

\begin{figure*}[!htbp]
    \centering
    \caption{\label{fig:gap_heatmaps_lapklrvssvm} \textbf{Laplacian KLR vs Laplacian SVM:} $\ell_2$ and $\ell_\infty$ perturbation for FashionMNIST, KMNIST and QMNIST. (Top) row corresponds to the results for FashionMNIST dataset, (Middle) row corresponds to the KMNIST dataset, and (Bottom) row corresponds to the QMNIST dataset. Of these, (First) column corresponds to the robust accuracies for $\ell_\infty$ adversaries while (Second) column portrays the robust accuracies for $\ell_2$ adversaries. PGD attack is used to attack the models.}
\includegraphics[width=1\textwidth]{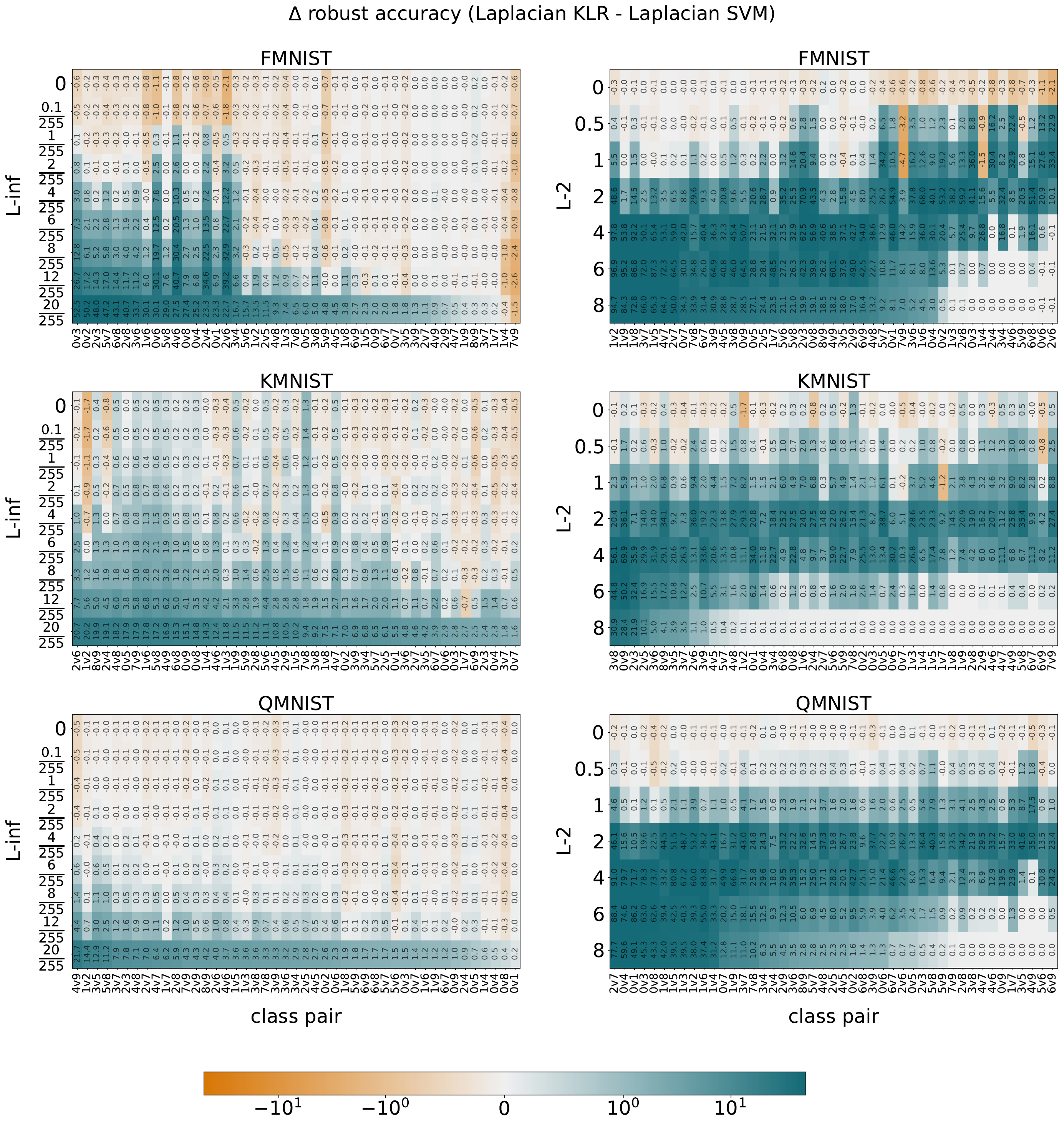}
\end{figure*}

\begin{figure*}[!htbp]
    \centering
    \caption{\label{fig:gap_heatmaps_lapkervsrfm} \textbf{Laplacian KRR vs Laplacian RFM:} $\ell_2$ and $\ell_\infty$ perturbation for FashionMNIST, KMNIST and QMNIST. (Top) row corresponds to the results for FashionMNIST dataset, (Middle) row corresponds to the KMNIST dataset, and (Bottom) row corresponds to the QMNIST dataset. Of these, (First) column corresponds to the robust accuracies for $\ell_\infty$ adversaries while (Second) column portrays the robust accuracies for $\ell_2$ adversaries. PGD attack is used to attack the models.}
\includegraphics[width=1\textwidth]{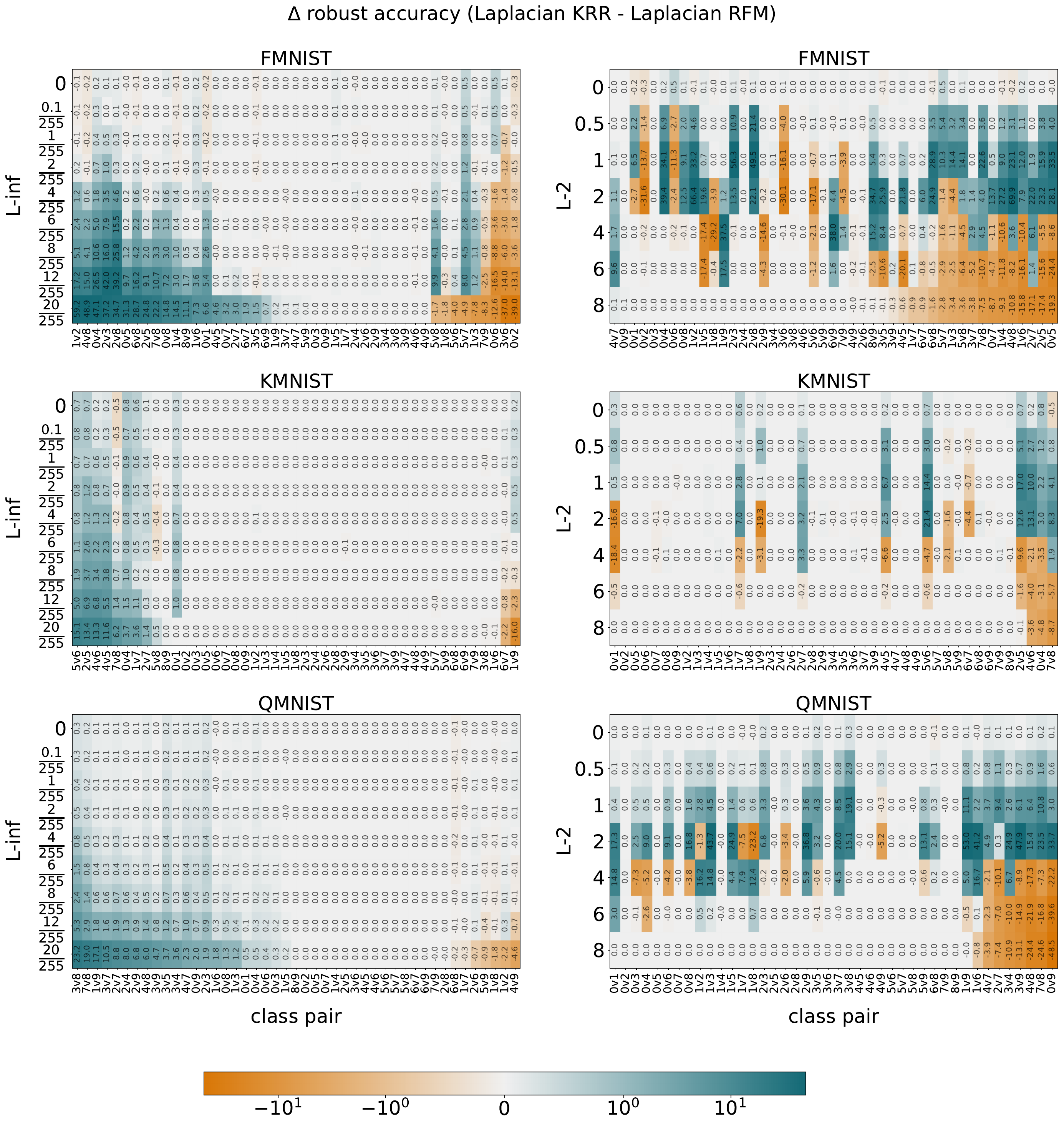}

\end{figure*}

\begin{figure*}[!htbp]
    \centering
    \caption{\label{fig:gap_heatmaps_lapkrr_vs_svm} \textbf{Laplacian KRR vs Laplacian SVM:} $\ell_2$ and $\ell_\infty$ perturbation for FashionMNIST, KMNIST and QMNIST. (Top) row corresponds to the results for FashionMNIST dataset, (Middle) row corresponds to the KMNIST dataset, and (Bottom) row corresponds to the QMNIST dataset. Of these, (First) column corresponds to the robust accuracies for $\ell_\infty$ adversaries while (Second) column portrays the robust accuracies for $\ell_2$ adversaries. PGD attack is used to attack the models.}
\includegraphics[width=1\textwidth]{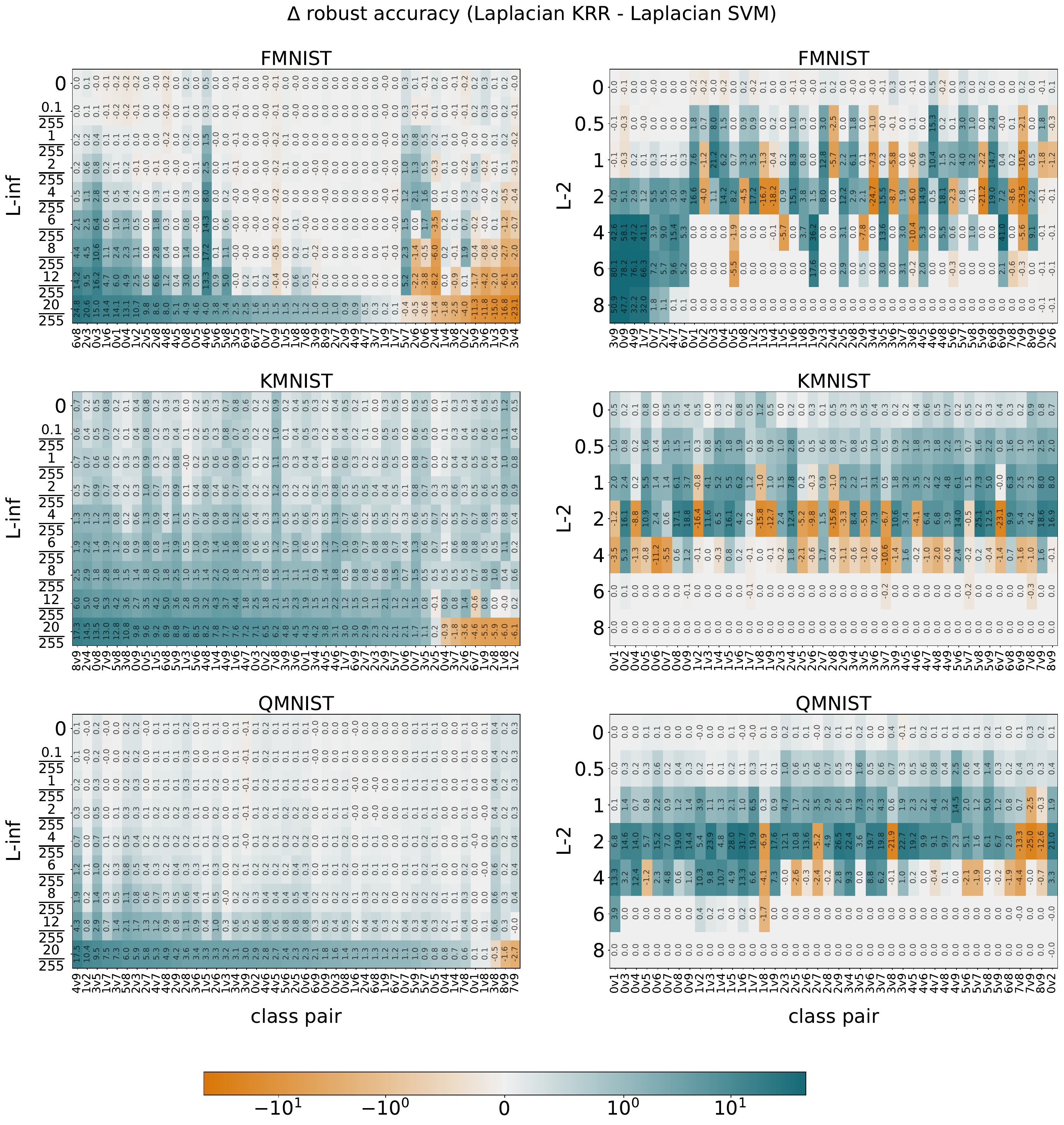}

\end{figure*}

\begin{figure*}[!htbp]
    \centering
    \caption{\label{fig:gap_heatmaps_laprfm_vs_svm} \textbf{Laplacian RFM vs Laplacian SVM:} $\ell_2$ and $\ell_\infty$ perturbation for FashionMNIST, KMNIST and QMNIST. (Top) row corresponds to the results for FashionMNIST dataset, (Middle) row corresponds to the KMNIST dataset, and (Bottom) row corresponds to the QMNIST dataset. Of these, (First) column corresponds to the robust accuracies for $\ell_\infty$ adversaries while (Second) column portrays the robust accuracies for $\ell_2$ adversaries. PGD attack is used to attack the models.}
\includegraphics[width=1\textwidth]{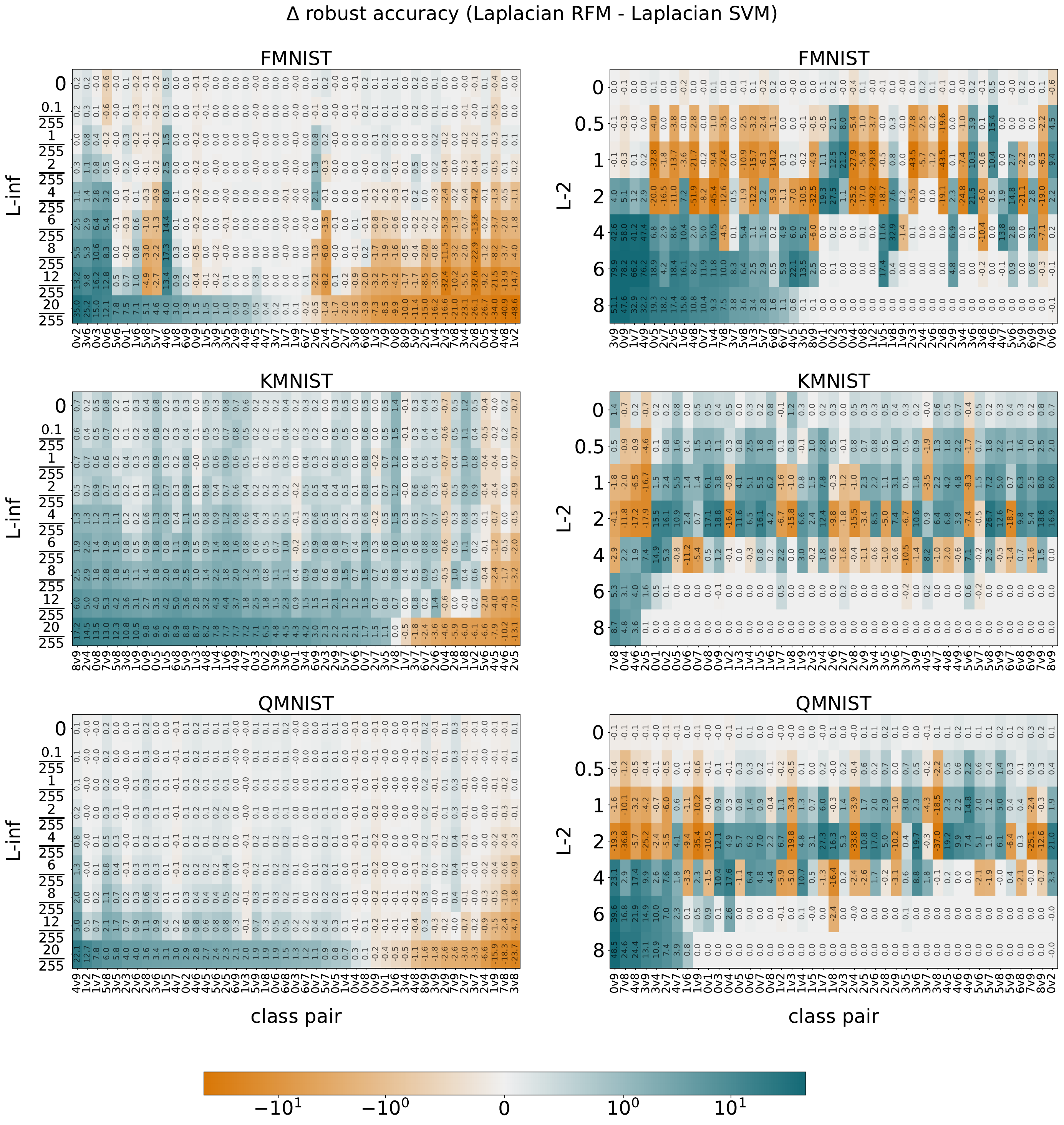}

\end{figure*}

\clearpage

\begin{figure*}[!htbp]
    \centering
    \caption{\label{fig:gap_heatmaps_rff5kvsfcnn} \textbf{Laplacian RFC (5k features) versus FCNN:} $\ell_2$ and $\ell_\infty$ perturbation for FashionMNIST, KMNIST and QMNIST. (Top) row corresponds to the results for FashionMNIST dataset, (Middle) row corresponds to the KMNIST dataset, and (Bottom) row corresponds to the QMNIST dataset. Of these, (First) column corresponds to the robust accuracies for $\ell_\infty$ adversaries while (Second) column portrays the robust accuracies for $\ell_2$ adversaries. Laplace RFC is trained using square loss. PGD attack is used to attack the models.}
\includegraphics[width=1\textwidth]{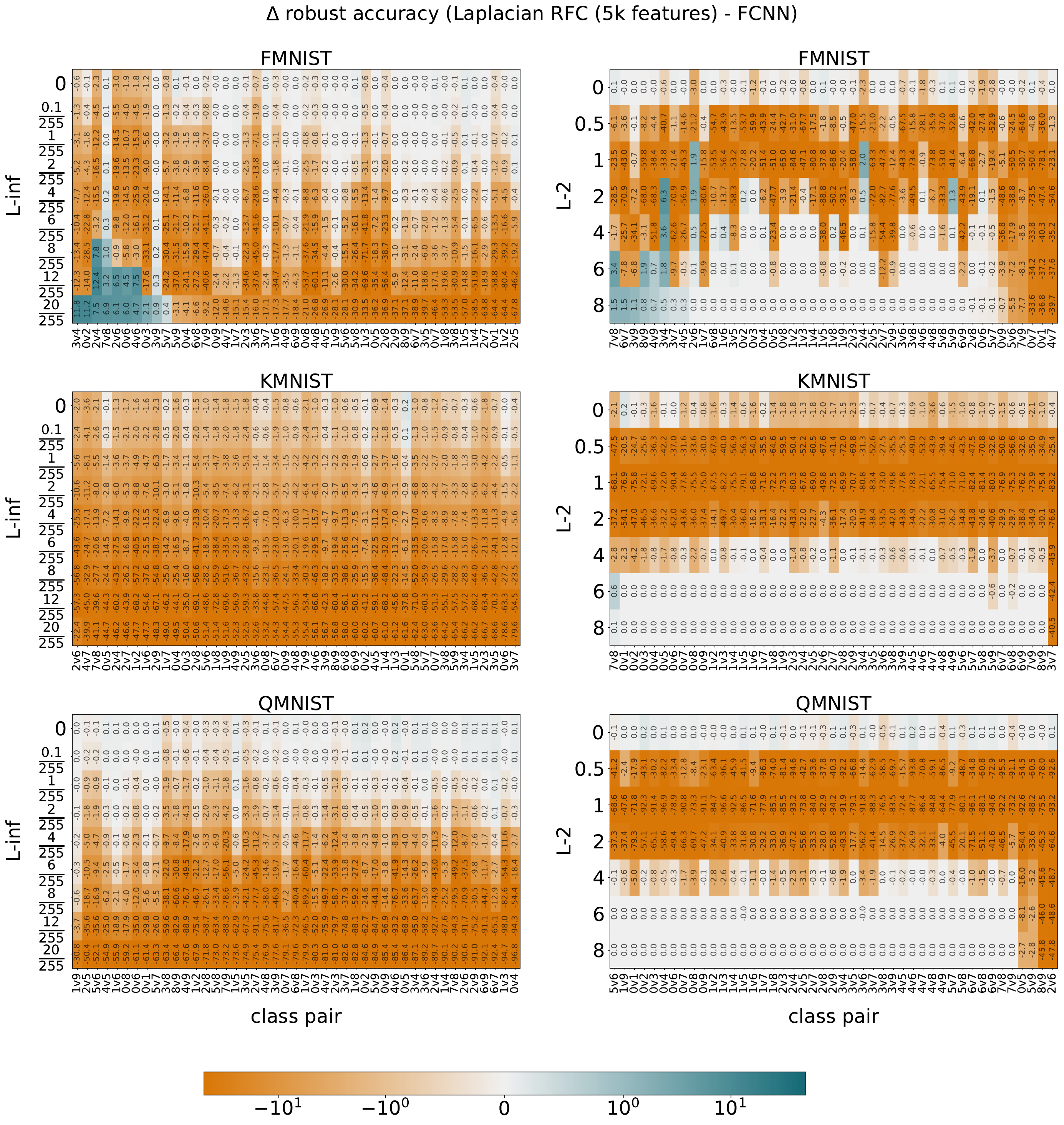}

\end{figure*}

\begin{figure*}[!htbp]
    \centering
    \caption{\label{fig:gap_heatmaps_rff20kvsfcnn} \textbf{Laplacian RFC (20k features) versus FCNN:} $\ell_2$ and $\ell_\infty$ perturbation for FashionMNIST, KMNIST and QMNIST. (Top) row corresponds to the results for FashionMNIST dataset, (Middle) row corresponds to the KMNIST dataset, and (Bottom) row corresponds to the QMNIST dataset. Of these, (First) column corresponds to the robust accuracies for $\ell_\infty$ adversaries while (Second) column portrays the robust accuracies for $\ell_2$ adversaries. Laplace RFC is trained using square loss. PGD attack is used to attack the models.}
\includegraphics[width=1\textwidth]{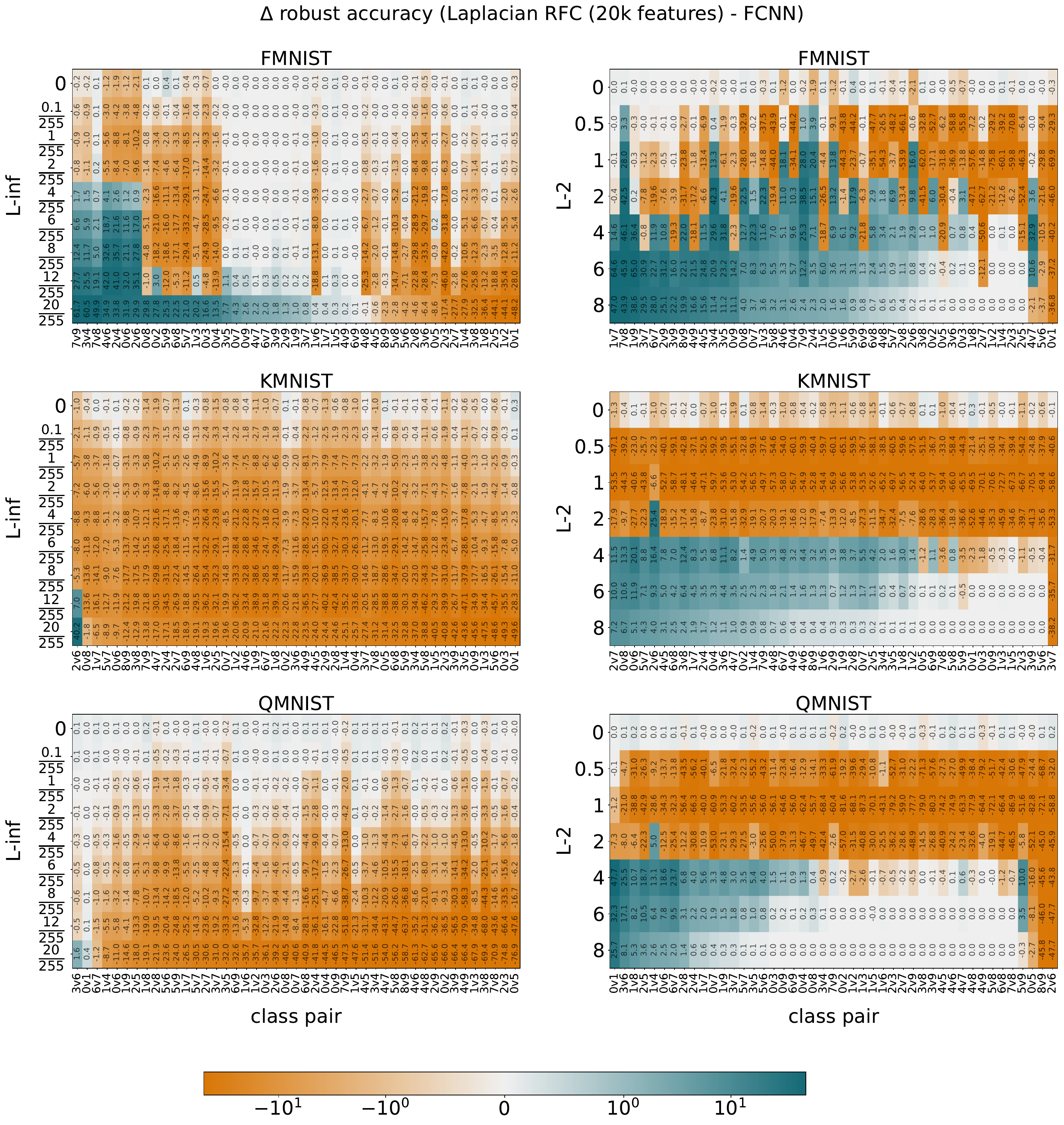}

\end{figure*}

\begin{figure*}[!htbp]
\centering
    \caption{\label{fig:gap_lapheatmaps_klrvsrff5k} \textbf{Laplacian KLR vs Laplacian RFC (5k features):} $\ell_2$ and $\ell_\infty$ perturbation for FashionMNIST, KMNIST and QMNIST. (Top) row corresponds to the results for FashionMNIST dataset, (Middle) row corresponds to the KMNIST dataset, and (Bottom) row corresponds to the QMNIST dataset. Of these, (First) column corresponds to the robust accuracies for $\ell_\infty$ adversaries while (Second) column portrays the robust accuracies for $\ell_2$ adversaries. PGD attack is used to attack the models.}
\includegraphics[width=1\textwidth]{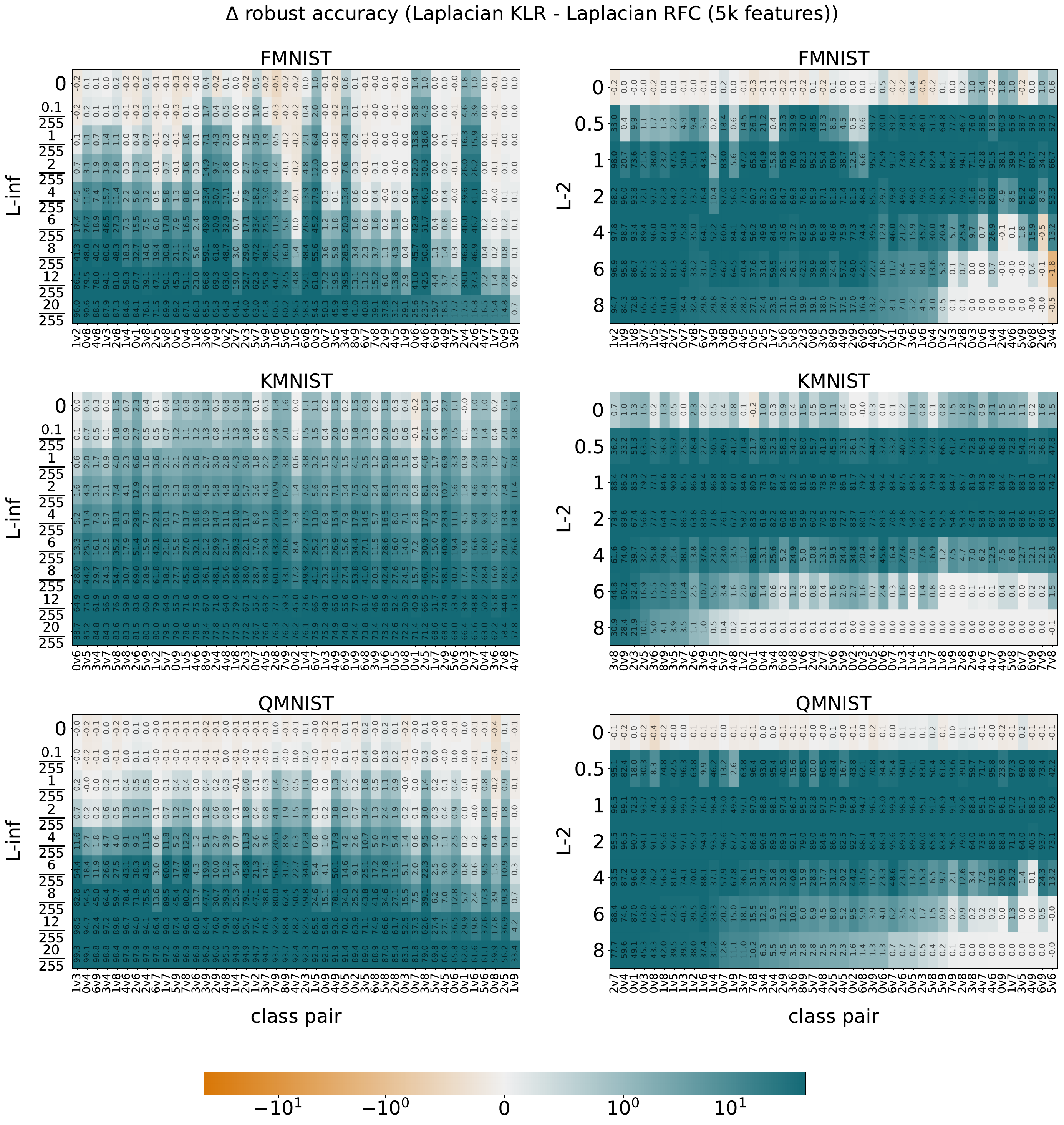}
\end{figure*}

\begin{figure*}[!htbp]
\centering
    \caption{\label{fig:gap_lapheatmaps_klrvsrff20k} \textbf{Laplacian KLR vs Laplacian RFC (20k features):} $\ell_2$ and $\ell_\infty$ perturbation for FashionMNIST, KMNIST and QMNIST. (Top) row corresponds to the results for FashionMNIST dataset, (Middle) row corresponds to the KMNIST dataset, and (Bottom) row corresponds to the QMNIST dataset. Of these, (First) column corresponds to the robust accuracies for $\ell_\infty$ adversaries while (Second) column portrays the robust accuracies for $\ell_2$ adversaries. PGD attack is used to attack the models.}
\includegraphics[width=1\textwidth]{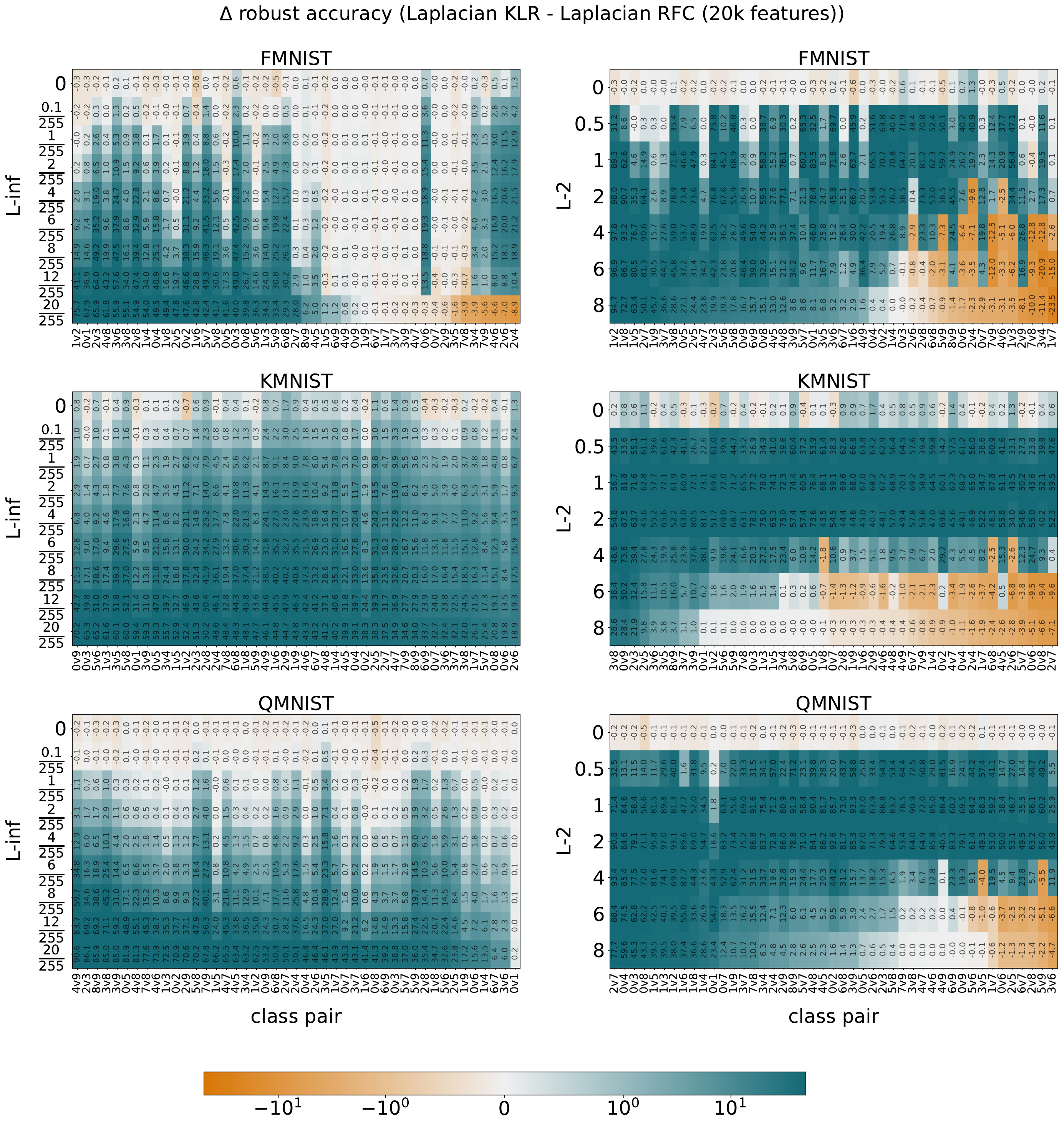}
\end{figure*}

\begin{figure*}[!htbp]
\centering
    \caption{\label{fig:gap_heatmaps_rff5kvsker} \textbf{Laplacian KRR vs Laplacian RFC (5k features):} $\ell_2$ and $\ell_\infty$ perturbation for FashionMNIST, KMNIST and QMNIST. (Top) row corresponds to the results for FashionMNIST dataset, (Middle) row corresponds to the KMNIST dataset, and (Bottom) row corresponds to the QMNIST dataset. Of these, (First) column corresponds to the robust accuracies for $\ell_\infty$ adversaries while (Second) column portrays the robust accuracies for $\ell_2$ adversaries. PGD attack is used to attack the models.}
\includegraphics[width=1\textwidth]{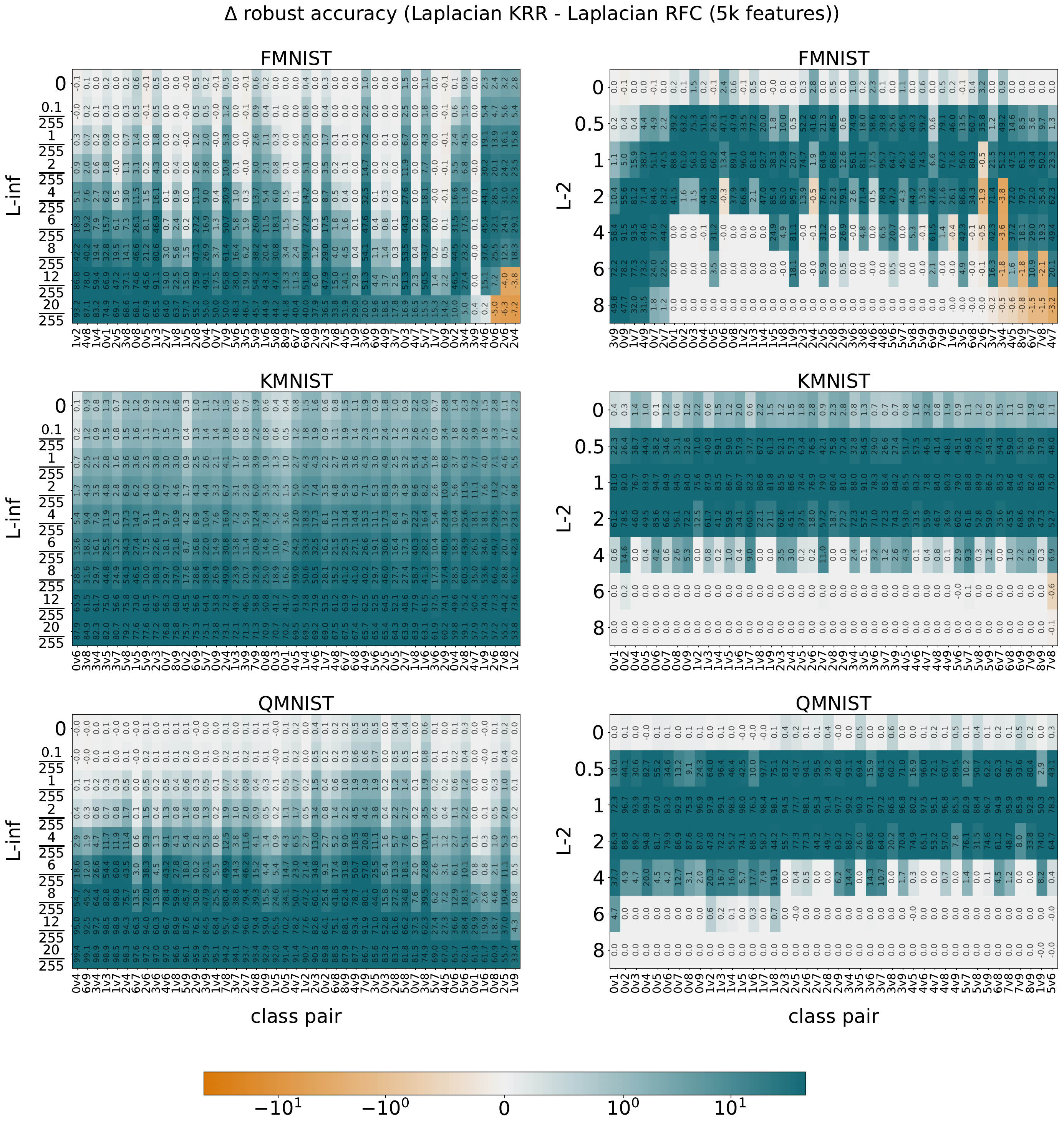}

\end{figure*}

\begin{figure*}[!htbp]
\centering
    \caption{\label{fig:gap_heatmaps_rff20kvsker} \textbf{Laplacian KRR vs Laplacian RFC (20k features):} $\ell_2$ and $\ell_\infty$ perturbation for FashionMNIST, KMNIST and QMNIST. (Top) row corresponds to the results for FashionMNIST dataset, (Middle) row corresponds to the KMNIST dataset, and (Bottom) row corresponds to the QMNIST dataset. Of these, (First) column corresponds to the robust accuracies for $\ell_\infty$ adversaries while (Second) column portrays the robust accuracies for $\ell_2$ adversaries. PGD attack is used to attack the models.}
\includegraphics[width=1\textwidth]{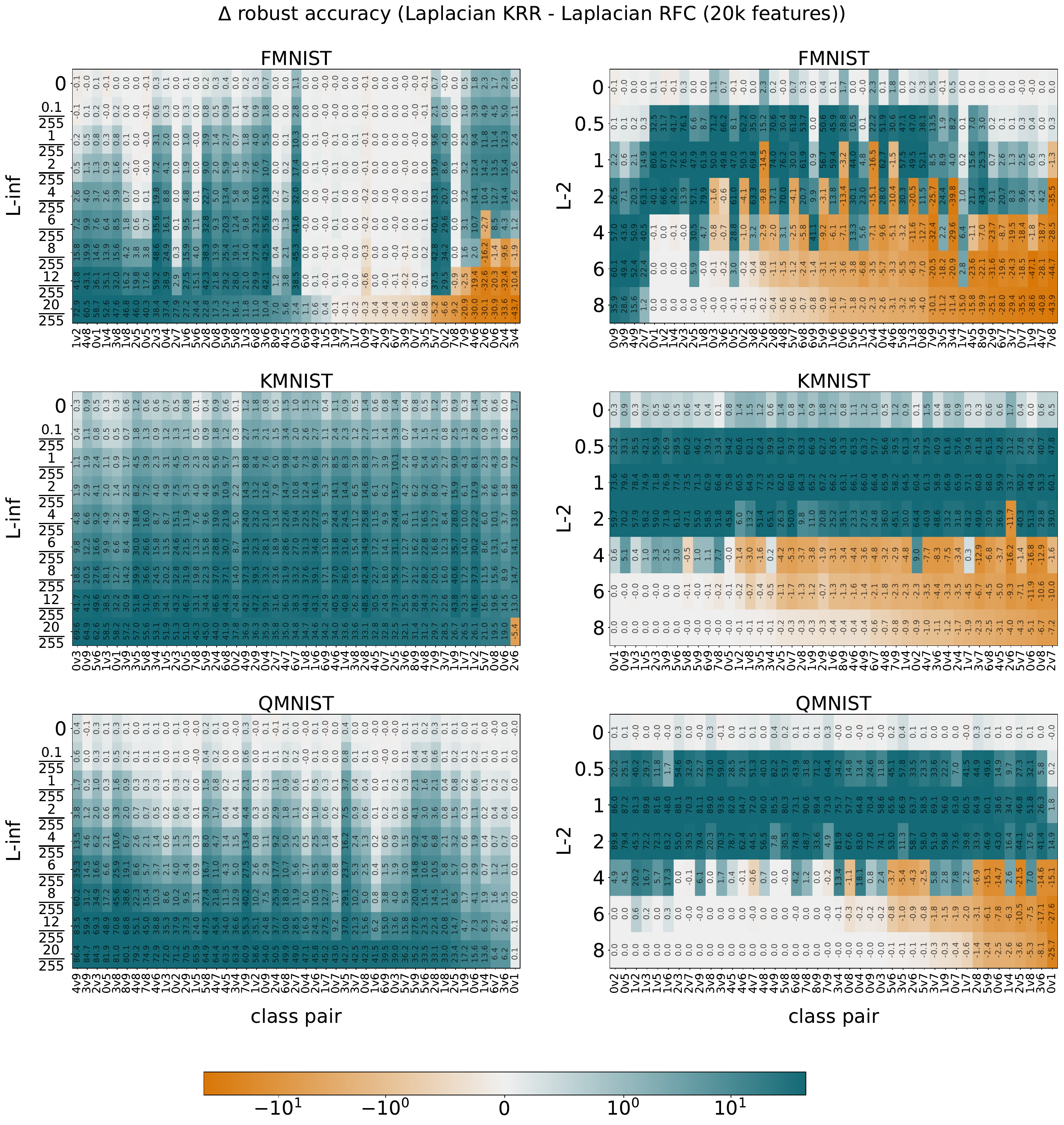}

\end{figure*}

\begin{figure*}[!htbp]
\centering
    \caption{\label{fig:gap_heatmaps_rff_5k_vsrfm} \textbf{Laplacian RFM vs Laplacian RFC (5k features):} $\ell_2$ and $\ell_\infty$ perturbation for FashionMNIST, KMNIST and QMNIST. (Top) row corresponds to the results for FashionMNIST dataset, (Middle) row corresponds to the KMNIST dataset, and (Bottom) row corresponds to the QMNIST dataset. Of these, (First) column corresponds to the robust accuracies for $\ell_\infty$ adversaries while (Second) column portrays the robust accuracies for $\ell_2$ adversaries. PGD attack is used to attack the models.}
\includegraphics[width=1\textwidth]{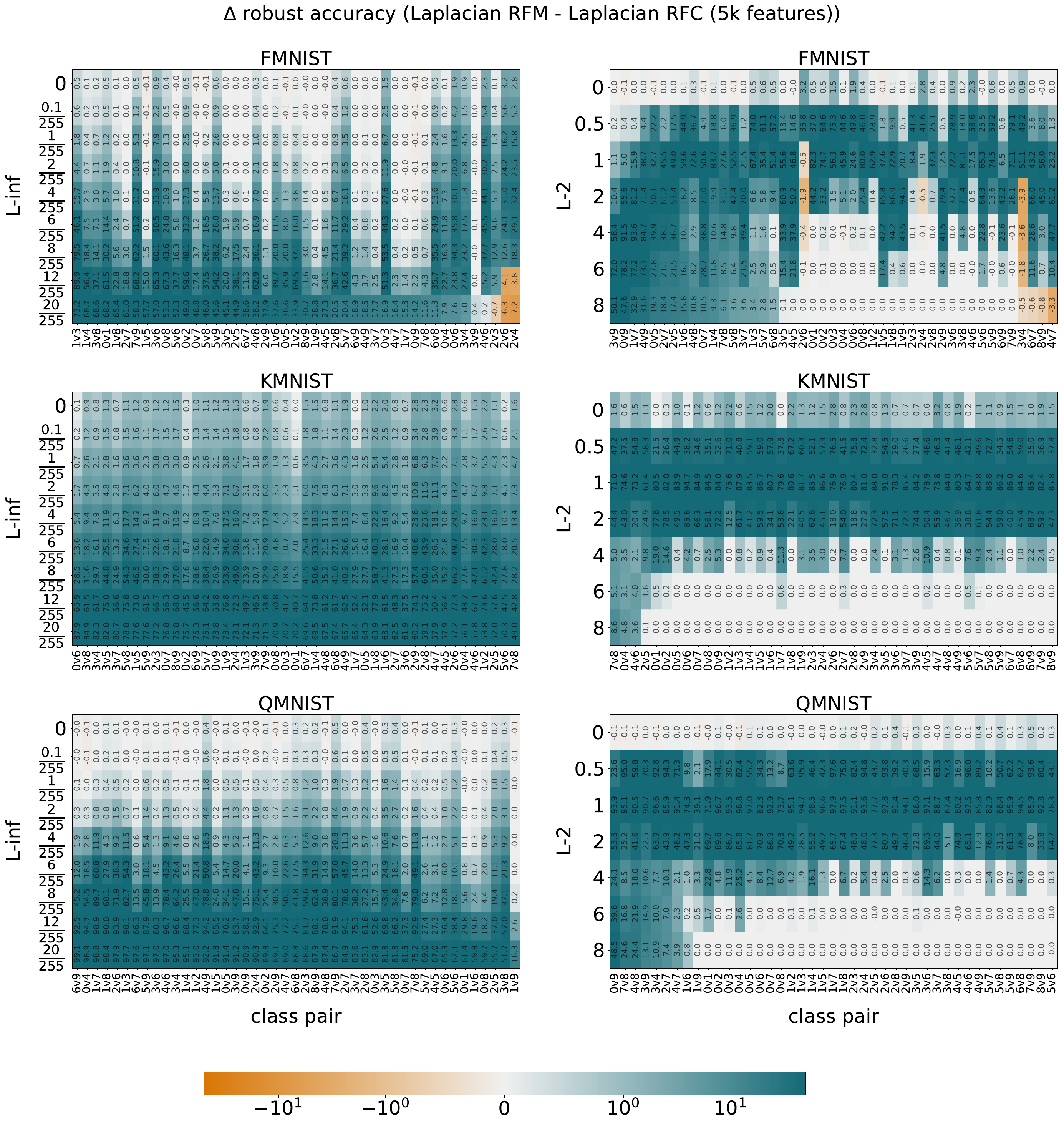}
\end{figure*}

\begin{figure*}[!htbp]
\centering
    \caption{\label{fig:gap_heatmaps_rff_20k_vsrfm} \textbf{Laplacian RFM vs Laplacian RFC (20k features):} $\ell_2$ and $\ell_\infty$ perturbation for FashionMNIST, KMNIST and QMNIST. (Top) row corresponds to the results for FashionMNIST dataset, (Middle) row corresponds to the KMNIST dataset, and (Bottom) row corresponds to the QMNIST dataset. Of these, (First) column corresponds to the robust accuracies for $\ell_\infty$ adversaries while (Second) column portrays the robust accuracies for $\ell_2$ adversaries. PGD attack is used to attack the models.}
\includegraphics[width=1\textwidth]{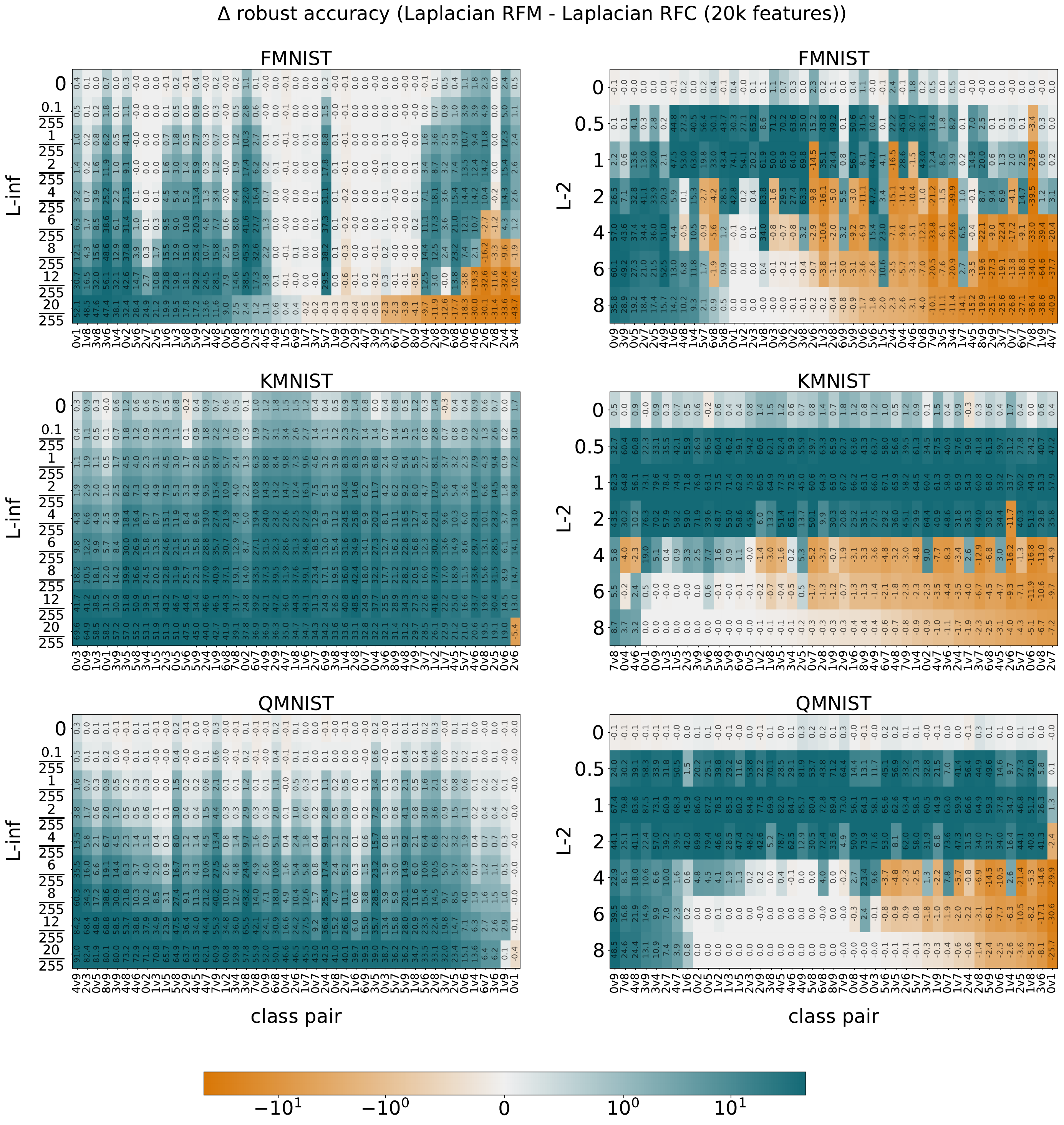}
\end{figure*}

\begin{figure*}[!htbp]
\centering
    \caption{\label{fig:gap_heatmaps_rff5k_vssvm} \textbf{Laplacian SVM vs Laplacian RFC (5k features):} $\ell_2$ and $\ell_\infty$ perturbation for FashionMNIST, KMNIST and QMNIST. (Top) row corresponds to the results for FashionMNIST dataset, (Middle) row corresponds to the KMNIST dataset, and (Bottom) row corresponds to the QMNIST dataset. Of these, (First) column corresponds to the robust accuracies for $\ell_\infty$ adversaries while (Second) column portrays the robust accuracies for $\ell_2$ adversaries. PGD attack is used to attack the models.}
\includegraphics[width=1\textwidth]{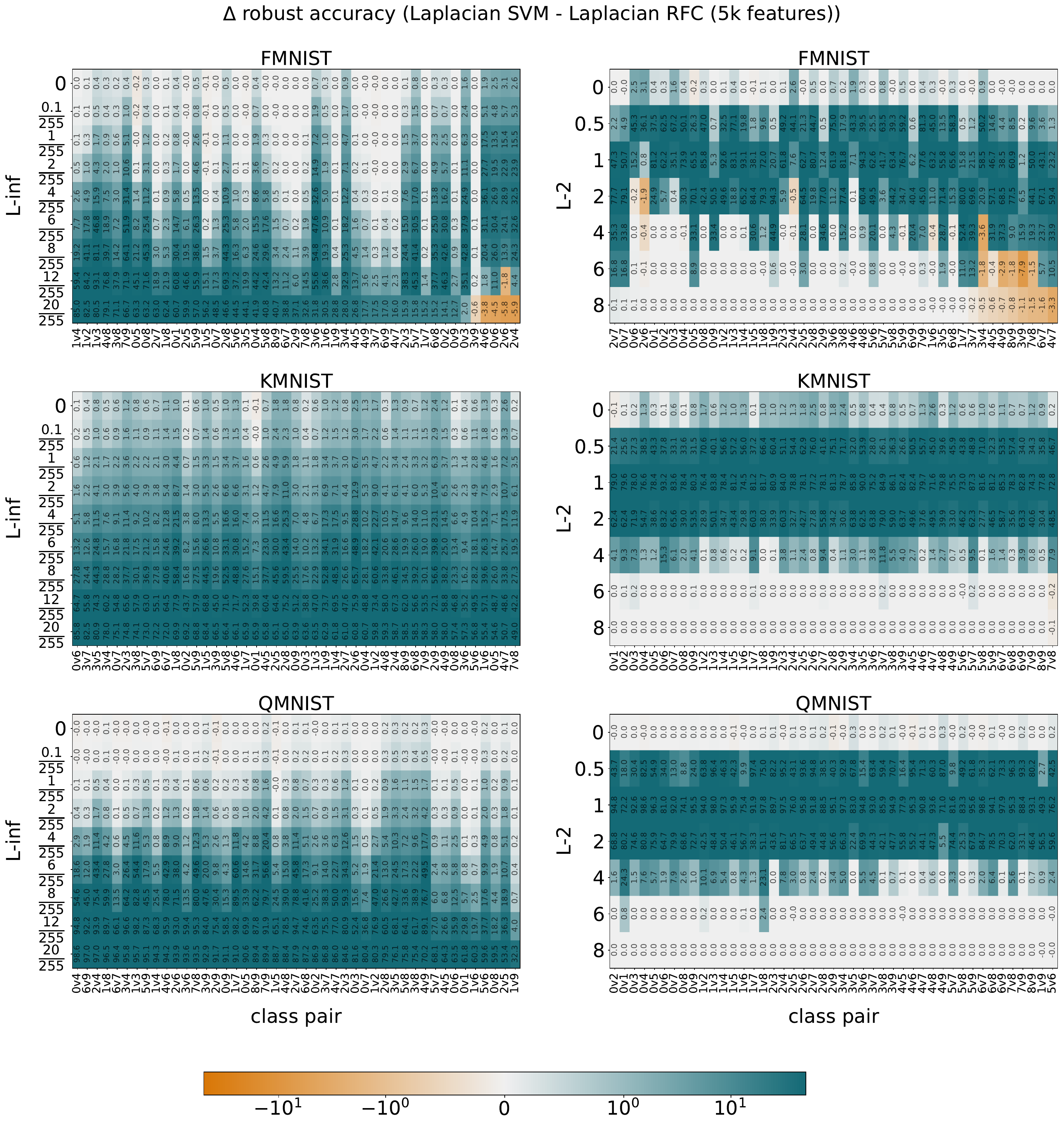}
\end{figure*}

\begin{figure*}[!htbp]
\centering
    \caption{\label{fig:gap_heatmaps_rff20k_vssvm} \textbf{Laplacian SVM vs Laplacian RFC (20k features):} $\ell_2$ and $\ell_\infty$ perturbation for FashionMNIST, KMNIST and QMNIST. (Top) row corresponds to the results for FashionMNIST dataset, (Middle) row corresponds to the KMNIST dataset, and (Bottom) row corresponds to the QMNIST dataset. Of these, (First) column corresponds to the robust accuracies for $\ell_\infty$ adversaries while (Second) column portrays the robust accuracies for $\ell_2$ adversaries. PGD attack is used to attack the models.}
\includegraphics[width=1\textwidth]{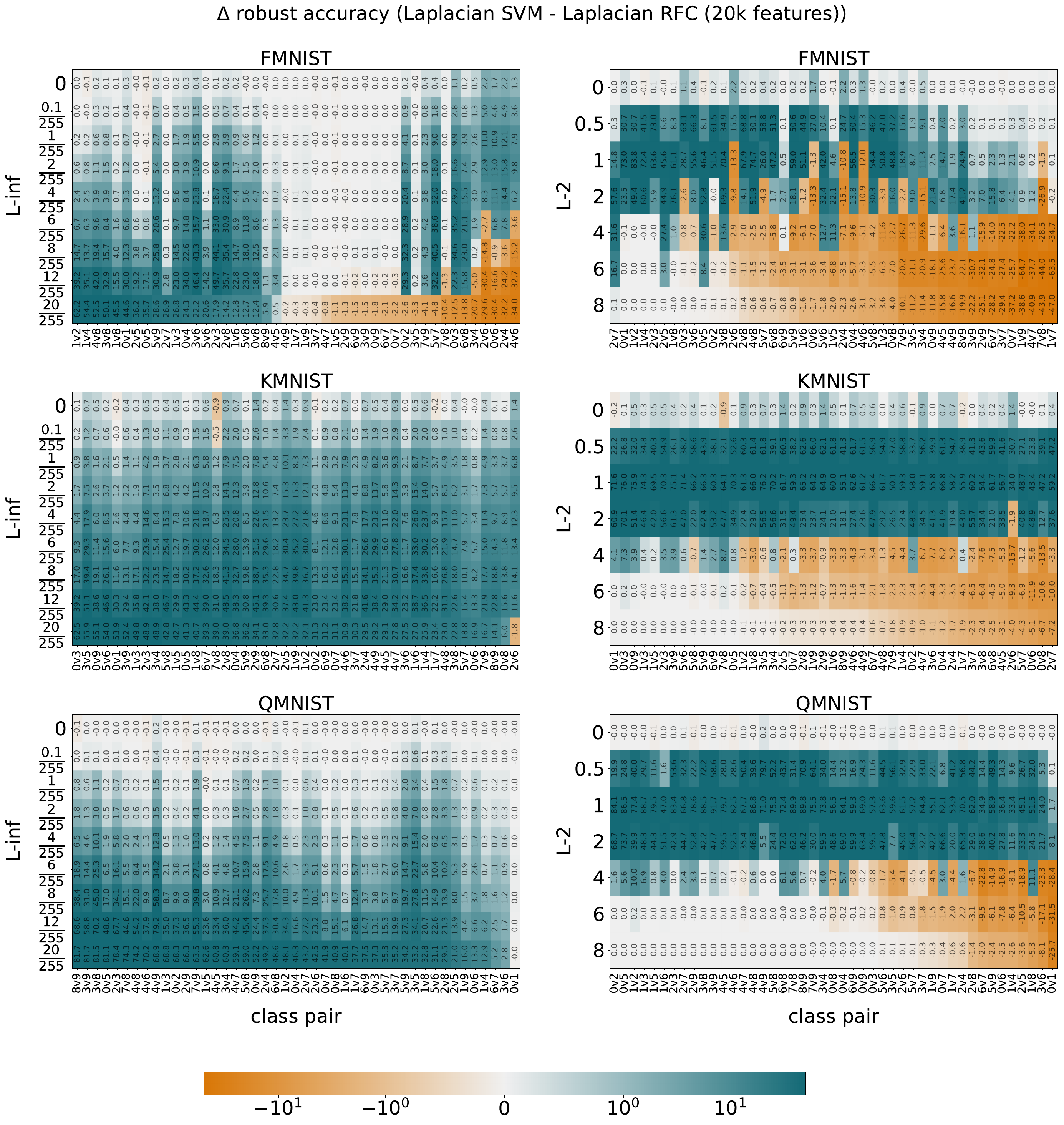}
\end{figure*}

\begin{figure*}[!htbp]
    \centering
    \caption{\label{fig:gap_heatmaps_laprff_5k_vslenet} \textbf{Laplacian RFC (5k features) versus LeNet:} $\ell_2$ and $\ell_\infty$ perturbation for FashionMNIST, KMNIST and QMNIST. (Top) row corresponds to the results for FashionMNIST dataset, (Middle) row corresponds to the KMNIST dataset, and (Bottom) row corresponds to the QMNIST dataset. Of these, (First) column corresponds to the robust accuracies for $\ell_\infty$ adversaries while (Second) column portrays the robust accuracies for $\ell_2$ adversaries. PGD attack is used to attack the models.}
\includegraphics[width=1\textwidth]{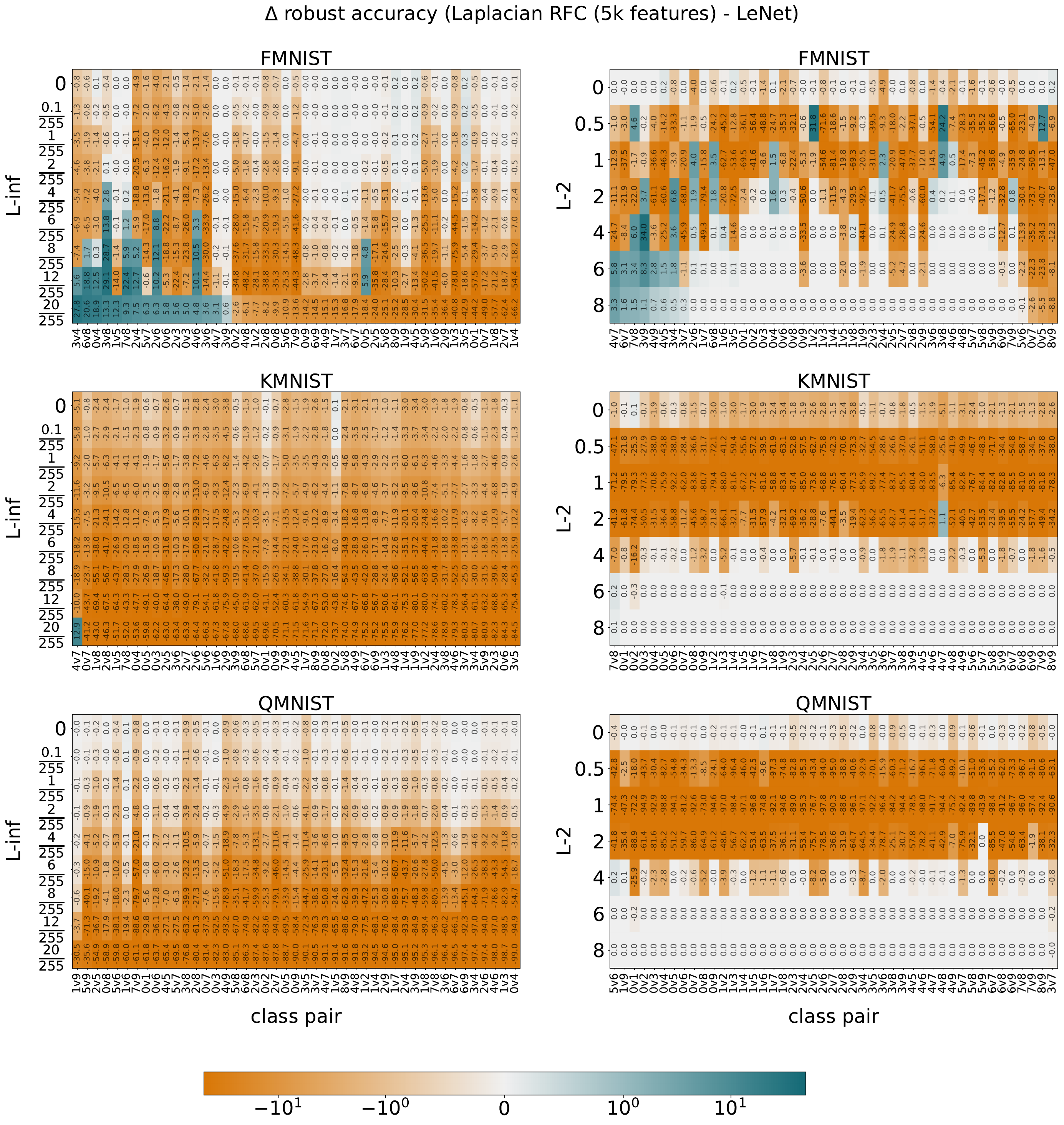}

\end{figure*}

\begin{figure*}[!htbp]
    \centering
    \caption{\label{fig:gap_heatmaps_laprff_20k_vslenet} \textbf{Laplacian RFC (20k features) versus LeNet:} $\ell_2$ and $\ell_\infty$ perturbation for FashionMNIST, KMNIST and QMNIST. (Top) row corresponds to the results for FashionMNIST dataset, (Middle) row corresponds to the KMNIST dataset, and (Bottom) row corresponds to the QMNIST dataset. Of these, (First) column corresponds to the robust accuracies for $\ell_\infty$ adversaries while (Second) column portrays the robust accuracies for $\ell_2$ adversaries. PGD attack is used to attack the models.}
\includegraphics[width=1\textwidth]{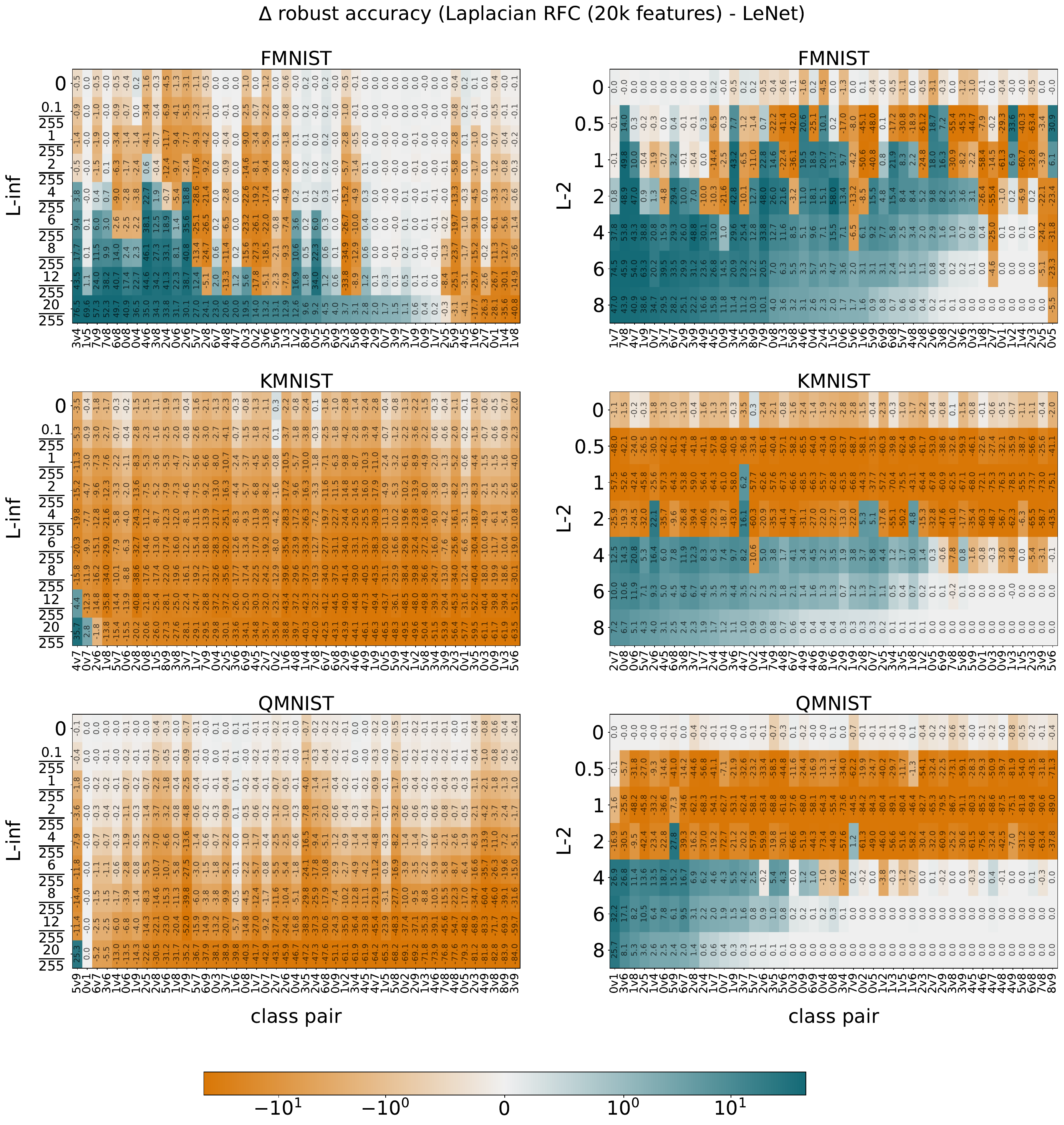}

\end{figure*}

\clearpage
\subsection{Gaussian Results for FMNIST, KMNIST, QMNIST}

Here we provide results comparing gaussian kernels in any type of model over the datasets FMNIST, KMNIST, QMNIST.

\vspace{2.5mm}

\begin{figure*}[!htbp]
	\centering
\caption{\label{fig:gap_heatmaps_gaussklr_vs_fcnn}\textbf{Gaussian KLR versus FCNN:} $\ell_2$ and $\ell_\infty$ perturbation for FashionMNIST, KMNIST and QMNIST. (Top) row corresponds to the results for FashionMNIST dataset, (Middle) row corresponds to the KMNIST dataset, and (Bottom) row corresponds to the QMNIST dataset. Of these, (First) column corresponds to the robust accuracies for $\ell_\infty$ adversaries while (Second) column portrays the robust accuracies for $\ell_2$ adversaries. PGD attack is used to attack the models.}
	\includegraphics[width=1\textwidth]{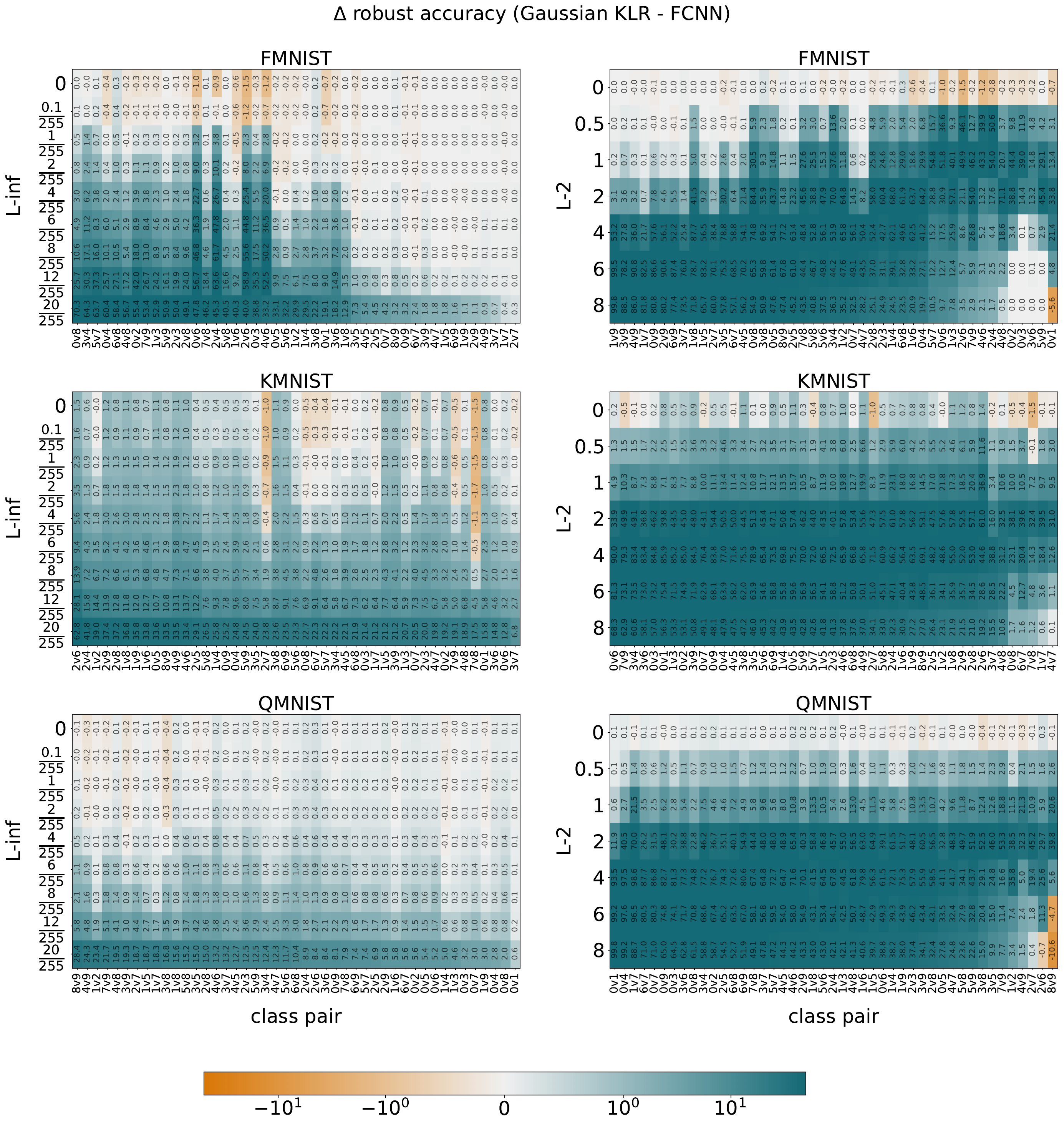}

\end{figure*}

\begin{figure*}[!htbp]
	\centering
\caption{\label{fig:gap_heatmaps_gauss_vs_fcnn}\textbf{Gaussian KRR versus FCNN:} $\ell_2$ and $\ell_\infty$ perturbation for FashionMNIST, KMNIST and QMNIST. (Top) row corresponds to the results for FashionMNIST dataset, (Middle) row corresponds to the KMNIST dataset, and (Bottom) row corresponds to the QMNIST dataset. Of these, (First) column corresponds to the robust accuracies for $\ell_\infty$ adversaries while (Second) column portrays the robust accuracies for $\ell_2$ adversaries. PGD attack is used to attack the models.}
	\includegraphics[width=1\textwidth]{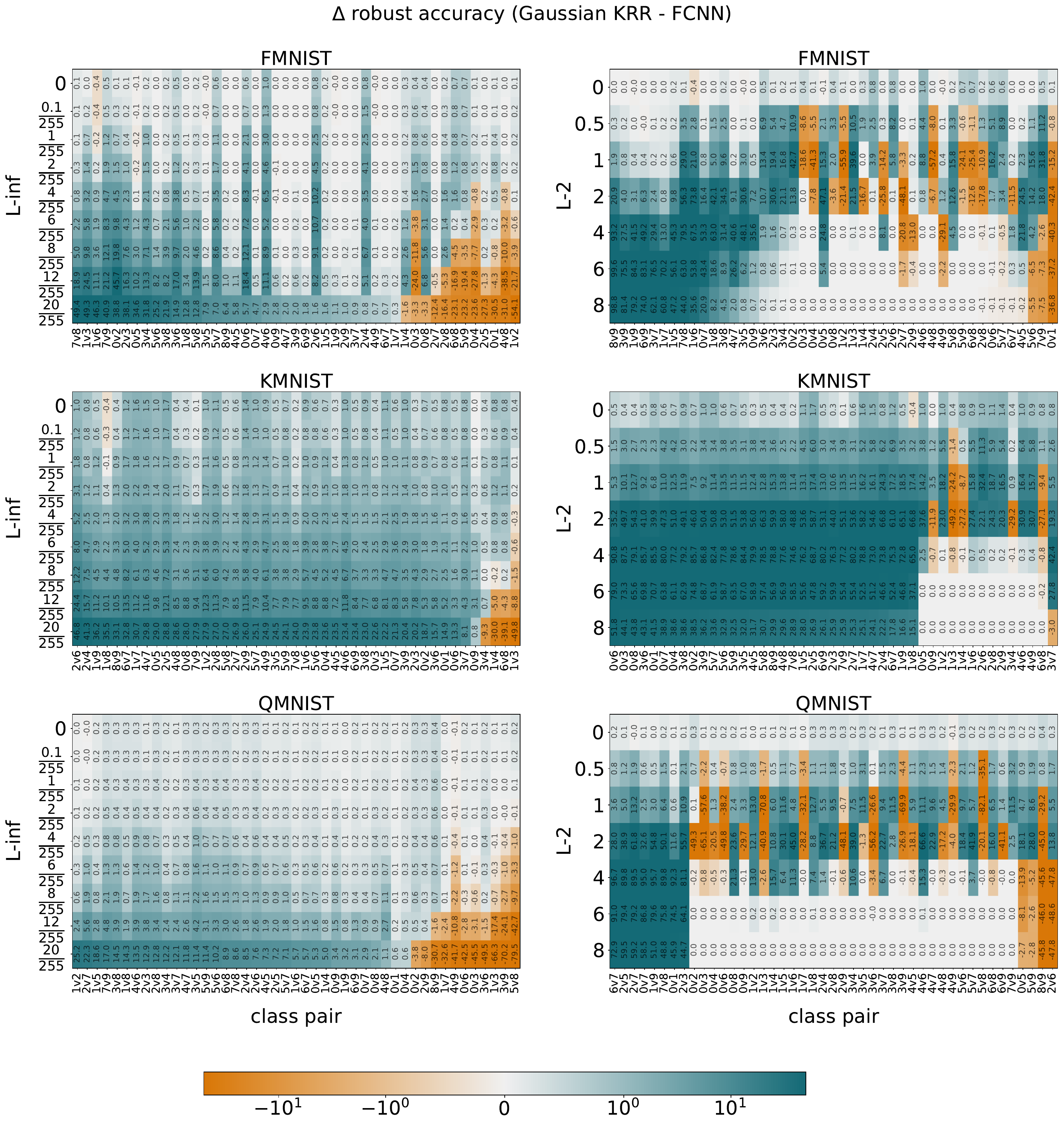}

\end{figure*}

\begin{figure*}[!htbp]
	\centering
    \caption{\label{fig:gap_heatmaps_gauss_rfmvsnn}\textbf{Gaussian RFM versus FCNN:} $\ell_2$ and $\ell_\infty$ perturbation for FashionMNIST, KMNIST and QMNIST. (Top) row corresponds to the results for FashionMNIST dataset, (Middle) row corresponds to the KMNIST dataset, and (Bottom) row corresponds to the QMNIST dataset. Of these, (First) column corresponds to the robust accuracies for $\ell_\infty$ adversaries while (Second) column portrays the robust accuracies for $\ell_2$ adversaries. PGD attack is used to attack the models.}
	\includegraphics[width=1\textwidth]{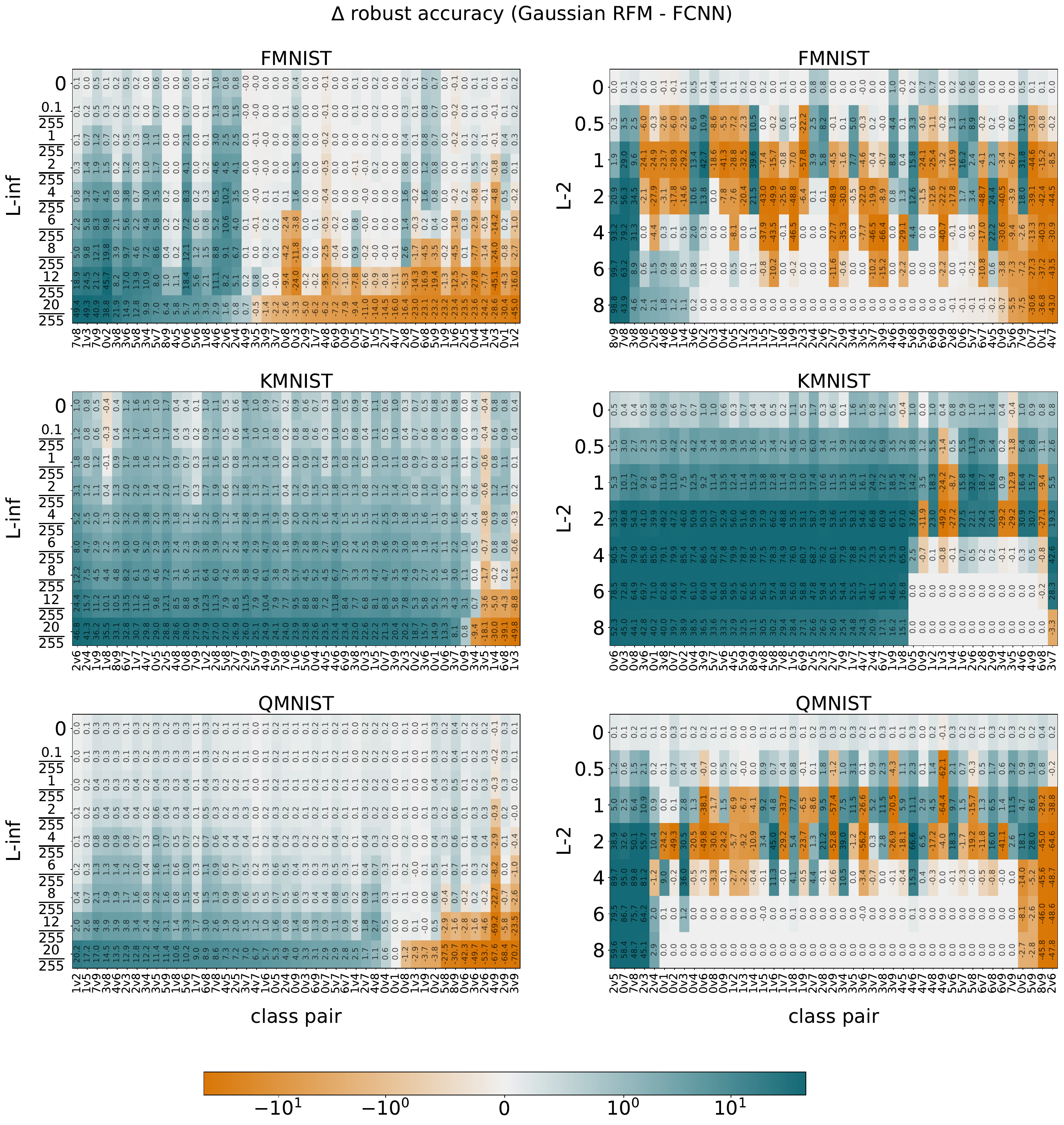}
\end{figure*}

\begin{figure*}[!htbp]
	\centering
    \caption{\label{fig:gap_heatmaps_gauss_svmvsfcnn}\textbf{Gaussian SVM versus FCNN:} $\ell_2$ and $\ell_\infty$ perturbation for FashionMNIST, KMNIST and QMNIST. (Top) row corresponds to the results for FashionMNIST dataset, (Middle) row corresponds to the KMNIST dataset, and (Bottom) row corresponds to the QMNIST dataset. Of these, (First) column corresponds to the robust accuracies for $\ell_\infty$ adversaries while (Second) column portrays the robust accuracies for $\ell_2$ adversaries. PGD attack is used to attack the models.}
	\includegraphics[width=1\textwidth]{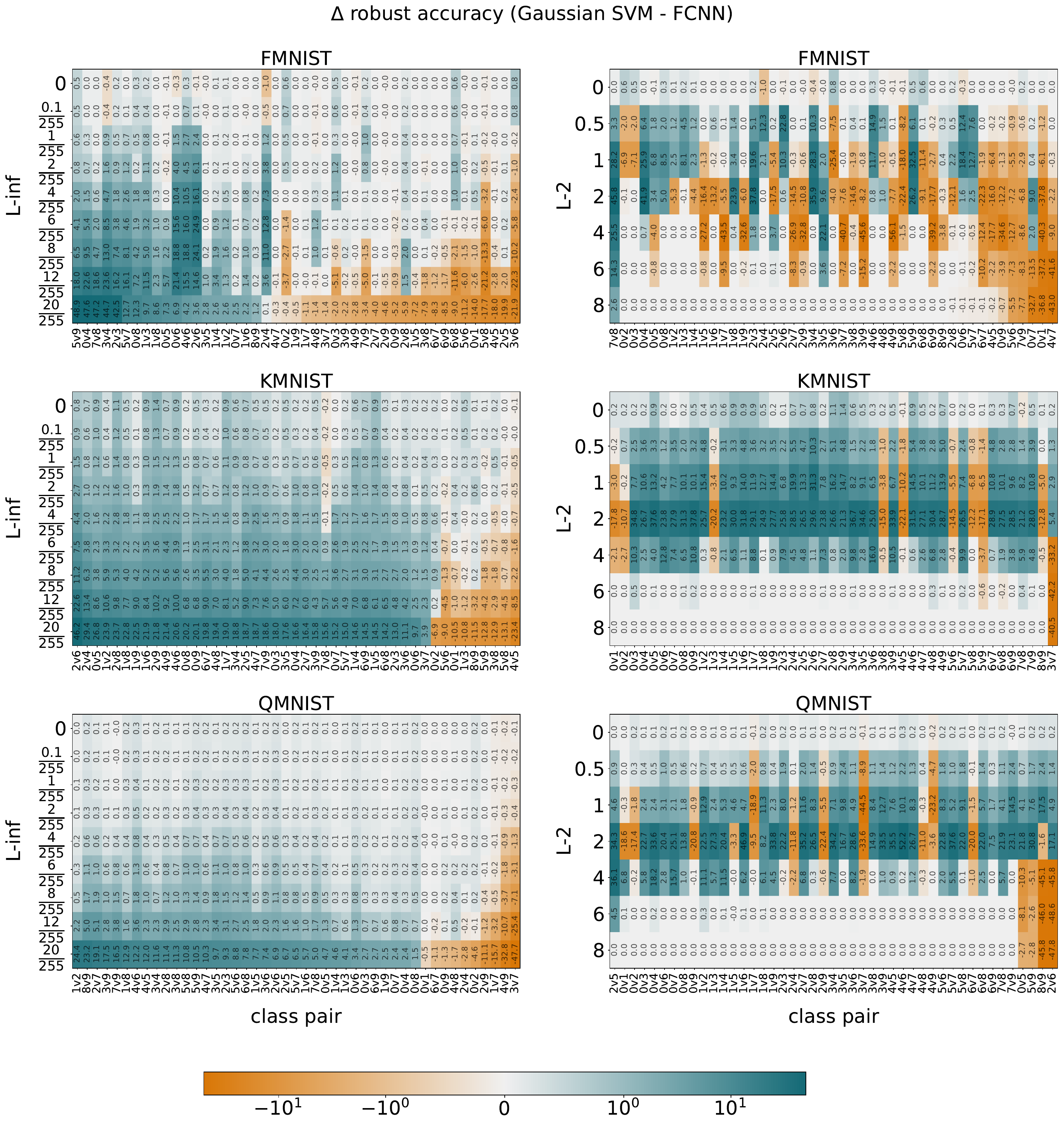}
\end{figure*}

\clearpage

\begin{figure*}[!htbp]
	\centering
    \caption{\label{fig:gap_heatmaps_gauss_klrvskrr}\textbf{Gaussian KLR versus Gaussian KRR:} $\ell_2$ and $\ell_\infty$ perturbation for FashionMNIST, KMNIST and QMNIST. (Top) row corresponds to the results for FashionMNIST dataset, (Middle) row corresponds to the KMNIST dataset, and (Bottom) row corresponds to the QMNIST dataset. Of these, (First) column corresponds to the robust accuracies for $\ell_\infty$ adversaries while (Second) column portrays the robust accuracies for $\ell_2$ adversaries. PGD attack is used to attack the models.}
	\includegraphics[width=1\textwidth]{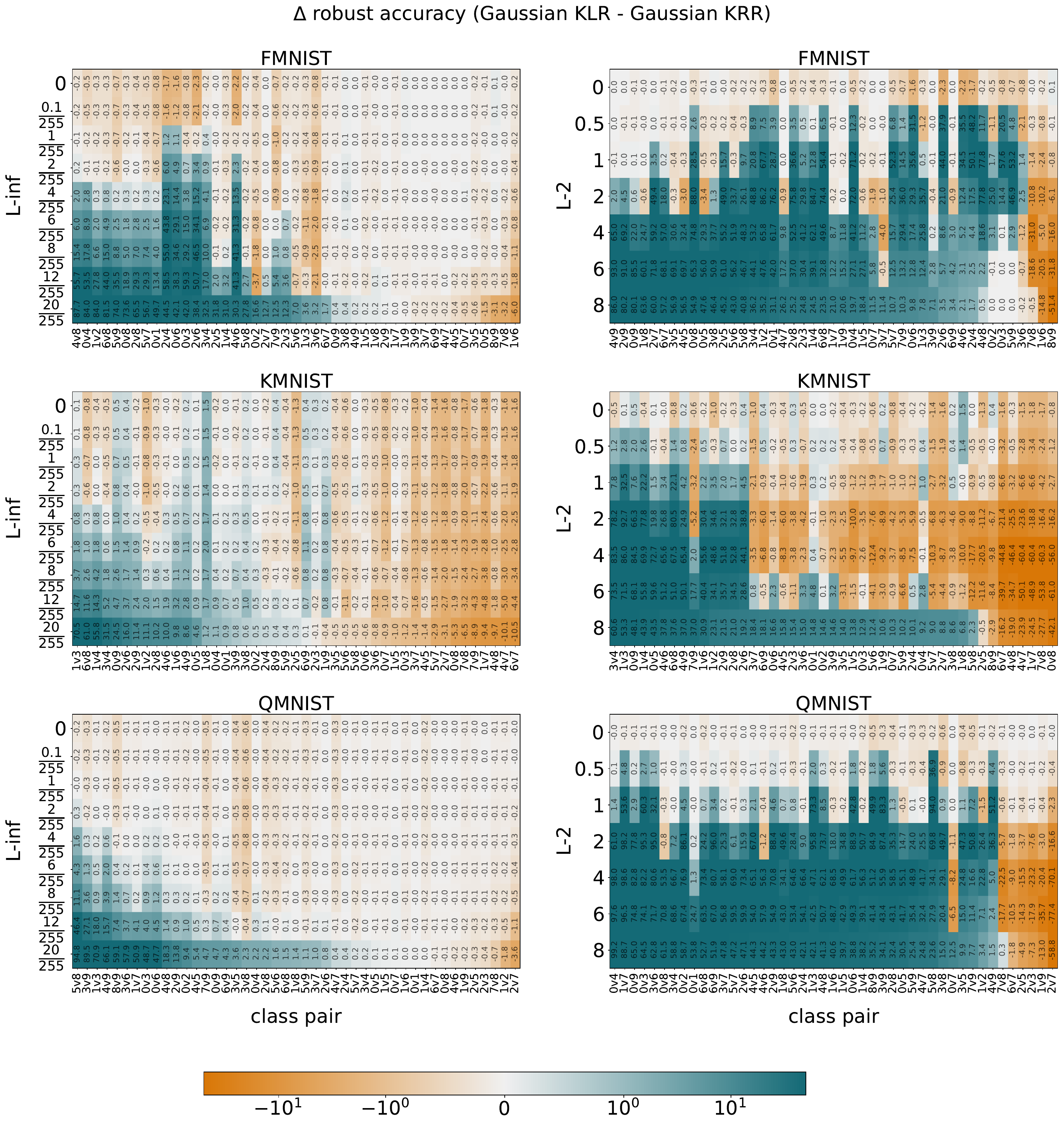}

\end{figure*}

\begin{figure*}[!htbp]
	\centering
    \caption{\label{fig:gap_heatmaps_gauss_klrvsrfm}\textbf{Gaussian KLR versus Gaussian RFM:} $\ell_2$ and $\ell_\infty$ perturbation for FashionMNIST, KMNIST and QMNIST. (Top) row corresponds to the results for FashionMNIST dataset, (Middle) row corresponds to the KMNIST dataset, and (Bottom) row corresponds to the QMNIST dataset. Of these, (First) column corresponds to the robust accuracies for $\ell_\infty$ adversaries while (Second) column portrays the robust accuracies for $\ell_2$ adversaries. PGD attack is used to attack the models.}
	\includegraphics[width=1\textwidth]{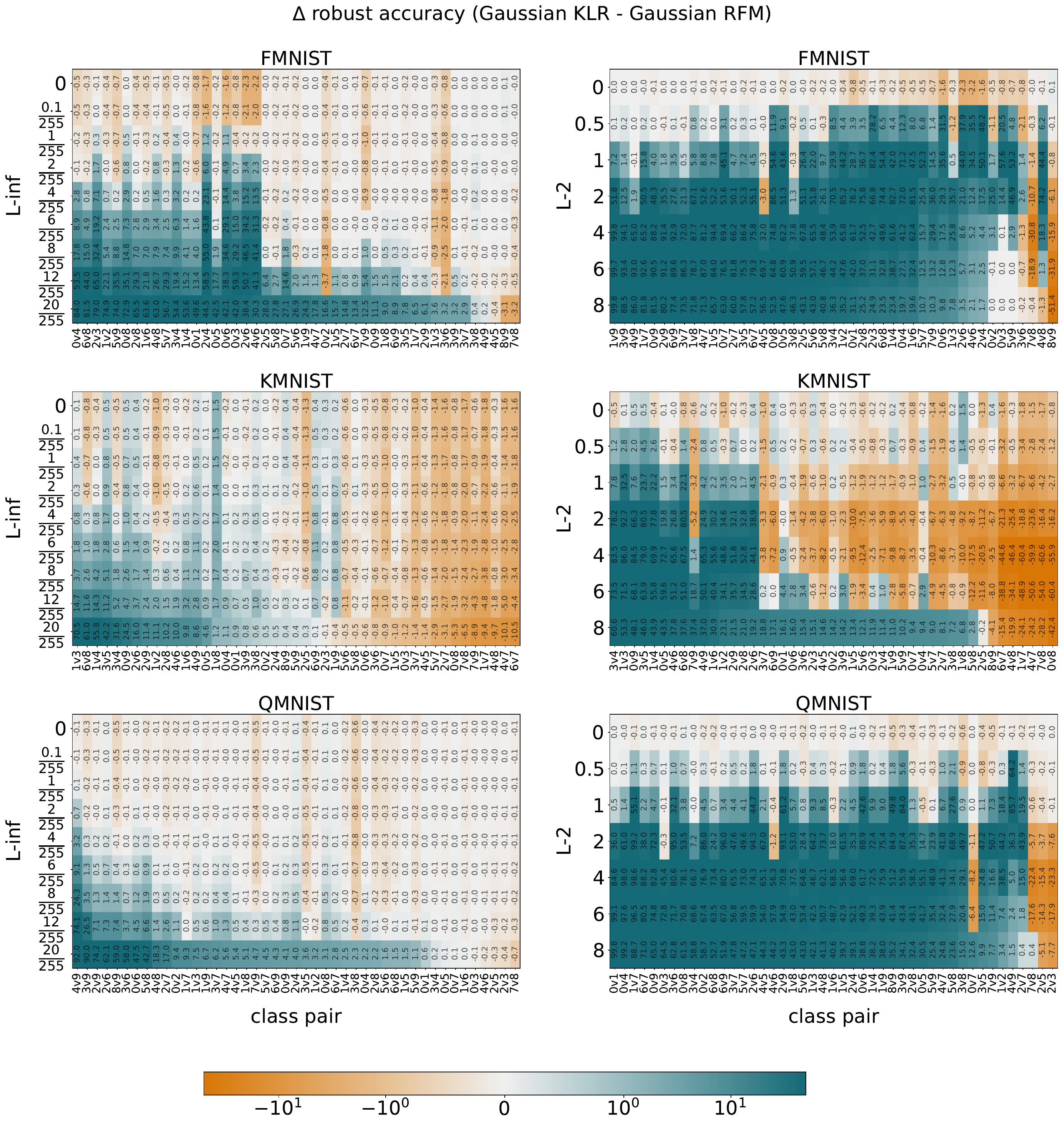}

\end{figure*}

\begin{figure*}[!htbp]
	\centering
    \caption{\label{fig:gap_heatmaps_gauss_klrvssvm}\textbf{Gaussian KLR versus Gaussian SVM:} $\ell_2$ and $\ell_\infty$ perturbation for FashionMNIST, KMNIST and QMNIST. (Top) row corresponds to the results for FashionMNIST dataset, (Middle) row corresponds to the KMNIST dataset, and (Bottom) row corresponds to the QMNIST dataset. Of these, (First) column corresponds to the robust accuracies for $\ell_\infty$ adversaries while (Second) column portrays the robust accuracies for $\ell_2$ adversaries. PGD attack is used to attack the models.}
	\includegraphics[width=1\textwidth]{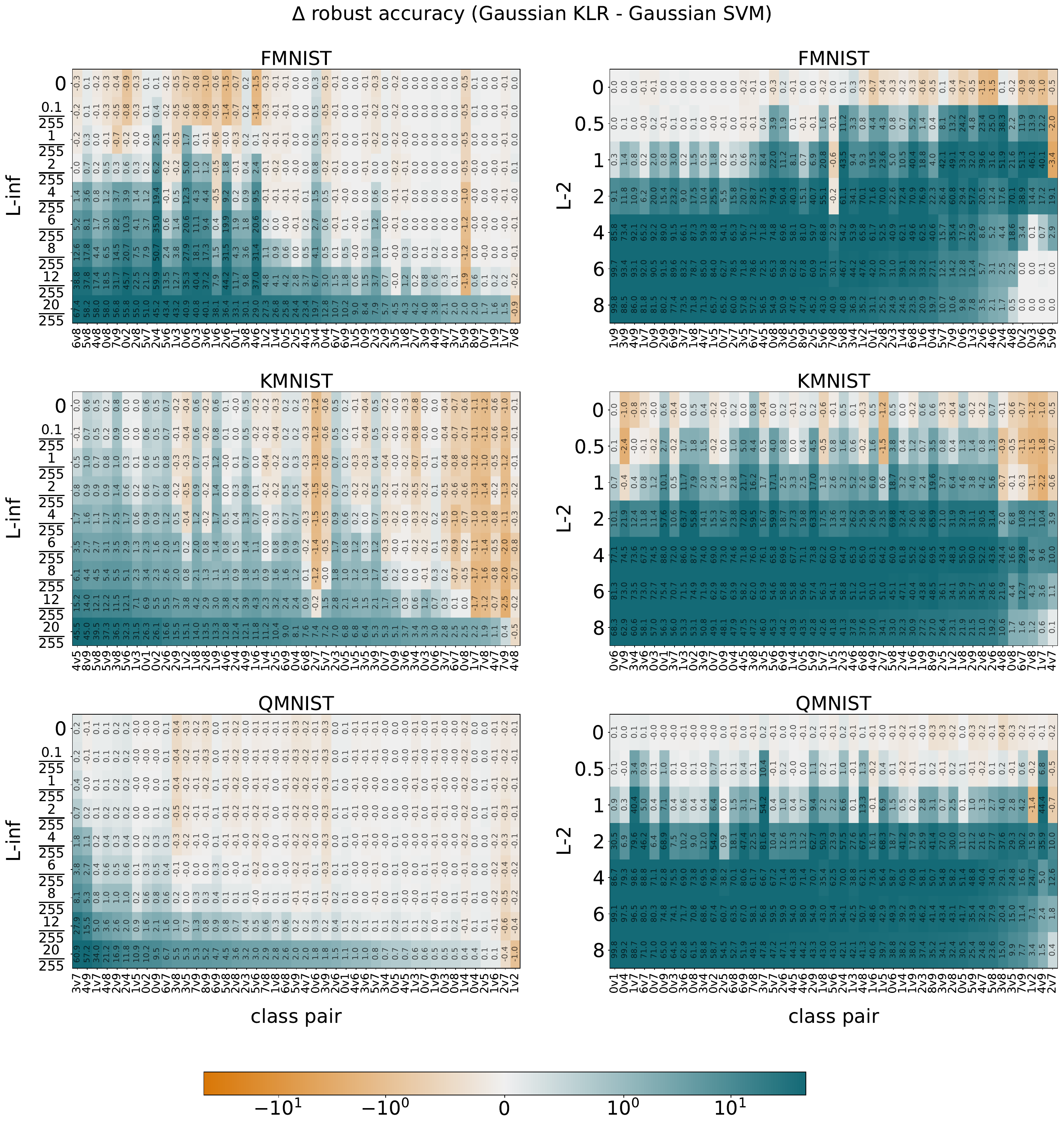}

\end{figure*}

\begin{figure*}[!htbp]
	\centering
    \caption{\label{fig:gap_heatmaps_gauss_rfmvsker}\textbf{Gaussian KRR versus Gaussian RFM:} $\ell_2$ and $\ell_\infty$ perturbation for FashionMNIST, KMNIST and QMNIST. (Top) row corresponds to the results for FashionMNIST dataset, (Middle) row corresponds to the KMNIST dataset, and (Bottom) row corresponds to the QMNIST dataset. Of these, (First) column corresponds to the robust accuracies for $\ell_\infty$ adversaries while (Second) column portrays the robust accuracies for $\ell_2$ adversaries. PGD attack is used to attack the models.}
	\includegraphics[width=1\textwidth]{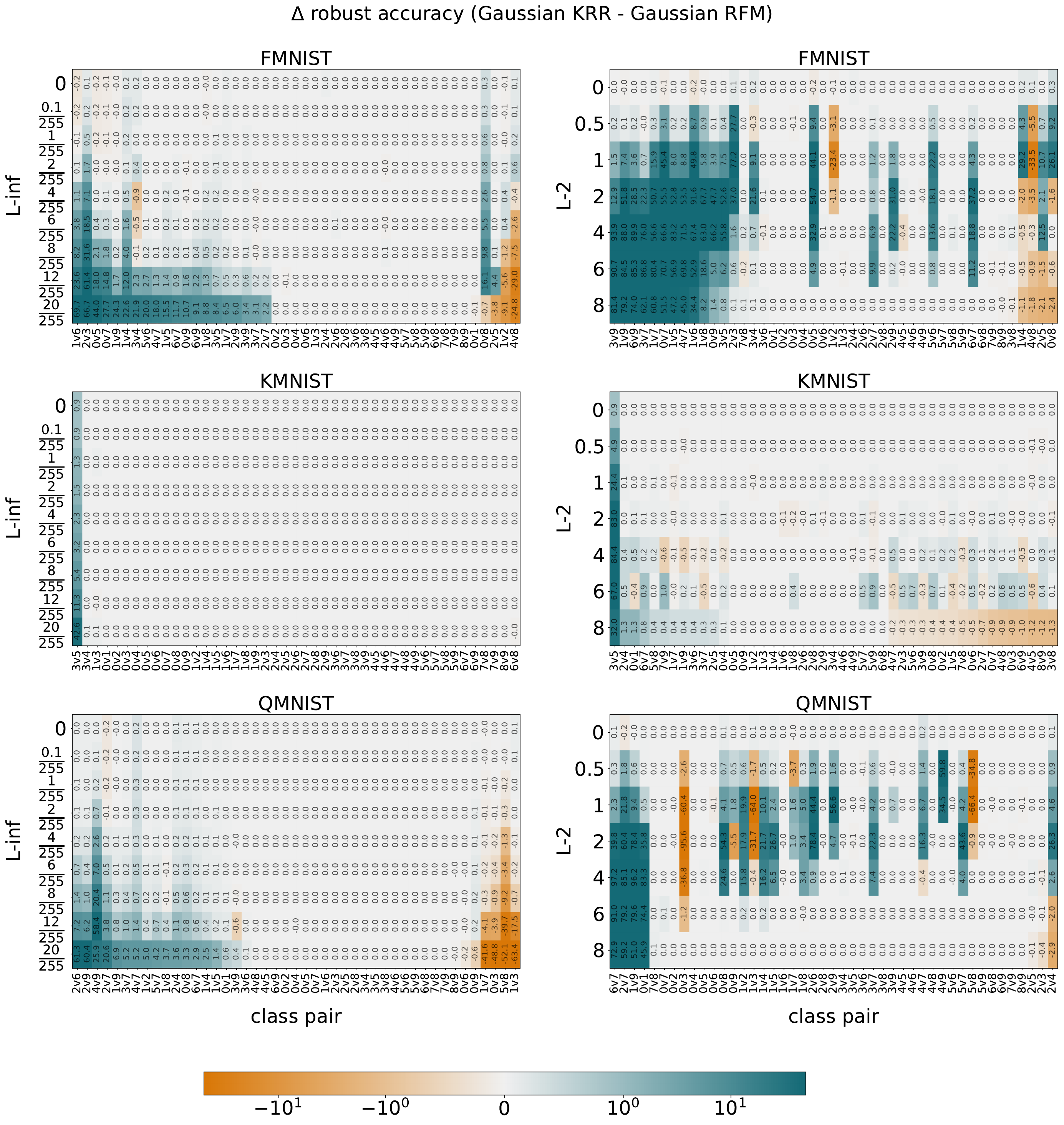}

\end{figure*}

\begin{figure*}[!htbp]
	\centering
    \caption{\label{fig:gap_heatmaps_gauss_kervssvm}\textbf{Gaussian KRR versus Gaussian SVM:} $\ell_2$ and $\ell_\infty$ perturbation for FashionMNIST, KMNIST and QMNIST. (Top) row corresponds to the results for FashionMNIST dataset, (Middle) row corresponds to the KMNIST dataset, and (Bottom) row corresponds to the QMNIST dataset. Of these, (First) column corresponds to the robust accuracies for $\ell_\infty$ adversaries while (Second) column portrays the robust accuracies for $\ell_2$ adversaries. PGD attack is used to attack the models.}
	\includegraphics[width=1\textwidth]{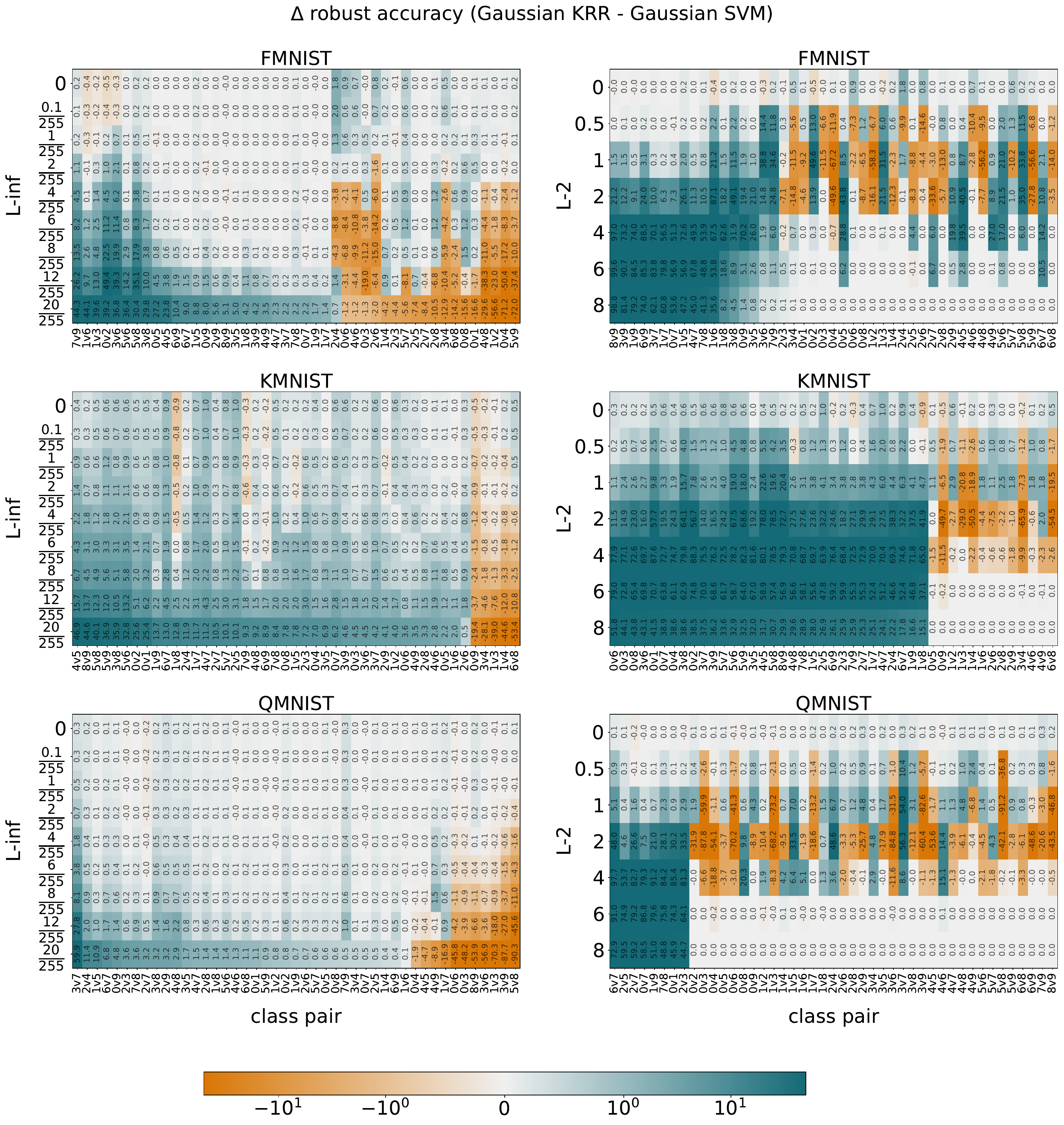}

\end{figure*}

\begin{figure*}[!htbp]
	\centering
    \caption{\label{fig:gap_heatmaps_gauss_rfmvssvm}\textbf{Gaussian RFM versus Gaussian SVM:} $\ell_2$ and $\ell_\infty$ perturbation for FashionMNIST, KMNIST and QMNIST. (Top) row corresponds to the results for FashionMNIST dataset, (Middle) row corresponds to the KMNIST dataset, and (Bottom) row corresponds to the QMNIST dataset. Of these, (First) column corresponds to the robust accuracies for $\ell_\infty$ adversaries while (Second) column portrays the robust accuracies for $\ell_2$ adversaries. PGD attack is used to attack the models.}
	\includegraphics[width=1\textwidth]{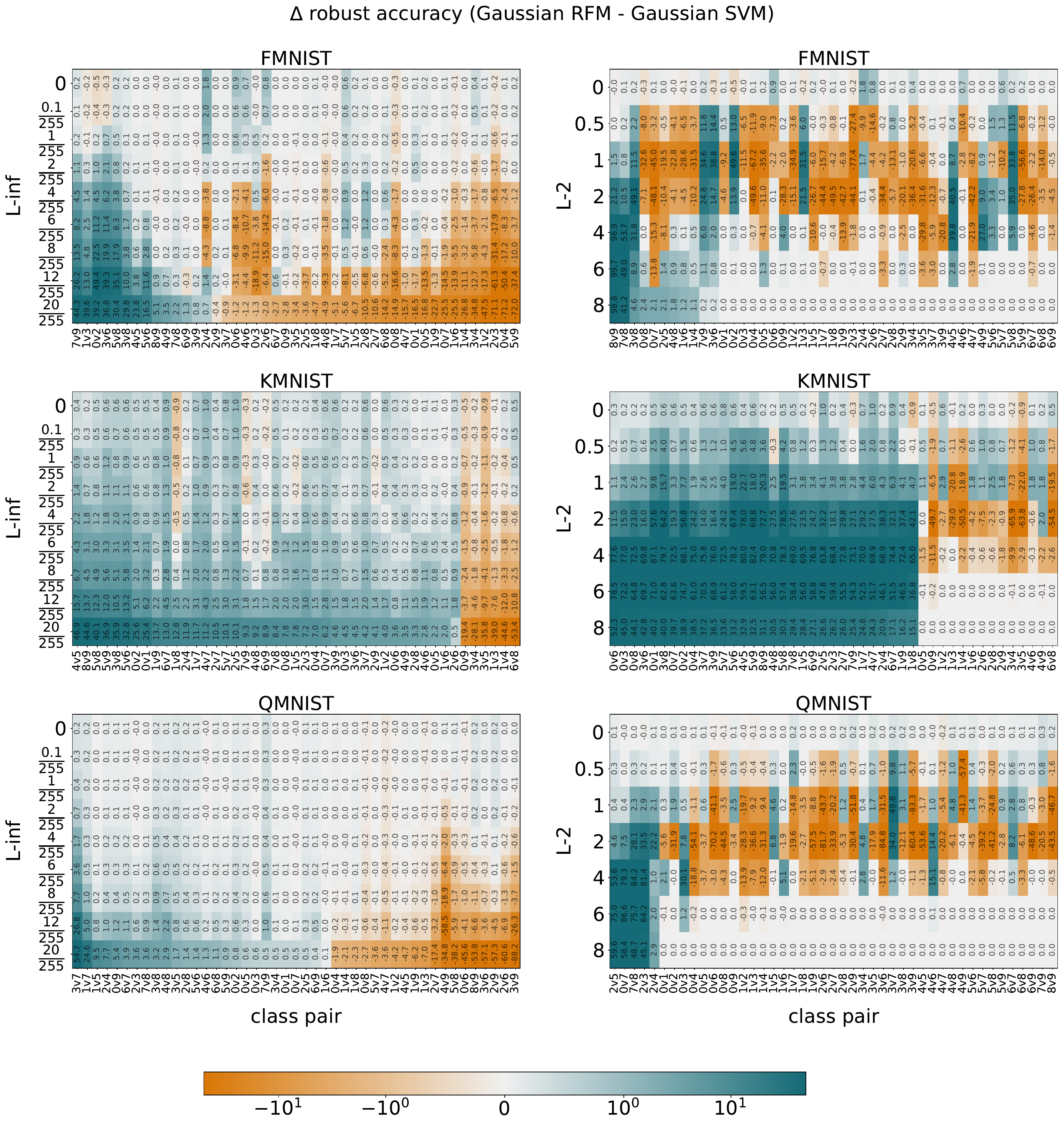}

\end{figure*}

\begin{figure*}[!htbp]
    \caption{ \label{fig:gap_rfc_summary_linf} 
    \textbf{Gap in robust accuracy (ra) for $\ell_\infty$ perturbation on 45 $1\text{v}1$ binary classification tasks from FashionMNIST} comparing 2 neural networks (columns) and 2 RFCs with different number of features (rows) with the Laplacian kernel. 
    Perturbation size increases as we go down each heatmap, and values correspond to \textrm{ra}(row model) - \textrm{ra}(col model).
    The \textcolor{teal}{teal} colour indicates that the RFC is more robust, whereas the \textcolor{Bittersweet}{orange} color indicates that the neural network is more robust. Each of the 4 heatmaps compares 45 pairs of models.
    For each heatmap, class pairs for binary classification are sorted according to values in the bottom row. 
    (Top) RFC with 5000 random features are a bit less robust compared to fully connected neural networks and convolutional neural networks. 
    (Bottom) RFC with 20000 random features are competing in robustness with fully connected neural networks and convolutional neural networks. Although, these results shouldn't be taken at face value as we see in~\cref{apd: num_of_ftr_vs_robust_results} that Laplace RFCs exhibit gradient masking. PGD attack is used to attack the models. \\
    }
\includegraphics[width=\textwidth,keepaspectratio]
{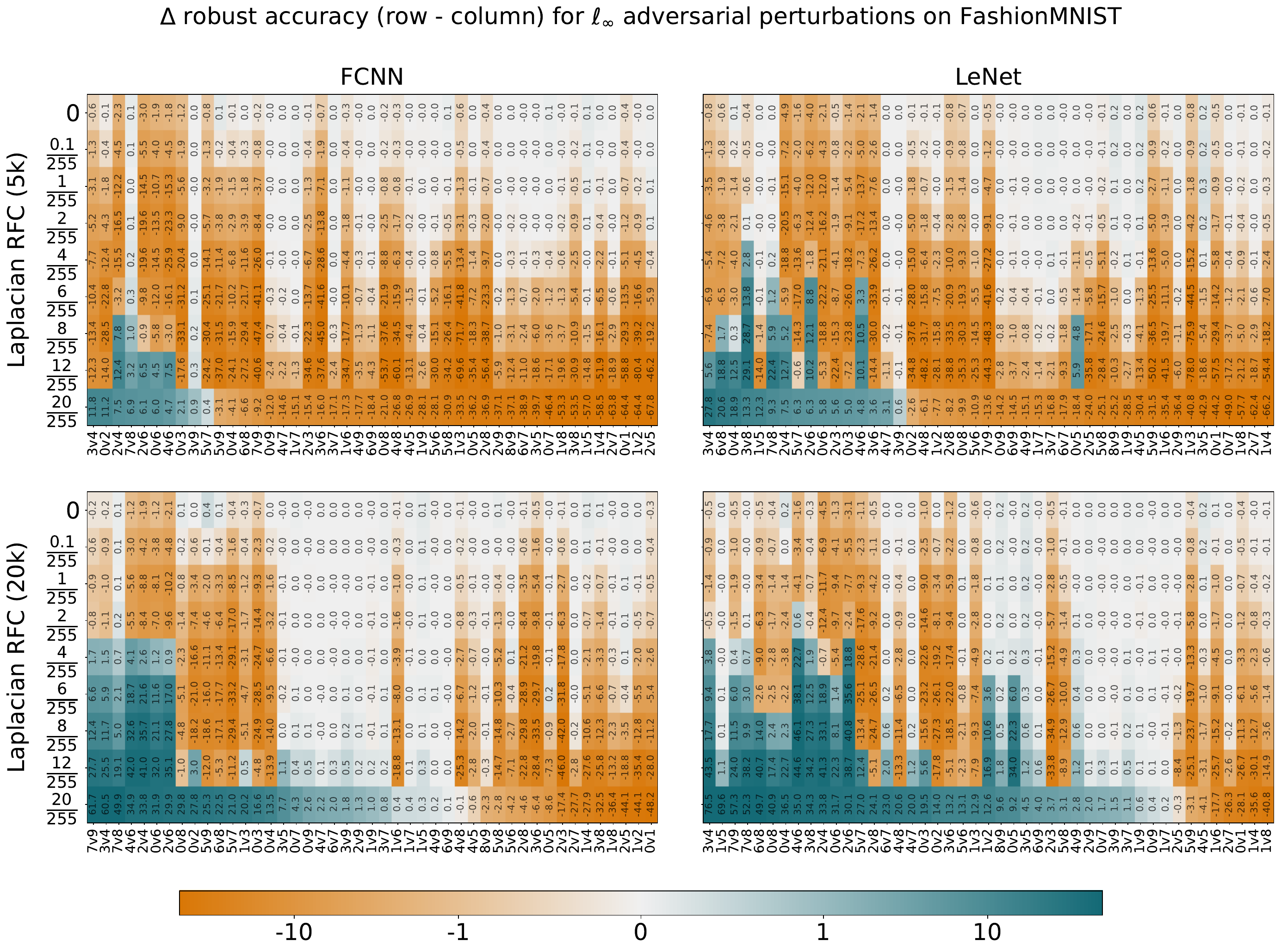}
\vspace{-6ex}
\end{figure*}

\begin{figure*}[!htbp] 
	\centering
    	\caption{\label{fig:gap_heatmaps_gauss_rff5kvsnn}	\textbf{Gaussian RFC (5k features) versus FCNN:} $\ell_2$ and $\ell_\infty$ perturbation for FashionMNIST, KMNIST and QMNIST. (Top) row corresponds to the results for FashionMNIST dataset, (Middle) row corresponds to the KMNIST dataset, and (Bottom) row corresponds to the QMNIST dataset. Of these, (First) column corresponds to the robust accuracies for $\ell_\infty$ adversaries while (Second) column portrays the robust accuracies for $\ell_2$ adversaries. PGD attack is used to attack the models.}
	\includegraphics[width=1\textwidth]{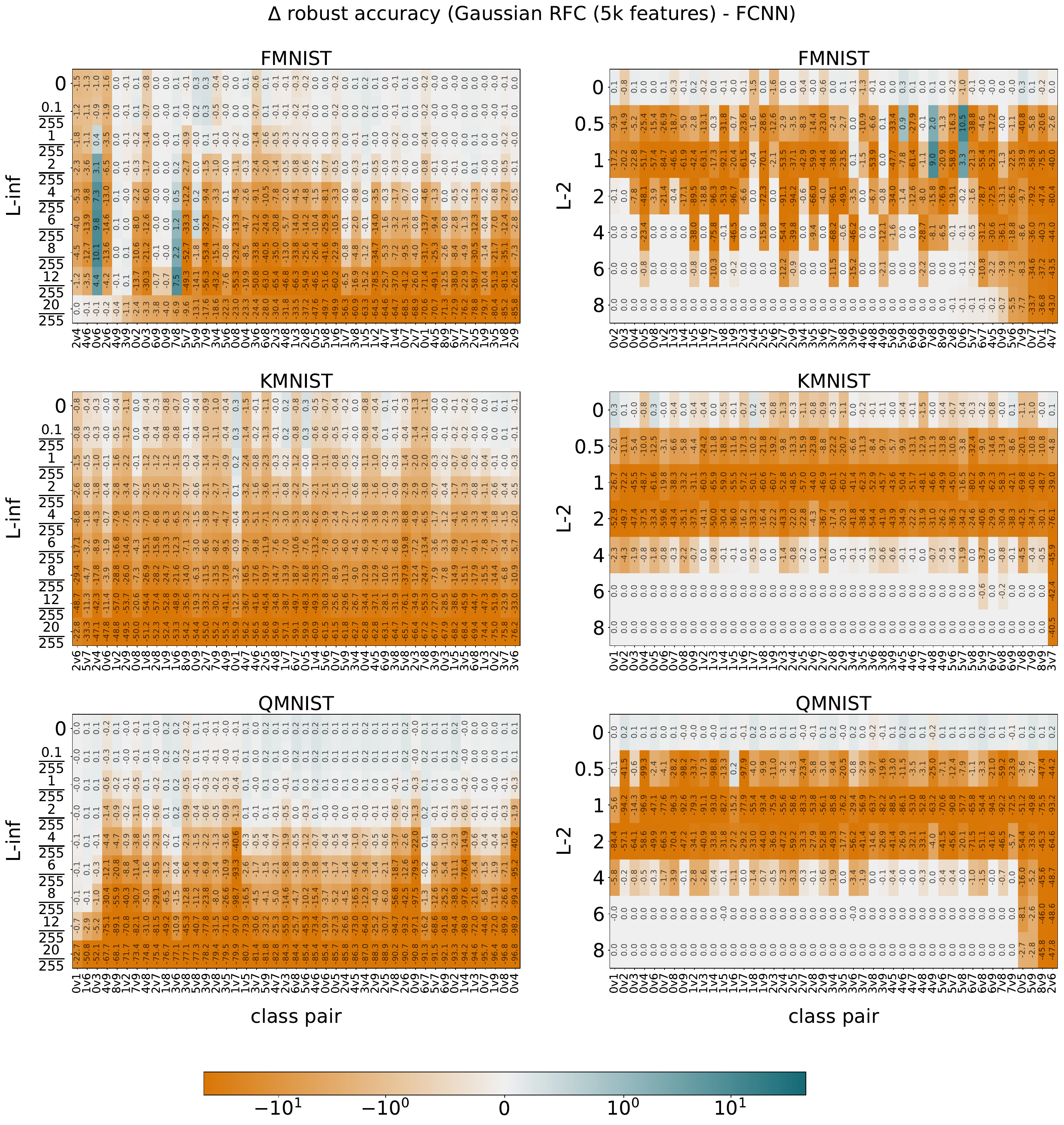}
\end{figure*}

\begin{figure*}[!htbp] 
	\centering
    	\caption{\label{fig:gap_heatmaps_gauss_rff20kvsnn}	\textbf{Gaussian RFC (20k features) versus FCNN:} $\ell_2$ and $\ell_\infty$ perturbation for FashionMNIST, KMNIST and QMNIST. (Top) row corresponds to the results for FashionMNIST dataset, (Middle) row corresponds to the KMNIST dataset, and (Bottom) row corresponds to the QMNIST dataset. Of these, (First) column corresponds to the robust accuracies for $\ell_\infty$ adversaries while (Second) column portrays the robust accuracies for $\ell_2$ adversaries. PGD attack is used to attack the models.}
	\includegraphics[width=1\textwidth]{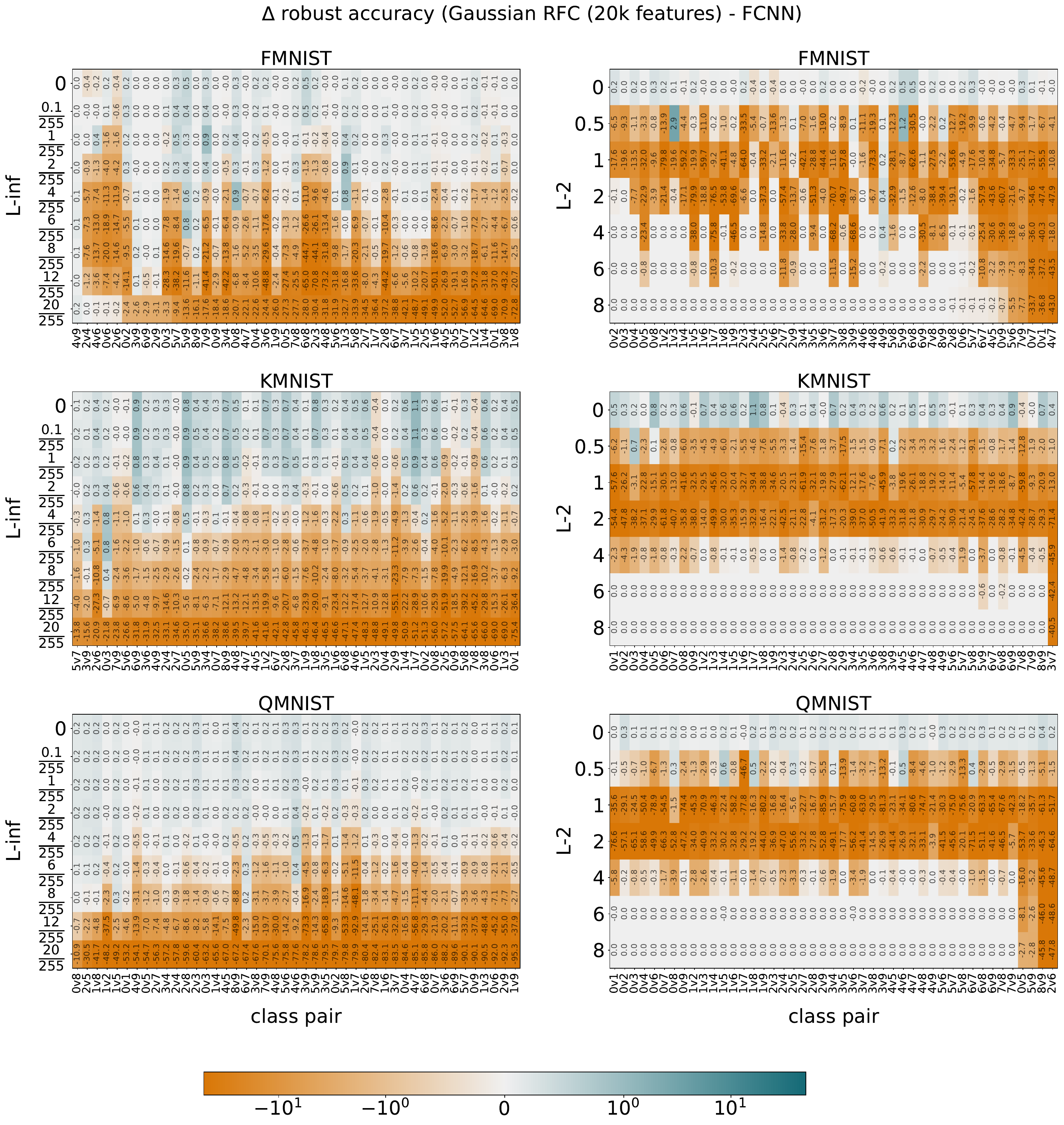}

\end{figure*}

\begin{figure*}[!htbp]
	\centering    \caption{\label{fig:gap_heatmaps_gauss_rff5kvsklr}\textbf{Gaussian KLR versus Gaussian RFC (5k features):} $\ell_2$ and $\ell_\infty$ perturbation for FashionMNIST, KMNIST and QMNIST. (Top) row corresponds to the results for FashionMNIST dataset, (Middle) row corresponds to the KMNIST dataset, and (Bottom) row corresponds to the QMNIST dataset. Of these, (First) column corresponds to the robust accuracies for $\ell_\infty$ adversaries while (Second) column portrays the robust accuracies for $\ell_2$ adversaries. PGD attack is used to attack the models.}
	\includegraphics[width=1\textwidth]{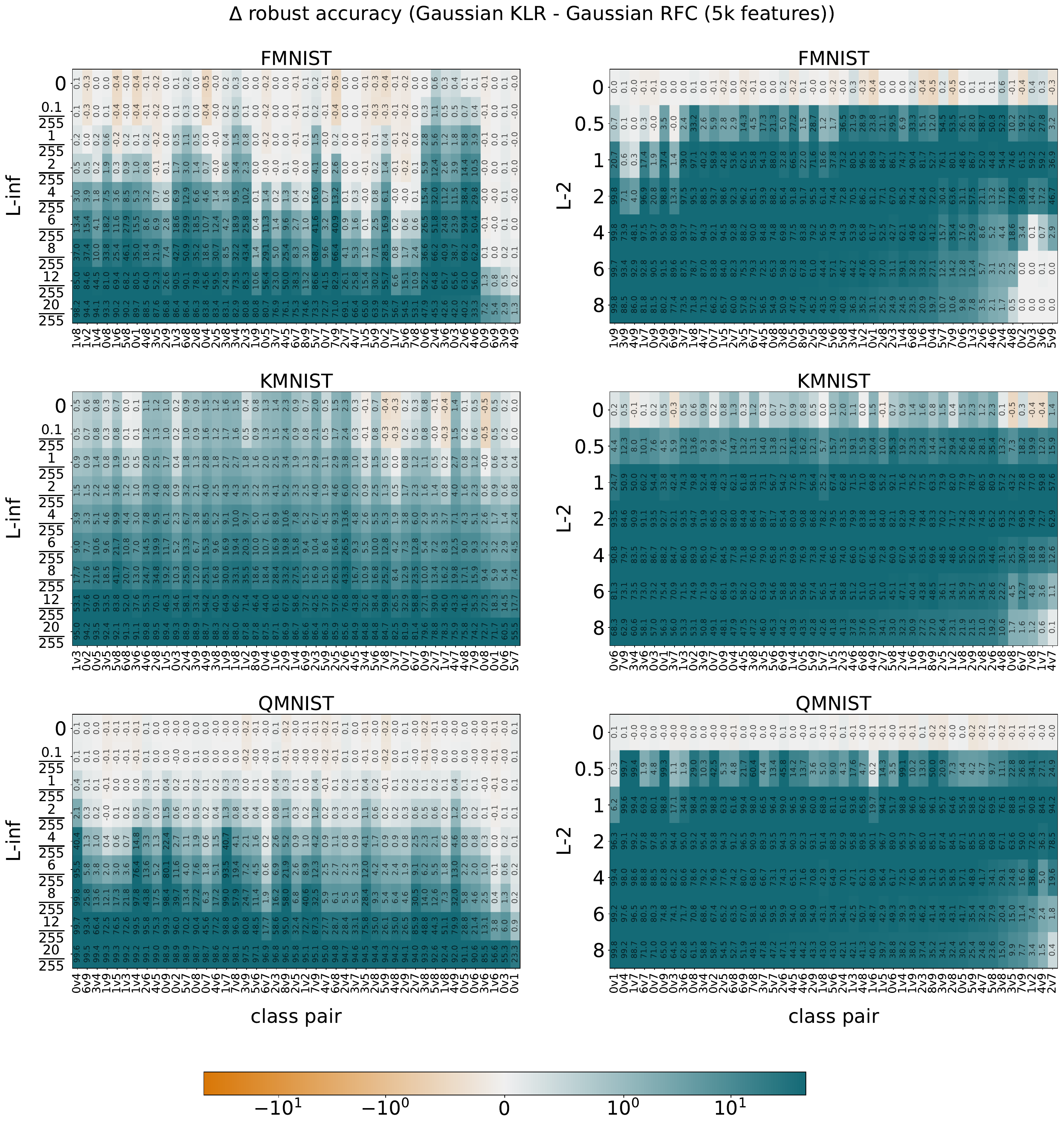}
\end{figure*}

\begin{figure*}[!htbp]
	\centering    \caption{\label{fig:gap_heatmaps_gauss_rff20kvsklr}\textbf{Gaussian KLR versus Gaussian RFC (20k features):} $\ell_2$ and $\ell_\infty$ perturbation for FashionMNIST, KMNIST and QMNIST. (Top) row corresponds to the results for FashionMNIST dataset, (Middle) row corresponds to the KMNIST dataset, and (Bottom) row corresponds to the QMNIST dataset. Of these, (First) column corresponds to the robust accuracies for $\ell_\infty$ adversaries while (Second) column portrays the robust accuracies for $\ell_2$ adversaries. PGD attack is used to attack the models.}
	\includegraphics[width=1\textwidth]{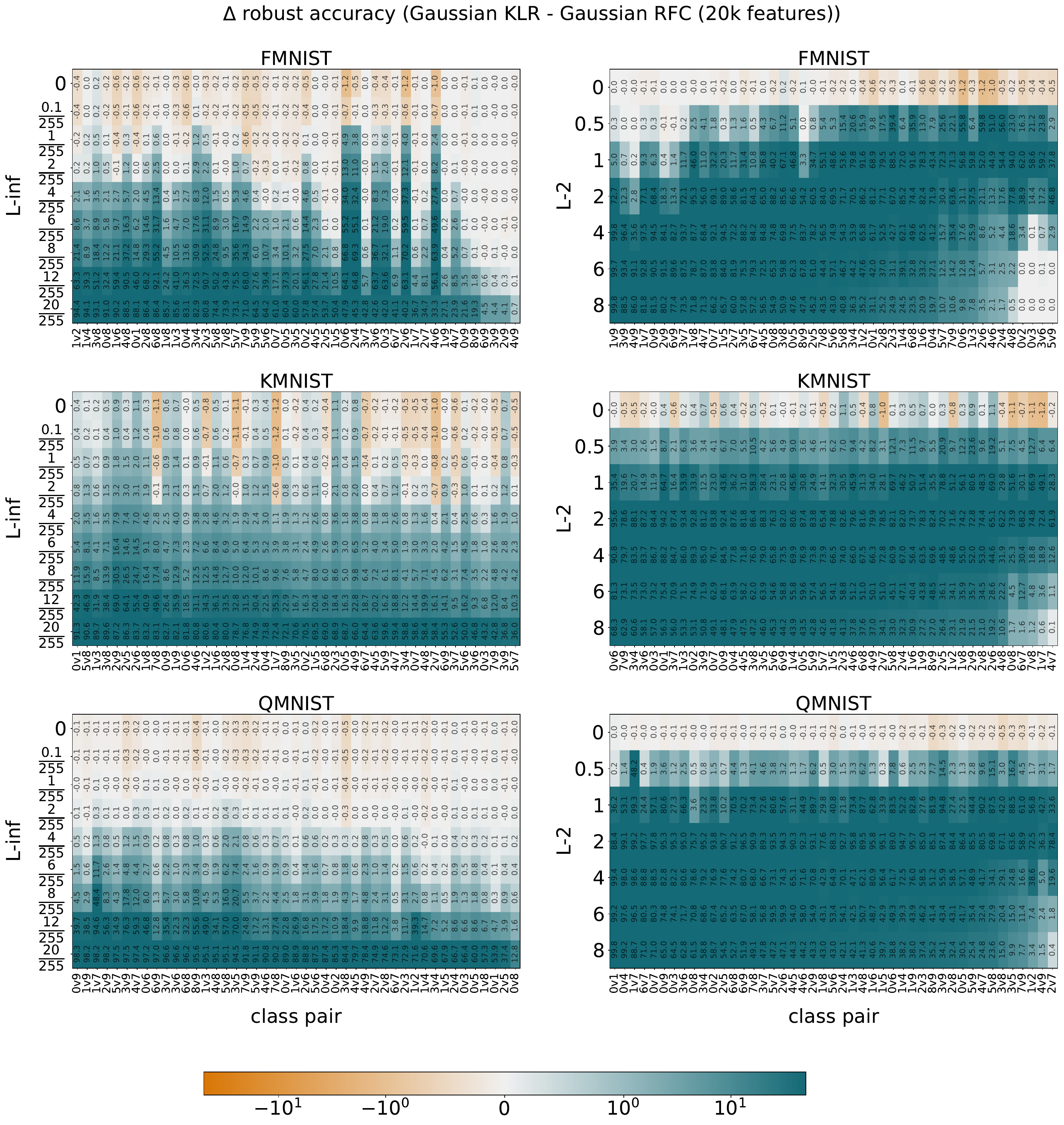}
\end{figure*}

\begin{figure*}[!htbp]
	\centering    \caption{\label{fig:gap_heatmaps_gauss_rff5kvsker}\textbf{Gaussian KRR versus Gaussian RFC (5k features):} $\ell_2$ and $\ell_\infty$ perturbation for FashionMNIST, KMNIST and QMNIST. (Top) row corresponds to the results for FashionMNIST dataset, (Middle) row corresponds to the KMNIST dataset, and (Bottom) row corresponds to the QMNIST dataset. Of these, (First) column corresponds to the robust accuracies for $\ell_\infty$ adversaries while (Second) column portrays the robust accuracies for $\ell_2$ adversaries. PGD attack is used to attack the models.}
	\includegraphics[width=1\textwidth]{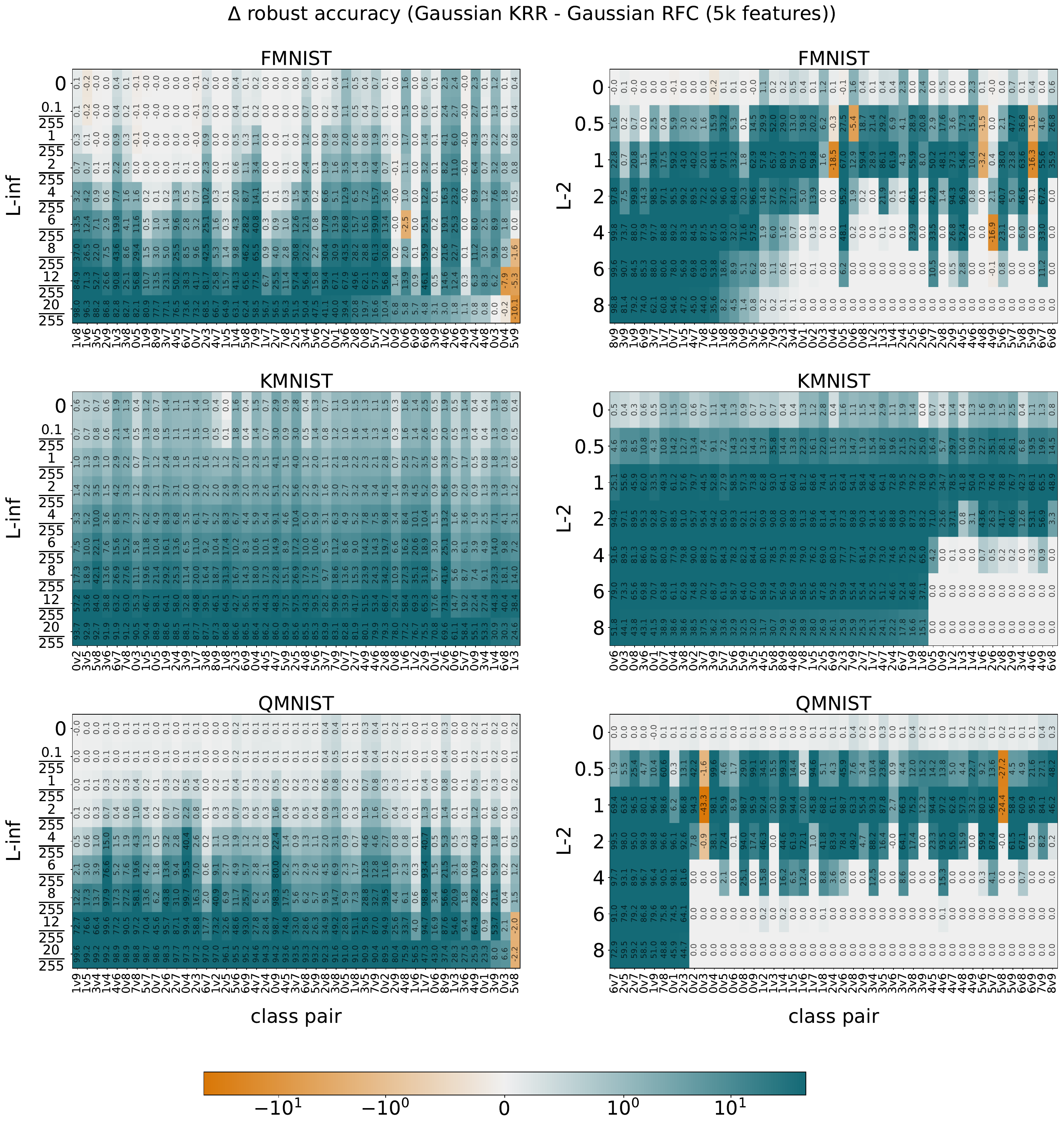}
\end{figure*}

\begin{figure*}[!htbp]
	\centering    \caption{\label{fig:gap_heatmaps_gauss_rff20kvsker}\textbf{Gaussian KRR versus Gaussian RFC (20k features):} $\ell_2$ and $\ell_\infty$ perturbation for FashionMNIST, KMNIST and QMNIST. (Top) row corresponds to the results for FashionMNIST dataset, (Middle) row corresponds to the KMNIST dataset, and (Bottom) row corresponds to the QMNIST dataset. Of these, (First) column corresponds to the robust accuracies for $\ell_\infty$ adversaries while (Second) column portrays the robust accuracies for $\ell_2$ adversaries. PGD attack is used to attack the models.}
	\includegraphics[width=1\textwidth]{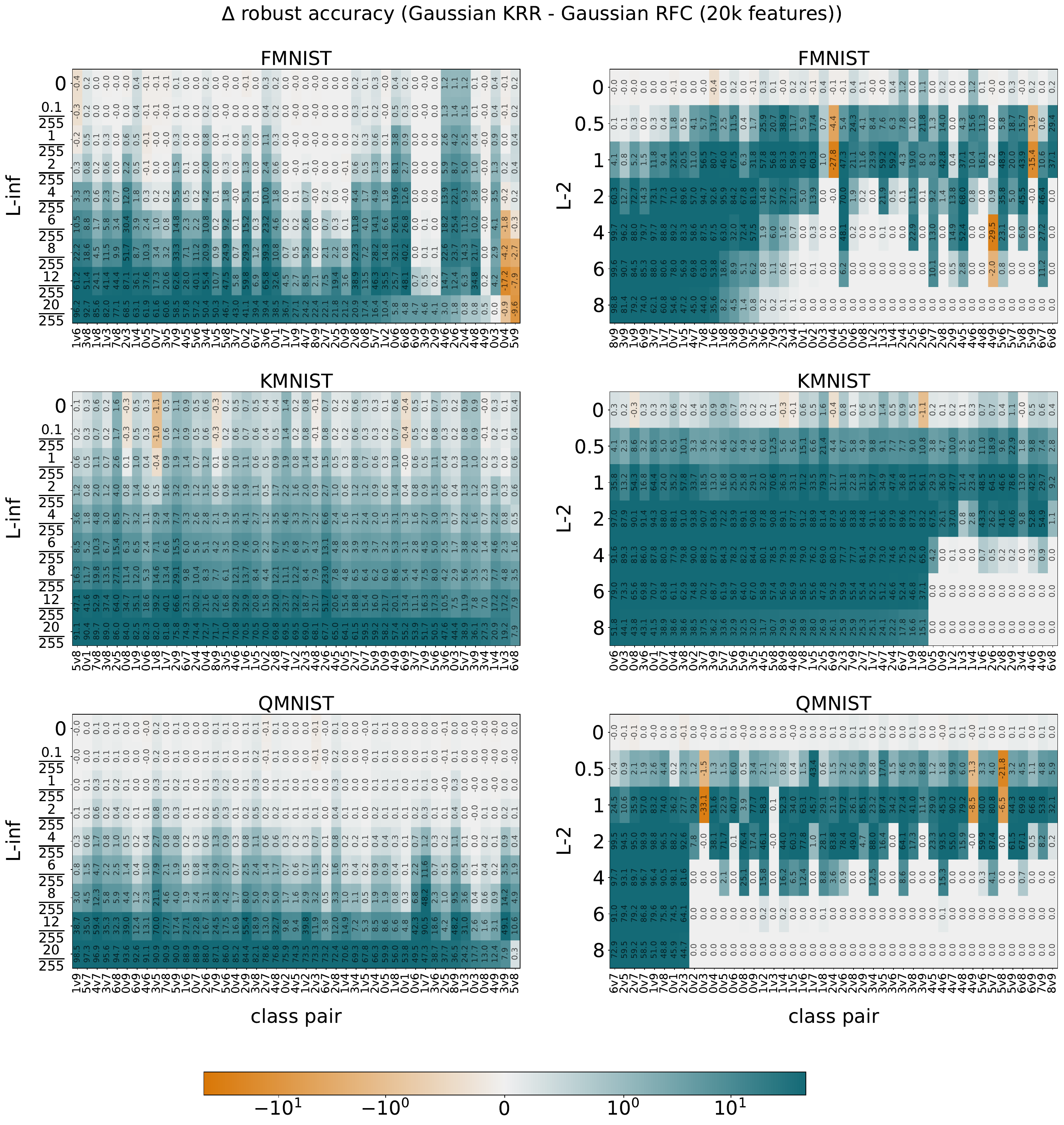}
	
\end{figure*}

\begin{figure*}[!htbp]
	\centering\caption{\label{fig:gap_heatmaps_gauss_rff5kvsrfm}\textbf{Gaussian RFM versus Gaussian RFC (5k features):} $\ell_2$ and $\ell_\infty$ perturbation for FashionMNIST, KMNIST and QMNIST. (Top) row corresponds to the results for FashionMNIST dataset, (Middle) row corresponds to the KMNIST dataset, and (Bottom) row corresponds to the QMNIST dataset. Of these, (First) column corresponds to the robust accuracies for $\ell_\infty$ adversaries while (Second) column portrays the robust accuracies for $\ell_2$ adversaries. PGD attack is used to attack the models.}
	\includegraphics[width=1\textwidth]{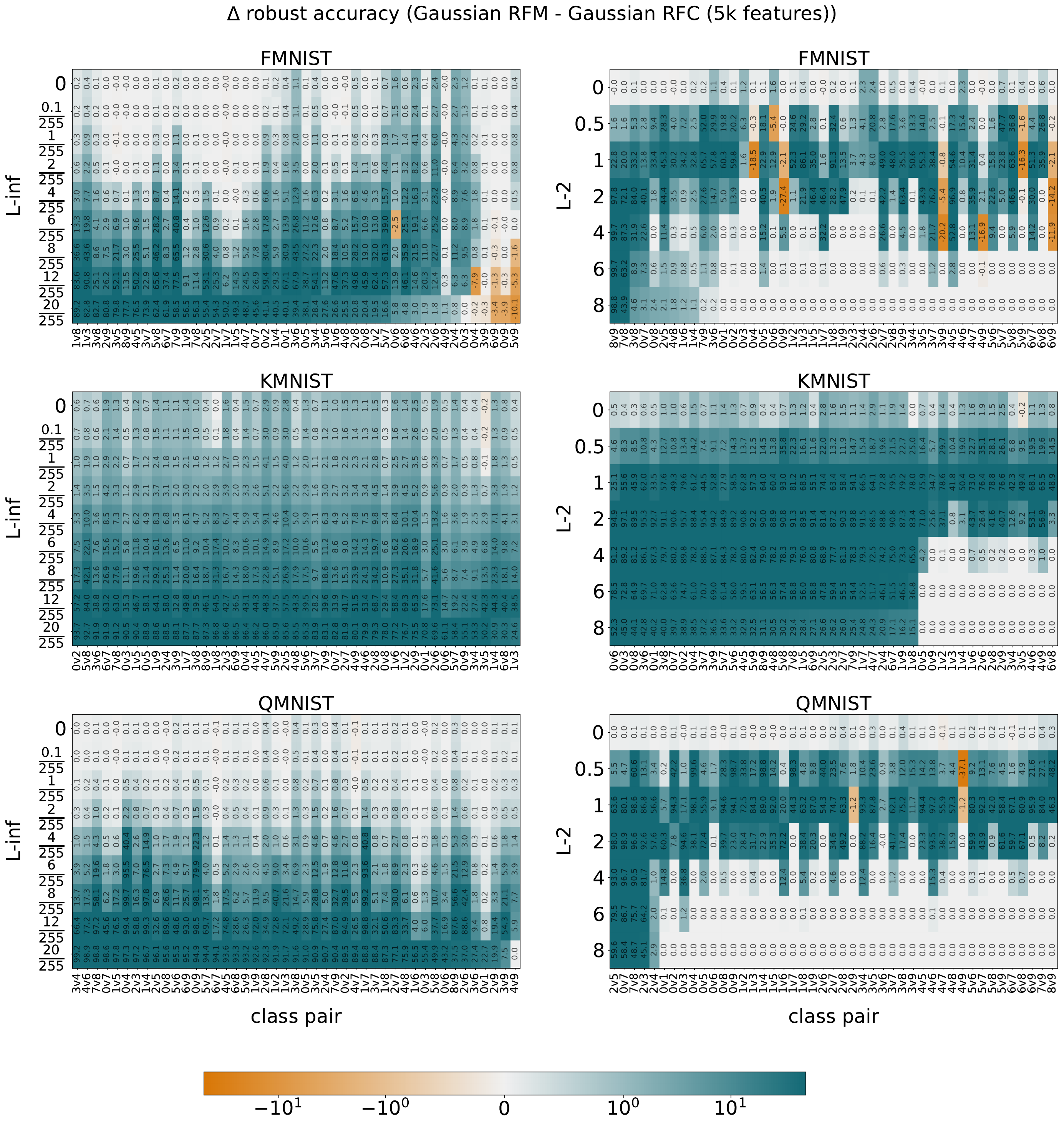}

\end{figure*}

\begin{figure*}[!htbp]
	\centering\caption{\label{fig:gap_heatmaps_gauss_rff20kvsrfm}\textbf{Gaussian RFM versus Gaussian RFC (20k features):} $\ell_2$ and $\ell_\infty$ perturbation for FashionMNIST, KMNIST and QMNIST. (Top) row corresponds to the results for FashionMNIST dataset, (Middle) row corresponds to the KMNIST dataset, and (Bottom) row corresponds to the QMNIST dataset. Of these, (First) column corresponds to the robust accuracies for $\ell_\infty$ adversaries while (Second) column portrays the robust accuracies for $\ell_2$ adversaries. PGD attack is used to attack the models.}
	\includegraphics[width=1\textwidth]{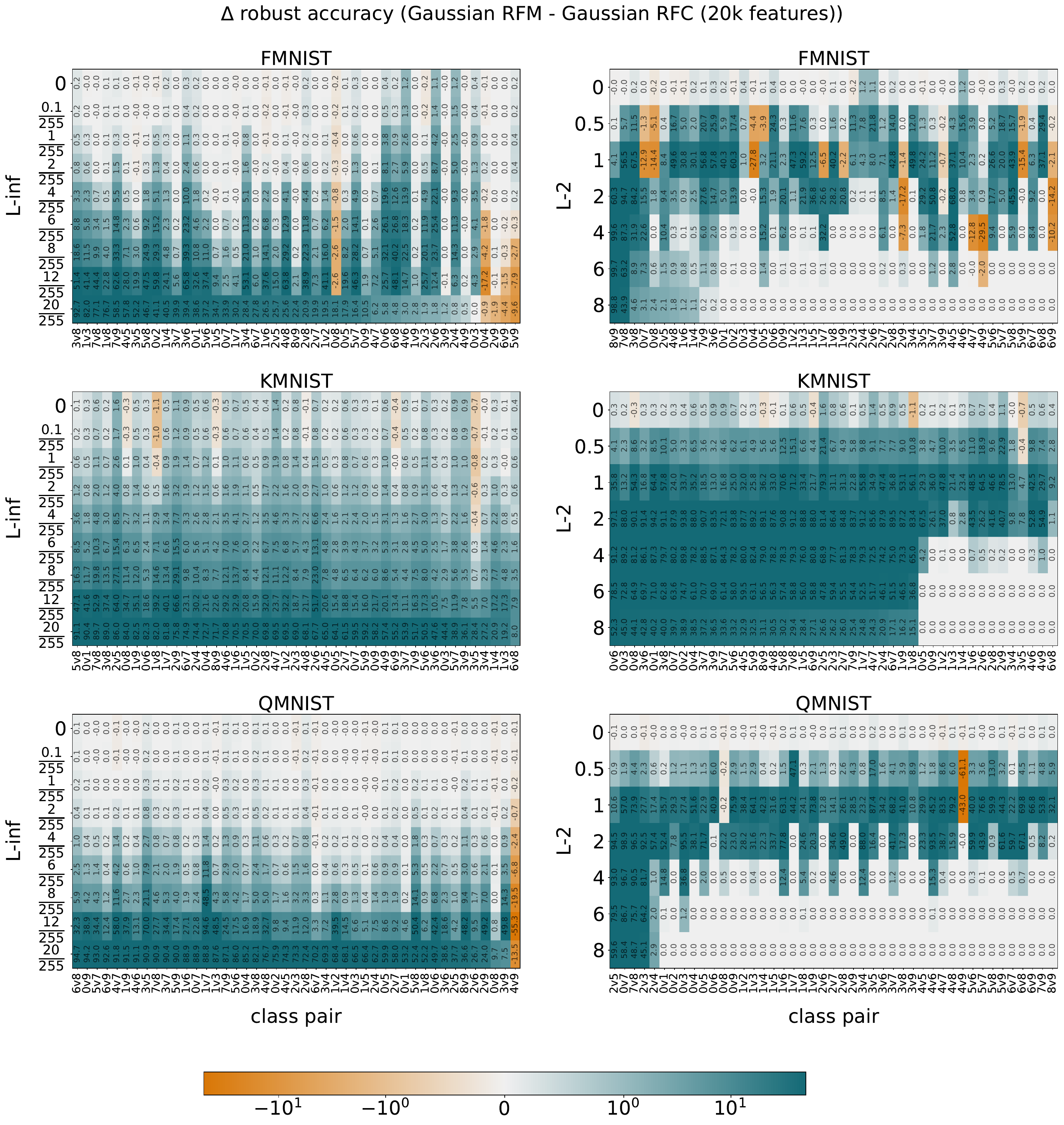}

\end{figure*}

\begin{figure*}[!htbp]
	\centering\caption{\label{fig:gap_heatmaps_gauss_rff5kvssvm}\textbf{Gaussian SVM versus Gaussian RFC (5k features):} $\ell_2$ and $\ell_\infty$ perturbation for FashionMNIST, KMNIST and QMNIST. (Top) row corresponds to the results for FashionMNIST dataset, (Middle) row corresponds to the KMNIST dataset, and (Bottom) row corresponds to the QMNIST dataset. Of these, (First) column corresponds to the robust accuracies for $\ell_\infty$ adversaries while (Second) column portrays the robust accuracies for $\ell_2$ adversaries. PGD attack is used to attack the models.}
	\includegraphics[width=1\textwidth]{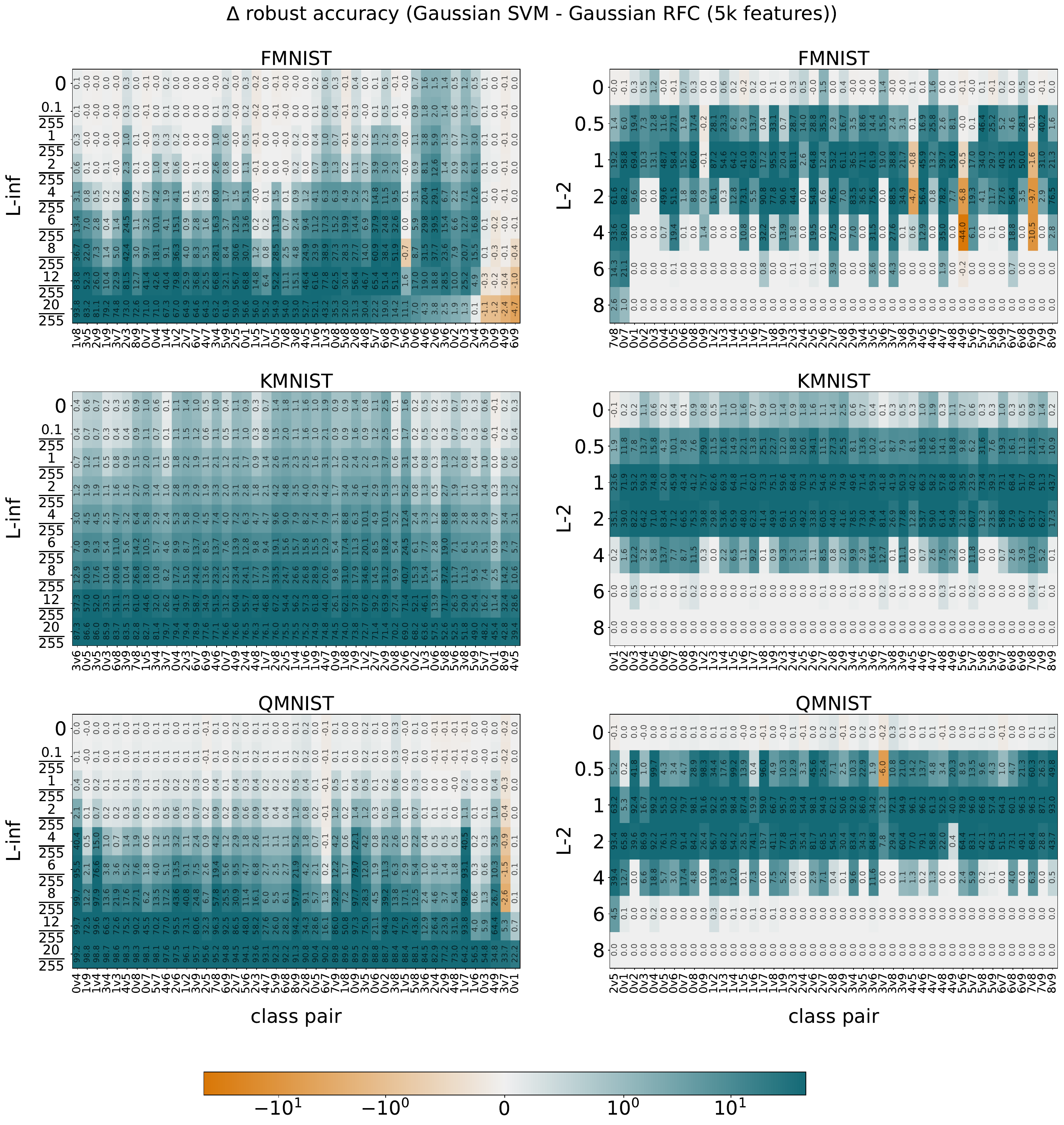}

\end{figure*}

\begin{figure*}[!htbp]
	\centering\caption{\label{fig:gap_heatmaps_gauss_rff20kvssvm}\textbf{Gaussian SVM versus Gaussian RFC (20k features):} $\ell_2$ and $\ell_\infty$ perturbation for FashionMNIST, KMNIST and QMNIST. (Top) row corresponds to the results for FashionMNIST dataset, (Middle) row corresponds to the KMNIST dataset, and (Bottom) row corresponds to the QMNIST dataset. Of these, (First) column corresponds to the robust accuracies for $\ell_\infty$ adversaries while (Second) column portrays the robust accuracies for $\ell_2$ adversaries. PGD attack is used to attack the models.}
	\includegraphics[width=1\textwidth]{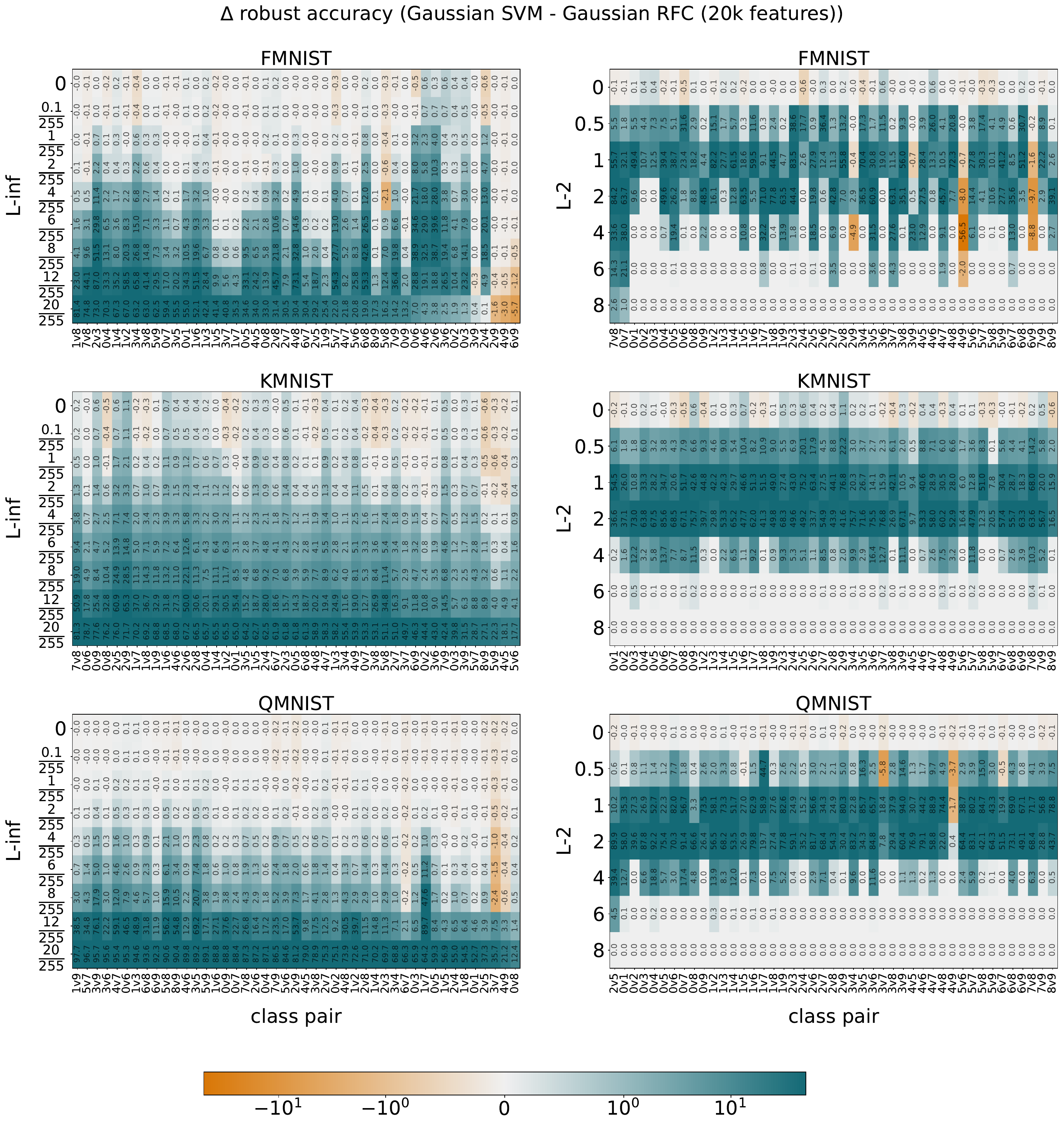}

\end{figure*}

\begin{figure*}
    \centering
    \caption{\label{fig:gap_heatmaps_gaussklr_vs_lenet} \textbf{Gaussian KLR versus LeNet:} $\ell_2$ and $\ell_\infty$ perturbation for FashionMNIST, KMNIST and QMNIST. (Top) row corresponds to the results for FashionMNIST dataset, (Middle) row corresponds to the KMNIST dataset, and (Bottom) row corresponds to the QMNIST dataset. Of these, (First) column corresponds to the robust accuracies for $\ell_\infty$ adversaries while (Second) column portrays the robust accuracies for $\ell_2$ adversaries. PGD attack is used to attack the models.}
\includegraphics[width=\textwidth]{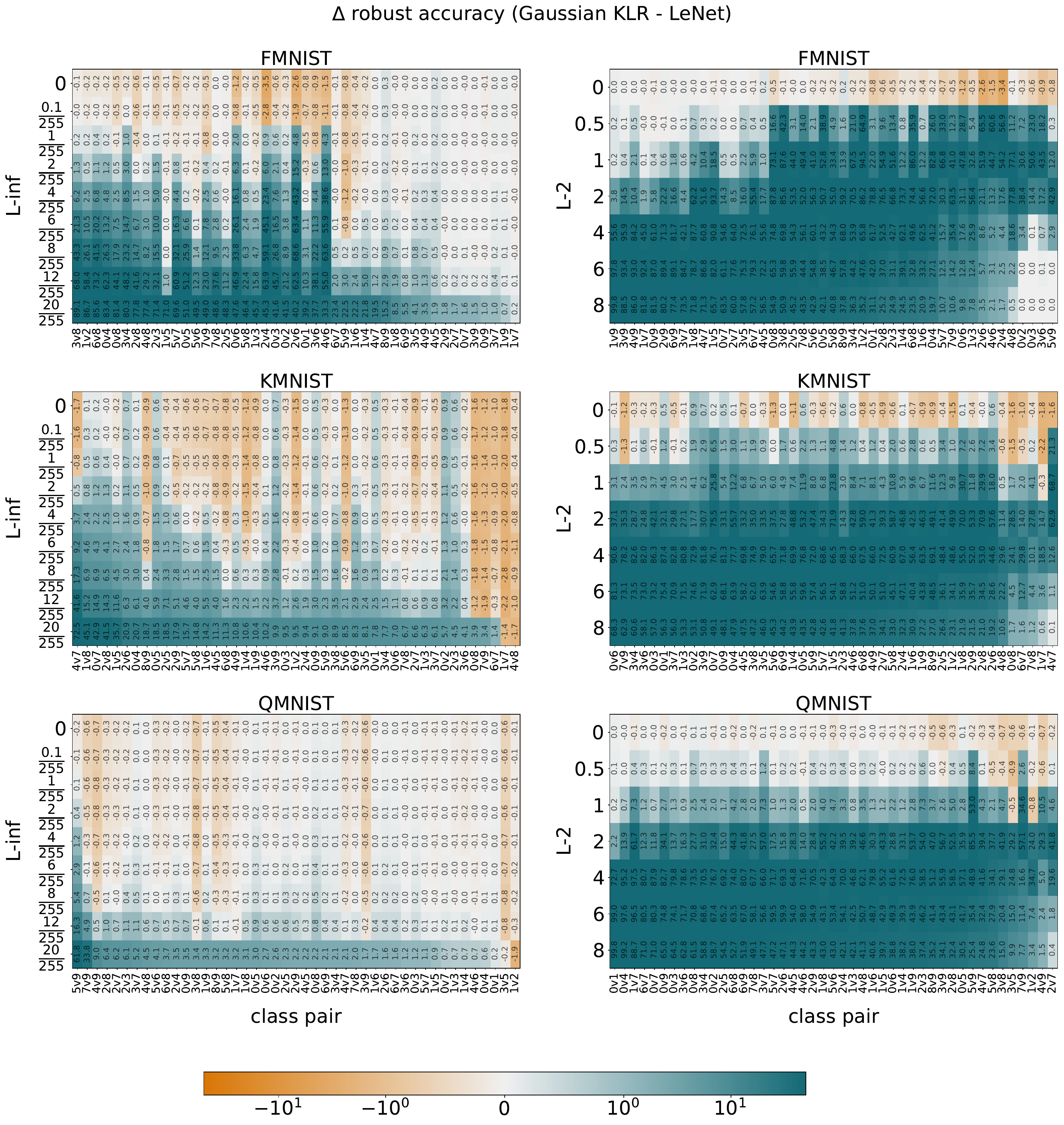}

\end{figure*}

\begin{figure*}
    \centering
    \caption{\label{fig:gap_heatmaps_gauss_vs_lenet} \textbf{Gaussian KRR versus LeNet:} $\ell_2$ and $\ell_\infty$ perturbation for FashionMNIST, KMNIST and QMNIST. (Top) row corresponds to the results for FashionMNIST dataset, (Middle) row corresponds to the KMNIST dataset, and (Bottom) row corresponds to the QMNIST dataset. Of these, (First) column corresponds to the robust accuracies for $\ell_\infty$ adversaries while (Second) column portrays the robust accuracies for $\ell_2$ adversaries. PGD attack is used to attack the models.}
\includegraphics[width=\textwidth]{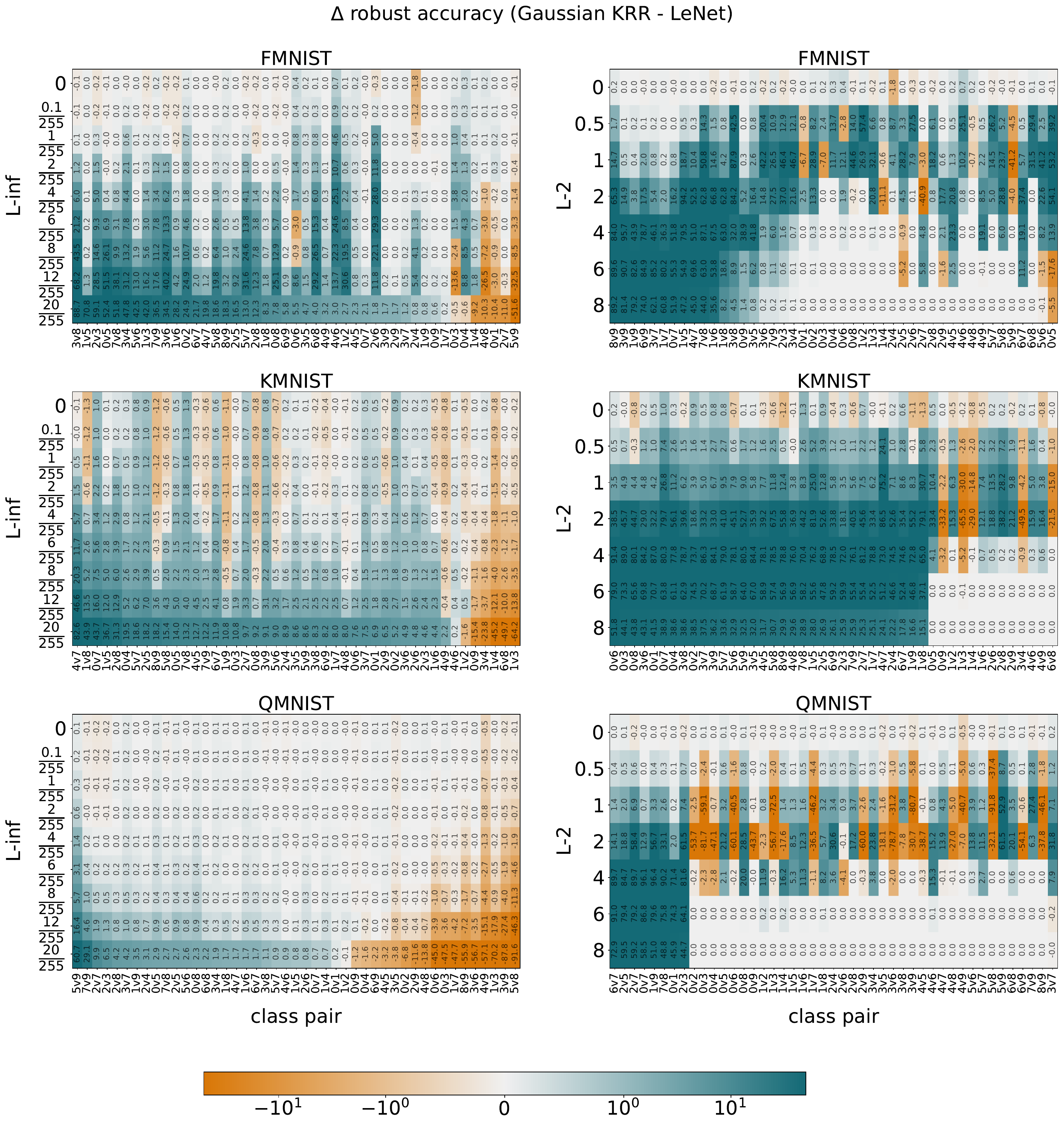}

\end{figure*}

\begin{figure*}[!htbp]
\caption{\label{fig:gap_heatmaps_leNetvsgaussrfm_ker}\textbf{Gaussian RFM versus LeNet:} $\ell_2$ and $\ell_\infty$ perturbation for FashionMNIST, KMNIST and QMNIST. (Top) row corresponds to the results for FashionMNIST dataset, (Middle) row corresponds to the KMNIST dataset, and (Bottom) row corresponds to the QMNIST dataset. Of these, (First) column corresponds to the robust accuracies for $\ell_\infty$ adversaries while (Second) column portrays the robust accuracies for $\ell_2$ adversaries. PGD attack is used to attack the models.} 
	\centering
	\includegraphics[width=1\textwidth]{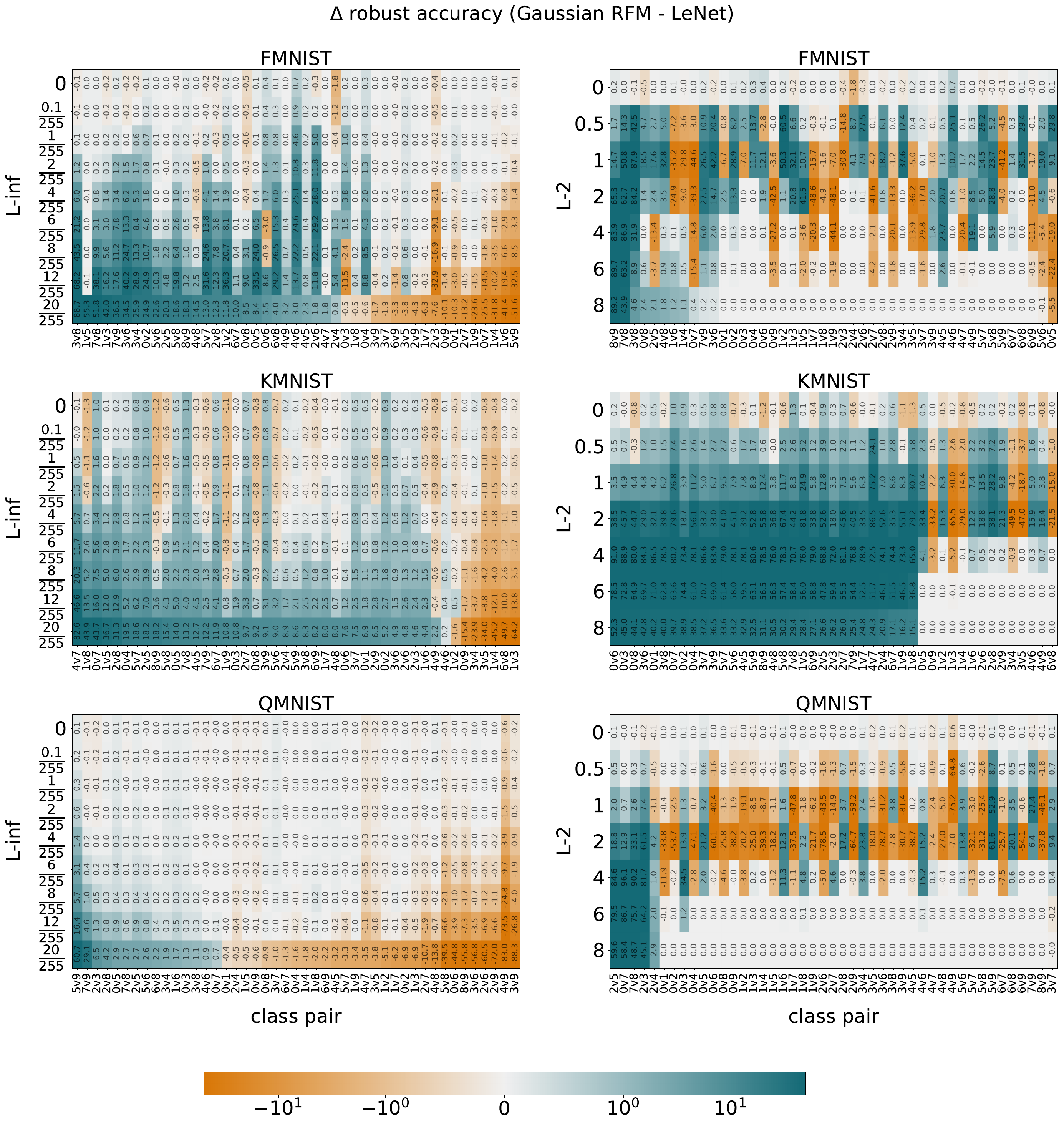}
	
\end{figure*}

\begin{figure*}[!htbp]
\caption{\label{fig:gap_heatmaps_leNetvsgausssvm}\textbf{Gaussian SVM versus LeNet:} $\ell_2$ and $\ell_\infty$ perturbation for FashionMNIST, KMNIST and QMNIST. (Top) row corresponds to the results for FashionMNIST dataset, (Middle) row corresponds to the KMNIST dataset, and (Bottom) row corresponds to the QMNIST dataset. Of these, (First) column corresponds to the robust accuracies for $\ell_\infty$ adversaries while (Second) column portrays the robust accuracies for $\ell_2$ adversaries. PGD attack is used to attack the models.} 
	\centering
	\includegraphics[width=1\textwidth]{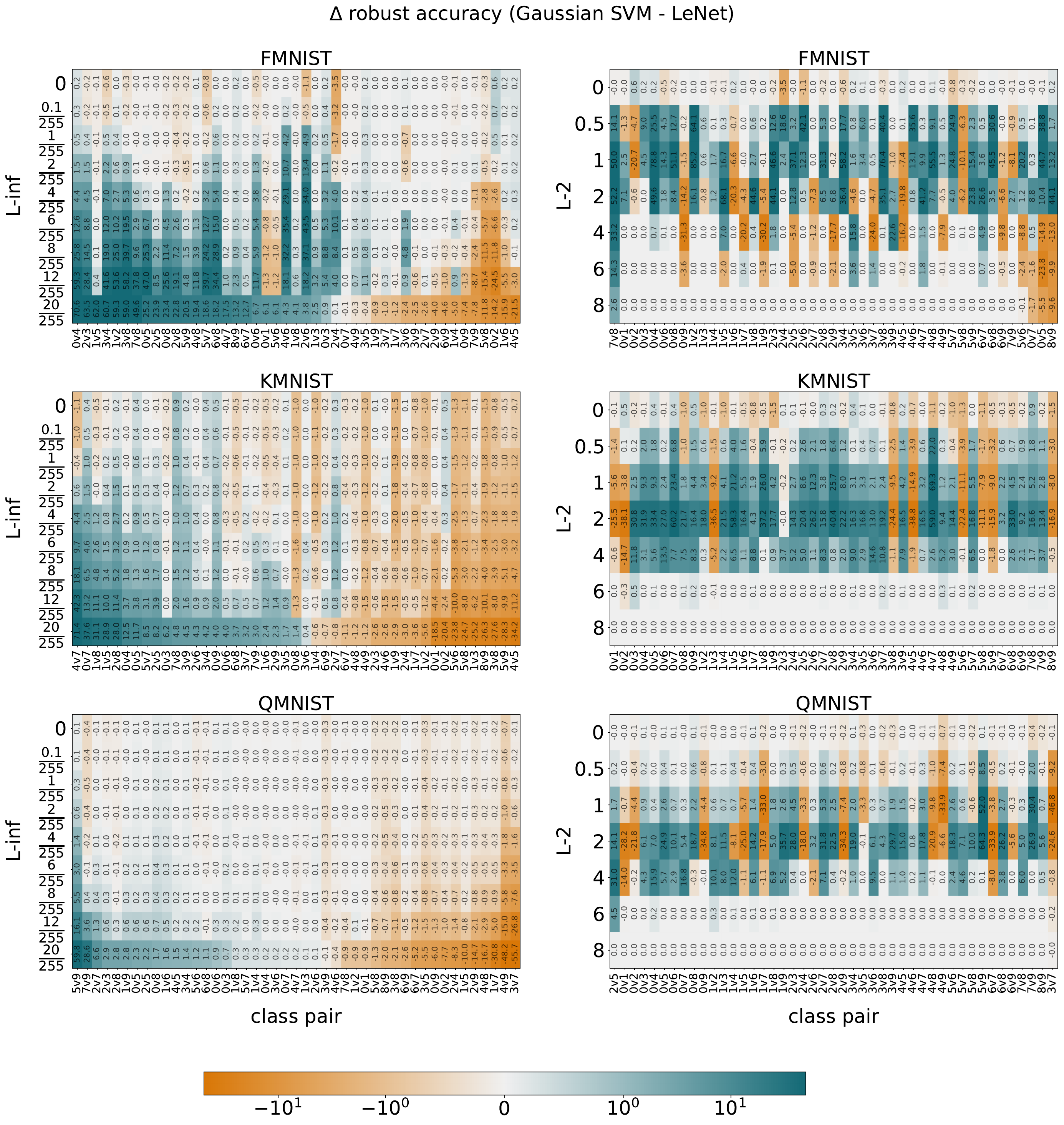}
	
\end{figure*}

\begin{figure*}[!htbp]
\caption{\label{fig:gap_heatmaps_leNetvsgaussrfc5k}\textbf{Gaussian RFC (5k features) versus LeNet:} $\ell_2$ and $\ell_\infty$ perturbation for FashionMNIST, KMNIST and QMNIST. (Top) row corresponds to the results for FashionMNIST dataset, (Middle) row corresponds to the KMNIST dataset, and (Bottom) row corresponds to the QMNIST dataset. Of these, (First) column corresponds to the robust accuracies for $\ell_\infty$ adversaries while (Second) column portrays the robust accuracies for $\ell_2$ adversaries. PGD attack is used to attack the models.} 
	\centering
	\includegraphics[width=1\textwidth]{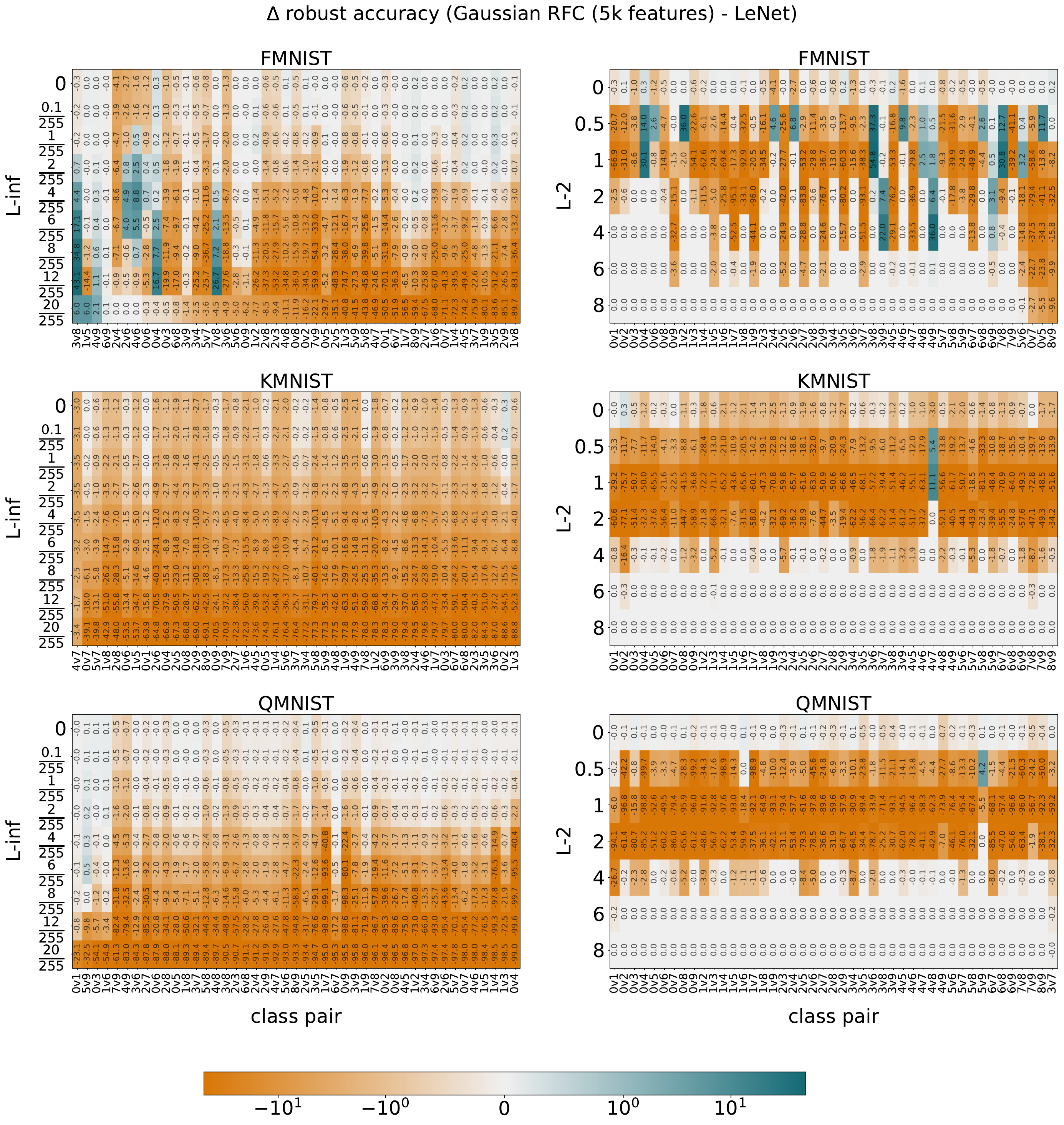}
	
\end{figure*}

\begin{figure*}[!htbp]
\caption{\label{fig:gap_heatmaps_leNetvsgaussrfc20k}\textbf{Gaussian RFC (20k features) versus LeNet:} $\ell_2$ and $\ell_\infty$ perturbation for FashionMNIST, KMNIST and QMNIST. (Top) row corresponds to the results for FashionMNIST dataset, (Middle) row corresponds to the KMNIST dataset, and (Bottom) row corresponds to the QMNIST dataset. Of these, (First) column corresponds to the robust accuracies for $\ell_\infty$ adversaries while (Second) column portrays the robust accuracies for $\ell_2$ adversaries. PGD attack is used to attack the models.} 
	\centering
	\includegraphics[width=1\textwidth]{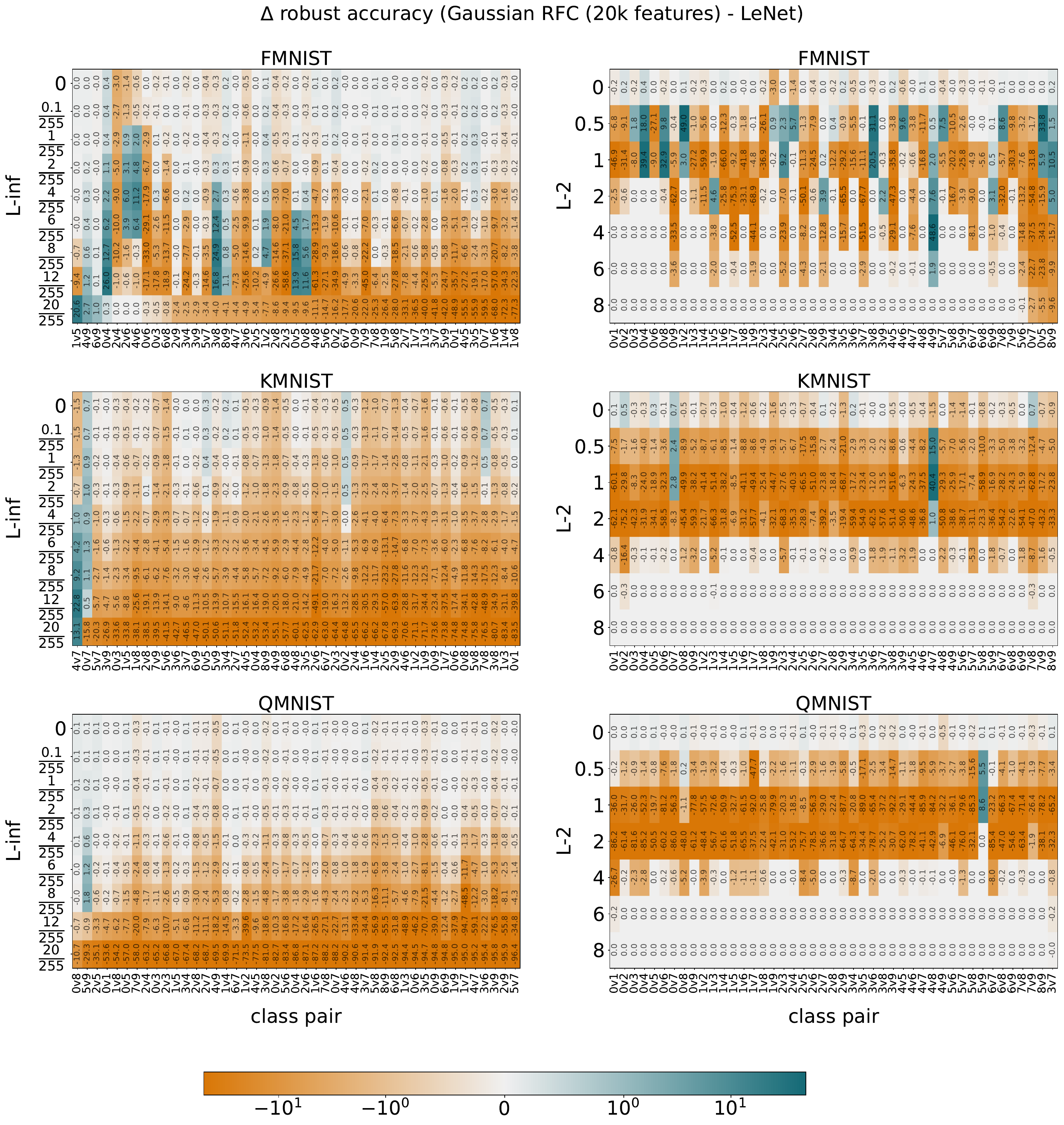}
	
\end{figure*}

\clearpage

\subsection{Kernels on SVHN and CIFAR-10}\label{apd:svhn_linf_l2}
\begin{figure*}[!htbp]
    \caption{
        $\ell_\infty$ and $\ell_2$ attacks on \textbf{SVHN}.
        (Left) corresponds to $\ell_\infty$ and (Right) corresponds to $\ell_2$ autoattacks.
    }
    \centering
    \includegraphics[width=\textwidth]{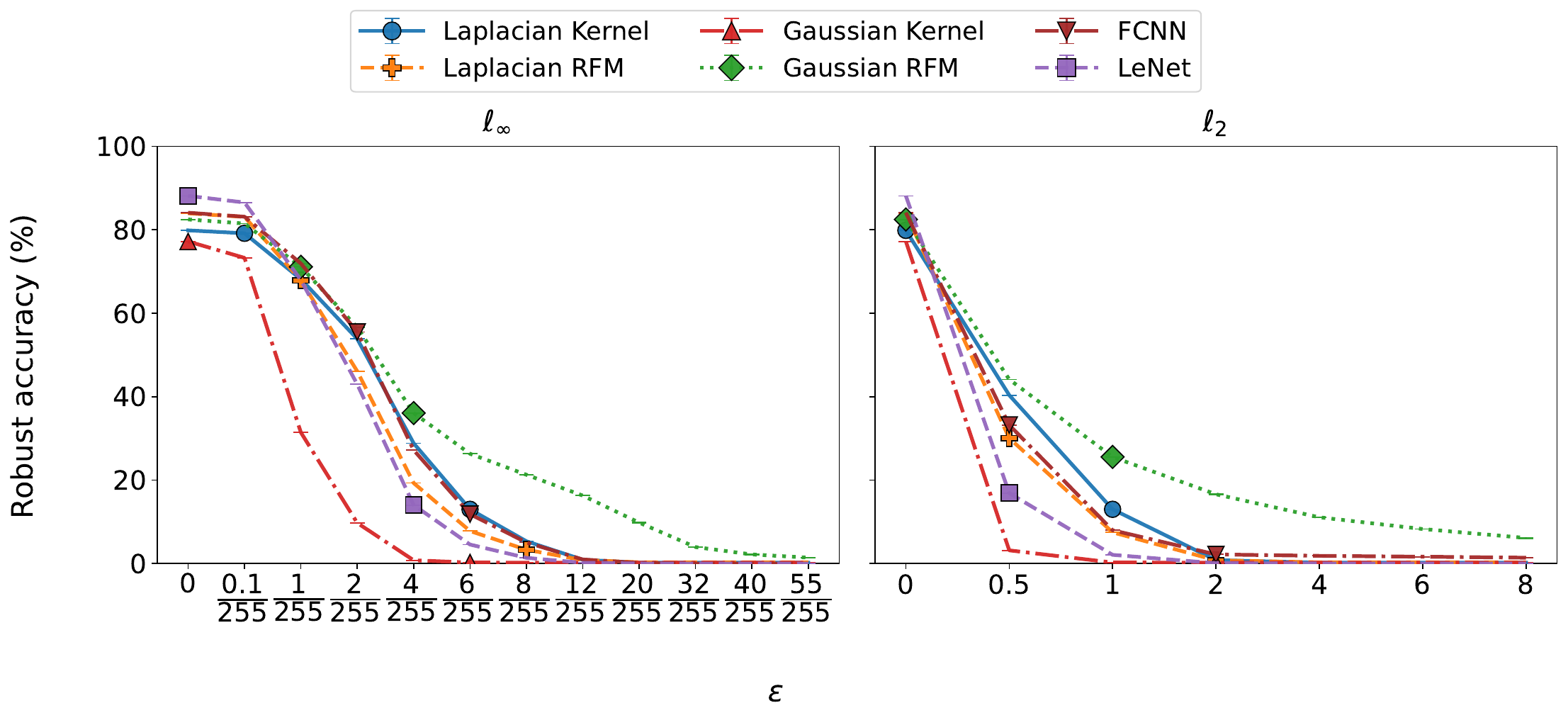}

    \label{fig:svhn_linf_l2}
\end{figure*}

\begin{figure*}[!htbp]
\centering
    \caption{\label{fig:cifar_lap_linf} \textbf{Gap in robust accuracy (ra) for $\ell_\infty$ perturbation on 5 $1\text{v}1$ binary classification tasks from CIFAR-10}. (Top) row has Laplace Kernel, and (Bottom) row has Laplace RFM. (First) Column is FCNN, and (Second) column is LeNet. PGD is used to attack the models.}
\includegraphics[width=1\textwidth]{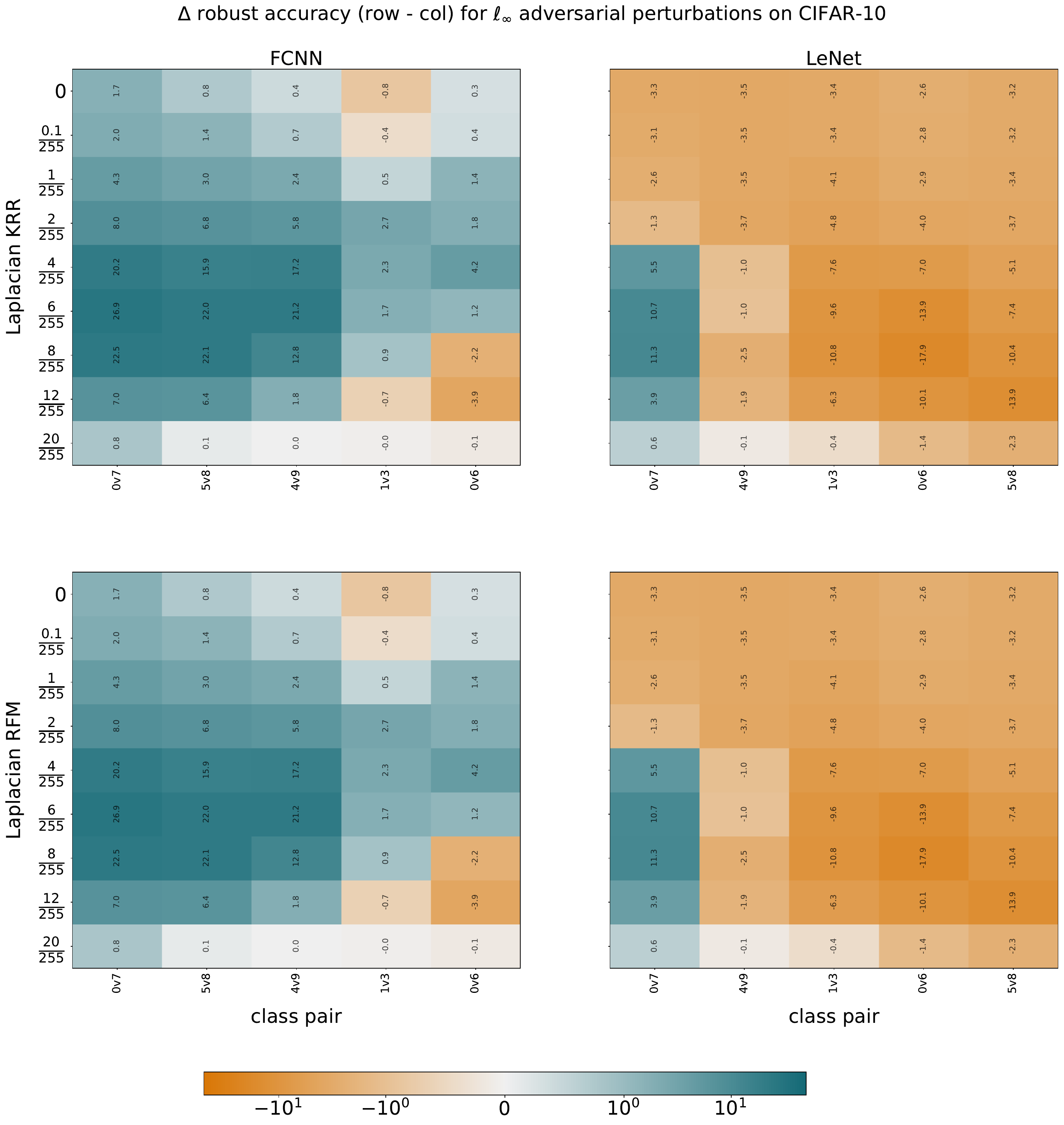}
\end{figure*}

\begin{figure*}[!htbp]
\centering
    \caption{\label{fig:cifar_lap_l2} \textbf{Gap in robust accuracy (ra) for $\ell_2$ perturbation on 5 $1\text{v}1$ binary classification tasks from CIFAR-10}. (Top) row has Laplace Kernel, and (Bottom) row has Laplace RFM. (First) column is FCNN, and (Second) column is LeNet. PGD is used to attack the models.}
\includegraphics[width=1\textwidth]{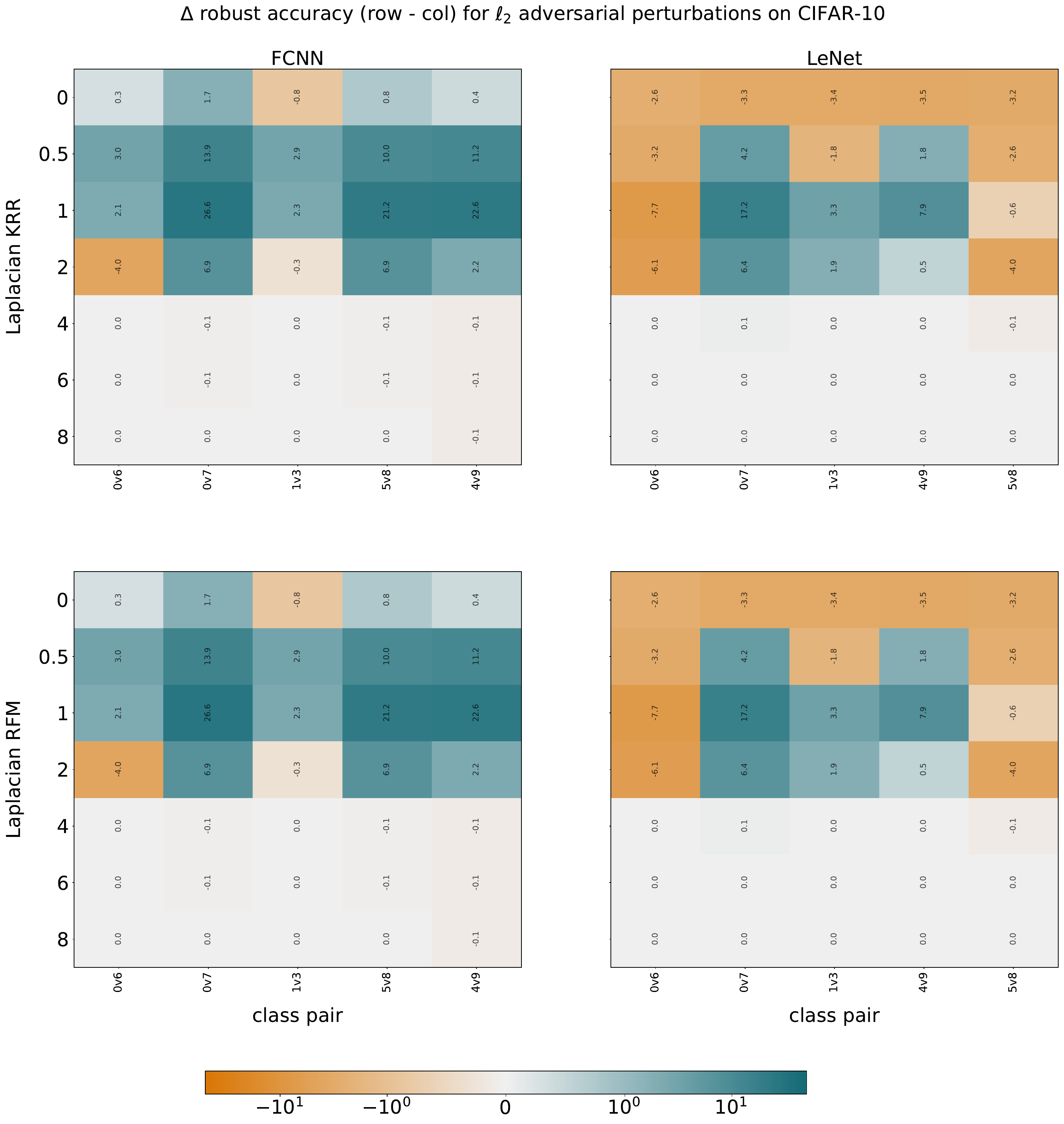}
\end{figure*}

\begin{figure*}[!htbp]
\centering
    \caption{\label{fig:cifar_gaus_linf} \textbf{Gap in robust accuracy (ra) for $\ell_\infty$ perturbation on 5 $1\text{v}1$ binary classification tasks from CIFAR-10}. (Top) row has Gaussian Kernel, and (Bottom) row has Gaussian RFM. (First) column is FCNN, and (Second) column is LeNet. PGD is used to attack the models.}
\includegraphics[width=1\textwidth]{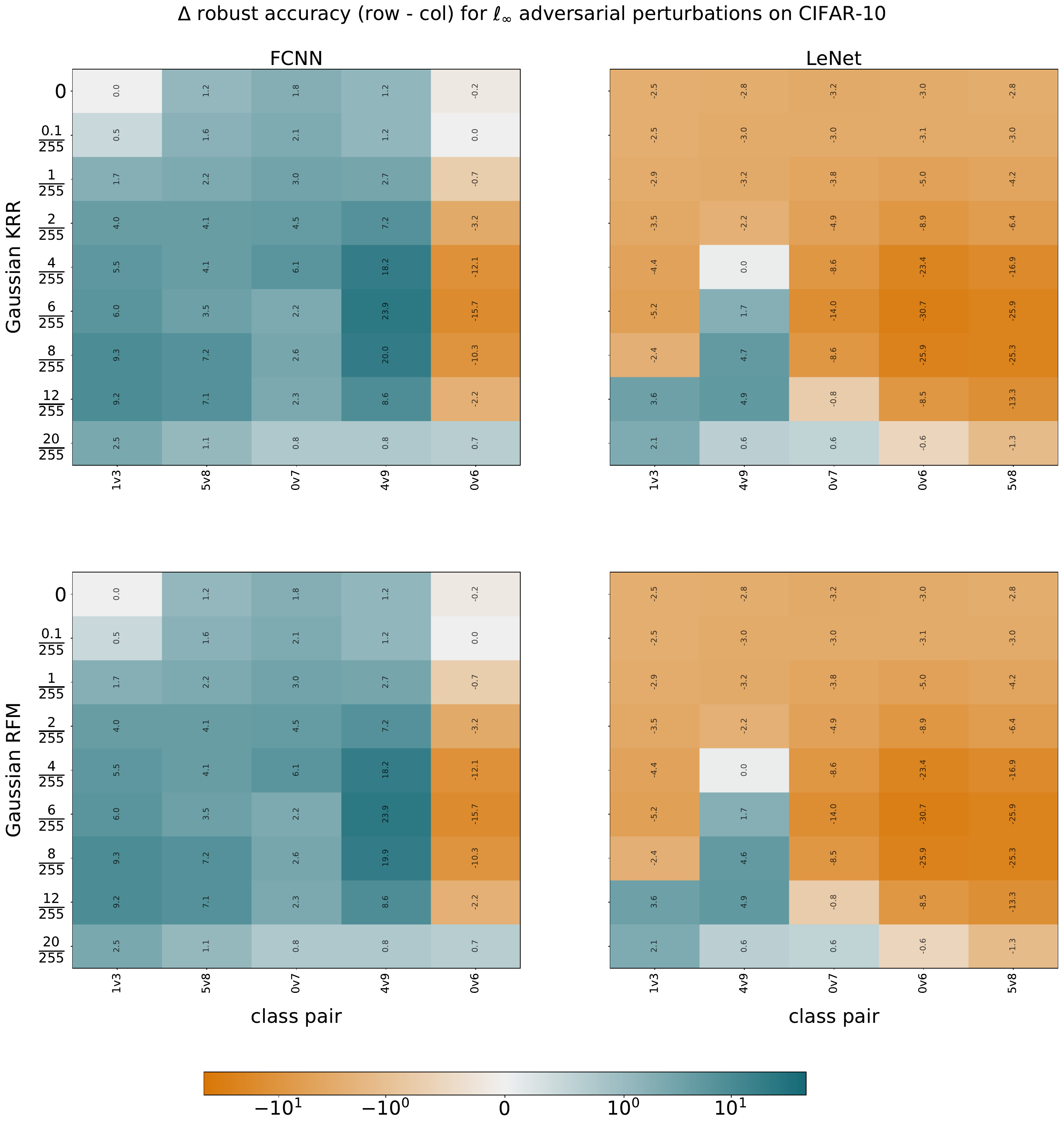}
\end{figure*}

\begin{figure*}[!htbp]
\centering
    \caption{\label{fig:cifar_gaus_l2} \textbf{Gap in robust accuracy (ra) for $\ell_2$ perturbation on 5 $1\text{v}1$ binary classification tasks from CIFAR-10}. (Top) row has Gaussian Kernel, and (Bottom) row has Gaussian RFM. (First) column is FCNN, and (Second) column is LeNet. PGD is used to attack the models.}
\includegraphics[width=1\textwidth]{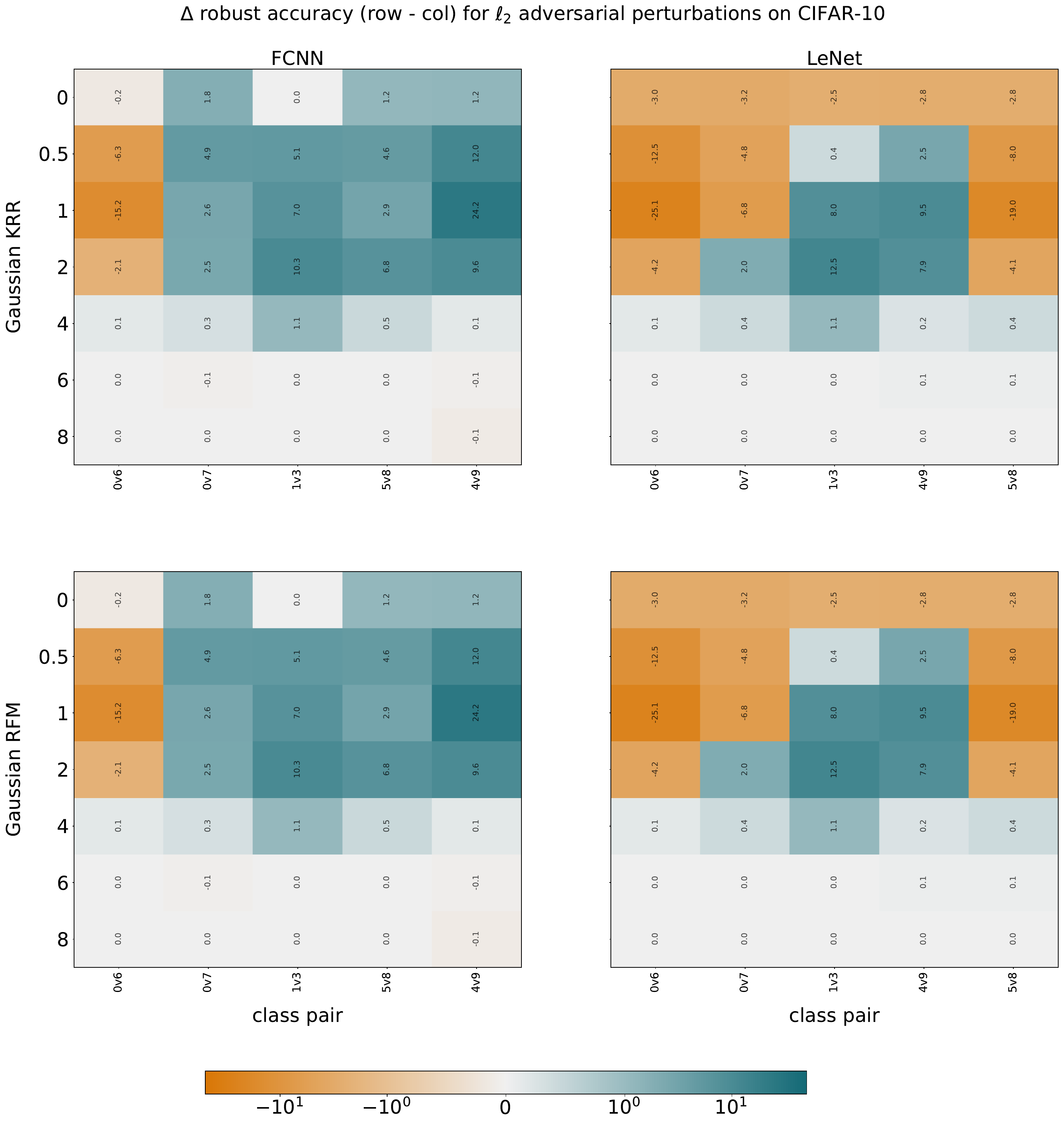}
\end{figure*}

We trained Gaussian KRR, Gaussian RFM, Laplacian KRR, Laplacian RFM, FCNN, and LeNet on $10$ class SVHN and different binary problems of CIFAR-10 (namely Aeroplane vs frog (0v6), Aeroplane vs Horse (0v7), Automobile vs Cat (1v3), Deer vs Truck (4v9) and Dog vs Ship (5v8)). Hyperparameter tuning of models used same grid search as for the other three datasets namely FMNIST, KMNIST, QMNIST. As in~\cref{fig:svhn_linf_l2}, Laplacian KRR and FCNN match in robustness with Gaussian RFM beating them all. The reason behind Gaussian RFM being the best model can be inferred from the fact that its base kernel despite being poor in benign accuracy is quite robust~(\cref{tab:svhn_gaussian_pgd}). Hence, even though further iterations in RFM generally lose the robustness (as we will see in~\cref{apd: rfm_iter_results}), the loss in robustness is not much to make it brittle.
Here, the base KM for Gaussian RFM is having different hyperparameters than the Gaussian KRR in~\cref{fig:svhn_linf_l2}.
Results for CIFAR-10 are in ~\cref{fig:cifar_lap_linf}, ~\cref{fig:cifar_lap_l2}, ~\cref{fig:cifar_gaus_linf}, ~\cref{fig:cifar_gaus_l2}. 

\vspace{10mm}
\begin{table*}[!htbp]
\centering
\caption{Robust accuracies for the Gaussian kernel and Gaussian RFM
with bandwidth $0.05$ on SVHN}
\label{tab:svhn_gaussian_pgd}
\begin{tabular}{c|cc}
\hline
$\epsilon$ ($L_\infty$)
& Gaussian Kernel ($\%$)
& Gaussian RFM ($\%$)\\
\hline
$0$
    & 62.56 & 82.46 \\
$0.1/255$
    & 62.11 & 81.48 \\
$1/255$
    & 58.22 & 71.10 \\
$2/255$
    & 53.86 & 56.59 \\
$4/255$
    & 46.01 & 36.08 \\
$6/255$
    & 39.48 & 26.30 \\
$8/255$
    & 34.44 & 21.29 \\
$12/255$
    & 27.60 & 16.26 \\
$20/255$
    & 21.27 & 9.77 \\
$32/255$
    & 17.25 & 3.91 \\
$40/255$
    & 15.81 & 2.25 \\
$55/255$
    & 13.58 & 1.40 \\
\hline
\end{tabular}
\end{table*}

\clearpage

\subsection{WideResNet vs LeNet}\label{subsec:wrn_vs_lenet}
We compare the PGD robust accuracy of WideResNet-28-10 (WRN) and LeNet-5 on multiclass FMNIST, KMNIST, and QMNIST datasets.~\cref{tab:pgd_combined} reports the accuracy (\%) for $\ell_\infty$ perturbation budgets $\epsilon$. Across all datasets, WRN attains higher clean accuracy than LeNet-5 (93.57\% vs.\ 90.54\% on FMNIST, 98.61\% vs.\ 95.49\% on KMNIST, and 99.46\% vs.\ 98.91\% on QMNIST). However, this advantage does not persist as $\varepsilon$ grows. Hence, our comparison of KMs against LeNets is transferable to larger CNN models like WideResNets as well.

\begin{table*}[!htbp]
\vspace{35mm}
\centering
\caption{PGD robust accuracy (\%) of WRN-28-10 and LeNet-5 on FMNIST, KMNIST, and QMNIST. Bold marks the better model per dataset.}
\label{tab:pgd_combined}
\renewcommand{\arraystretch}{1.1}
\begin{tabular*}{\textwidth}{@{\extracolsep{\fill}} c cc cc cc}
\toprule
& \multicolumn{2}{c}{FMNIST} & \multicolumn{2}{c}{KMNIST} & \multicolumn{2}{c}{QMNIST} \\
\cmidrule(lr){2-3} \cmidrule(lr){4-5} \cmidrule(lr){6-7}
$\epsilon$ & WRN & LeNet & WRN & LeNet & WRN & LeNet \\
\midrule
0        & \textbf{93.57} & 90.54 & \textbf{98.61} & 95.49 & \textbf{99.46} & 98.91 \\
0.1/255  & \textbf{92.64} & 90.03 & \textbf{98.58} & 95.42 & \textbf{99.46} & 98.90 \\
1/255    & 77.60 & \textbf{84.04} & \textbf{98.45} & 94.92 & \textbf{99.37} & 98.75 \\
2/255    & 55.64 & \textbf{74.82} & \textbf{98.19} & 94.20 & \textbf{99.27} & 98.54 \\
4/255    & 13.54 & \textbf{55.67} & \textbf{97.60} & 92.40 & \textbf{99.03} & 98.00 \\
6/255    & 0.80  & \textbf{39.68} & \textbf{96.61} & 90.35 & \textbf{98.70} & 97.39 \\
8/255    & 0.02  & \textbf{26.88} & \textbf{94.99} & 87.34 & \textbf{98.31} & 96.42 \\
12/255   & 0.00  & \textbf{8.95}  & \textbf{84.18} & 80.23 & \textbf{96.72} & 93.43 \\
20/255   & 0.00  & \textbf{0.18}  & 30.65 & \textbf{57.57} & 75.85 & \textbf{79.66} \\
32/255   & 0.00  & 0.00  & 1.01  & \textbf{17.58} & 1.68  & \textbf{35.52} \\
40/255   & 0.00  & 0.00  & 0.05  & \textbf{4.06}  & 0.07  & \textbf{11.61} \\
55/255   & 0.00  & 0.00  & 0.00  & \textbf{0.07}  & 0.00  & \textbf{0.32} \\
\bottomrule
\end{tabular*}
\end{table*}

\clearpage
\subsection{KMs and adversarially trained KMs vs adversarially trained nets}\label{apd: kms_vs_adv_training}
\begin{figure*}[!htbp]
\centering
    \caption{\label{fig:adv_klr_minus_klr} \textbf{Adv-KLR versus KLR for FashionMNIST and KMNIST. (Top) row corresponds to the results for FashionMNIST dataset, (bottom) row corresponds to the KMNIST dataset. PGD is used to attack the models.}}
\includegraphics[width=1\textwidth]{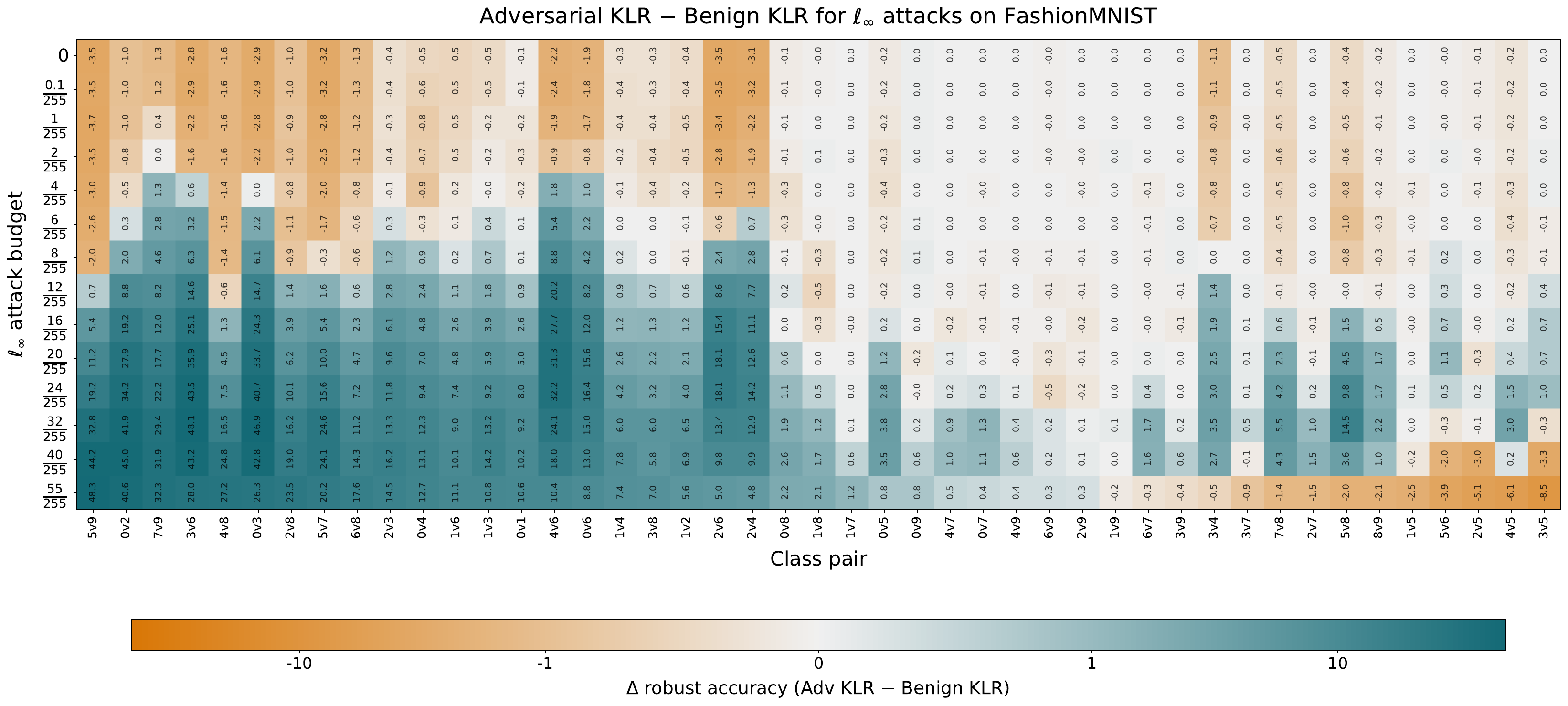}
\includegraphics[width=1\textwidth]{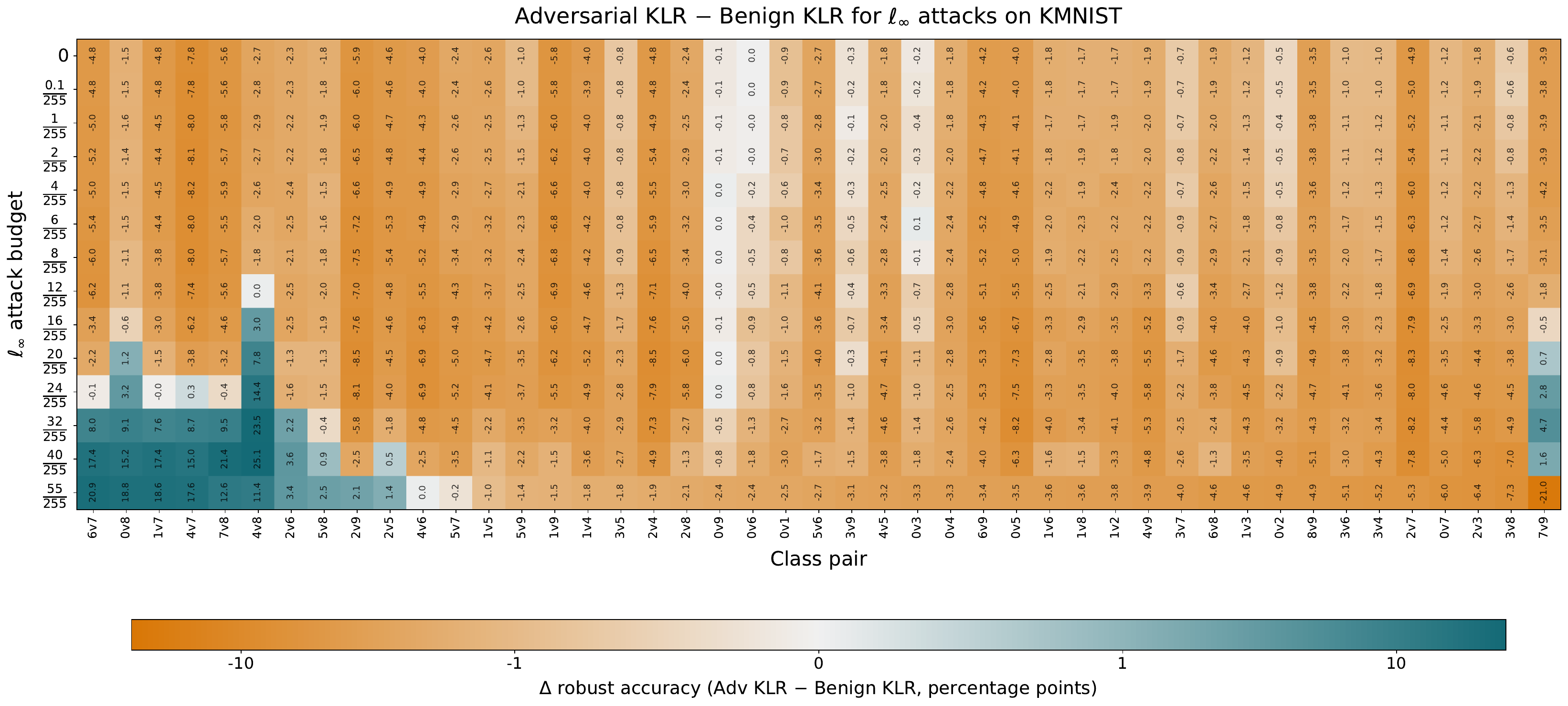}
\end{figure*}

\begin{figure*}[!htbp]
\centering
    \caption{\label{fig:adv_klr_fcnn} \textbf{Adv-KLR versus FCNN for FashionMNIST and KMNIST. (Top) row corresponds to the results for FashionMNIST dataset, (bottom) row corresponds to the KMNIST dataset. PGD is used to attack the models.}}
\includegraphics[width=1\textwidth]{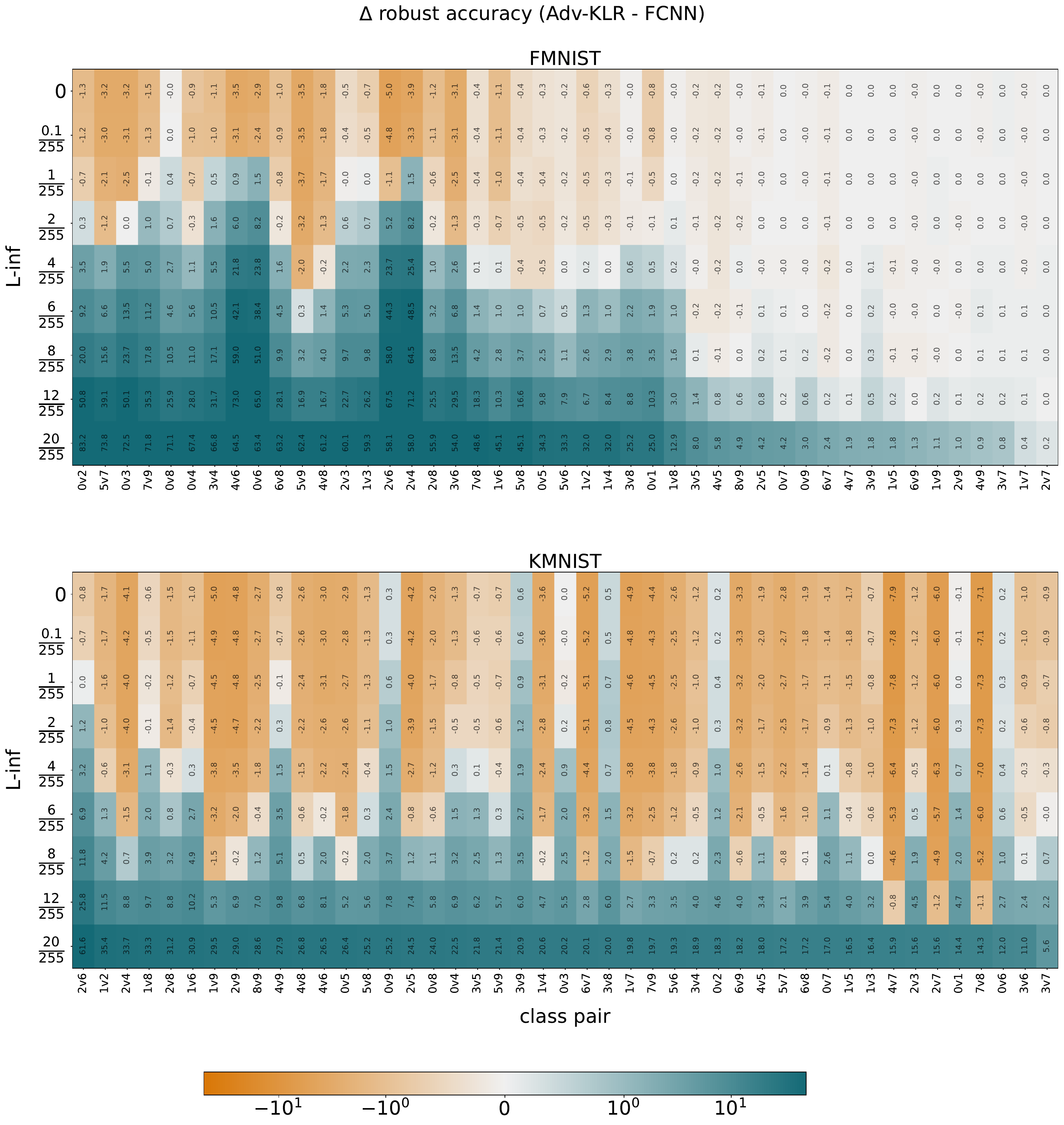}
\end{figure*}

\begin{figure*}[!htbp]
\centering
    \caption{\label{fig:gap_adv_klr_adv_fcnn} \textbf{Adv-KLR versus Adv-FCNN trained at norm $\ell_\infty = \frac{12}{255}$:} for FashionMNIST and KMNIST. (Top) row corresponds to the results for FashionMNIST dataset, (Bottom) row corresponds to the KMNIST dataset. PGD is used to attack the models.}
\includegraphics[width=1\textwidth]{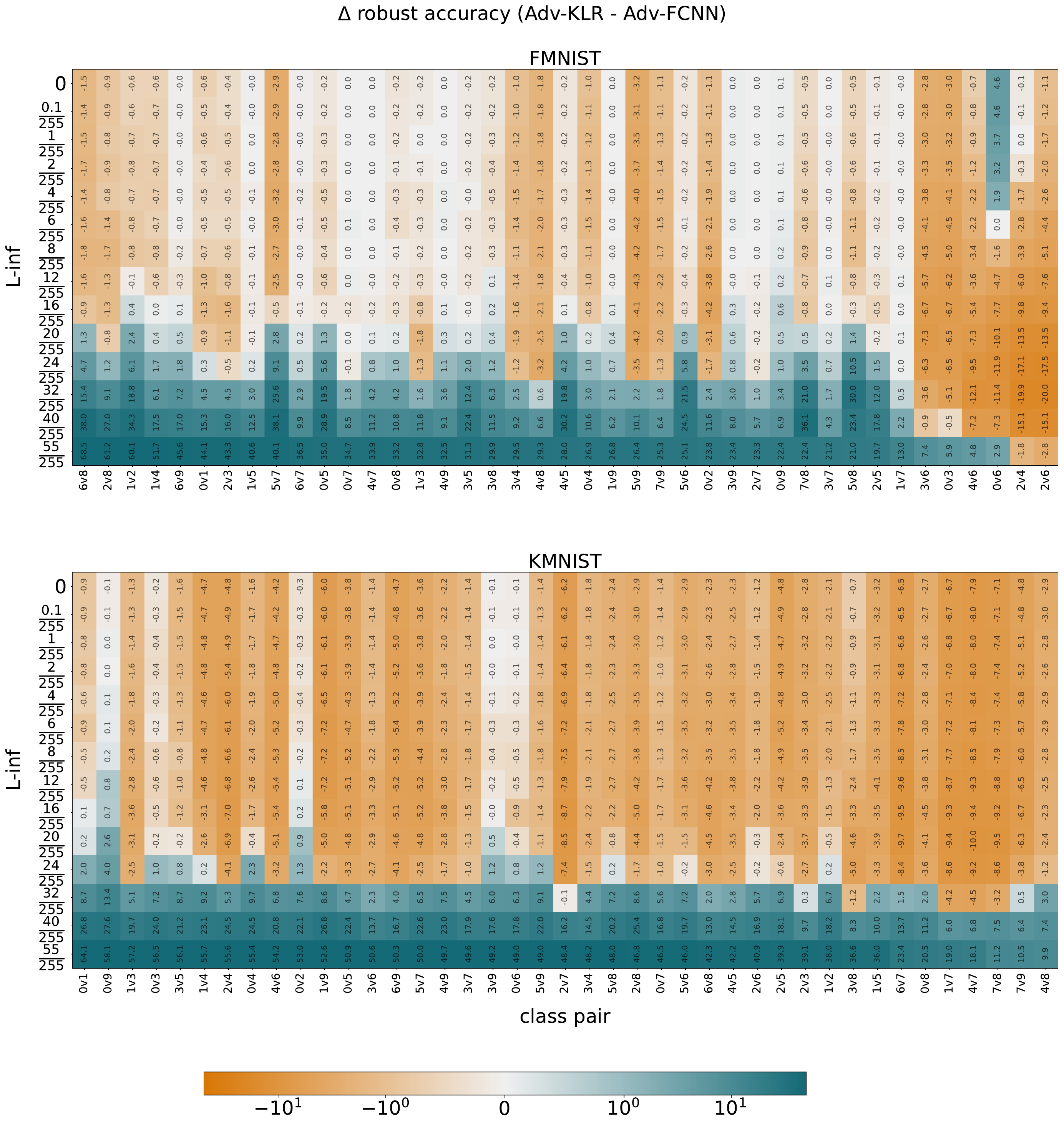}
\label{fig:adv_klr_adv_fcnn}
\end{figure*}

\begin{figure*}[!htbp]
\centering
    \caption{\label{fig:gap_lapvsadvnn_l2_1} \textbf{Laplacian KRR versus $\ell_2=1.0$ norm and $\ell_\infty = \frac{6}{255}$ adversarially trained FCNNs:} $\ell_2$ and $\ell_\infty$ norm adversarially trained FCNNs were attacked at their respective norm adversarial perturbations for FashionMNIST, KMNIST and QMNIST. (Top) row corresponds to the results for FashionMNIST dataset, (Middle) row corresponds to the KMNIST dataset, and (Bottom) row corresponds to the QMNIST dataset. Of these, (First) column corresponds to the robust accuracies for $\ell_\infty$ adversaries while (Second) column portrays the robust accuracies for $\ell_2$ adversaries. PGD is used to attack the models.}
\includegraphics[width=1\textwidth]{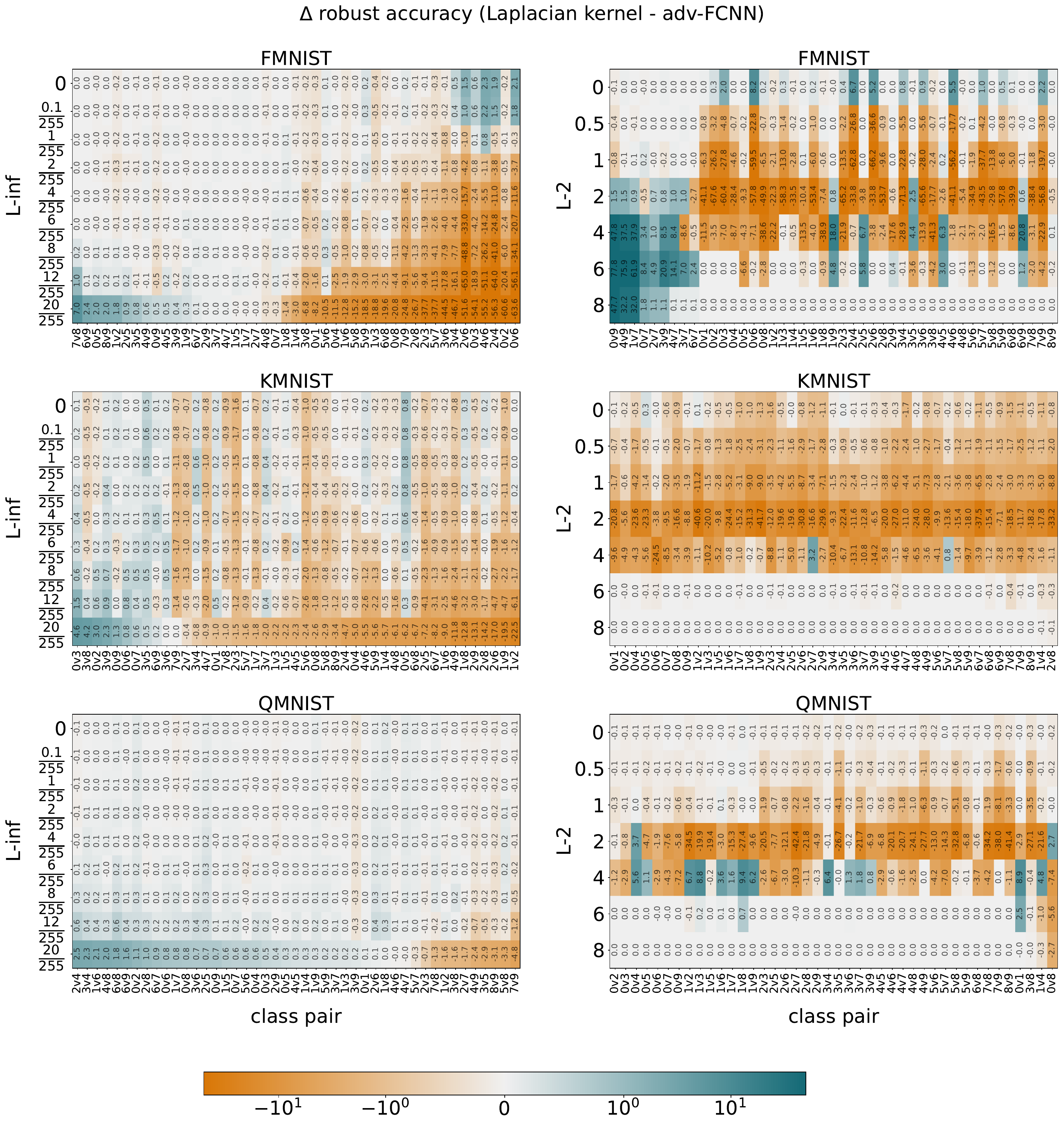}

\end{figure*}

\cref{fig:adv_klr_minus_klr,fig:adv_klr_fcnn,fig:gap_adv_klr_adv_fcnn,fig:gap_lapvsadvnn_l2_1} contains the fine-grained $1\text{v}1$ results. The multiclass results are in~\cref{fig:fmnist_mc,fig:kmnist_mc,fig:qmnist_mc,fig:fmnist_kmnist_qmnist_mc_l2}.

\vspace{2.5mm}

\vspace{2.5mm}

\vspace{2.5mm}

\clearpage
\section{Further Ablations}\label{appsec: further_ablations}

\subsection{Results for gradient-free (black-box) attacks}\label{apd: bb_results}

\begin{figure*}[!htbp]
\centering
    \caption{\label{fig:gap_spsa_vs_pgd_lap} \textbf{SPSA attacked Laplacian KRR versus PGD attacked Laplacian KRR}. (Top) row corresponds to the results for FashionMNIST dataset, (Middle) row corresponds to the KMNIST dataset, and (Bottom) row corresponds to the QMNIST dataset. Of these, (First) column corresponds to the robust accuracies for $\ell_\infty$ adversaries while (Second) column portrays the robust accuracies for $\ell_2$ adversaries.}
\includegraphics[width=1\textwidth]{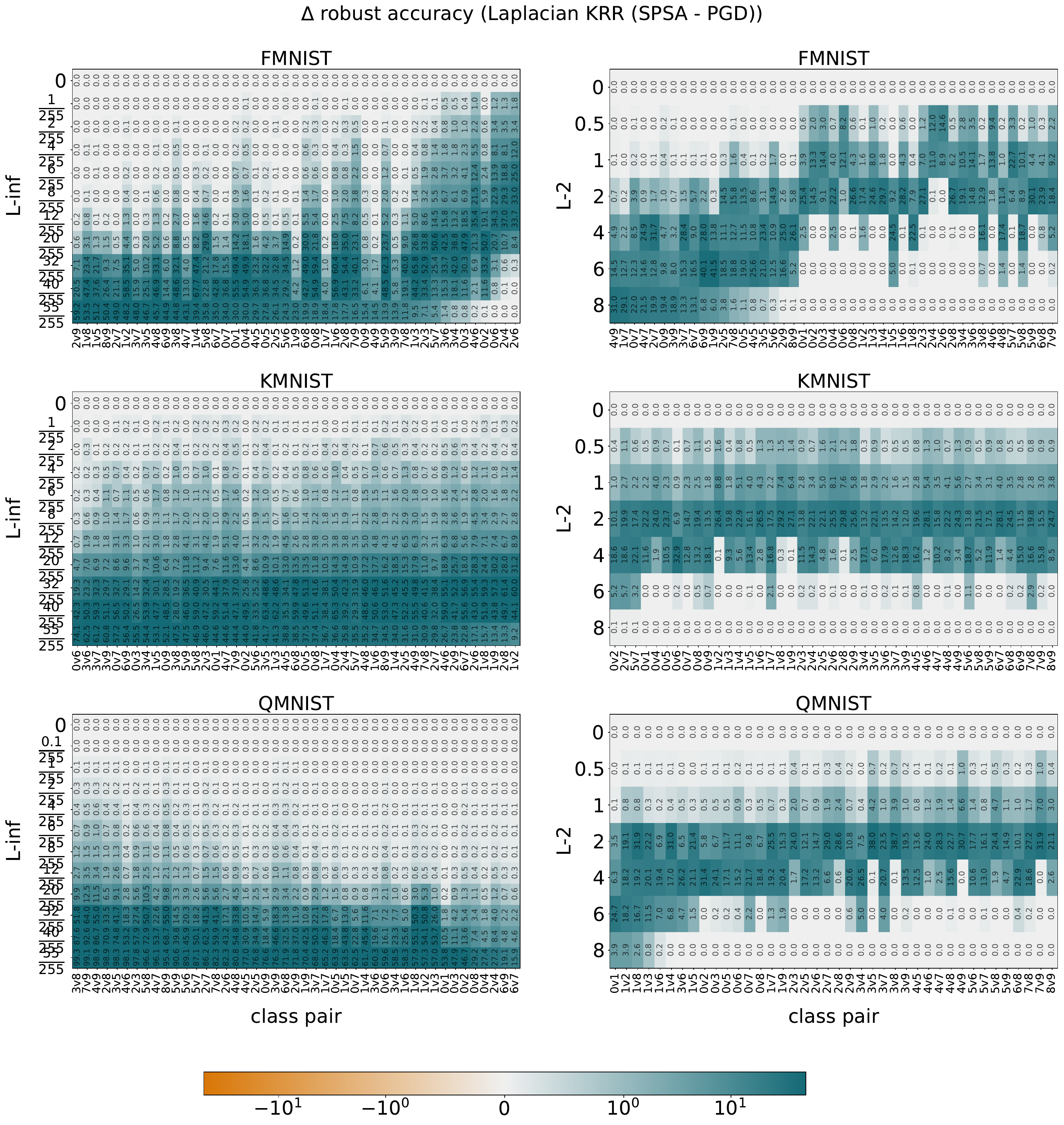}

\end{figure*}

\begin{figure*}[!htbp]
\centering
    \caption{\label{fig:gap_spsa_vs_pgd_lap_rfm} \textbf{SPSA attacked Laplacian RFM versus PGD attacked Laplacian RFM}. (Top) row corresponds to the results for FashionMNIST dataset, (Middle) row corresponds to the KMNIST dataset, and (Bottom) row corresponds to the QMNIST dataset. Of these, (First) column corresponds to the robust accuracies for $\ell_\infty$ adversaries while (Second) column portrays the robust accuracies for $\ell_2$ adversaries.}
\includegraphics[width=1\textwidth]{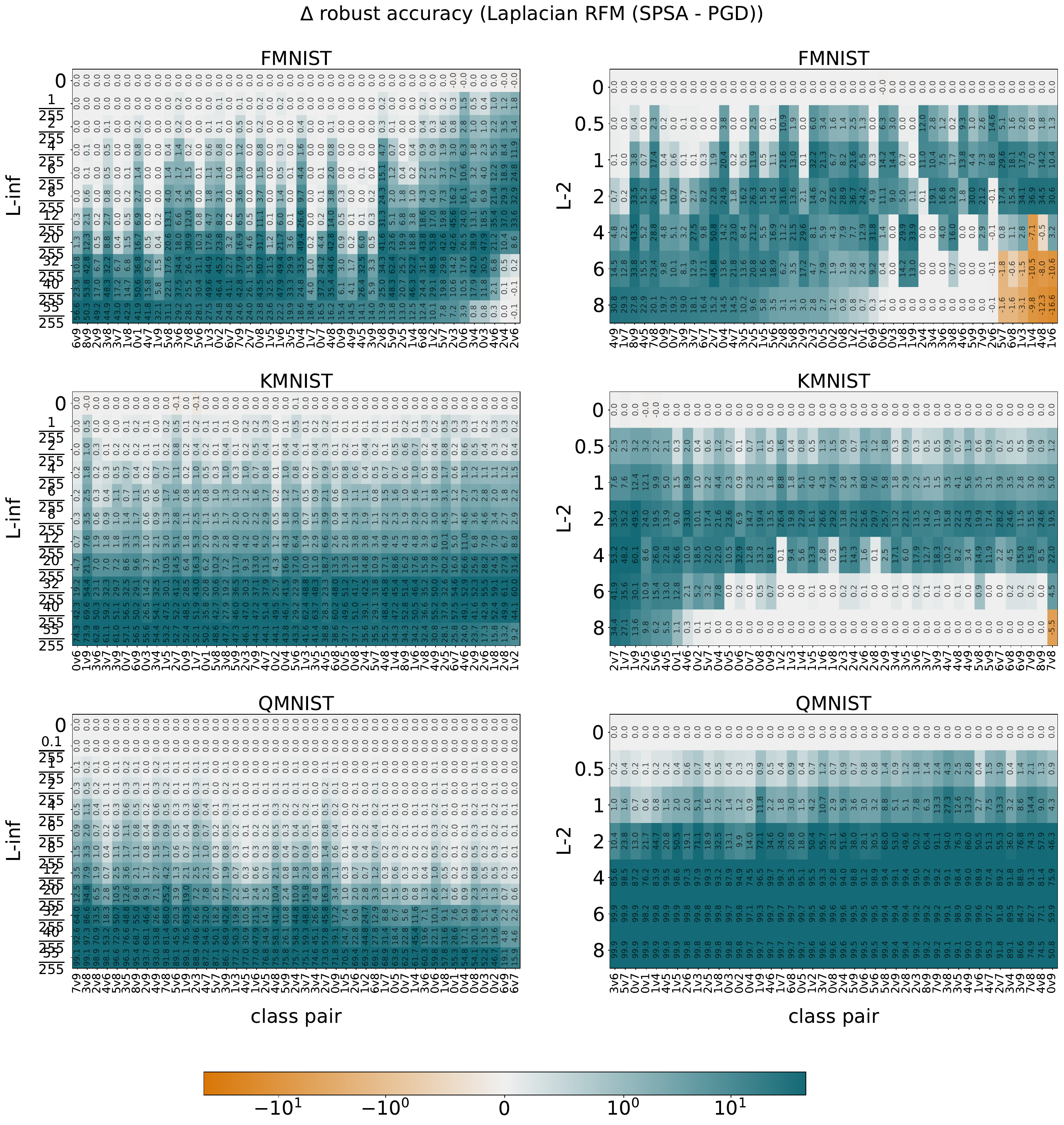}

\end{figure*}

\begin{figure*}[!htbp]
\centering
    \caption{\label{fig:gap_spsa_vs_pgd_gaus} \textbf{SPSA attacked Gaussian KRR versus PGD attacked Gaussian KRR}. (Top) row corresponds to the results for FashionMNIST dataset, (Middle) row corresponds to the KMNIST dataset, and (Bottom) row corresponds to the QMNIST dataset. Of these, (First) column corresponds to the robust accuracies for $\ell_\infty$ adversaries while (Second) column portrays the robust accuracies for $\ell_2$ adversaries.}
\includegraphics[width=1\textwidth]{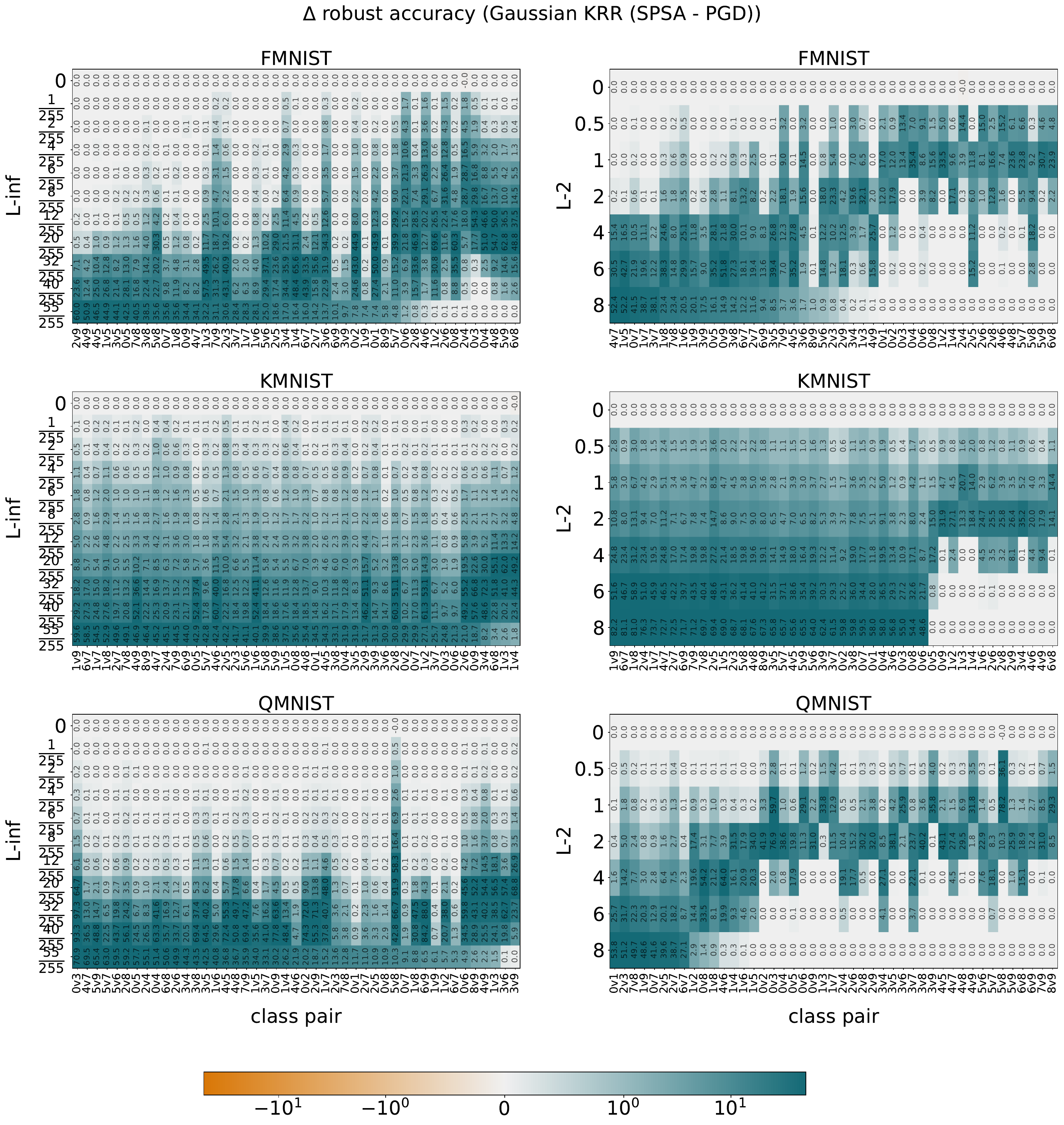}

\end{figure*}

\begin{figure*}[!htbp]
\centering
    \caption{\label{fig:gap_spsa_vs_pgd_gaus_rfm} \textbf{SPSA attacked Gaussian RFM versus PGD attacked Gaussian RFM}. (Top) row corresponds to the results for FashionMNIST dataset, (Middle) row corresponds to the KMNIST dataset, and (Bottom) row corresponds to the QMNIST dataset. Of these, (First) column corresponds to the robust accuracies for $\ell_\infty$ adversaries while (Second) column portrays the robust accuracies for $\ell_2$ adversaries.}
\includegraphics[width=1\textwidth]{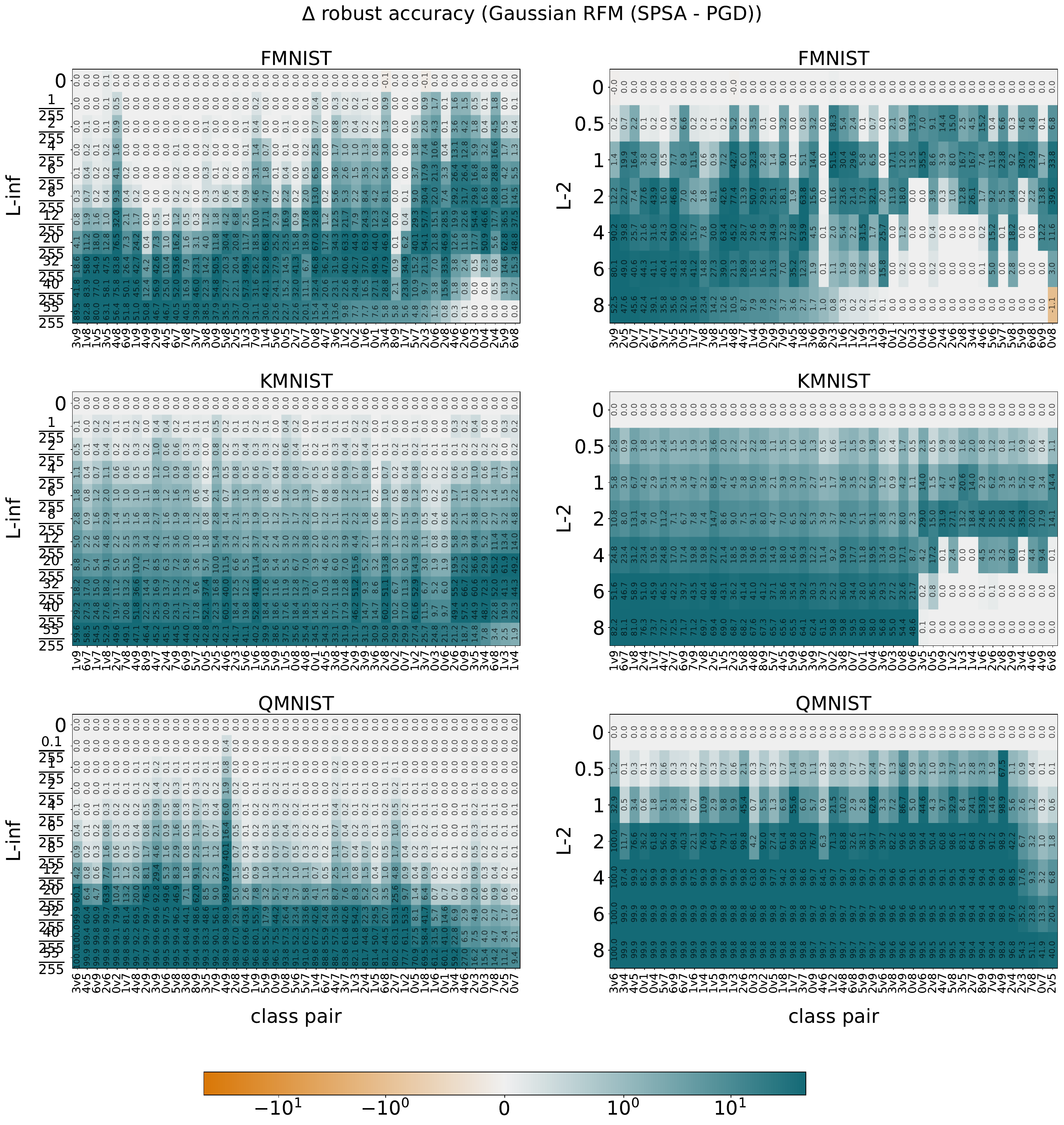}

\end{figure*}

\begin{figure*}[!htbp]
\centering
    \caption{\label{fig:gap_spsa_vs_pgd_fcnn} \textbf{SPSA attacked FCNN versus PGD attacked FCNN}. (Top) row corresponds to the results for FashionMNIST dataset, (Middle) row corresponds to the KMNIST dataset, and (Bottom) row corresponds to the QMNIST dataset. Of these, (First) column corresponds to the robust accuracies for $\ell_\infty$ adversaries while (Second) column portrays the robust accuracies for $\ell_2$ adversaries.}
\includegraphics[width=1\textwidth]{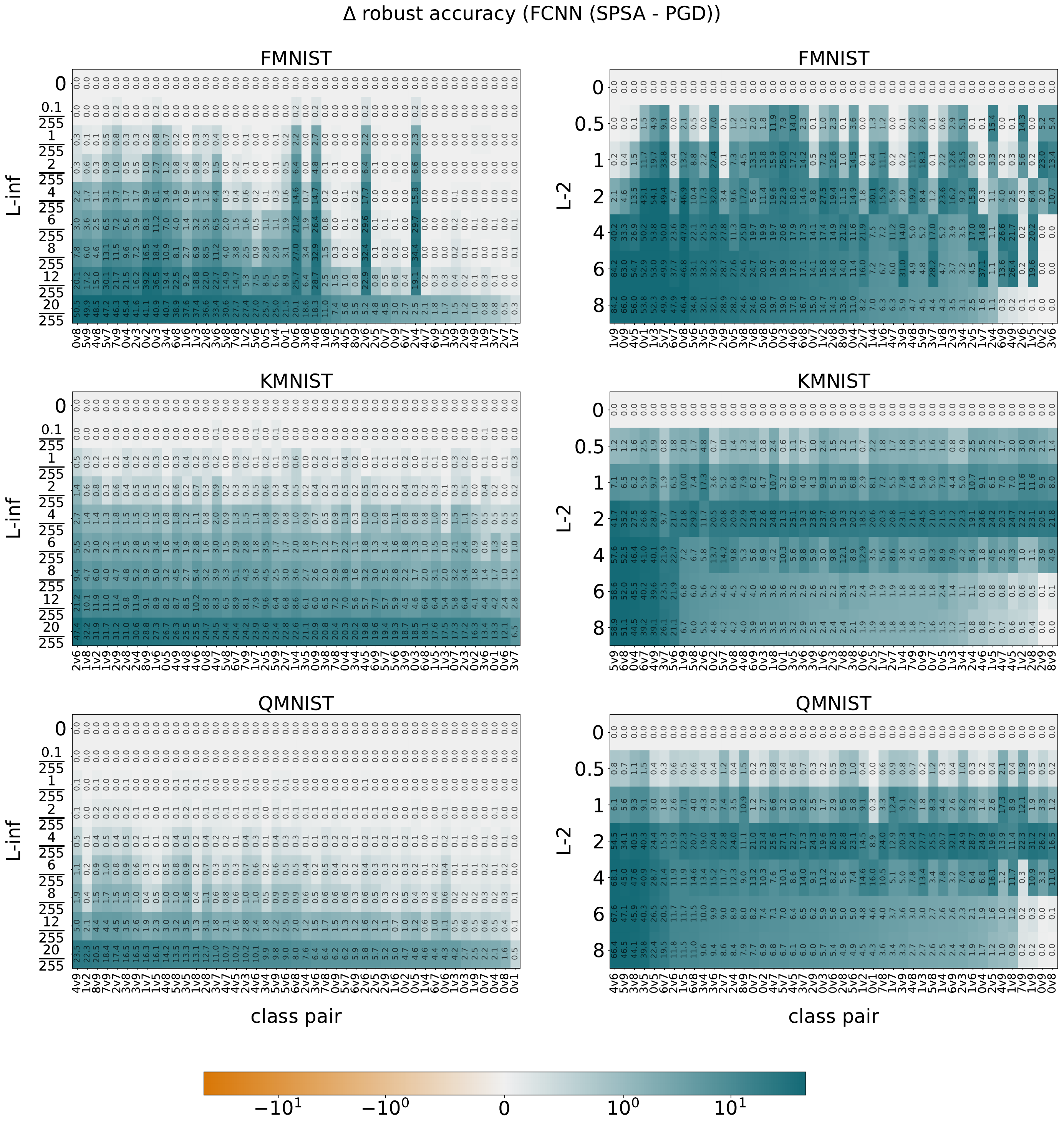}
\end{figure*}

\begin{figure*}[!htbp]
\centering
    \caption{\label{fig:gap_spsa_vs_pgd_lenet} \textbf{SPSA attacked LeNet versus PGD attacked LeNet}. (Top) row corresponds to the results for FashionMNIST dataset, (Middle) row corresponds to the KMNIST dataset, and (Bottom) row corresponds to the QMNIST dataset. Of these, (First) column corresponds to the robust accuracies for $\ell_\infty$ adversaries while (Second) column portrays the robust accuracies for $\ell_2$ adversaries.}
\includegraphics[width=1\textwidth]{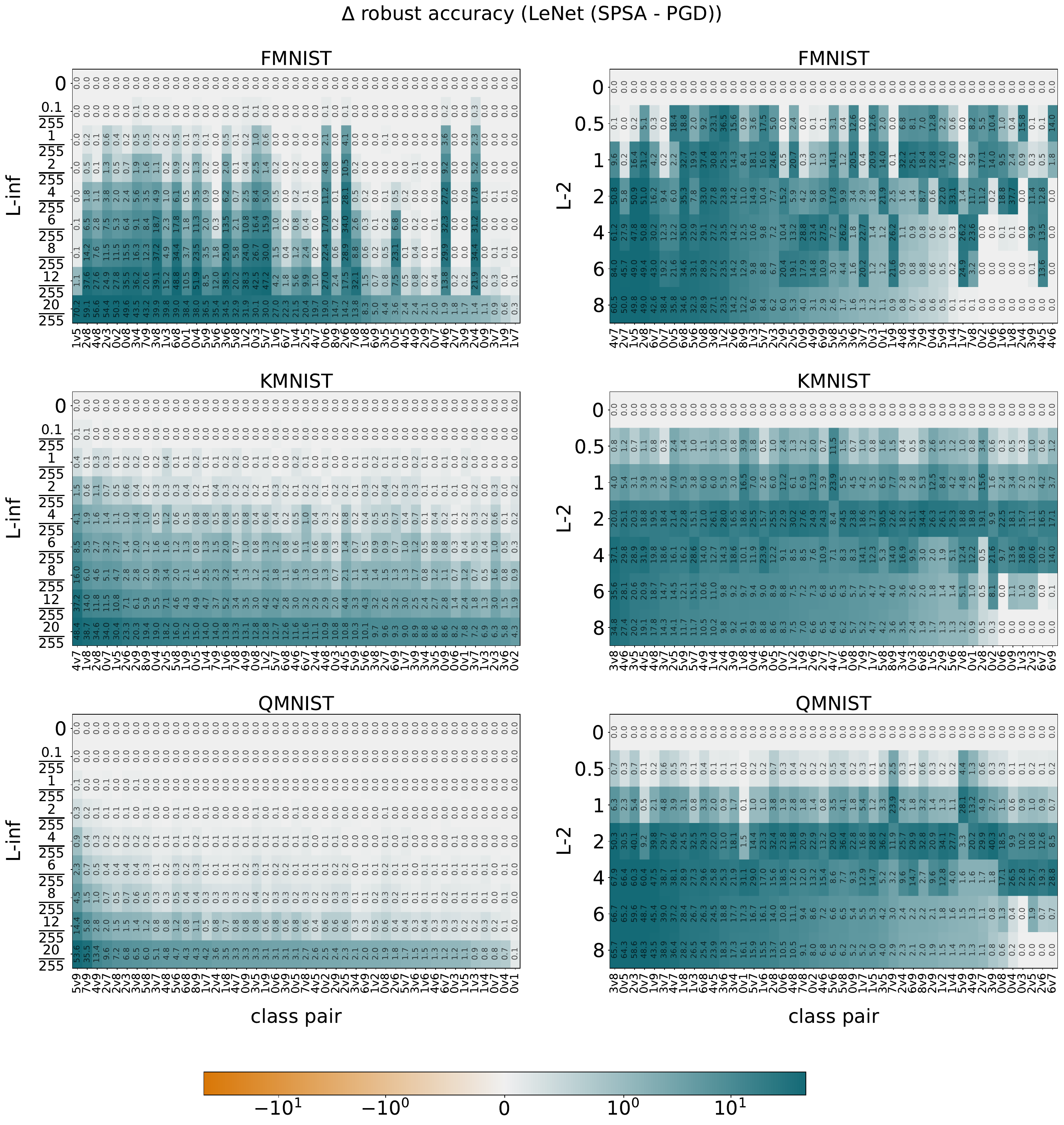}
\end{figure*}

Carlini et al.~\citep{carlini2019evaluating} strongly recommend the use of gradient-free attacks to ensure that robustness is not over-estimated due to gradient masking. We use SPSA as recommended by Uesato et. al.~\citep{uesato2018adversarial}. The number of samples used to approximate the gradient is $16$. For KLR the number of samples used is $32$. For both $\ell_\infty$ and $\ell_2$ attacks, the step size in the finite difference step is chosen as $\alpha=\frac{\eps}{2}$, with lower bound and upper bound being $0.0001$, and $0.001$ respectively. In the case of $\ell_2$ attacks and $\ell_\infty$, we set PGD step size to be $\frac{\eps}{10}$. Our results in \cref{tab:attack_comparison_spsa_pgd_lap,tab:attack_comparison_spsa_autoattack_gaussiansvc,tab:attack_comparison_spsa_pgd_adv_klr,tab:attack_comparison_spsa_pgd_klr,tab:attack_comparison_spsa_pgd_lap_klr,tab:attack_comparison_spsa_pgd_laplacian_svc} and~\cref{fig:gap_spsa_vs_pgd_lap}, ~\cref{fig:gap_spsa_vs_pgd_lap_rfm}, ~\cref{fig:gap_spsa_vs_pgd_gaus}, ~\cref{fig:gap_spsa_vs_pgd_gaus_rfm}, ~\cref{fig:gap_spsa_vs_pgd_fcnn},
~\cref{fig:gap_spsa_vs_pgd_lenet} demonstrate that SPSA is a weaker attack than PGD, indicating \emph{no gradient masking}, except for the case of Laplace RFC.

We attack the Laplacian KRR, Laplacian RFM, Gaussian KRR, Gaussian RFM, Laplacian KLR, Gaussian KLR, Laplacian SVM, Gaussian SVM, Adversarially trained Gaussian KLR, FCNN and LeNet on FashionMNIST, KMNIST, and QMNIST using SPSA and compare its robust accuracy to that under PGD attacks. We can see the SPSA attack is weaker than the standard PGD, hence confirming there is no gradient masking for the aforementioned models. For the multiclass case, \cref{tab:attack_comparison_spsa_pgd_lap} summarizes this result for the $\ell_\infty$ attack for Laplacian KRR.
~\cref{tab:attack_comparison_spsa_pgd_klr} has the SPSA vs. PGD results for $\ell_\infty$ attack for Gaussian KLR.~\cref{tab:attack_comparison_spsa_pgd_lap_klr} contains SPSA results for Laplacian KLR.~\cref{tab:attack_comparison_spsa_pgd_adv_klr} contains the SPSA vs. PGD results for Adversarially trained Gaussian KLR. ~\cref{tab:attack_comparison_spsa_pgd_laplacian_svc} contains the SPSA vs. PGD results for Laplacian SVC.~\cref{tab:attack_comparison_spsa_autoattack_gaussiansvc} contains the SPSA vs. PGD results for Gaussian SVC.~\cref{tab:fmnist_gaussian_rff_pgd_spsa,tab:kmnist_laplacian_rff_pgd_spsa} contains preliminary results for Gaussian RFC and Laplacian RFC respectively. From~\cref{tab:kmnist_laplacian_rff_pgd_spsa}, we can see that \emph{Laplace RFCs exhibit gradient masking}. Comprehensive results corresponding to RFCs are given in~\cref{apd: num_of_ftr_vs_robust_results}. 

\begin{table*}[!htbp]
\centering
\caption{Comparison of robust accuracies of Laplacian KRR in the multi-class setting for SPSA, PGD, and the difference of the two across datasets for $\ell_\infty$ perturbations. The difference is computed as SPSA minus PGD.}
\label{tab:attack_comparison_spsa_pgd_lap}
\begin{tabular*}{\textwidth}{@{\extracolsep{\fill}}l*{9}{c}}
\toprule
 & \multicolumn{3}{c}{FMNIST}
 & \multicolumn{3}{c}{KMNIST}
 & \multicolumn{3}{c}{QMNIST} \\
\cmidrule(lr){2-4}
\cmidrule(lr){5-7}
\cmidrule(lr){8-10}
$\varepsilon$
& \textrm{ra}(SPSA) & \textrm{ra}(PGD) & diff
& \textrm{ra}(SPSA) & \textrm{ra}(PGD) & diff
& \textrm{ra}(SPSA) & \textrm{ra}(PGD) & diff \\
\midrule
$0$ & $90.26$ & $90.26$ & $0.00$  & $92.62$ & $92.62$ & $0.00$  & $98.23$ & $98.23$ & $0.00$ \\

$\frac{1}{255}$ & $89.28$ & $88.57$ & $0.71$ & $92.23$ & $91.95$ & $0.28$ & $98.12$ & $98.05$ & $0.07$ \\
$\frac{2}{255}$ & $88.16$ & $86.14$ & $2.02$ & $91.92$ & $91.33$ & $0.59$ & $98.02$ & $97.91$ & $0.11$ \\
$\frac{4}{255}$ & $85.25$ & $81.19$ & $4.06$ & $91.3$ & $89.95$ & $1.35$ & $97.86$ & $97.51$ & $0.35$ \\
$\frac{6}{255}$ & $82.34$  & $74.02$ & $8.32$ & $90.47$ & $88.29$ & $2.18$ & $97.62$ & $97.06$ & $0.56$ \\
$\frac{8}{255}$ & $78.63$ & $66.33$ & $12.30$ & $89.73$ & $86.67$ & $3.06$ & $97.35$ & $96.52$ & $0.83$ \\
$\frac{12}{255}$ & $69.83$ & $51.74$ & $18.09$ & $87.93$ & $82.49$ & $5.44$ & $96.74$ & $95.11$ & $1.63$ \\
$\frac{20}{255}$ & $51.83$ & $30.35$ & $21.48$ & $83.35$ & $69.53$ & $13.82$ & $95.02$ & $89.59$ & $5.43$ \\
$\frac{32}{255}$ & $31.68$  & $11.73$ & $19.95$ & $72.79$ & $35.47$ & $37.32$ & $89.17$ & $61.80$ & $27.37$ \\
$\frac{40}{255}$ & $22.36$ & $3.20$ & $19.16$ & $61.42$ & $13.52$ & $47.90$ & $79.32$ & $30.93$ & $48.40$ \\
\bottomrule
\end{tabular*}
\end{table*}

\begin{table*}[!htbp]
\centering
\caption{Comparison of robust accuracies of Gaussian KLR in the
multi-class setting for SPSA, PGD, and the difference between the two across datasets for $\ell_\infty$ perturbations. The difference is computed as SPSA minus PGD.}
\label{tab:attack_comparison_spsa_pgd_klr}
\begin{tabular*}{\textwidth}{@{\extracolsep{\fill}}l*{6}{c}}
\toprule
 & \multicolumn{3}{c}{FMNIST}
 & \multicolumn{3}{c}{QMNIST} \\
\cmidrule(lr){2-4}
\cmidrule(lr){5-7}
$\varepsilon$
& \textrm{ra}(SPSA) & \textrm{ra}(PGD) & diff
& \textrm{ra}(SPSA) & \textrm{ra}(PGD) & diff \\
\midrule
$0$ & $86.95$ & $86.95$ & $0.00$ & $97.47$ & $97.47$ & $0.00$ \\
$\tfrac{1}{255}$ & $86.10$ & $85.56$ & $0.54$ & $97.37$ & $97.33$ & $0.05$ \\
$\tfrac{8}{255}$ & $78.51$ & $72.30$ & $6.21$ & $96.32$ & $95.66$ & $0.66$ \\
$\tfrac{55}{255}$ & $43.25$ & $35.07$ & $8.18$ & $82.31$ & $65.17$ & $17.13$ \\
\bottomrule
\end{tabular*}
\end{table*}

\begin{table*}[!htbp]
\centering
\caption{Comparison of robust accuracies of Laplacian KLR in the
multi-class setting under SPSA and PGD attacks across FMNIST
and KMNIST for $\ell_\infty$ perturbations.
The difference is computed as SPSA minus PGD.
All values are reported in percentages.}
\label{tab:attack_comparison_spsa_pgd_lap_klr}

\begin{tabular*}{\textwidth}{@{\extracolsep{\fill}}l*{6}{c}}
\toprule
 & \multicolumn{3}{c}{FMNIST}
 & \multicolumn{3}{c}{KMNIST} \\

\cmidrule(lr){2-4}
\cmidrule(lr){5-7}

$\varepsilon$
& \textrm{ra}(SPSA)
& \textrm{ra}(PGD)
& diff
& \textrm{ra}(SPSA)
& \textrm{ra}(PGD)
& diff \\

\midrule

$0$
& $88.18$ & $88.18$ & $0.00$
& $92.26$ & $92.26$ & $0.00$ \\

$\frac{1}{255}$
& $87.10$ & $86.64$ & $0.46$
& $91.82$ & $91.56$ & $0.26$ \\

$\frac{2}{255}$
& $85.86$ & $84.99$ & $0.87$
& $91.39$ & $90.92$ & $0.47$ \\

$\frac{4}{255}$
& $83.56$ & $81.44$ & $2.12$
& $90.29$ & $89.31$ & $0.98$ \\

$\frac{8}{255}$
& $77.93$ & $73.03$ & $4.90$
& $88.00$ & $85.08$ & $2.92$ \\

$\frac{12}{255}$
& $71.72$ & $63.73$ & $7.99$
& $85.05$ & $79.24$ & $5.81$ \\

$\frac{20}{255}$
& $59.20$ & $48.19$ & $11.01$
& $76.55$ & $62.61$ & $13.94$ \\

$\frac{32}{255}$
& $45.04$ & $33.83$ & $11.21$
& $58.34$ & $29.30$ & $29.04$ \\

$\frac{40}{255}$
& $38.01$ & $30.61$ & $7.40$
& $42.70$ & $12.49$ & $30.21$ \\

$\frac{55}{255}$
& $30.38$ & $26.00$ & $4.38$
& $20.02$ & $1.31$ & $18.71$ \\

\bottomrule
\end{tabular*}
\end{table*}

\begin{table*}[!htbp]
\centering
\caption{Comparison of robust accuracies of adversarially trained
Gaussian KLR in the multi-class setting under SPSA and PGD
attacks for $\ell_\infty$ perturbations on KMNIST. The difference
is computed as $\mathrm{ra(SPSA)}-\mathrm{ra(PGD)}$.}
\label{tab:attack_comparison_spsa_pgd_adv_klr}
\begin{tabular*}{\textwidth}{@{\extracolsep{\fill}}l*{3}{c}}
\toprule
& \multicolumn{3}{c}{KMNIST} \\
\cmidrule(lr){2-4}
$\varepsilon$
& \textrm{ra}(SPSA)
& \textrm{ra}(PGD)
& diff \\
\midrule
$0$               & $88.40$ & $88.40$ & $0.00$ \\
$\tfrac{1}{255}$   & $88.35$ & $88.24$ & $0.11$ \\
$\tfrac{8}{255}$   & $87.26$ & $86.27$ & $0.99$ \\
$\tfrac{20}{255}$  & $85.24$ & $82.40$ & $2.84$ \\
$\tfrac{32}{255}$  & $82.78$ & $77.63$ & $5.15$ \\
$\tfrac{55}{255}$  & $77.17$ & $66.61$ & $10.56$ \\
\bottomrule
\end{tabular*}
\end{table*}

\begin{table*}[!htbp]
\centering
\renewcommand{\arraystretch}{1.5}
\caption{Comparison of robust accuracies of Laplacian SVM in the
multi-class setting for SPSA, PGD, and the difference between
the two across datasets for $\ell_\infty$ perturbations.}
\label{tab:attack_comparison_spsa_pgd_laplacian_svc}
\begin{tabular*}{\textwidth}{@{\extracolsep{\fill}}l*{6}{c}}
\toprule
 & \multicolumn{3}{c}{FMNIST}
 & \multicolumn{3}{c}{KMNIST} \\
\cmidrule(lr){2-4}
\cmidrule(lr){5-7}
$\varepsilon$
& \textrm{ra}(SPSA) & \textrm{ra}(PGD) & diff
& \textrm{ra}(SPSA) & \textrm{ra}(PGD) & diff \\
\midrule
$0$               & $90.01$ & $90.01$ & $0.00$ & $90.54$ & $90.54$ & $0.00$ \\
$\tfrac{0.1}{255}$ & $89.90$ & $89.87$ & $0.03$  & $90.53$ & $90.47$ & $0.06$ \\
$\tfrac{1}{255}$   & $89.17$ & $88.39$ & $0.78$  & $90.27$ & $89.87$ & $0.40$ \\
$\tfrac{2}{255}$   & $88.38$ & $86.63$ & $1.75$  & $90.03$ & $89.02$ & $1.01$ \\
$\tfrac{4}{255}$   & $86.76$ & $82.18$ & $4.58$  & $89.43$ & $87.34$ & $2.09$ \\
$\tfrac{6}{255}$   & $84.86$ & $76.01$ & $8.85$  & $88.85$ & $85.41$ & $3.44$ \\
$\tfrac{8}{255}$   & $82.87$ & $69.11$ & $13.76$ & $88.32$ & $82.89$ & $5.43$ \\
$\tfrac{12}{255}$  & $78.20$ & $55.21$ & $22.99$ & $87.02$ & $76.84$ & $10.18$ \\
$\tfrac{20}{255}$  & $66.61$ & $34.76$ & $31.85$ & $83.66$ & $60.43$ & $23.23$ \\
$\tfrac{32}{255}$  & $49.28$ & $12.67$ & $36.61$ & $76.26$ & $26.91$ & $49.35$ \\
$\tfrac{40}{255}$  & $40.51$ & $4.50$  & $36.01$ & $69.24$ & $11.37$ & $57.87$ \\
$\tfrac{55}{255}$  & $26.59$ & $0.27$  & $26.32$ & $51.21$ & $0.91$  & $50.30$ \\
\bottomrule
\end{tabular*}
\end{table*}

\begin{table*}[!htbp]
\centering
\caption{Comparison of robust accuracies of Gaussian SVM in the
multi-class setting under SPSA and PGD attack,
along with their differences across FMNIST and KMNIST datasets
for $\ell_\infty$ perturbations.}
\label{tab:attack_comparison_spsa_autoattack_gaussiansvc}
\begin{tabular*}{\textwidth}{@{\extracolsep{\fill}}l*{6}{c}}
\toprule
 & \multicolumn{3}{c}{FMNIST}
 & \multicolumn{3}{c}{KMNIST} \\
\cmidrule(lr){2-4}
\cmidrule(lr){5-7}
$\varepsilon$
& \textrm{ra}(SPSA) & \textrm{ra}(PGD) & diff
& \textrm{ra}(SPSA) & \textrm{ra}(PGD) & diff \\
\midrule
$0$                 & $89.62$ & $89.62$ & $0.00$  & $93.14$ & $93.14$ & $0.00$ \\
$\tfrac{0.1}{255}$   & $89.49$ & $89.29$ & $0.20$  & $93.09$ & $93.00$ & $0.09$ \\
$\tfrac{1}{255}$     & $88.42$ & $86.28$ & $2.14$  & $92.76$ & $92.11$ & $0.65$ \\
$\tfrac{2}{255}$     & $86.91$ & $82.41$ & $4.50$  & $92.40$ & $91.02$ & $1.38$ \\
$\tfrac{4}{255}$     & $84.02$ & $73.17$ & $10.85$ & $91.72$ & $88.74$ & $2.98$ \\
$\tfrac{6}{255}$     & $80.98$ & $61.94$ & $19.04$ & $90.88$ & $86.09$ & $4.79$ \\
$\tfrac{8}{255}$     & $77.25$ & $50.67$ & $26.58$ & $90.10$ & $82.99$ & $7.11$ \\
$\tfrac{12}{255}$    & $68.34$ & $32.45$ & $35.89$ & $88.26$ & $74.43$ & $13.83$ \\
$\tfrac{20}{255}$    & $50.23$ & $8.29$  & $41.94$ & $83.44$ & $49.51$ & $33.93$ \\
$\tfrac{32}{255}$    & $28.78$ & $0.19$  & $28.59$ & $71.08$ & $13.68$ & $57.40$ \\
$\tfrac{40}{255}$    & $18.90$ & $0.01$  & $18.89$ & $58.86$ & $3.89$  & $54.97$ \\
$\tfrac{55}{255}$    & $7.60$  & $0.00$  & $7.60$  & $35.80$ & $0.13$  & $35.67$ \\
\bottomrule
\end{tabular*}
\end{table*}

\begin{table}[htbp]
\centering
\caption{
Comparison of PGD and SPSA robust accuracy (\%) for
Gaussian Random Feature Classifiers (RFC) trained with
cross-entropy loss on Fashion-MNIST.
The model uses 50,000 random features and random seed 150.
}
\label{tab:fmnist_gaussian_rff_pgd_spsa}
\begin{tabular}{c|cc}
\hline
\textbf{$\varepsilon$ ($L_\infty$)}
& \textbf{PGD (\%)}
& \textbf{SPSA (\%)} \\
\hline
$0$        & 89.66 & 89.66 \\
$0.1/255$  & 89.28 & 89.34 \\
$1/255$    & 84.63 & 85.69 \\
$2/255$    & 78.54 & 81.16 \\
$4/255$    & 63.72 & 69.58 \\
$6/255$    & 51.09 & 58.07 \\
$8/255$    & 42.91 & 49.19 \\
$12/255$   & 28.26 & 37.12 \\
$20/255$   &  6.85 & 15.96 \\
$32/255$   &  0.07 &  1.41 \\
$40/255$   &  0.00 &  0.17 \\
$55/255$   &  0.00 &  0.07 \\
\hline
\end{tabular}
\end{table}

\begin{table}[htbp]
\centering
\caption{
Comparison of PGD and SPSA robust accuracy (\%) for
Laplacian Random Feature Classifiers (RFC) trained with
cross-entropy loss on KMNIST.
The model uses 50,000 random features and random seed 150.
}
\label{tab:kmnist_laplacian_rff_pgd_spsa}
\begin{tabular}{c|cc}
\hline
\textbf{$\varepsilon$ ($L_\infty$)}
& \textbf{PGD (\%)}
& \textbf{SPSA (\%)} \\
\hline
$0$        & 89.43 & 89.43 \\
$0.1/255$  & 88.46 & 87.57 \\
$1/255$    & 87.22 & 77.11 \\
$2/255$    & 86.66 & 69.48 \\
$4/255$    & 85.84 & 56.25 \\
$6/255$    & 85.20 & 42.82 \\
$8/255$    & 84.62 & 30.12 \\
$12/255$   & 83.59 & 15.28 \\
$20/255$   & 81.08 &  8.55 \\
$32/255$   & 75.82 &  8.08 \\
$40/255$   & 71.15 &  8.07 \\
$55/255$   & 59.37 &  8.04 \\
\hline
\end{tabular}
\end{table}

\subsection{Robustness on Tabular datasets}
\label{apd: tabular_data}
Recent work~\citep{radhakrishnan2024mechanism,beaglehole2025xrfm} has shown that RFMs and their variants are state-of-the-art on tabular data benchmarks. Given this, it is instructive to determine whether kernel-based classifiers are also \emph{more robust} than neural networks on tabular data. However, since tabular data has dimension-wise semantic constraints, we use the Cost-aware Feasible Attack (CaFA) method from~\citep{bentov2024cafa} to ensure that all generated adversarial examples remain feasible. Further details about the attack are in~\cref{apd: cafa_attack}.

\noindent \textbf{Results:} When running CaFA, it marks points as infeasible if they do not satisfy the automatically determined integrity constraints. All accuracies in~\cref{tab:tabular_cafa_attacks} are only reported on the portion of the dataset found to be feasible. We find that on the Phishing and Bank datasets, KMs are indeed more robust. However, the FCNN is comfortably more robust on the Adult dataset. This is the only dataset where we find KMs to be less robust, and we plan to investigate the reason for this in future work.
\begin{table*}[!htbp]
\centering
\caption{\textbf{CaFA attacks on FCNN, KRR and RFM} on the Phishing, Bank Marketing and Adult datasets. Robust accuracy numbers are reported only on compliant samples after each stage. The accuracy values are rounded to the nearest tenth.}
\small
\label{tab:tabular_cafa_attacks}
\begin{tabular}{ccccccccccc}
\toprule
&& \multicolumn{3}{c}{Phishing} 
& \multicolumn{3}{c}{Bank Marketing}
& \multicolumn{3}{c}{Adult}
\\
\cmidrule(lr){3-5}
\cmidrule(lr){6-8}
\cmidrule(lr){9-11}
$\varepsilon$ & stage & FCNN & KRR & RFM 
&  FCNN & KRR & RFM
&  FCNN & KRR & RFM
\\
\midrule
\multirow{3}{*}{0.01}
& Benign & 
94.8 & 92.6 & \textbf{95.2} & 
87.8 & \textbf{89.7} & \textbf{89.7} & 
\textbf{84.3} & 72.4 & 80.8 \\
& Post CaFA & 
53.0 & 52.4 & \textbf{70.5} & 
\textbf{90.3} & 89.5 & 78.6 & 
\textbf{73.0} &61.2 & 43.1 \\
& Post projection   & 
42.0 & 31.1 & \textbf{63.5} & 
84.3 & \textbf{85.7} & 75.0 & 
\textbf{64.2} & 51.2& 38.0 \\
\midrule

\multirow{3}{*}{0.03}
& Benign & 
94.8 & 92.6 & \textbf{95.2} & 
87.8 & \textbf{89.7} & \textbf{89.7} & 
\textbf{84.3} & 72.4 & 80.8\\
& Post CaFA & 
46.7 & 44.3 & \textbf{60.6} & 
89.5 & \textbf{90.5} & 79.9 & 
\textbf{69.6} & 55.6 & 39.2\\ 
& Post projection & 38.1 & 36.8 & \textbf{56.4} & 
80.8 & \textbf{85.6} & 66.2& 
\textbf{61.3}& 45.4& 33.7\\
\midrule

\multirow{3}{*}{0.075}
& Benign & 94.8 & 92.6 & \textbf{95.2} & 
87.8 & \textbf{89.7} & \textbf{89.7} & 
\textbf{84.3} &72.4& 80.8\\
& Post CaFA & 42.8 & 42.3 & \textbf{49.6} & 
78.5& \textbf{91.9} & 76.4& 
\textbf{68.5} & 49.1 & 36.6 \\
& Post projection & 31.5 & 32.2 & \textbf{41.8} & 
64.6 & \textbf{85.9} & 70.8& 
\textbf{60.0} & 40.0& 30.4\\
\bottomrule
\end{tabular}
\vspace{-4ex}
\end{table*}

\clearpage
\subsection{Results for Mat\'ern Kernel \texorpdfstring{$\nu$}{nu} versus robust accuracy}\label{apd: mat_results}
\begin{figure*}[!htbp]
    \centering
    \includegraphics[width=\textwidth]{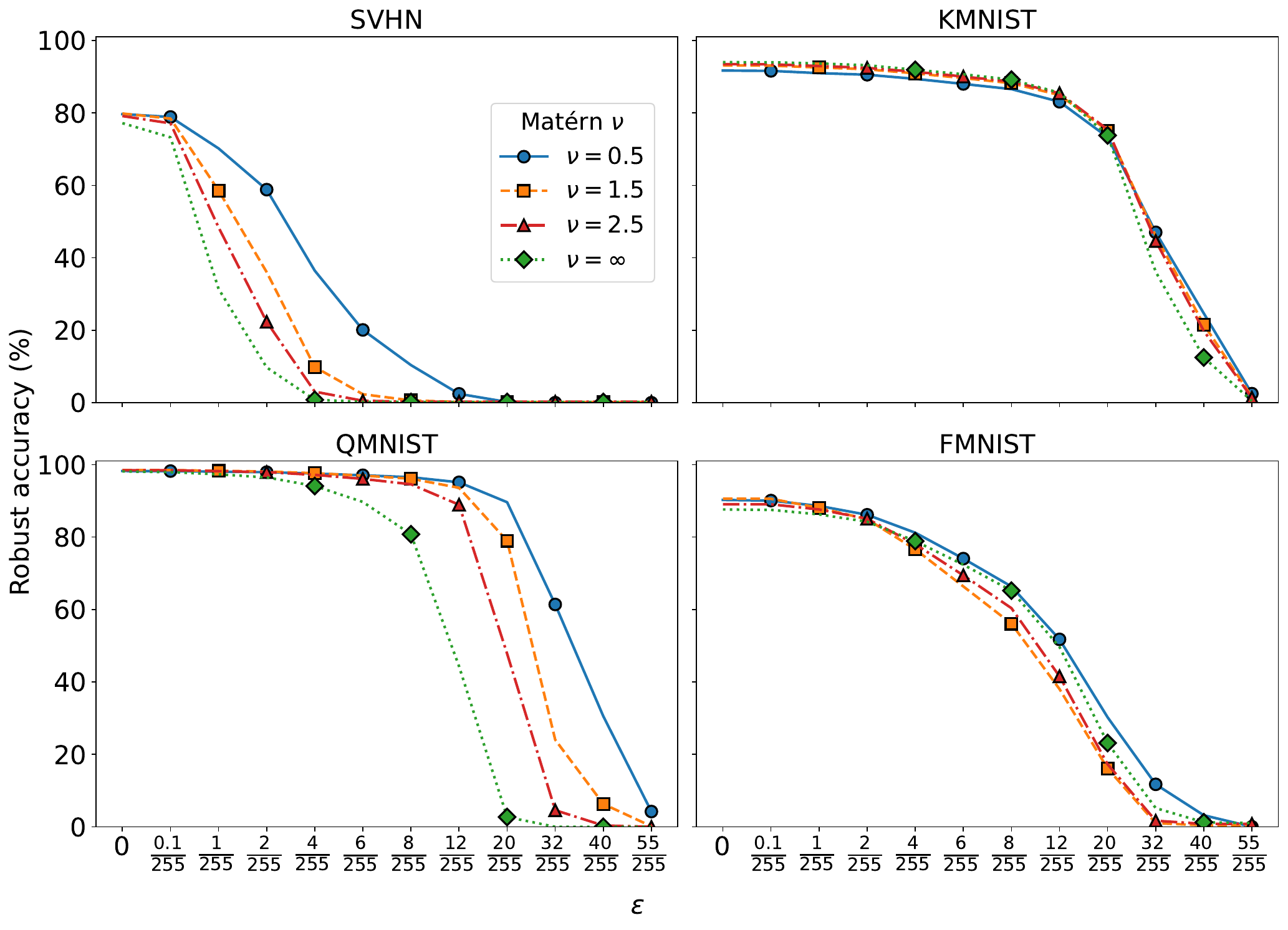}
    \caption{
        \textbf{Robust Accuracy} variation with \textbf{Mat\'ern Kernel} smoothness parameter \textbf{$\nu$} for \textbf{SVHN, KMNIST, QMNIST, FMNIST.}
    }
    \label{fig:matern}
\end{figure*}

We observe that the robust accuracy of Mat\'ern Kernel decreases with increase in the smoothness parameter($\nu$) value in SVHN, KMNIST, QMNIST. In FashionMNIST robust accuracies stay quite close to each other, although not following a clear pattern. Further details about the experiment are in ~\cref{apd:mat_kernel}, with results in ~\cref{fig:matern}. PGD is used to attack the Kernel.

\vspace{15mm}

\clearpage
\subsection{Robust Accuracy variation with iteration in RFMs}\label{apd: rfm_iter_results}
\begin{figure*}[!htbp]
    \centering
    \includegraphics[width=\textwidth]{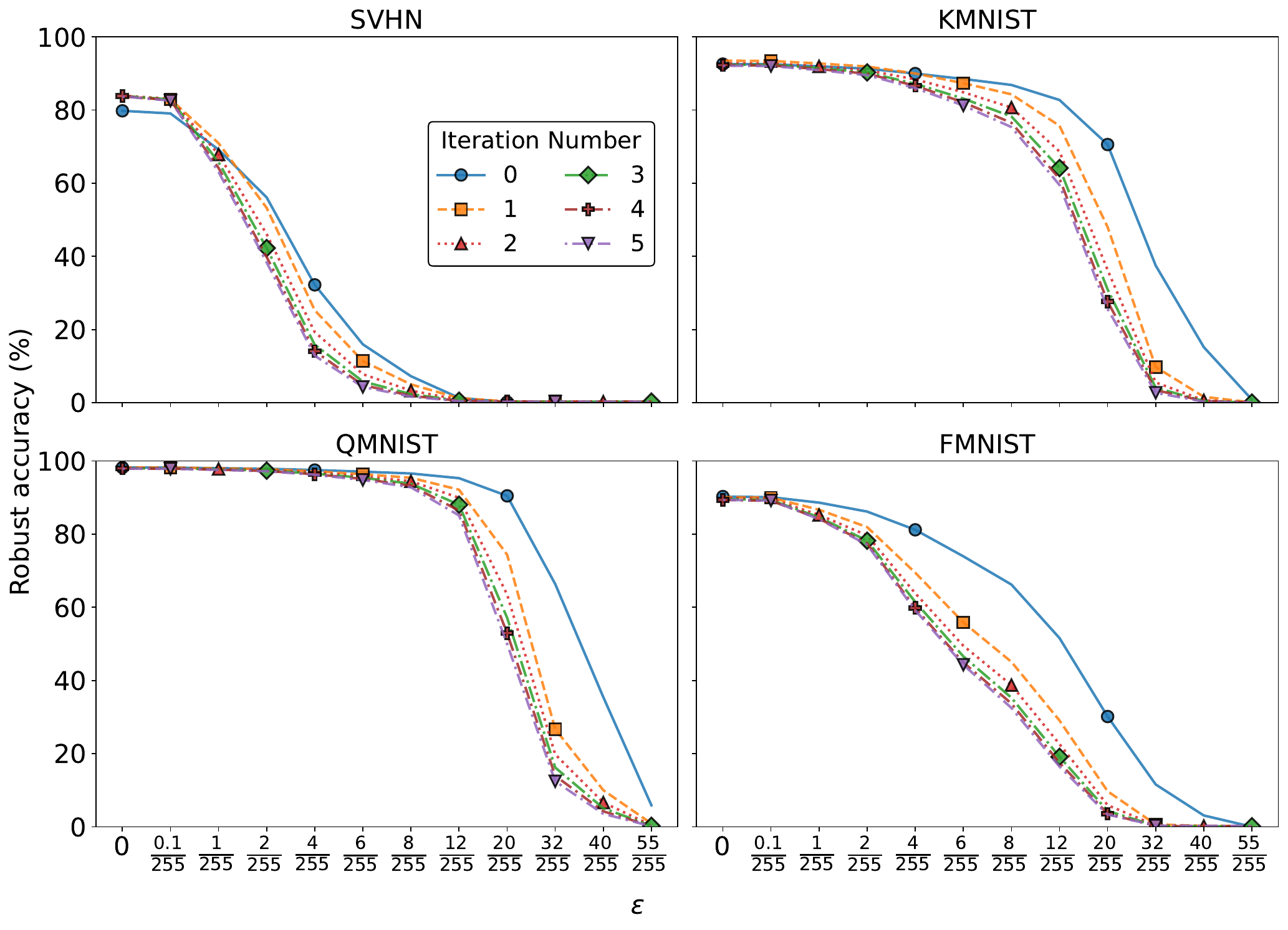}
    \caption{
        \textbf{Robust Accuracy} variation with \textbf{iterations in Laplacian RFM} for \textbf{SVHN, KMNIST, QMNIST, FMNIST.}
    }
    \label{fig:rfm_v_iter}
\end{figure*}

RFMs are KM variant that learn features iteratively. This compels us to ask what happens to the robust accuracy as RFM learns features. The results in~\cref{fig:rfm_v_iter} show that RFMs' robust accuracy decreases with an increase in the feature learning (or with number of iterations), specifically for Laplacian RFMs. We attacked the models using PGD.

\vspace{18mm}

\clearpage
\subsection{Robust Accuracy variation with length scale (\texorpdfstring{$\sigma$}{sigma})}\label{apd: sigma_robust_results}

\begin{figure*}[!htbp]
    \centering
    \includegraphics[width=\textwidth]{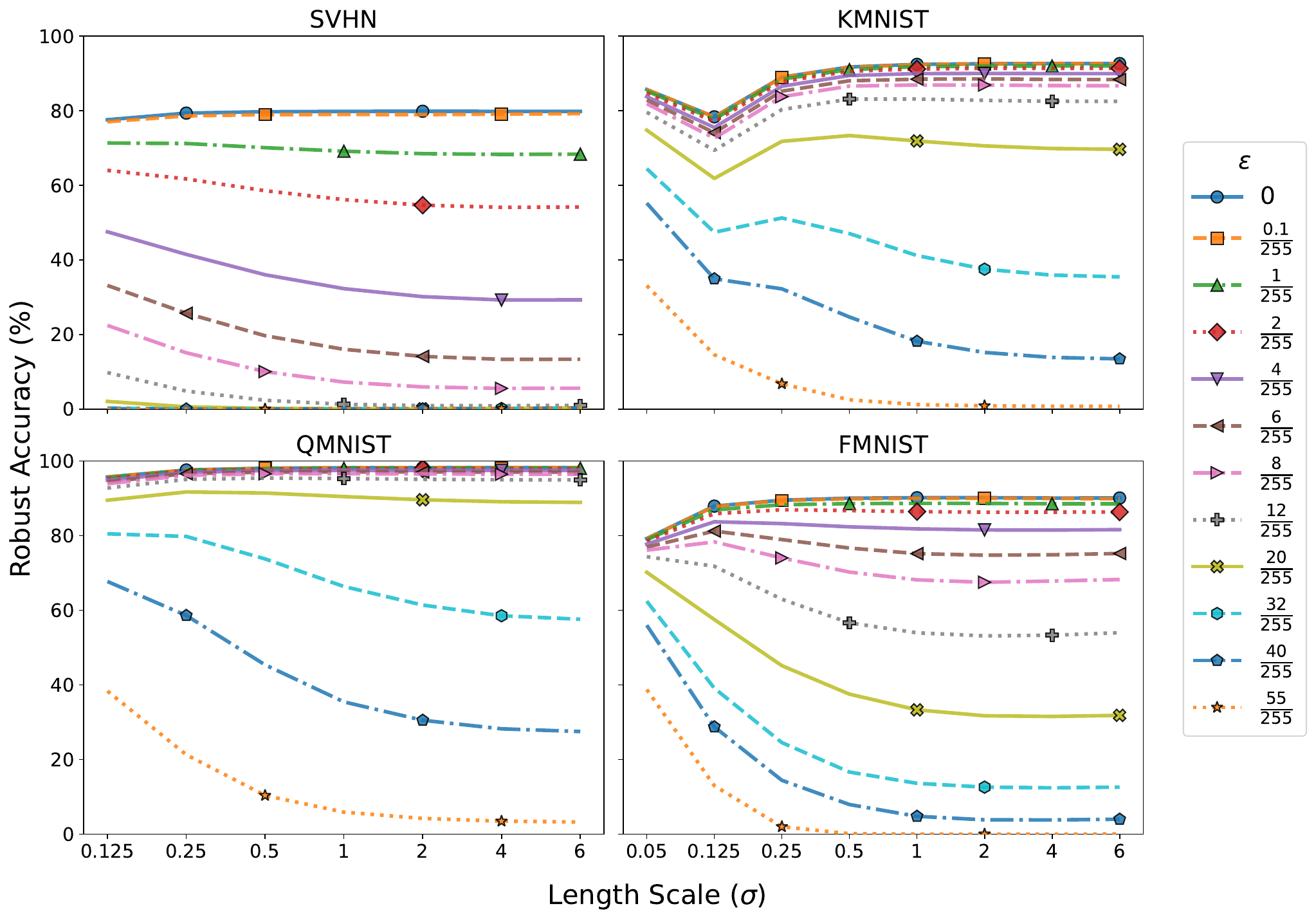}
    \caption{
        \textbf{Robust Accuracy} variation with \textbf{length scale ($\sigma$) in Laplacian Kernel} for \textbf{SVHN, KMNIST, QMNIST, FMNIST} across different attack $\varepsilon$.
    }
    \label{fig:length_scale_vs_robustness}
\end{figure*}

\begin{figure*}[!htbp]
    \centering
    \includegraphics[width=\textwidth]{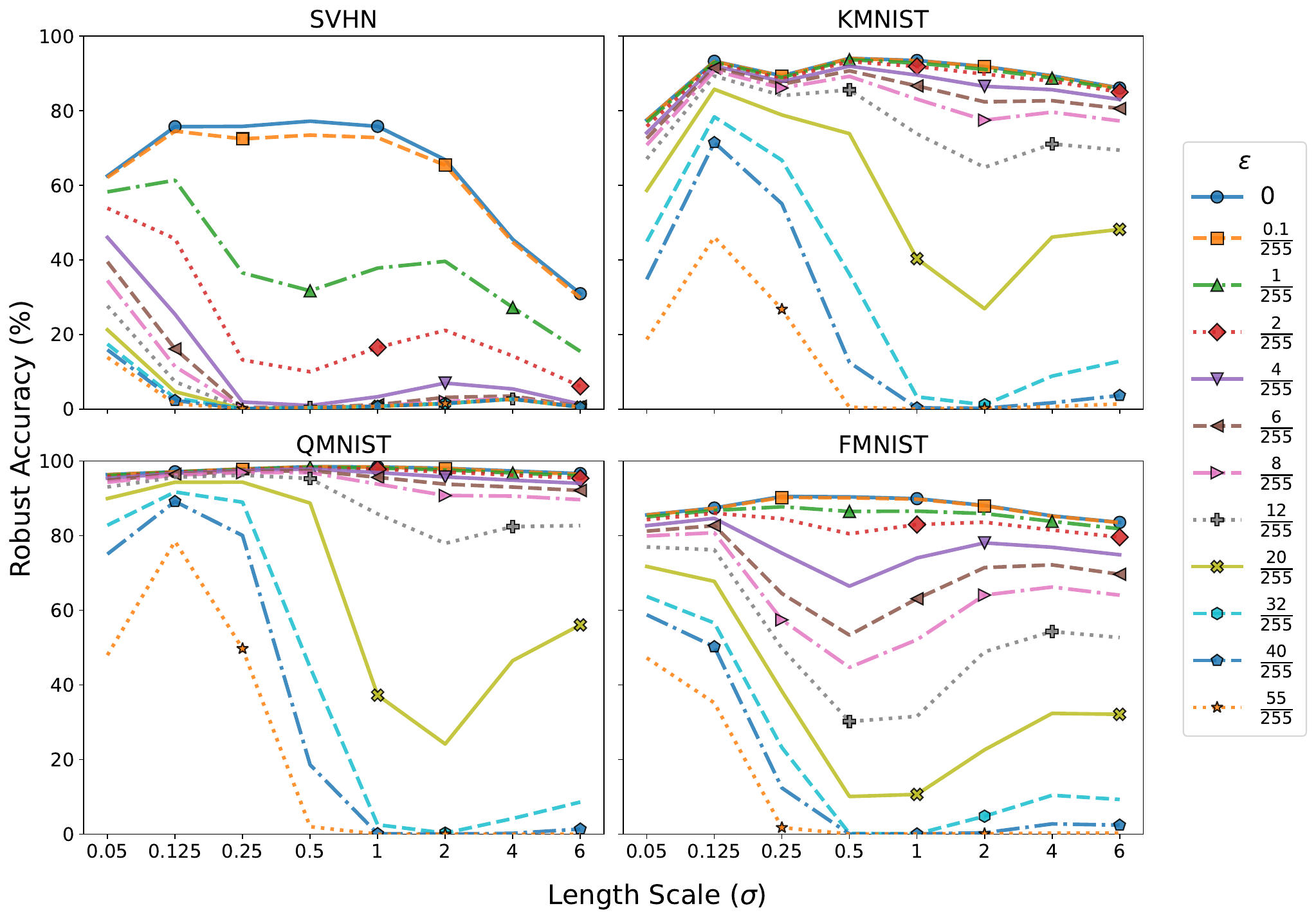}
    \caption{
        \textbf{Robust Accuracy} variation with \textbf{length scale ($\sigma$) in Gaussian Kernel} for \textbf{SVHN, KMNIST, QMNIST, FMNIST} across different attack $\varepsilon$.
    }
    \label{fig:length_scale_vs_robustness_gaus}
\end{figure*}

 \cref{fig:length_scale_vs_robustness} shows that, for the Laplacian kernel, adversarial robustness generally favours smaller length scales \(\sigma\). Even though smaller \(\sigma\) can yield lower benign accuracy, it often achieves higher robust accuracy as the attack strength increases, indicating a trade-off between benign performance and robustness. The corresponding Gaussian-kernel results are shown in~\cref{fig:length_scale_vs_robustness_gaus}, where the dependence on \(\sigma\) is less monotonic and more dataset dependent. Ridge ($\lambda$) was kept the same across different length scales ($\sigma$). Results are given for the KRR. Models are attacked using PGD.

\vspace{25mm}

\clearpage
\subsection{Robust Accuracy variation with ridge (\texorpdfstring{$\lambda$}{lambda})}\label{apd: ridge_robust_results}
\begin{figure*}[!htbp]
    \centering
    \includegraphics[width=\textwidth]{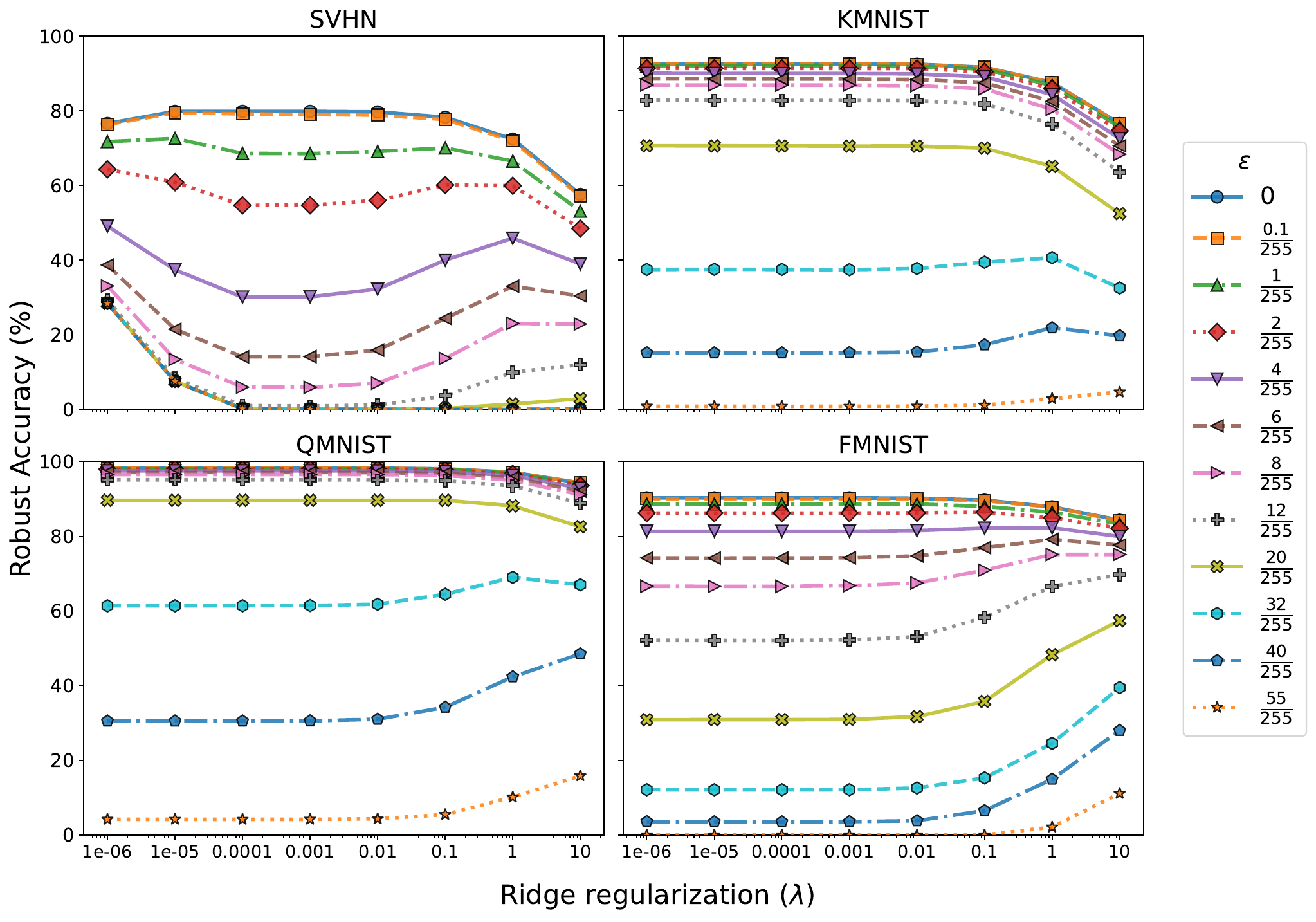}
    \caption{
        \textbf{Robust Accuracy} variation with \textbf{ridge ($\lambda$) in Laplacian KRR} for \textbf{SVHN, KMNIST, QMNIST, FMNIST} across different attack $\varepsilon$.
    }
    \label{fig:ridge_vs_robust}
\end{figure*}

We observe in~\cref{fig:ridge_vs_robust} that the effect of ridge regularization on KRR is dataset dependent. While small-to-moderate ridge values largely preserve performance, stronger regularization can improve robustness under larger perturbations for FMNIST, KMNIST and QMNIST, but does not provide a consistent benefit for SVHN. This indicates a trade-off between benign accuracy and adversarial robustness. No single ridge value comes out to be optimal across all datasets and attack strengths. Length Scale ($\sigma$) was kept the same across different ridge ($\lambda$) values. PGD is used to attack KRR>

\vspace{25mm}

\clearpage
\subsection{Number of features in RFC versus robust accuracy}\label{apd: num_of_ftr_vs_robust_results}

\begin{figure*}[!htbp]
    \centering
    \includegraphics[width=\textwidth]{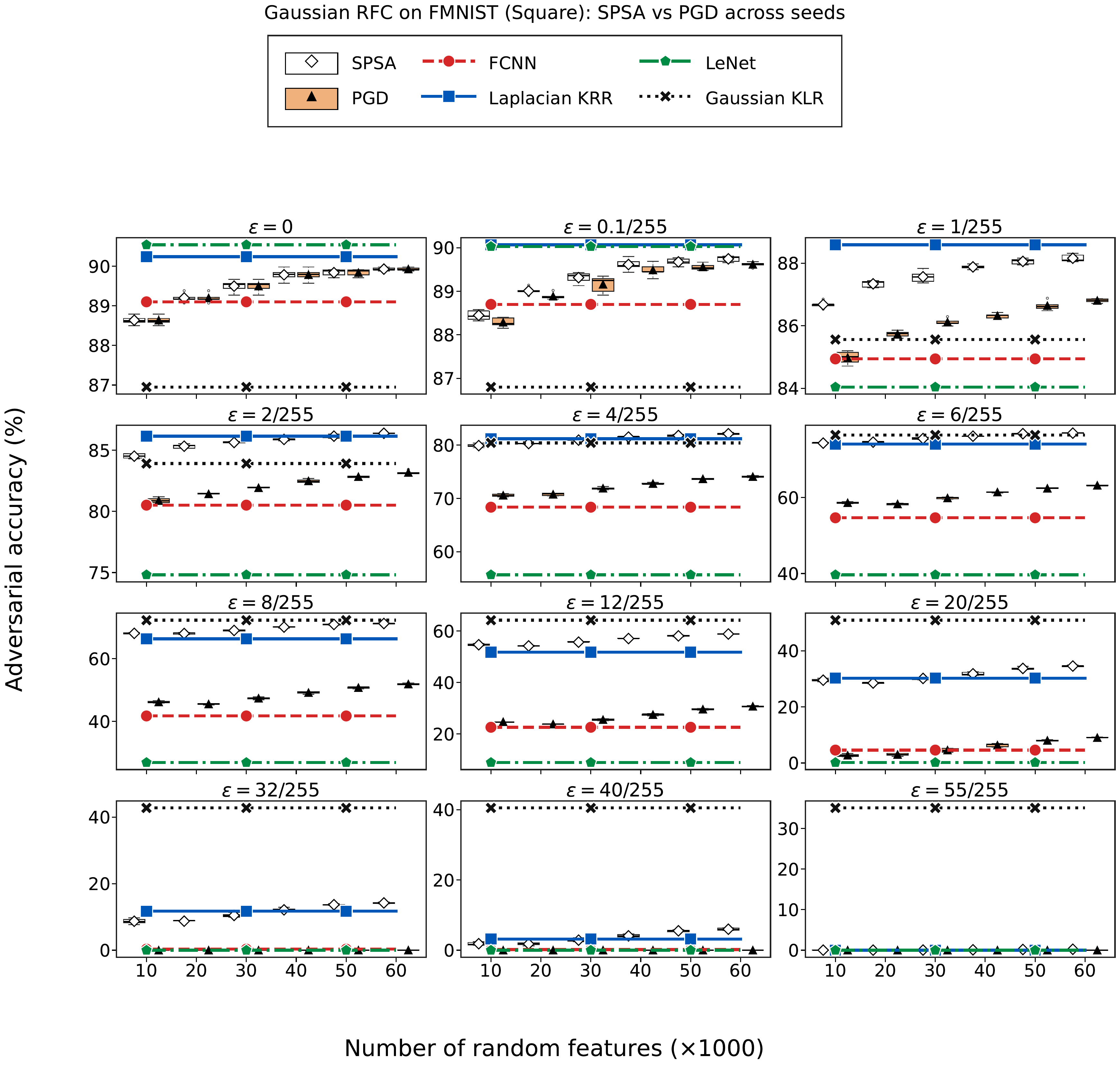}
    \caption{
        \textbf{Robust Accuracy} variation with \textbf{Number of Features in Gaussian RFC (trained using Square Loss)} for \textbf{FMNIST} across different seeds per attack. SPSA and PGD denote attacks on the RFC, each line shows the robust accuracy of a model at a given perturbation radius ($\varepsilon$).
    }
    \label{fig:num_of_robust_gaus_sq_fmnist}
\end{figure*}

\begin{figure*}[!htbp]
    \centering
    \includegraphics[width=\textwidth]{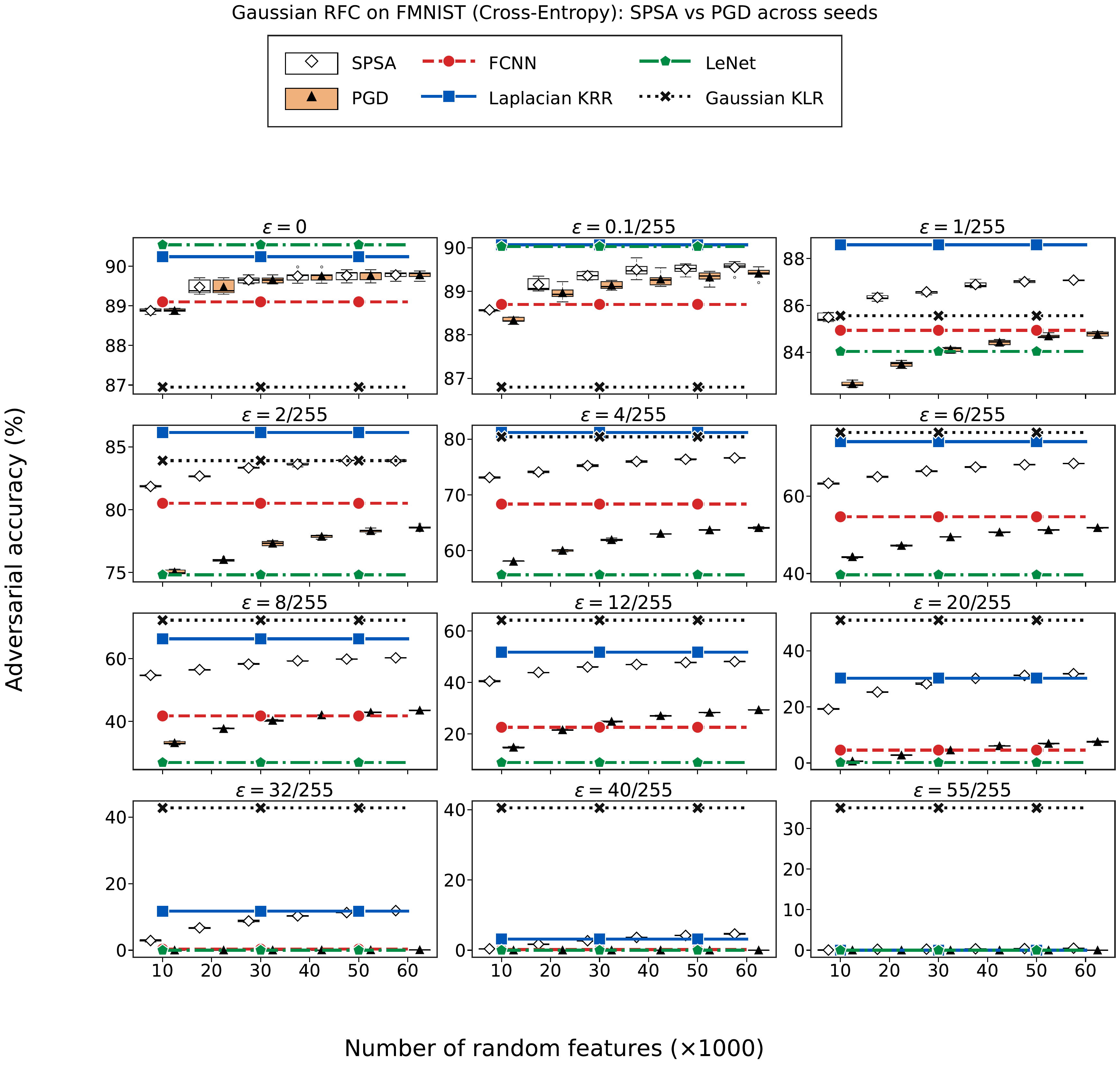}
    \caption{
        \textbf{Robust Accuracy} variation with \textbf{Number of Features in Gaussian RFC (trained using Cross Entropy)} for \textbf{FMNIST} across different seeds per attack. SPSA and PGD denote attacks on the RFC, each line shows the robust accuracy of a model at a given perturbation radius ($\varepsilon$).
    }
    \label{fig:num_of_robust_gaus_ce_fmnist}
\end{figure*}

\begin{figure*}[!htbp]
    \centering
    \includegraphics[width=\textwidth]{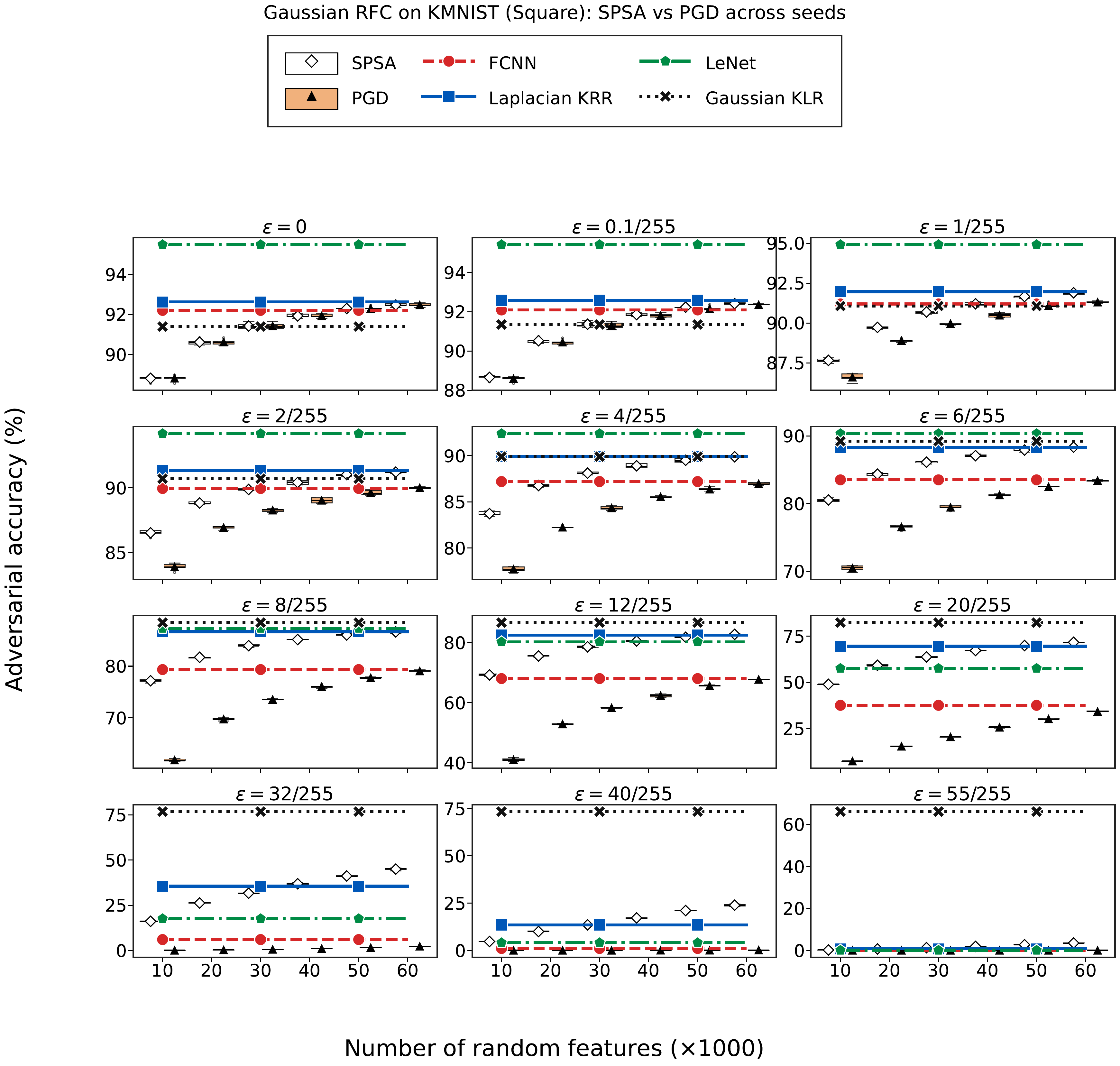}
    \caption{
        \textbf{Robust Accuracy} variation with \textbf{Number of Features in Gaussian RFC (trained using Square Loss)} for \textbf{KMNIST} across different seeds per attack. SPSA and PGD denote attacks on the RFC, each line shows the robust accuracy of a model at a given perturbation radius ($\varepsilon$).
    }
    \label{fig:num_of_robust_gaus_sq_kmnist}
\end{figure*}

\begin{figure*}[!htbp]
    \centering
    \includegraphics[width=\textwidth]{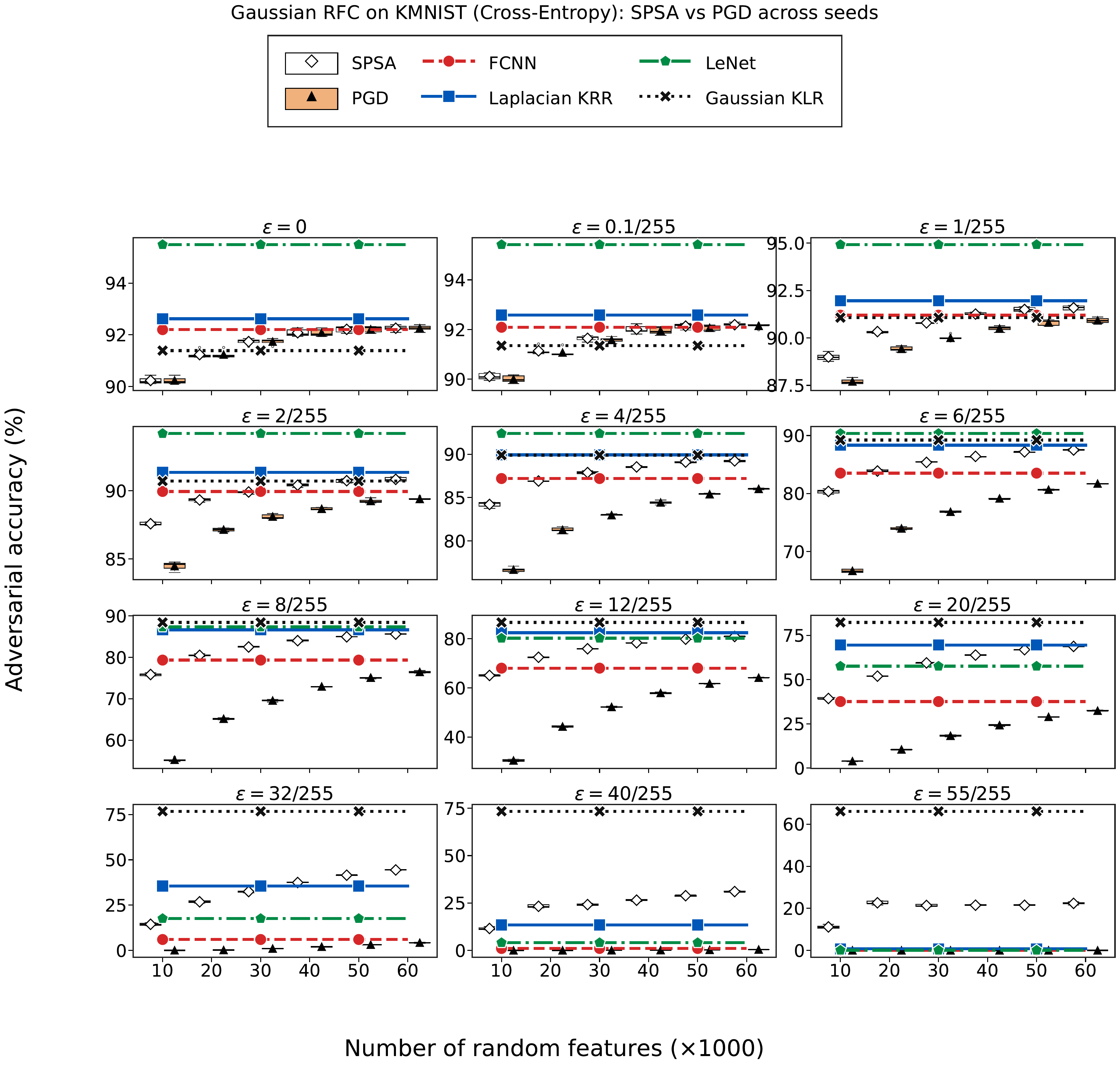}
    \caption{
        \textbf{Robust Accuracy} variation with \textbf{Number of Features in Gaussian RFC (trained using Cross Entropy)} for \textbf{KMNIST} across different seeds per attack. SPSA and PGD denote attacks on the RFC, each line shows the robust accuracy of a model at a given perturbation radius ($\varepsilon$).
    }
    \label{fig:num_of_robust_gaus_ce_kmnist}
\end{figure*}

\begin{figure*}[!htbp]
    \centering
    \includegraphics[width=\textwidth]{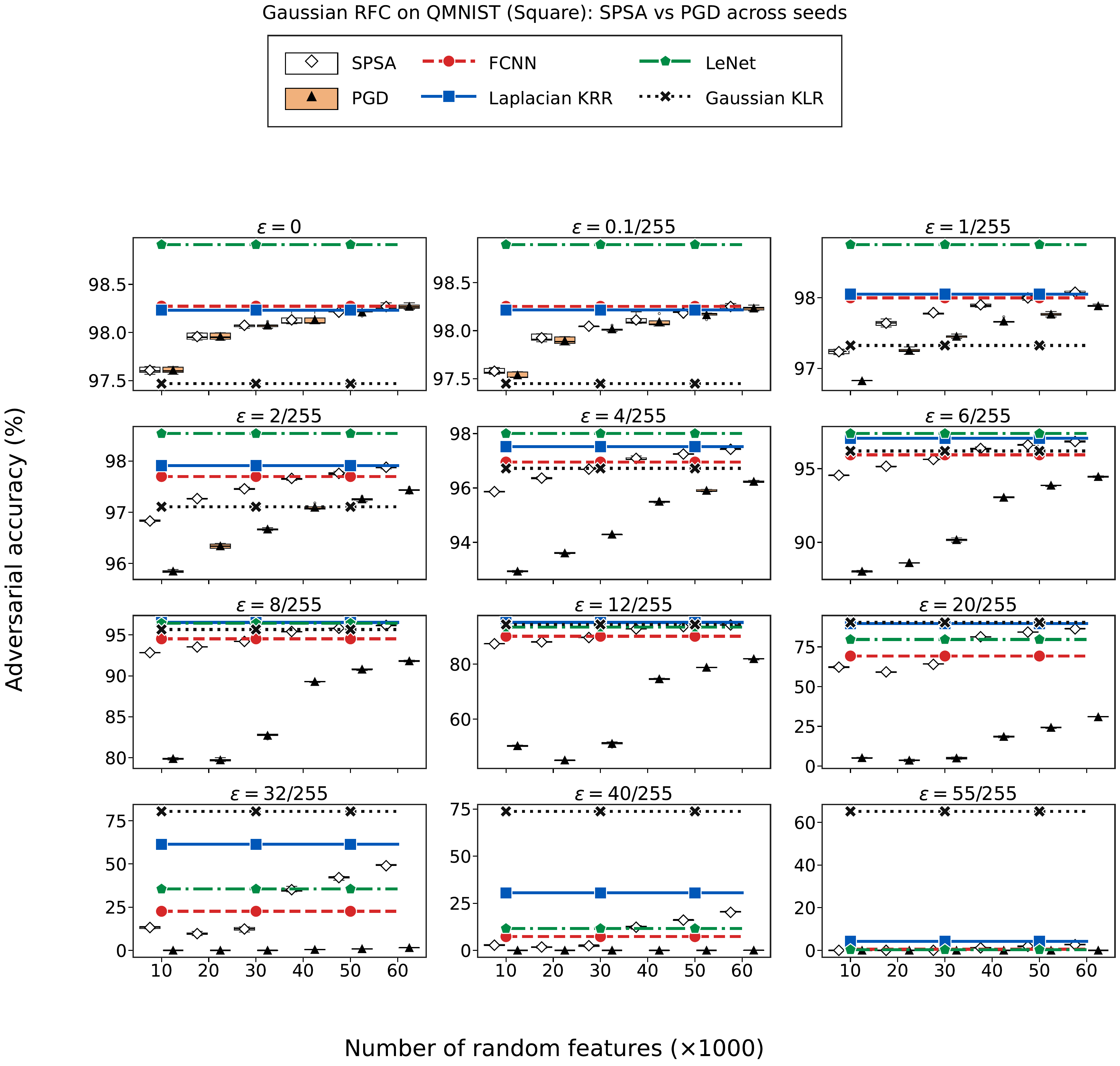}
    \caption{
        \textbf{Robust Accuracy} variation with \textbf{Number of Features in Gaussian RFC (trained using Square Loss)} for \textbf{QMNIST} across different seeds per attack. SPSA and PGD denote attacks on the RFC, each line shows the robust accuracy of a model at a given perturbation radius ($\varepsilon$).
    }
    \label{fig:num_of_robust_gaus_sq_qmnist}
\end{figure*}

\begin{figure*}[!htbp]
    \centering
    \includegraphics[width=\textwidth]{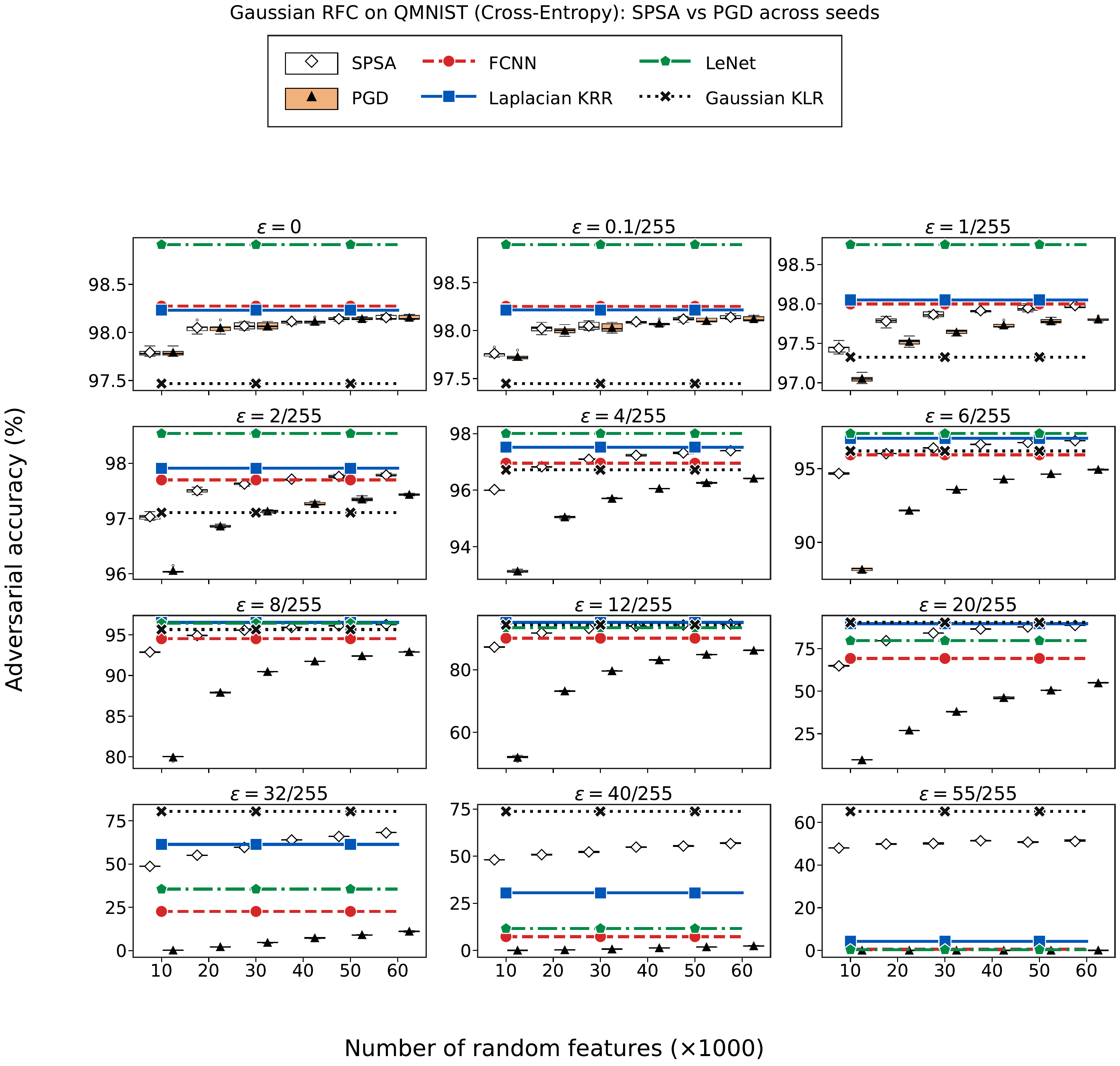}
    \caption{
        \textbf{Robust Accuracy} variation with \textbf{Number of Features in Gaussian RFC (trained using Cross Entropy Loss)} for \textbf{QMNIST} across different seeds per attack. SPSA and PGD denote attacks on the RFC, each line shows the robust accuracy of a model at a given perturbation radius ($\varepsilon$).
    }
    \label{fig:num_of_robust_gaus_ce_qmnist}
\end{figure*}

\begin{figure*}[!htbp]
    \centering
    \includegraphics[width=\textwidth]{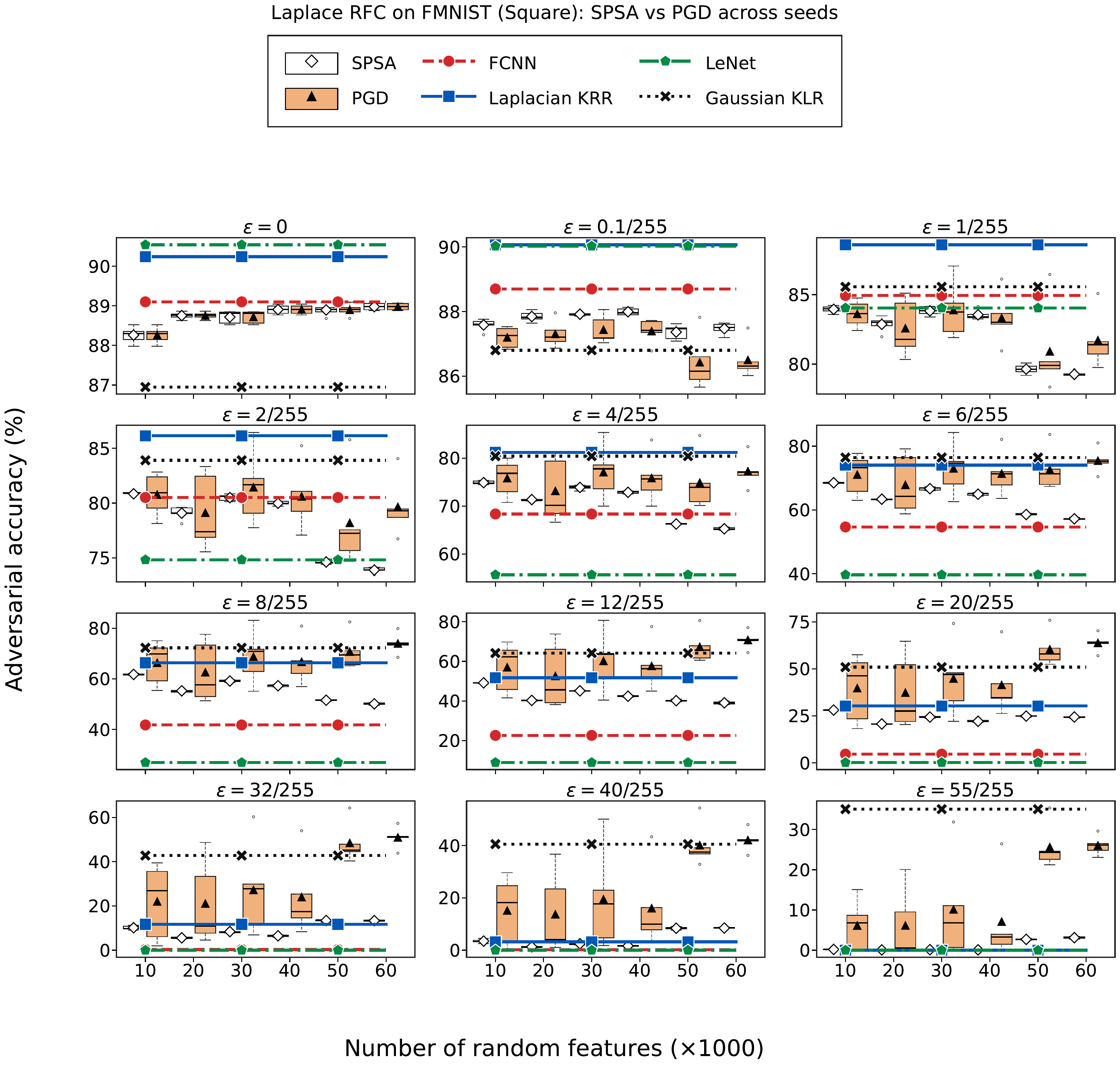}
    \caption{
        \textbf{Robust Accuracy} variation with \textbf{Number of Features in Laplace RFC (trained using Square Loss)} for \textbf{FMNIST} across different seeds per attack. SPSA and PGD denote attacks on the RFC, each line shows the robust accuracy of a model at a given perturbation radius ($\varepsilon$).
    }
    \label{fig:num_of_robust_lap_sq_fmnist}
\end{figure*}

\begin{figure*}[!htbp]
    \centering
    \includegraphics[width=\textwidth]{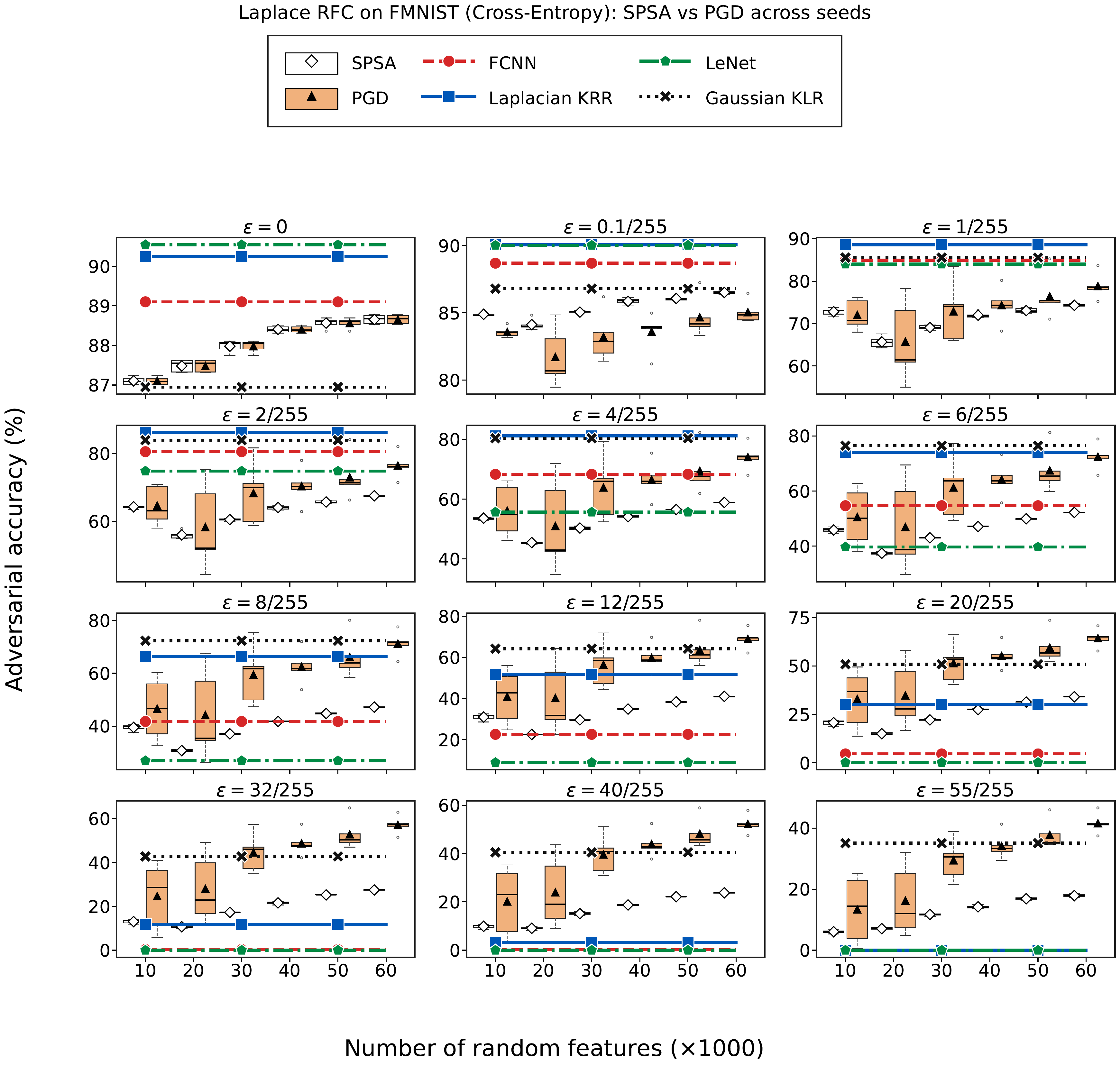}
    \caption{
        \textbf{Robust Accuracy} variation with \textbf{Number of Features in Laplacian RFC (trained using Cross Entropy)} for \textbf{FMNIST} across different seeds per attack. SPSA and PGD denote attacks on the RFC, each line shows the robust accuracy of a model at a given perturbation radius ($\varepsilon$).
    }
    \label{fig:num_of_robust_lap_ce_fmnist}
\end{figure*}

\begin{figure*}[!htbp]
    \centering
    \includegraphics[width=\textwidth]{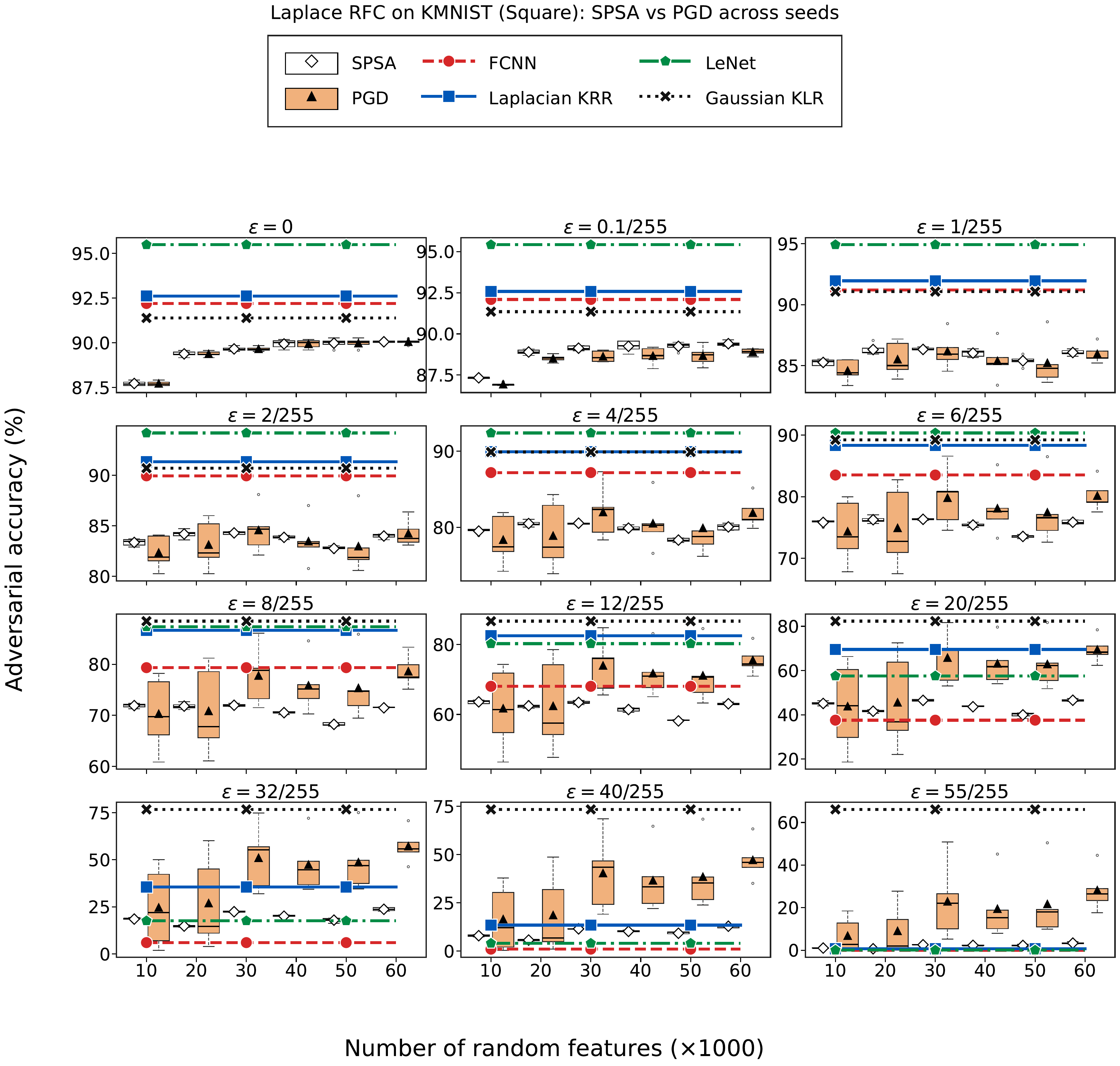}
    \caption{
        \textbf{Robust Accuracy} variation with \textbf{Number of Features in Laplace RFC (trained using Square Loss)} for \textbf{KMNIST} across different seeds per attack. SPSA and PGD denote attacks on the RFC, each line shows the robust accuracy of a model at a given perturbation radius ($\varepsilon$).
    }
    \label{fig:num_of_robust_lap_sq_kmnist}
\end{figure*}

\begin{figure*}[!htbp]
    \centering
    \includegraphics[width=\textwidth]{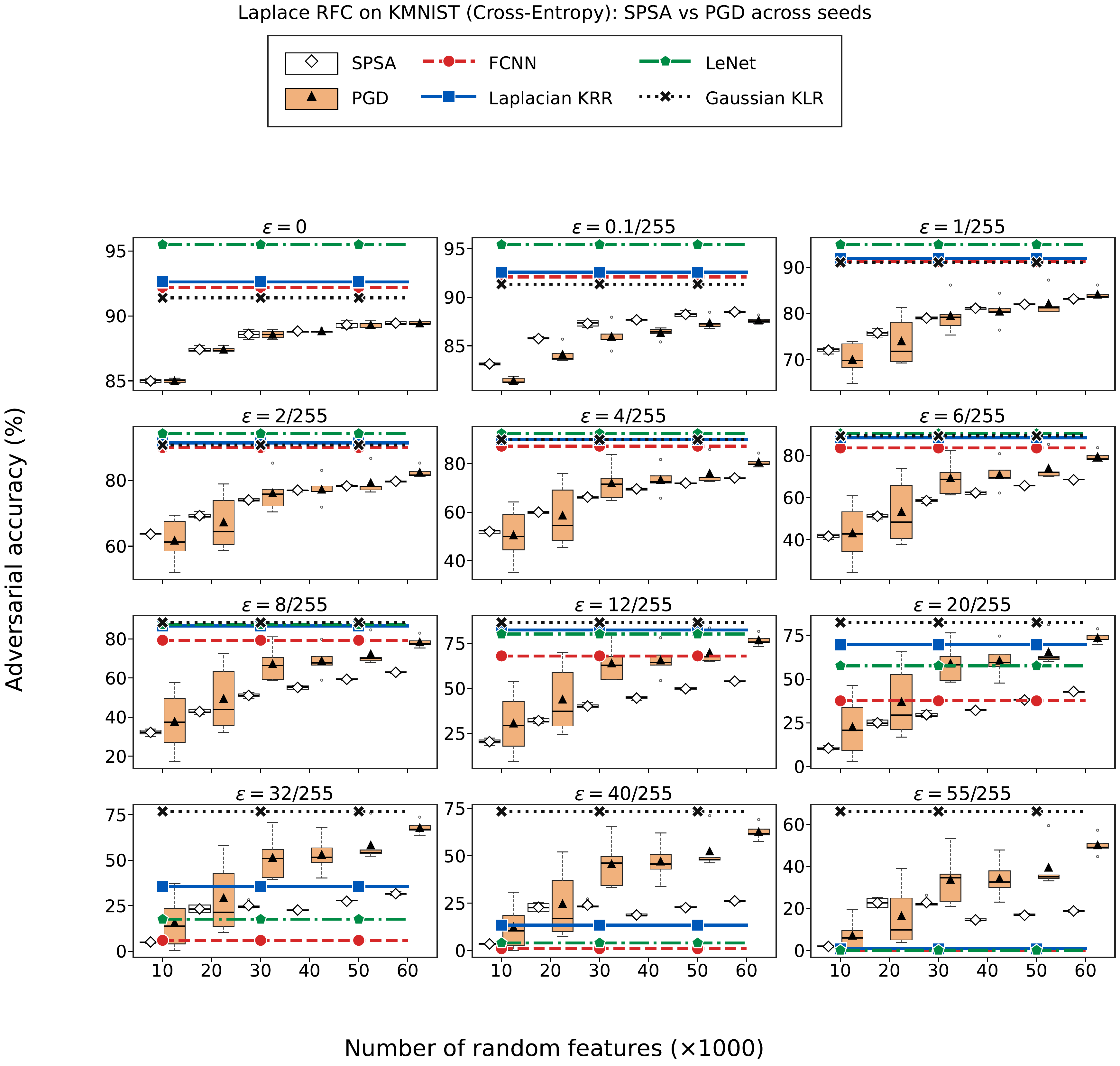}
    \caption{
        \textbf{Robust Accuracy} variation with \textbf{Number of Features in Laplacian RFC (trained using Cross Entropy)} for \textbf{KMNIST} across different seeds per attack. SPSA and PGD denote attacks on the RFC, each line shows the robust accuracy of a model at a given perturbation radius ($\varepsilon$).
    }
    \label{fig:num_of_robust_lap_ce_kmnist}
\end{figure*}

\begin{figure*}[!htbp]
    \centering
    \includegraphics[width=\textwidth]{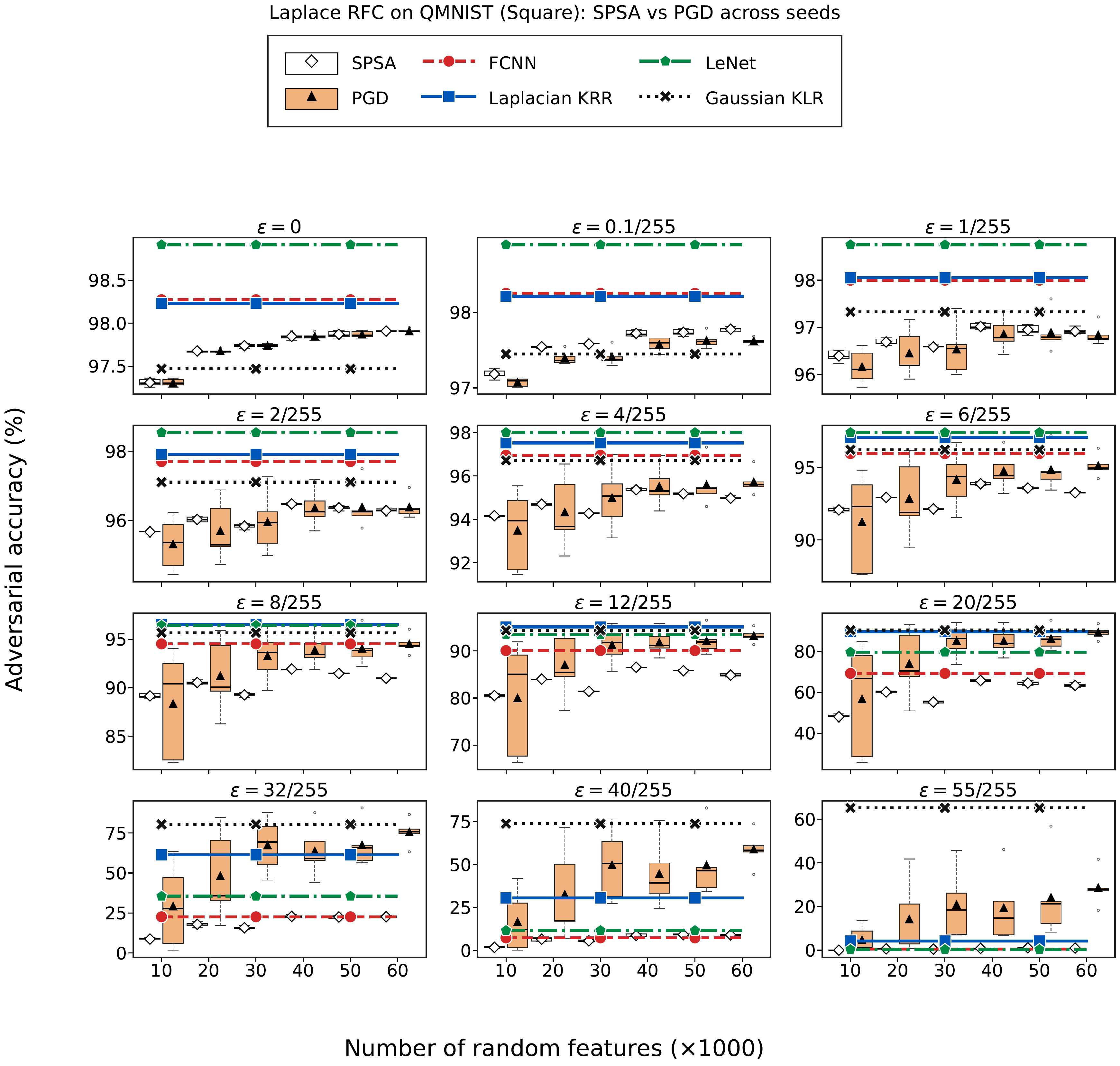}
    \caption{
        \textbf{Robust Accuracy} variation with \textbf{Number of Features in Laplace RFC (trained using Square Loss)} for \textbf{QMNIST} across different seeds per attack. SPSA and PGD denote attacks on the RFC, each line shows the robust accuracy of a model at a given perturbation radius ($\varepsilon$).
    }
    \label{fig:num_of_robust_lap_sq_qmnist}
\end{figure*}

\begin{figure*}[!htbp]
    \centering
    \includegraphics[width=\textwidth]{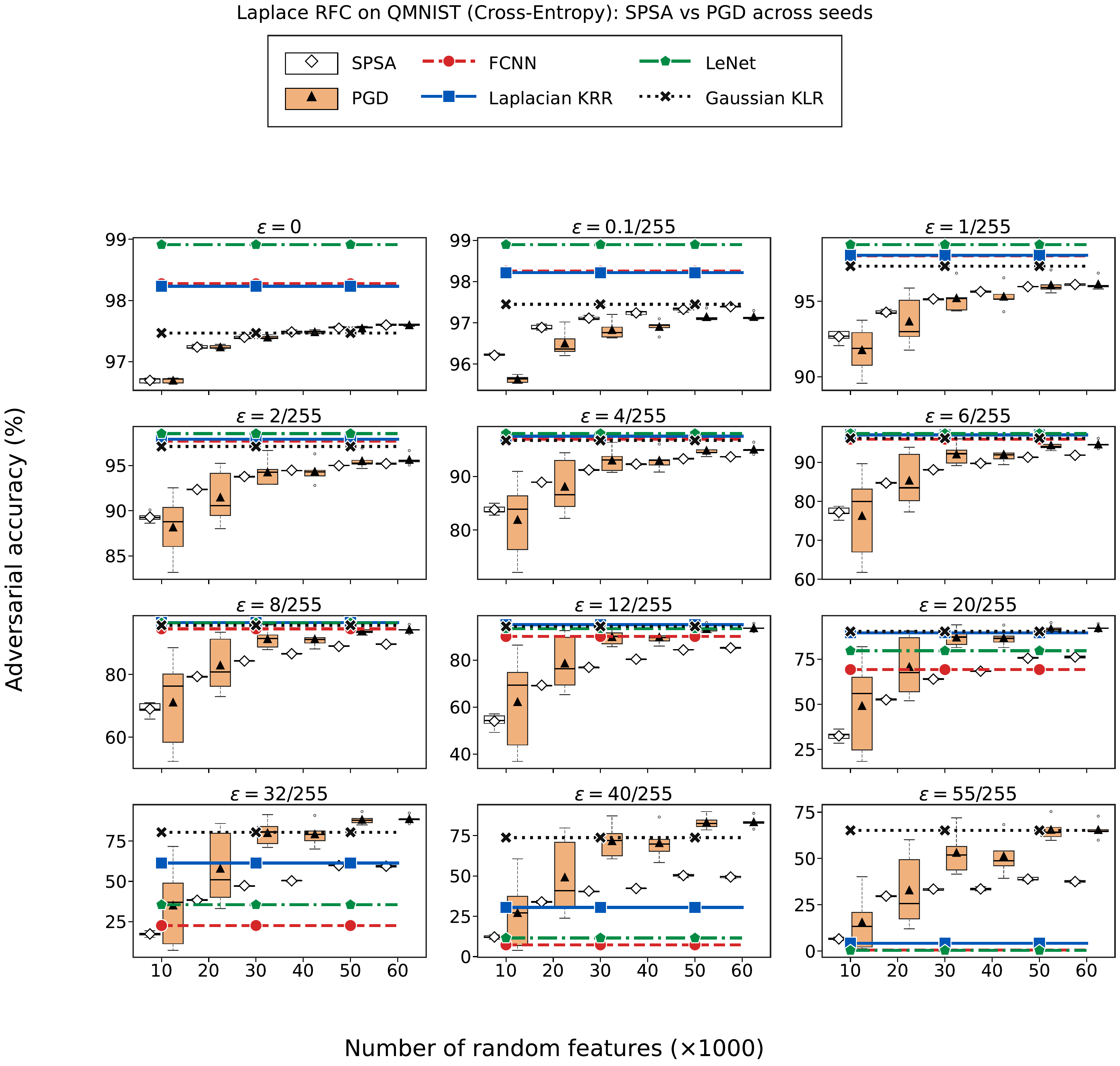}
    \caption{
        \textbf{Robust Accuracy} variation with \textbf{Number of Features in Laplace RFC (trained using Cross Entropy Loss)} for \textbf{QMNIST} across different seeds per attack. SPSA and PGD denote attacks on the RFC, each line shows the robust accuracy of a model at a given perturbation radius ($\varepsilon$).
    }
    \label{fig:num_of_robust_lap_ce_qmnist}
\end{figure*}

As is evident from ~\cref{fig:num_of_robust_gaus_sq_fmnist} to ~\cref{fig:num_of_robust_lap_ce_qmnist}, both the benign accuracy as well as robust accuracy improves with an increase in the number of features, except when we train Laplace RFC with square loss. In Laplace RFCs, trained with square loss, benign accuracy improves with an increase in the number of features, but that is not the case with robust accuracy. We observe gradient masking in the case of Laplace RFC, no matter the training loss, as is seen in~\cref{fig:num_of_robust_lap_sq_fmnist,fig:num_of_robust_lap_ce_fmnist,fig:num_of_robust_lap_sq_kmnist,fig:num_of_robust_lap_ce_kmnist,fig:num_of_robust_lap_sq_qmnist,fig:num_of_robust_lap_ce_qmnist}. 5 different seeds are used for RFCs.

Ablation results for number of features vs. robust accuracy for Gaussian RFC on FMNIST, KMNIST, QMNIST for square and cross-entropy loss is given in~\cref{fig:num_of_robust_gaus_sq_fmnist} to~\cref{fig:num_of_robust_gaus_ce_qmnist}.
Results for Laplacian Kernel trained using square and cross entropy training loss range from~\cref{fig:num_of_robust_lap_sq_fmnist} to~\cref{fig:num_of_robust_lap_ce_qmnist}.
\subsubsection{FMNIST Results for Gaussian RFC}\label{apd: num_of_ftr_vs_acc_fmnist_gaus}
\vspace{10mm}

\subsubsection{KMNIST Results for Gaussian RFC}\label{apd: num_of_ftr_vs_acc_kmnist_gaus}
\vspace{10mm}

\clearpage
\subsubsection{QMNIST Results for Gaussian RFC}\label{apd: num_of_ftr_vs_acc_qmnist_gaus}
\vspace{10mm}

\clearpage
\subsubsection{FMNIST Results for Laplace RFC}\label{apd: num_of_ftr_vs_acc_fmnist_lap}
\vspace{10mm}

\clearpage
\subsubsection{KMNIST Results for Laplace RFC}\label{apd: num_of_ftr_vs_acc_kmnist_lap}
\vspace{10mm}

\clearpage
\subsubsection{QMNIST Results for Laplace RFC}\label{apd: num_of_ftr_vs_acc_qmnist_lap}
\vspace{10mm}

\end{document}